\documentclass[10pt]{article}
\usepackage[preprint]{tmlr}

\usepackage{amsmath,amsfonts,bm}

\def\eqref#1{equation~\ref{#1}}

\def\1{\bm{1}}

\DeclareMathAlphabet{\mathsfit}{\encodingdefault}{\sfdefault}{m}{sl}
\SetMathAlphabet{\mathsfit}{bold}{\encodingdefault}{\sfdefault}{bx}{n}

\usepackage{hyperref}
\usepackage{url}

\usepackage[utf8]{inputenc}
\usepackage[T1]{fontenc}
\usepackage{hyperref}
\usepackage{url}
\usepackage{booktabs}
\usepackage{amsfonts}
\usepackage{nicefrac}
\usepackage{microtype}
\usepackage{xcolor}
\usepackage{bibunits}
\usepackage{hyperref}
\usepackage[capitalize]{cleveref}

\usepackage{graphicx}
\usepackage{enumitem}
\usepackage{array}
\usepackage{wrapfig}
\usepackage{listings}
\usepackage{subcaption}
\usepackage{amsmath}
\usepackage{amssymb}
\usepackage[capitalize]{cleveref}
\usepackage[breakable]{tcolorbox}

\newcommand{\myparagraph}[1]{\vspace{0.5pt}\noindent{\bf{#1}}}

\definecolor{matchgreen}{HTML}{00DC00}
\definecolor{llmbg}{HTML}{f0f6fc}
\definecolor{llmbgframe}{HTML}{cccccc}

\definecolor{sbblue}{HTML}{0066cc}
\definecolor{sbgrey}{HTML}{666666}

\newcommand{\callout}[1]{%
\vspace{-0.1em}
    \begin{center}
    \begin{tcolorbox}[colframe=llmbgframe, colback=llmbg, coltext=sbblue!50!black, width=\columnwidth, left=.7mm, right=.7mm, top=1mm, bottom=1mm, boxrule=.5mm]
        #1
    \end{tcolorbox}
    \end{center}
    \vspace{-0.8em}
}

\newcounter{finding}
\newcommand{\finding}[1]{%
    \stepcounter{finding}
    \callout{\textbf{Observation~\thefinding:}\hspace{.5em}#1}
}

\newcommand{\mysummary}[1]{%
    \callout{#1}
}

\makeatletter
\def\onedot{\futurelet\@let@token\@onedot}
\def\@onedot{\ifx\@let@token.\else.\null\fi\xspace}
\makeatother
\usepackage{xspace}
\def\etal{\emph{et al}\onedot}

\title{Back into Plato's Cave: Examining Cross-modal\\ Representational Convergence at Scale}

\author{%
  A. Sophia Koepke$^{1,2,3}$ \quad
  Daniil Zverev$^{2}$ \quad
  Shiry Ginosar$^{4}$ \quad
  Alexei A. Efros$^{1}$ \\ [1.8ex]
  $^{1}$UC Berkeley \quad
  $^{2}$Technical University of Munich, MCML \\ [1ex]
  $^{3}$University of T\"ubingen, T\"ubingen AI Center \quad
  $^{4}$Toyota Technical Institute at Chicago
}

\def\month{MM}
\def\year{YYYY}
\def\openreview{\url{https://openreview.net/forum?id=XXXX}}

\begin{document}
\begin{bibunit}[tmlr]

\maketitle

\begin{abstract}
The Platonic Representation Hypothesis~\citep{huh2024prh} posits that neural networks trained on different modalities (e.g., text and images) converge toward a shared representation of reality. If true, this has significant implications for whether modality choice matters at all. In this paper, we show that the evidence for this claim is substantially weaker than subsequent work suggests. The mutual $k$-nearest-neighbor metric used on 1024 text--image pairs in the original study captures only coarse structure.
To keep the alignment from collapsing as one scales up the data, $k$ has to grow proportionally, undercutting the argument for fine-grained representational convergence.
The reported increase in alignment with language model strength saturates for recent models.
Moreover, the one-to-one text--image pairing favors alignment, while alignment decreases with non-bijective data. We further find that image and text representations indeed share coarse semantic structure, but neither stronger language models nor richer captions yield fine-grained alignment. Thus, multimodal representations share coarse structure without evidence of convergence to a shared representation -- arguably, full representational convergence would require fine-grained alignment.
\\[2.5px]
\textbf{Project page: \url{https://akoepke.github.io/cave_umwelten}}
\end{abstract}

\section{Introduction}\label{sec:intro}

The success of Large Language Models (LLMs) is causing hand-wringing in the computer vision community: do we need pixels to build machines that understand our world, or is language ``all you need''?

Several works have demonstrated that models trained only on text data successfully solve what were thought to be fundamentally visual problems, such as visual question answering~\citep{gu2023can,hu2023promptcap}, visual reasoning~\citep{chollet2019measureintelligence,hu2023context,wang2023hypothesis,akyurek2024surprising}, or embodied robotics applications~\citep{ahn2022can,liang2023code}.
This resonates with the suggestion that text data may make other modalities redundant~\citep{Sutskever2023}, on the premise that the part of the world that is relevant to humans is manifest in language.
On the other hand, it is argued that linguistic data alone cannot yield genuine understanding~\citep{Yann22} or actual embodiment. After all, there is a reason we visit art museums rather than just read descriptions of paintings in a catalogue.
This raises a central question: how do models trained on different modalities represent reality?

The Platonic Representation Hypothesis~\citep{huh2024prh} offers a compelling answer: it suggests that as neural networks grow larger and consume more data, their learned representations will eventually converge toward a common representation of the underlying world, no matter which data modality (text, vision, audio, touch, etc.) was used for training.
Proponents of language-only learning have interpreted this as validation of their approach.
Since, according to the Platonic Representation Hypothesis, the choice of modality does not matter, as they all lead to the same shared representation, one might as well use language as the most convenient source of data.\footnote{But analogously, the same argument could be made for vision-only learning~\citep{phil_communication}.}
However, a hypothesis is only as strong as its evidence.
Here, the evidence rests on methodological choices that have largely gone unexamined.

\begin{figure}[t]
    \centering
    \includegraphics[width=0.49\linewidth,trim=240 270 5 60, clip]%
    {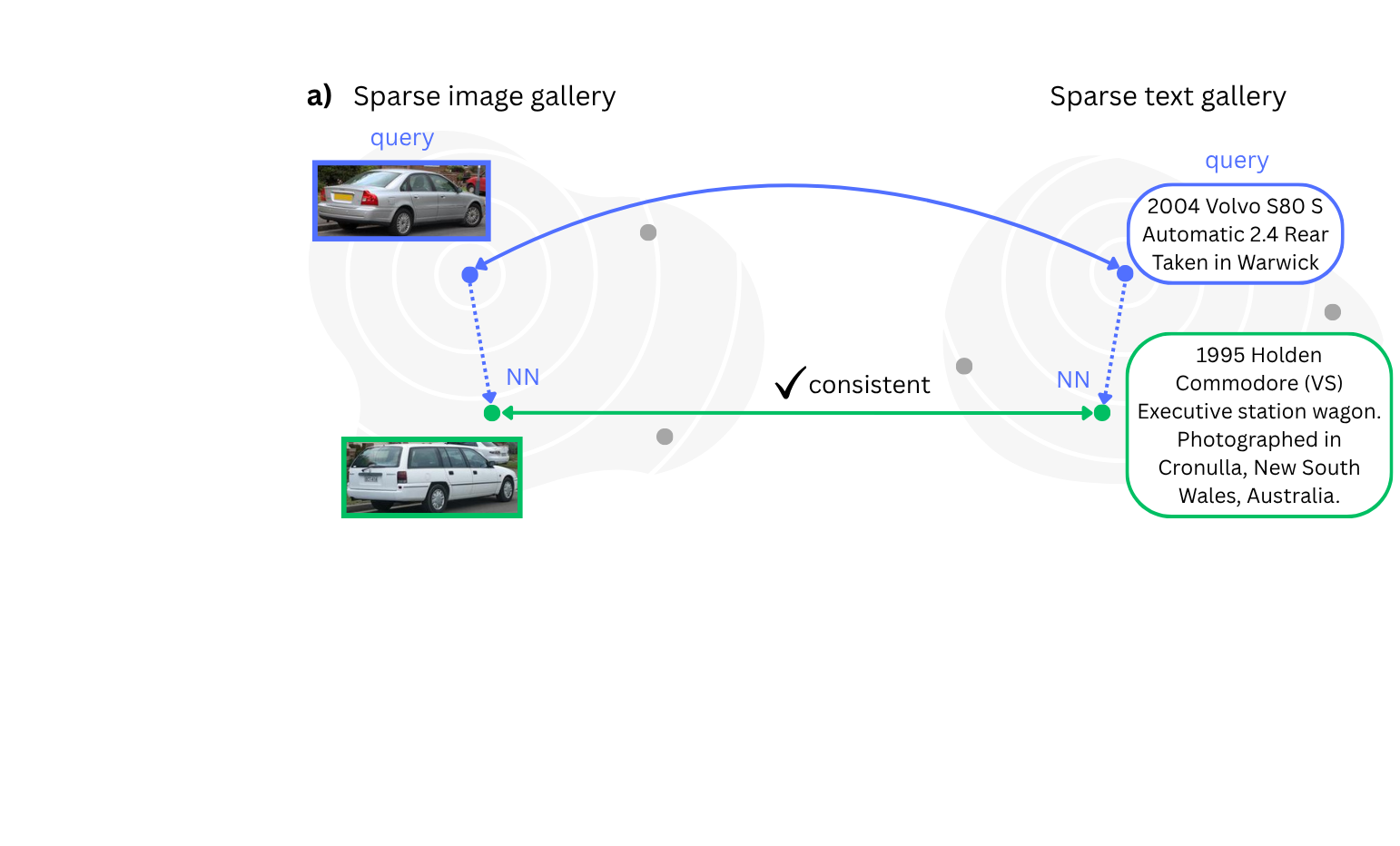}
    \hfill
    \rule{0.1pt}{0.19\linewidth}
    \hfill
    \includegraphics[width=0.49\linewidth,trim=240 270 5 60, clip]%
    {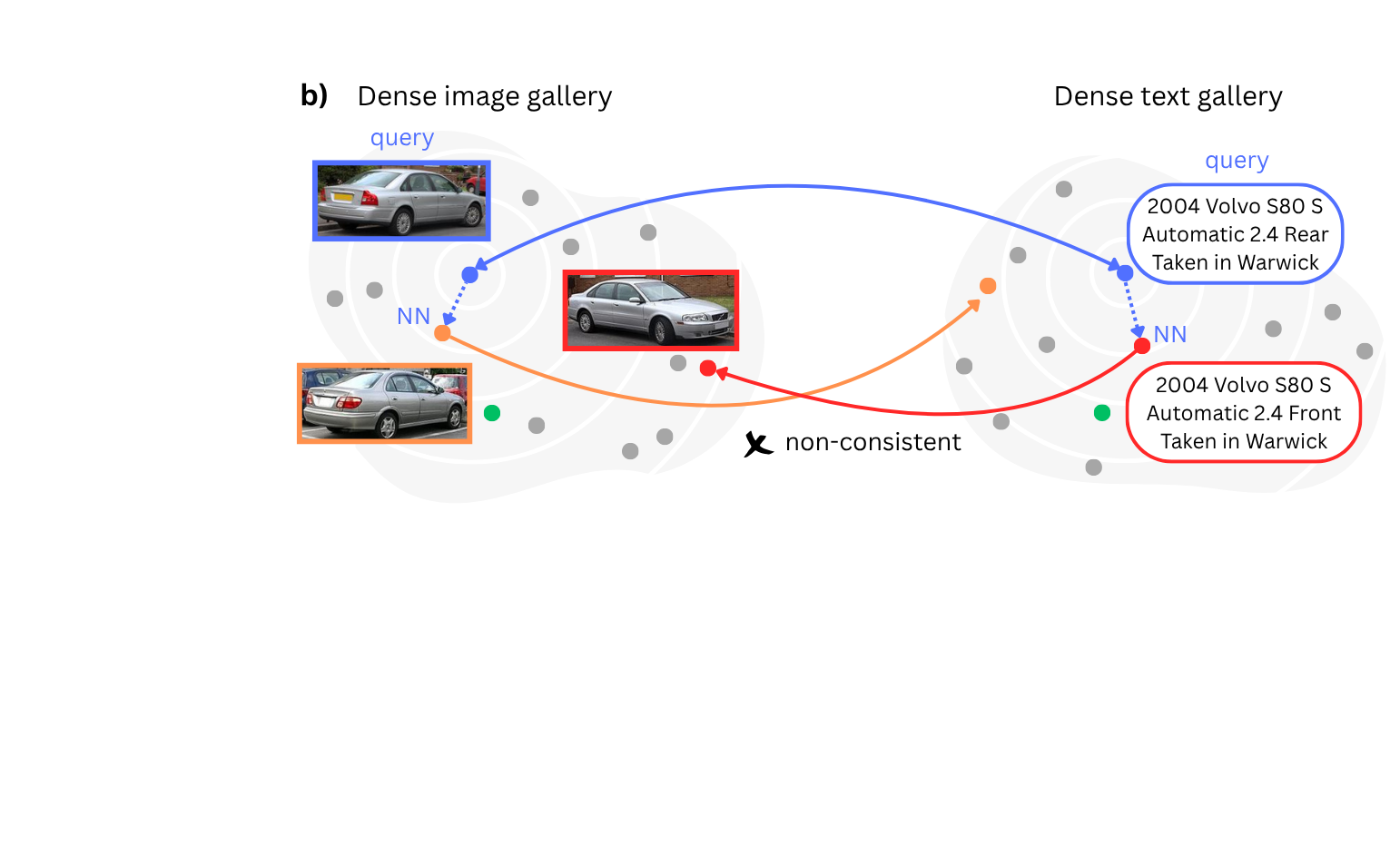}
   \vspace{-0.5em}
    \caption{Illustration of the mutual nearest neighbor metric used by \cite{huh2024prh} to measure cross-modal alignment.
    (a) Sparse regime: given a query image and caption (blue), nearest neighbors (NN) are retrieved independently in image and text embedding spaces. Mutual NN alignment measures whether the NNs are consistent across modalities. (b) Dense regime: as the dataset size increases, NNs within each modality get better. The vision model retrieves a car in the same pose, and the language model retrieves a caption of the same car model regardless of pose. Within-modality organization does not necessarily agree across modalities.}
    \label{fig:teaser}
    \vspace{-1.3em}
\end{figure}

In the first part of this paper, we take a close look at the experimental setting of \cite{huh2024prh}. They conducted their analysis on small, sparse datasets with one-to-one correspondences between modalities. However, real-world multi-modal data is large, dense, and inherently many-to-many. One image has many valid descriptions, and a single caption can correspond to many plausible images. These differences change what it means for two representations to ``align'' under the mutual nearest neighbor metrics.

In a sparse dataset, models can appear aligned simply because no better alternatives exist (\cref{fig:teaser}a). Here, two models can agree despite organizing their representations differently. As the dataset grows (i.e.\ the gallery used for retrieving neighbors gets denser), both models find closer neighbors (\cref{fig:teaser}b).
A vision model may retrieve an image of a car in the same pose as the query, while the language model retrieves a caption describing the query's car model. Both are individually valid, but produce a mismatch across modalities that the mutual nearest-neighbor metric penalizes.
Mutual nearest-neighbor agreement with fixed $k$ as used in \citet{huh2024prh} is an increasingly fine-grained measure for alignment as the gallery grows.

Indeed, when we increase the gallery from $1024$ to millions of samples, cross-modal mutual $k$NN agreement at fixed small $k$ drops, while same-modality pairs remain stable under identical densification. Rerunning the original setup from \cite{huh2024prh} with recent models and benchmarks, we further find that the reported alignment trend for increasingly strong language models saturates. Allowing multiple valid correspondences per sample reduces mutual $k$NN agreement, reflecting a limitation of the mutual $k$NN protocol.

In the second part of the paper, we characterize the cross-modal agreement that remains. We find that image and text models agree on what the world contains at a categorical level, while organizing fine-grained within-category structure differently. Neither stronger language models nor richer captions close this gap. The picture that emerges is less a shared Platonic reality than a set of overlapping Umwelten~\citep{uexkull1934streifzuge}. Each modality agrees on the broad semantic categories yet perceives and arranges them in its own way. The evidence shows shared coarse semantics, not shared representations.

In this paper, we make the following observations:

\vspace{-1em}
\mysummary{
\begin{itemize}[leftmargin=0em, itemsep=0.35em, topsep=0.15em, parsep=0pt, after=\vspace{-0.2em}]
    \renewcommand\labelitemi{}
    \item\colorbox[RGB]{214,228,245}{\parbox{\dimexpr\linewidth-2\fboxsep}{\textbf{Mutual $k$NN alignment for fixed small $k$ degrades with scale:} Increasing the gallery from $1024$ to millions of samples causes a drop in cross-modal mutual nearest-neighbor agreement for fixed small neighborhood sizes, while agreement at $k{=}n/100$ remains stable (\cref{sec:alignment_across_scales}).}}
    \item\colorbox[RGB]{214,228,245}{\parbox{\dimexpr\linewidth-2\fboxsep}{\textbf{Previously reported trends do not continue under the original setup:} The claim that stronger language models align better with vision weakens for more recent models and benchmarks (\cref{sec:trend-check}).}}
    \item\colorbox[RGB]{214,228,245}{\parbox{\dimexpr\linewidth-2\fboxsep}{\textbf{Many-to-many correspondence reduces mutual $k$NN alignment:} Allowing multiple valid correspondences per sample leads to decay in mutual $k$NN agreement for fixed $k$ (\cref{sec:no-bijecton-experriments}).}}
    \item\colorbox[RGB]{214,228,245}{\parbox{\dimexpr\linewidth-2\fboxsep}{\textbf{Modalities share coarse structure, not global organization:} Cross-modal agreement on coarse structure approaches the same-modality ceiling. Global organization saturates well below it, and unimodal agreement stays high throughout (\cref{sec:diagnosis}).
    }}
\end{itemize}
}
\vspace{0.3em}

\section{Related Work}\label{sec:related}

\textbf{One Platonic Ideal vs many {\em Umwelten}.}
In his ``Theory of Forms'', \cite{plato_forms} argued that every physical object we perceive is a flawed imitation (a shadow) of some eternal, abstract ``ideal'' form, and only by escaping from the tyranny of our physical senses (leaving the cave of shadows), we can achieve true understanding.  But in the 20th century, this argument for a single, unified  Platonic Ideal representation has been repeatedly undercut by biologists, psychologists, and philosophers. \cite{uexkull1934streifzuge} argued that every organism inhabits its own perceptual environment, or {\em Umwelt}: a tick lives in a world of thermal gradients, a bat in a world of echoes. The different Umwelten might have only limited overlap with each other.
\footnote{For a tour of von Uexküll's ideas, see Koenderink's delightful book~\citep{koenderink2019sentience}.}.
\cite{gibson1979ecological} pushed this further, proposing that perception is shaped by what an organism can {\em do} in its environment, not by an observer-independent reality.
Wittgenstein, thinking about language, arrived at a strikingly similar conclusion.  He famously argued: ``If a lion could talk, we wouldn't be able to understand it''~\citep{Wittgenstein1953-WITPI-4}, meaning that the lion's world (goals, instincts, perceived reality) is so utterly different from our own, that even if it spoke English, we would not comprehend the meaning.
\footnote{For a great treatment of Wittgenstein's argument in popular culture, see the episode {\em Darmok} of the American TV series {\em Star Trek: The Next Generation}.}
Building on Wittgenstein,  \cite{Rosch1978} developed her Prototype Theory of Categorization, arguing against a single Platonic ideal as a representation of object categories, proposing a data-driven clustering-based model instead.

\myparagraph{Representational alignment.} The question of representational similarity has been studied extensively in the neurosciences~\citep{edelman1998representation,haxby2001distributed,kriegeskorte2008representational}. In machine learning, the parallel question of whether independently trained networks learn similar internal structure has received growing attention.
\cite{lenc2015understandingimagerepresentationsmeasuring} investigated the equivalence of representations from different trained models and found that early convolutional layers are more interchangeable than later ones. This task, also referred to as ``model stitching'', was later revisited by \cite{bansal2021revisitingmodelstitchingcompare}. Related to this, \cite{li2016convergentlearningdifferentneural} proposed methods to align neurons across independently trained networks. More recently, \cite{dravid2023rosetta} introduced ``Rosetta Neurons'', showing that different vision models share common units corresponding to similar visual concepts across architectures, tasks, and training data.
\cite{tasker2026universalnormalembedding} propose the Universal Normal Embedding hypothesis, where encoder and generative latents are interpreted as noisy views of an approximately Gaussian shared space.

Furthermore, alignment has been linked with shared model capabilities measured by task performances~\citep{balestriero2018spline,jiang2024tracing,huh2024prh,morcos2018insightsrepresentationalsimilarityneural,bansal2021revisitingmodelstitchingcompare}. To directly quantify representational similarities, several metrics have been used to measure correlations between features~\citep{morcos2018insightsrepresentationalsimilarityneural,hotelling1992relations}. \cite{kornblith2019similarity} introduced Centered Kernel Alignment (CKA) as a robust measure invariant to orthogonal transformations and isotropic scaling. \cite{huh2024prh} found the CKA metric to reveal only a ``weak trend of alignment between models'' and proposed the use of the mutual $k$NN metric that measures the overlap of two sets of neighborhoods of size $k$.

\myparagraph{Multi-modal alignment.} Early efforts to connect images and text utilized human annotations~\citep{von2007improving}. The curation of large-scale paired image-caption datasets, such as MS-COCO~\citep{lin2014microsoft} and Visual Genome~\citep{krishna2017visual}, facilitated the systematic study of cross-modal correspondence and models. The CLIP model~\citep{radford2021learningtransferablevisualmodels} formed a turning point by demonstrating that contrastive learning on web-scale image-text pairs could produce shared embedding spaces.

Since then, a growing body of work has investigated whether such alignment arises even without explicit joint training. \cite{merullo2023linearlymappingimagetext} showed that a simple learned linear transformation could map between frozen vision encoders and LLMs. \cite{moschella2023relativerepresentationsenablezeroshot} use similarities to an anchor set.
\cite{maniparambil2024visionlanguageencodersrepresent} demonstrated that even unaligned unimodal encoders possess high semantic similarity. \cite{li2024vision} study alignment at the level of shared categories. Complementary to that, \cite{lu2026indrarepresentationhypothesismultimodal} define each sample by its angular distances to others, improving training-free alignment across vision, language, and audio.

Fully unsupervised approaches include blind vision-language matching~\citep{schnaus2025itsblindmatchvisionlanguage} and unpaired embedding translation via cycle-consistency~\citep{jha2025harnessinguniversalgeometryembeddings,zhu2017unpairedimagetoimagetranslationusing}.
Finally, \cite{gupta2026canonicalizingmultimodalcontrastiverepresentation} show that an orthogonal map can map between independently trained multi-modal contrastive models. These results are often seen as evidence for representational convergence, also supported by \cite{lu2025representationpotentialsfoundationmodels} in their survey of multi-modal alignment.
However, those results are obtained in restricted settings (e.g.\ \cite{schnaus2025itsblindmatchvisionlanguage} experiments on CIFAR-100 and ImageNet-100) and do not scale to real-world multi-modal data. Our work examines whether alignment survives beyond these constraints, showing that cross-modal alignment reflects coarse agreement rather than shared global structure.

\myparagraph{Limits and measurement of emergent cross-modal structure.}
Several analyses show that alignment between independently trained unimodal encoders depends strongly on data, architecture, and evaluation protocol. \cite{tjandrasuwita2025understandingemergencemultimodalrepresentation} find that alignment varies with modality similarity and the balance of shared versus unique information, while \cite{hadgi2025escapingplatoscavealignment} report weaker alignment for ``pure'' 3D encoders without careful subspace selection. \cite{zhu2026dynamicreflectionsprobingvideo} further show that video–text alignment depends on temporal information and text availability.

\cite{groger2026revisitingplatonicrepresentationhypothesis} show that measures such as CKA are sensitive to network scale and can be altered via null calibration, largely removing evidence of global convergence while leaving local neighborhood similarity (e.g.\ mutual $k$NN) stable, though their evaluation remains in small-scale regimes and on the original terse captions, which cap measurable alignment (\cref{sec:supp_ceiling_bands_additional}). Our granularity analysis in \cref{sec:alignment_across_scales} suggests the calibration result and the local-persistence result are not two separate findings: at
$n{\approx}1$k, mutual $k$NN with $k{=}10$ probes coarse categorical structure, so the ``local''
alignment that survives their calibration and the ``global'' alignment
that vanishes are different views of the same coarse overlap. Beyond similarity metrics, \cite{pmlr-v267-smith25a} and \cite{kumar2025questioning} show that functional agreement can persist even when internal representations are misaligned, suggesting that behavioral compatibility does not imply shared structure.
These caveats echo arguments that text-only learning may be insufficient to recover perceptual structure~\citep{bender-koller-2020-climbing,lecun2022path}, and motivate multi-modal foundation models that integrate perception and language~\citep{huang2023languageneedaligningperception,baiQwen3VLTechnicalReport2025,meta_llama4_2025,vteam2026glm45vglm41vthinkingversatilemultimodal}.

\section{How much do representations align?}\label{sec:experiments}
In this section, we take a close look at the experimental evidence underpinning the Platonic Representation Hypothesis~\citep{huh2024prh}. The experiments in \cite{huh2024prh} rest on two foundations that warrant scrutiny: the use of mutual $k$NN alignment on a small evaluation set of only 1024 samples from the Wikipedia Image-Text (WIT) dataset~\citep{srinivasan2021wit} (WIT-1024) with
effectively bijective (one-to-one) image-text correspondences. This assumption is implicit in the WIT-1024 setup rather than stated explicitly in~\cite{huh2024prh}.
Typically, these choices are not acknowledged when the hypothesis is cited \citep{marcosmanchon2026sharedrepresentationsbrainsmodels,ruan2024ndpdistributionpredictionbroad,liu2025luminamgptilluminateflexiblephotorealistic,chai2025auroracapefficientperformantvideo,dar2026minivec2vecscalinguniversalgeometry}. The hypothesis is usually invoked in its broad, appealing form rather than in the narrower terms under which experimental support was provided.

\begin{figure}[t]
    \centering
    \begin{subfigure}[t]{0.39\linewidth}
        \centering
        \includegraphics[width=\linewidth]{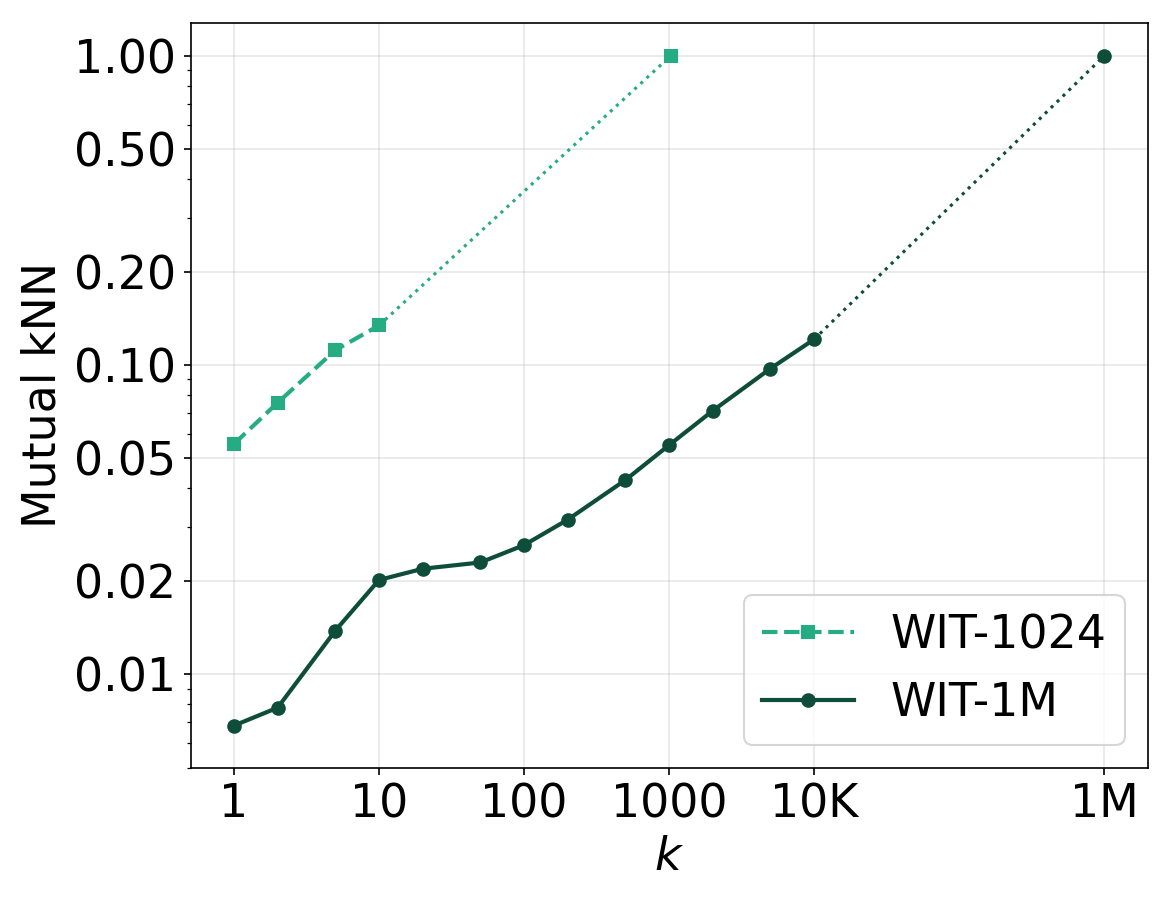}
        \vfill
        \caption{Effect of neighborhood sizes $k$ in mutual $k$NN (both axes log-scaled). Trivially, mutual $k$NN converges to 1 as $k$ approaches the full gallery. \cite{huh2024prh} utilized mutual $k$NN with $k=10$.}
        \label{fig:mutual_knn_metric}
    \end{subfigure}
    \hfill
    \begin{subfigure}[t]{0.58\linewidth}
        \centering
        \includegraphics[width=\linewidth]{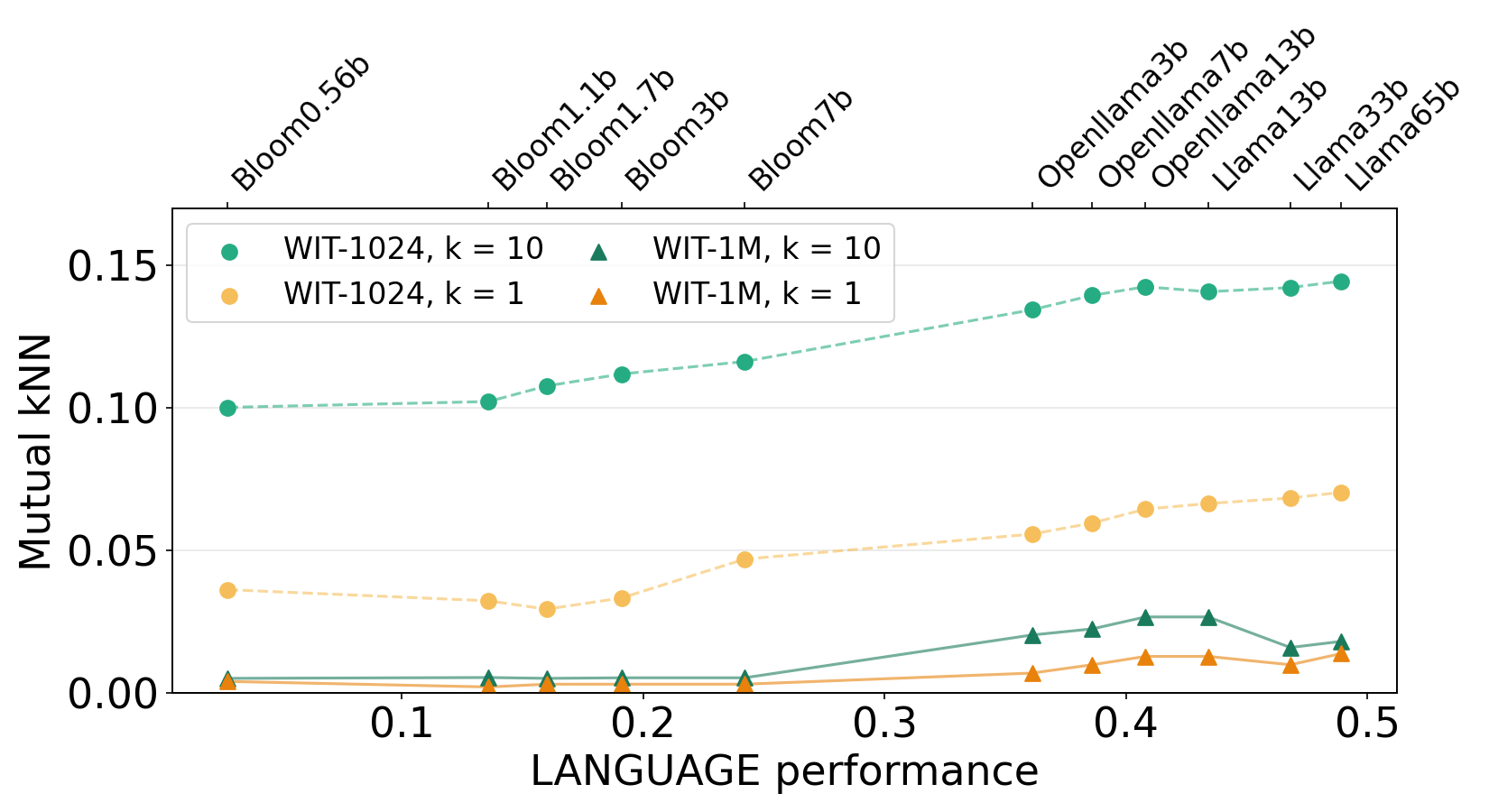}
        \vfill
        \caption{Alignment between DINOv2 and different LLMs, measured on WIT-1024 and WIT-1M. \cite{huh2024prh} observed that alignment (mutual $k$NN) increases with language performance (measured as $1-\text{bitsperbyte}$ on the 1024-sample set), but we find that this trend weakens with larger gallery size.}
        \label{fig:second_plot}
    \end{subfigure}
    \vspace{-0.5em}
    \caption{Mutual $k$NN text-image feature alignment when scaling from  WIT-1024 to WIT-1M. (a) shows the dependence on neighborhood size $k$, while (b) examines alignment for different LLMs. \textit{The observation from~\cite{huh2024prh}, that more capable language models align better with vision at fixed small $k$ is strongly weakened at WIT-1M scale.}}
    \label{fig:combined}
     \vspace{-1.3em}
\end{figure}

\subsection{Experimental setup}
\noindent \textbf{Mutual $k$NN metric.}
To measure alignment between representations from different models, we follow  \cite{huh2024prh} in using the mutual $k$-nearest-neighbor metric (illustrated in \cref{fig:teaser}).
Given a shared gallery set of $n$ datapoints (referred to as mini-batch sampled from the data distribution in \cite{huh2024prh}) encoded into feature vectors $\mathbf{a_i} \in \mathbb{R}^{d_1}$ and $\mathbf{b_i} \in \mathbb{R}^{d_2}$ by two models with $i \in \{1,\cdots,n\}$, we first L2-normalize each representation. The features are then clamped to the 0.95-quantile of the norm of the activations~\citep{hastings1947low}.
We then retrieve the $k$ nearest neighbors of every query point independently for each model (e.g.\ image and text query for vision and language encoders):
$$\mathcal{N}^\mathbf{a}_k(i) = \operatorname{argtopk}_{j \neq i} \mathbf{a}_i^\top \mathbf{a}_j, \qquad \mathcal{N}^\mathbf{b}_k(i) = \operatorname{argtopk}_{j \neq i} \mathbf{b}_i^\top \mathbf{b}_j.$$
The per-sample score is the number of overlapping samples normalized by $k$:
$$s_i = \frac{|\mathcal{N}^\mathbf{a}_k(i) \cap \mathcal{N}^\mathbf{b}_k(i)|}{k},$$
and the overall mutual-$k$NN score is the mean over all samples. A score of 1 means that every point's $k$ nearest neighbors are identical in both spaces, and a score of 0 means that the $k$ nearest neighbors do not overlap. In the sparse gallery in \cref{fig:teaser}, the query (blue) retrieves the same neighbor in both image and text spaces, giving a mutual $k$NN score of 1 for $k{=}1$.
A score of $\frac{k}{n}$ suggests chance-level expected overlap for independent random retrieval, which decreases for growing $n$.
Chance-level overlap is
${\approx}9.8{\times}10^{-3}$ for $k{=}10$ at $n{=}1024$, but only
$10^{-5}$ at $n{=}10^6$.
Correcting for chance,
$\tilde{s} = (s - \frac{k}{n})/(1 - \frac{k}{n})$, changes scores by
at most ${\approx}0.009$ at $n{=}1024$ and by less than $10^{-4}$ for
$n \geq 10^5$, and does not affect any of the trends we report. Enrichment over chance (${\sim}800\times$ at $n{=}15$M) is not the
quantity at issue. Identical representations agree on neighborhoods at
every granularity, so high enrichment with ${\sim}1\%$ absolute overlap
signals partial coarse agreement, not convergence.
Note that throughout this section, we report raw mutual $k$NN (as in \cite{huh2024prh}) using WIT-1024 as the set of queries. We report results for different galleries: WIT-1024, WIT-1M, and LAION-15M.

\begin{figure*}[t]
    \includegraphics[width=1\linewidth]{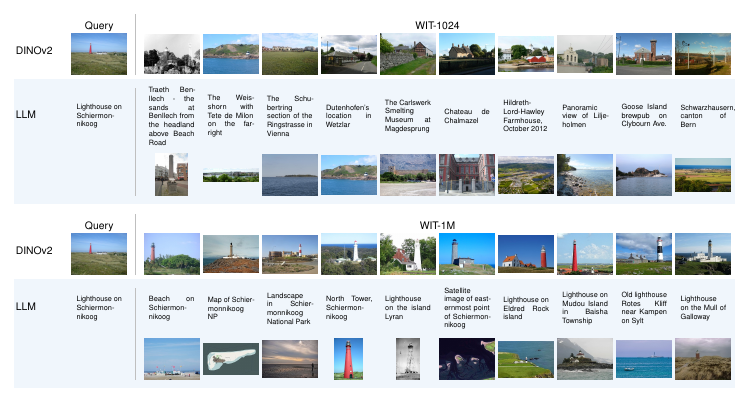}
\vspace{-1.8em}
\caption{
\textbf{Nearest-neighbor quality depends on data density.}
We show $10$ within-modality nearest neighbors for image (DINOv2) and text (LLM) embeddings on a sparse WIT-1024 gallery (top) and a denser WIT-1M gallery (bottom). For text queries, retrieved captions and their corresponding reference images are shown. At smaller scale, nearest neighbors are less semantically precise. Nearest-neighbor structure becomes more semantically refined as gallery density increases.
}
    \label{fig:sample376-k10-galleries}
     \vspace{-1.3em}
\end{figure*}

\myparagraph{Implementation details.}
For most experiments, we use DINOv2-base~\citep{oquab2024dinov2learningrobustvisual} as the vision encoder. We refer to this model as DINOv2 in the following. Our primary language model is OpenLlama3b~\citep{touvron2023llama,openlm2023openllama}, abbreviated as OpenLlama.
For each image and text sample, we extract the representations from all layers of their respective encoders and follow the experimental protocol from \citet{huh2024prh}. Additional model pairs and the full model list used in \cref{sec:experiments} are provided in \cref{sec:additional_models} and \cref{sec:all-llm-models}.
We use Faiss~\citep{douze2024faiss} for nearest neighbor computation at scale, specifically \texttt{IndexFlatL2}, which is equivalent to using cosine similarity on normalized vectors.

\subsection{Probing alignment as measured by mutual $k$NN}\label{sec:probing_alignment}
Here, we analyze how alignment as measured by mutual $k$NN behaves at different levels of granularity ($k{=}1$, $k{=}10$, $k{=}\frac{n}{100}$, and on a denser gallery (million(s of) instead of 1024 samples).
We then turn to the bijective assumption and examine what happens when it is relaxed. We further test whether these patterns extend beyond text-image to the text-audio and text-video settings. Finally, we perform a trend check to ask whether the predictions from~\cite{huh2024prh} have held up as models have improved.

\myparagraph{Sensitivity to $k$ in mutual $k$NN.} \cite{huh2024prh} reported mutual $k$NN alignment for $k{=}10$. As shown in \cref{fig:mutual_knn_metric}, the metric trivially converges to 1 as $k$ approaches the full gallery size $n$, since both neighbor sets then contain all samples. In our analyses at larger gallery scales, we perform deduplication to prevent near-duplicate samples from trivially inflating neighborhood overlap (see \cref{sec:wit_laion} for details).

\subsubsection{Alignment across dataset scales}\label{sec:alignment_across_scales}

\textbf{Nearest neighbors in sparse gallery.}
We now turn to the data and ask whether the 1024-sample gallery used in \cite{huh2024prh} is too sparse to capture more than coarse structural agreement.
\begin{wraptable}{r}{0.38\textwidth}
 \vspace{-0.5em}
  \centering
  \small
  \caption{As the gallery grows, nearest neighbors get closer to the query set in both DINOv2 and OpenLlama embedding spaces, facilitating a more fine-grained analysis of cross-modal alignment.}\label{tab:wit1024_structure}
  \vspace{-0.2em}
  \resizebox{0.36\textwidth}{!}{%
  \setlength{\tabcolsep}{6pt}
  \begin{tabular}{llcc}
    \toprule
    Gallery & Model & $k{=}1$ & $k{=}10$ \\
    \midrule
    WIT-1024 & DINOv2  & 0.799 & 0.717 \\
    WIT-1024 & OpenLlama & 0.502 & 0.400 \\
    \midrule
    WIT-1M & DINOv2   & 0.906 & 0.888 \\
    WIT-1M & OpenLlama    & 0.757 & 0.701 \\
    \bottomrule
    \bottomrule
  \end{tabular}%
  }
  \vspace{-1.6em}
\end{wraptable}
As shown in \cref{tab:wit1024_structure}, the mean cosine similarity between queries and nearest neighbors in terms of both image (DINOv2) and text features (OpenLlama) is significantly lower for the WIT-1024 gallery compared to WIT-1M (e.g.\ 0.799 compared to 0.906 for DINOv2 at $k{=}1$). For both, we use WIT-1024 as the query set.

We visualize the 10 nearest neighbors for image and text features on WIT-1024 and WIT-1M in \cref{fig:sample376-k10-galleries}.
At low density, semantically unrelated samples can appear as nearest neighbors as there is nothing closer available, meaning that measured mutual $k$NN agreement can reflect the shared lack of alternatives. To get more meaningful insights, we scale the density of the retrieval gallery in the following subsection.

\begin{figure}[t]
  \centering
  \includegraphics[width=0.49\textwidth]{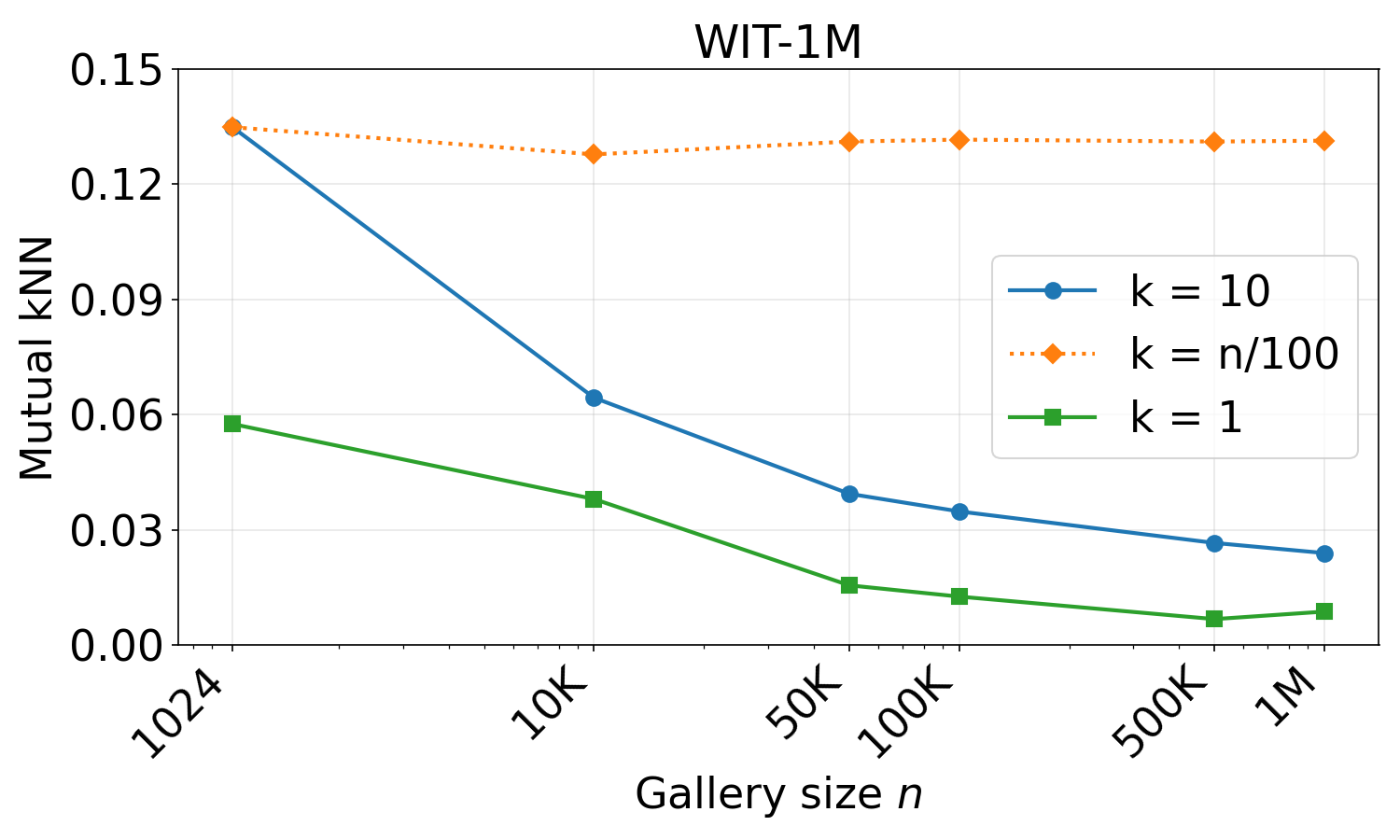}
  \hfill
  \includegraphics[width=0.49\textwidth]{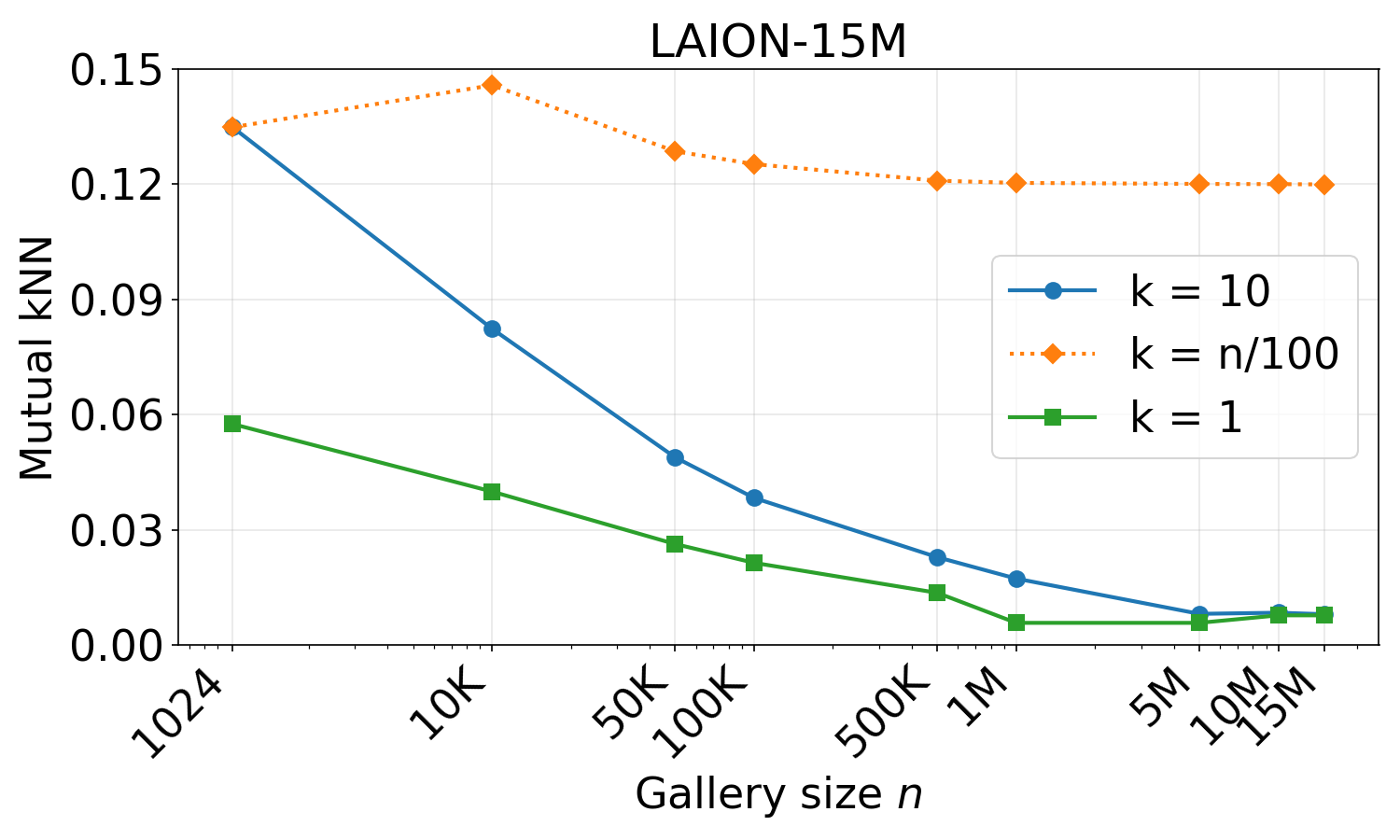}
  \vspace{-.1in}
  \caption{Scaling the gallery size to 1M (WIT) and 15M (LAION) shows a large drop in mutual $k$NN alignment for $k{=}1$ and $k{=}10$ for DINOv2 and OpenLlama features. Because $\frac{k}{n}$ shrinks as $n$ grows, we probe progressively finer granularity. Mutual $k$NN alignment for $k = \frac{n}{100}$ is stable.}

  \label{fig:both}
  \vspace{-1.3em}
\end{figure}

\myparagraph{Densification by scaling the gallery size.}
Having established that WIT-1024 captures mainly coarse structure, we test whether alignment persists as we densify the gallery up to 1M (English WIT~\citep{srinivasan2021wit}) and 15M samples (LAION-400M~\citep{schuhmann2021laion}). Best layer pairs are chosen following~\cite{huh2024prh} (see \cref{sec:supp_layer_selection}).

As shown in \cref{fig:both}, alignment scores decrease as gallery size $n$ grows for fixed $k$. On WIT-1024, mutual $k$NN scores are 0.135 ($k=10$) and 0.058 ($k=1$), but drop to 0.008 and 0.001 on LAION-15M. As the neighborhood considered by the metric shrinks by four orders of magnitude, these fixed-$k$ sweeps probe increasingly fine granularity.

Crucially, this decay is not a metric artifact. Under identical 1M and 15M densification, same-modality baselines (e.g.\ OpenLlama-3B vs.\ 13B, or DINOv2 vs.\ CLIP visual) keep strong alignment (\cref{fig:uni_dino_alignment}, $R_{k\text{NN}} \in [0.35, 0.62]$), whereas cross-modal pairs collapse to near zero. This is consistent with cross-modal misalignment reflecting genuine organizational divergence. At increasing data density, candidates get semantically closer, placing greater demand on models to agree on fine-grained distinctions.

Conversely, alignment at $k=\frac{n}{100}$ remains stable across scales.
This establishes broad category-level correspondence, expected from
overlapping internet data, but convergence to the \emph{same}
representation additionally requires agreement on fine-grained structure,
which is exactly what decays.
While \cite{groger2026revisitingplatonicrepresentationhypothesis} reported local neighborhood persistence, their effective neighborhood size ($k \approx n/100$) falls entirely within this coarse regime. When probing genuinely fine-grained neighborhoods ($k \ll n/100$), local agreement collapses.

We next re-examine the trend reported by \cite{huh2024prh}, where stronger LLMs exhibit higher alignment with visual features (DINOv2). On WIT-1024 ($k=10$), this trend is clear (\cref{fig:second_plot}). However, at the WIT-1M scale, the gap between LLMs narrows and the trend disappears for fixed small $k$.
The trend reappears on WIT-1M when restoring the proportional coarse ratio $k = \frac{n}{100}$ (\cref{fig:llm_trend_kpct}). As $k=10$ on $n=1024$ represents a coarse metric ($k \approx n/100$), the original evidence in \cite{huh2024prh} tracks coarse overlap rather than fine-grained representational convergence.

\begin{figure}[t]
  \centering
 \includegraphics[width=1\textwidth, trim=0 85 0 0, clip]{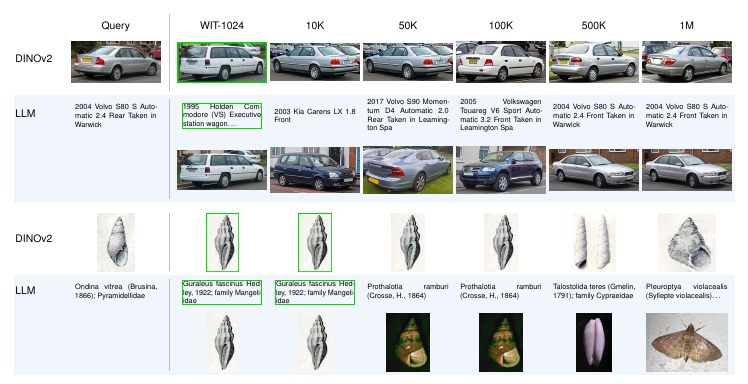}
  \vspace{-2em}
    \caption{Nearest-neighbor ($k{=}1$) examples with DINOv2 and OpenLlama across gallery scales on WIT-1M. Captions are shown with corresponding images. Mutual $k$NN matches across modalities are framed in green. \textit{At larger scales, each model finds closer but most often different matches.}}
    \label{fig:tum_transfer_examples}
    \vspace{-1.3em}
\end{figure}

The nearest-neighbor examples in \cref{fig:tum_transfer_examples} further illustrate this. Matches at small gallery sizes often disappear as candidates are added, with each modality finding better but divergent neighbors. The matches at 1M and 15M scale are mostly near-duplicates that our deduplication missed (e.g.\ a crop shifted by a few pixels). We show additional visualizations in \cref{fig:wit_transfer_examples,fig:wit_transfer_examples_2,fig:laion_transfer_examples_2,fig:laion_transfer_examples_3}.

\finding{Mutual $k$NN alignment at fixed small $k$ decays for denser galleries for cross-modal pairs, while same-modality pairs remain significantly more stable under identical densification.
The reported trend that stronger language models align better with vision is a property of the coarse regime. It persists at $k=\frac{n}{100}$ but vanishes at fixed $k=10$ on WIT-1M.}

\subsubsection{Text-audio and text-video alignment}\label{sec:text_audio_video}

\begin{figure}[t]
    \centering
    \begin{subfigure}[b]{0.25\linewidth}
        \centering
        \includegraphics[width=\linewidth,trim=7 7 5 5, clip]{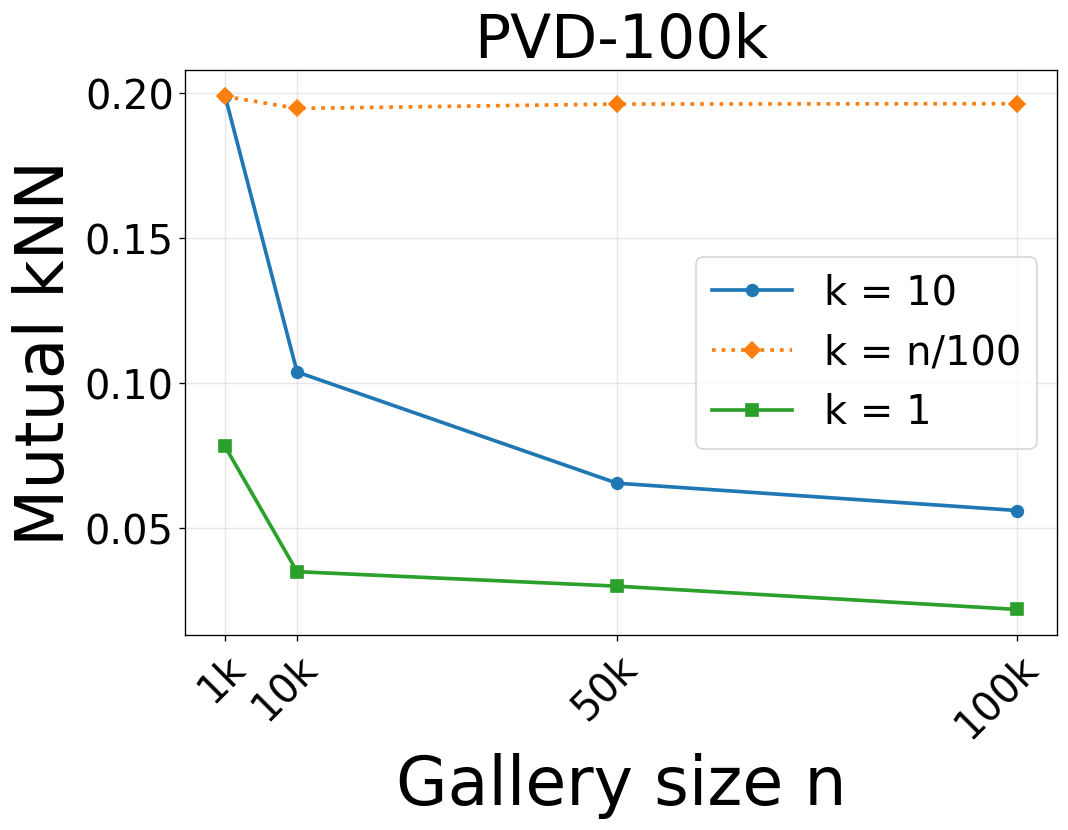}
        \caption{Text-video alignment.}\label{fig:video_scaling}
    \end{subfigure}
    \hfill
    \begin{subfigure}[b]{0.25\linewidth}
        \centering
       \includegraphics[width=\linewidth,trim=7 7 5 5, clip]{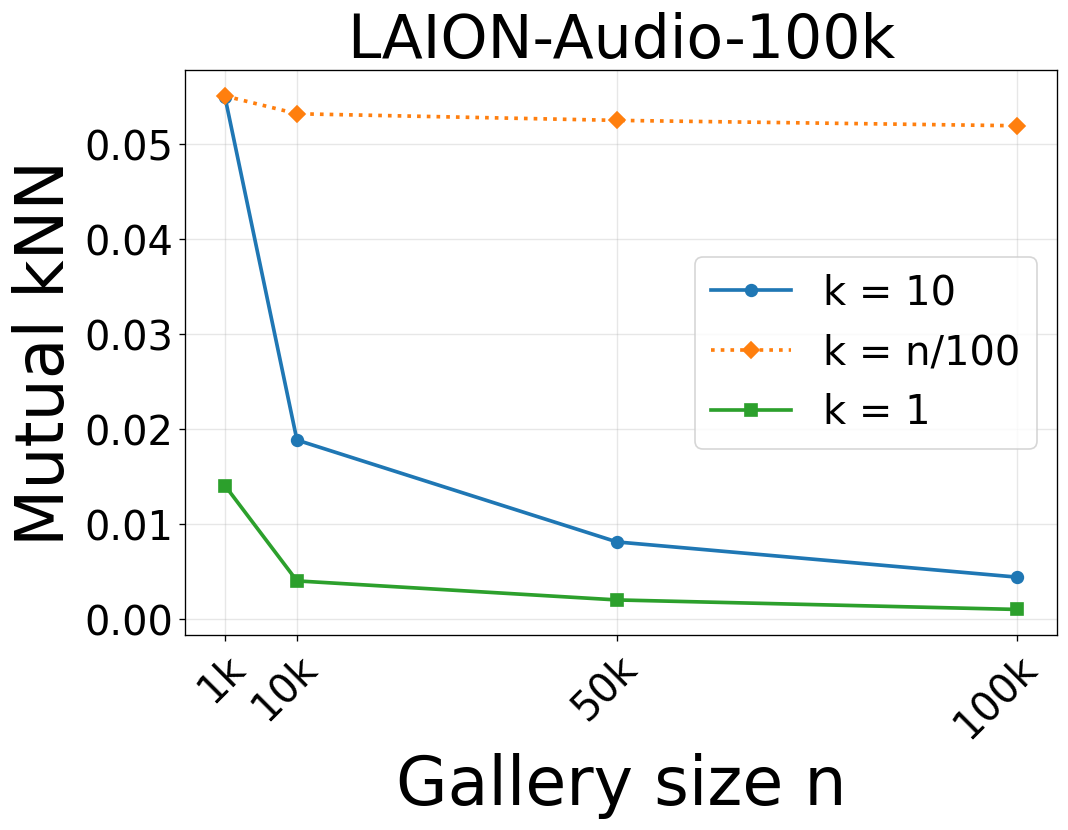}

        \caption{Text-audio alignment.}\label{fig:audio_scaling}
    \end{subfigure}
    \hfill
    \begin{subfigure}[b]{0.43\linewidth}
        \centering
        \includegraphics[width=\linewidth]{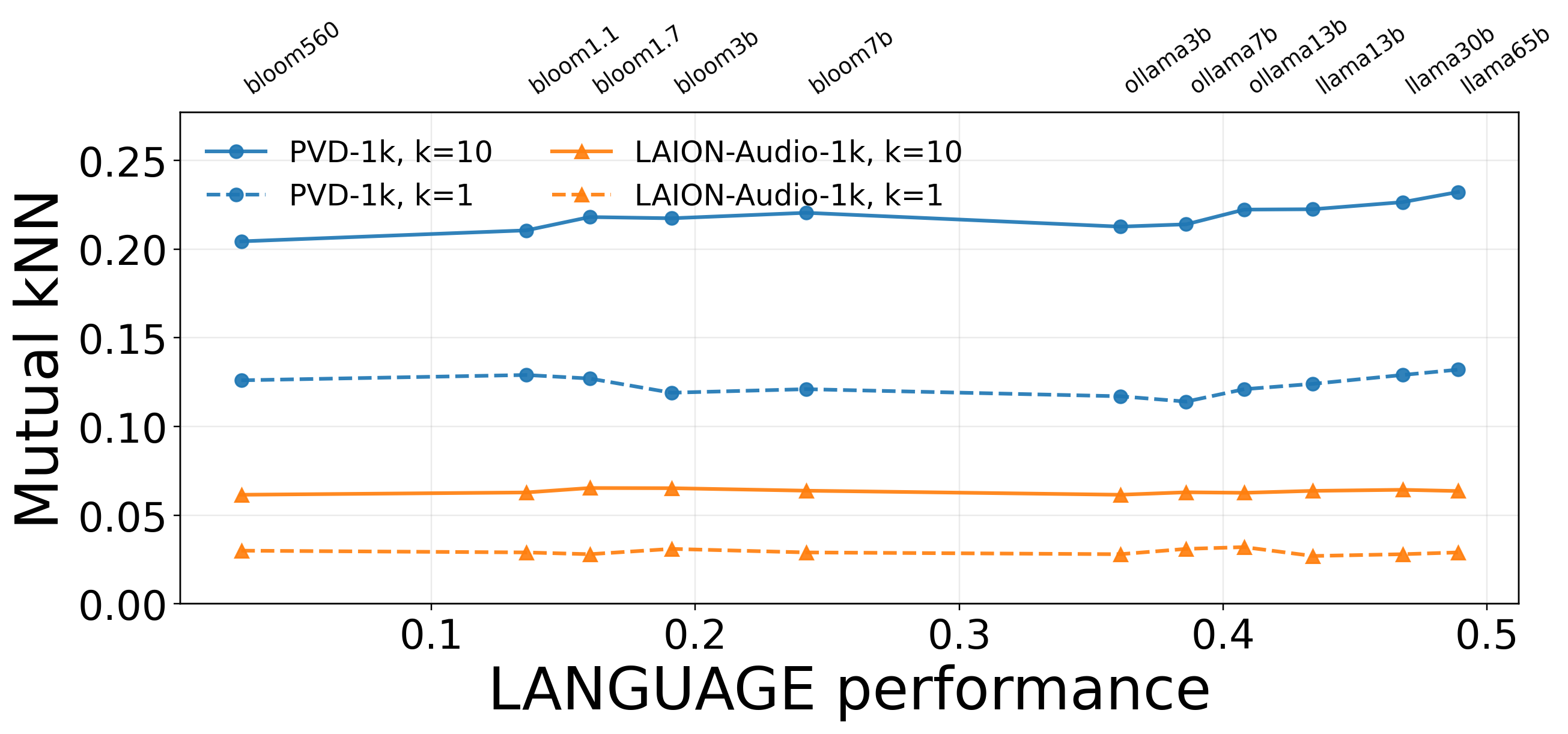}
        \caption{Alignment vs.\ LLM performance.}\label{fig:audio_video_llm_trend}
    \end{subfigure}
     \vspace{-0.5em}
    \caption{Cross-modal alignment beyond text-image. (a, b) Mutual $k$NN text-video (PVD-100k~\citep{bolya2025perception,wang2019vatex}) and text-audio alignment (LAION-Audio-100k~\citep{wu2023large}) scales similarly to the text-image setting. (c) Alignment vs.\ LLM performance at 1k: weak trend for video, flat for audio.}
    \label{fig:audio_video_scaling}
    \vspace{-1.3em}
\end{figure}

To check that our findings are not specific to text-image, we repeat the scaling analysis for a video encoder (VideoMAE-v2~\citep{wang2023videomaev2scalingvideo}) on the PVD-100k dataset, a 100k subset of PVD~\citep{bolya2025perception,wang2019vatex}, and an audio encoder (Dasheng from \cite{dinkel2024dasheng}) on a 100k subset of LAION-Audio-630k~\citep{wu2023large}.

\cref{fig:audio_video_scaling} shows that the same patterns observed for text-image alignment hold across these modality pairs. Here, the LLM used is Gemma-2-9b~\citep{gemma_2_2024}. Alignment at fixed small $k$ drops with gallery size, while alignment at $k{=}n/100$ remains stable across scales. The trend that stronger LLMs align better with the non-text modality (\cref{fig:audio_video_llm_trend}) is weak. We observe a slight positive slope for text-video and an essentially flat evolution for text-audio. This measurement sits in the coarse regime, where the text-image capability trend does persist.
The flat audio trend may reflect the terseness of the audio captions, consistent with our text--image finding that terse text caps cross-modal alignment regardless of which language model reads it (\cref{fig:ceiling_bands}).
We note the narrower scope of this experiment relative to our text-image analysis in terms of models and gallery sizes.

\begin{figure}[t]
    \centering
    \begin{minipage}[t]{0.48\textwidth}
      \centering
      \includegraphics[width=\linewidth]{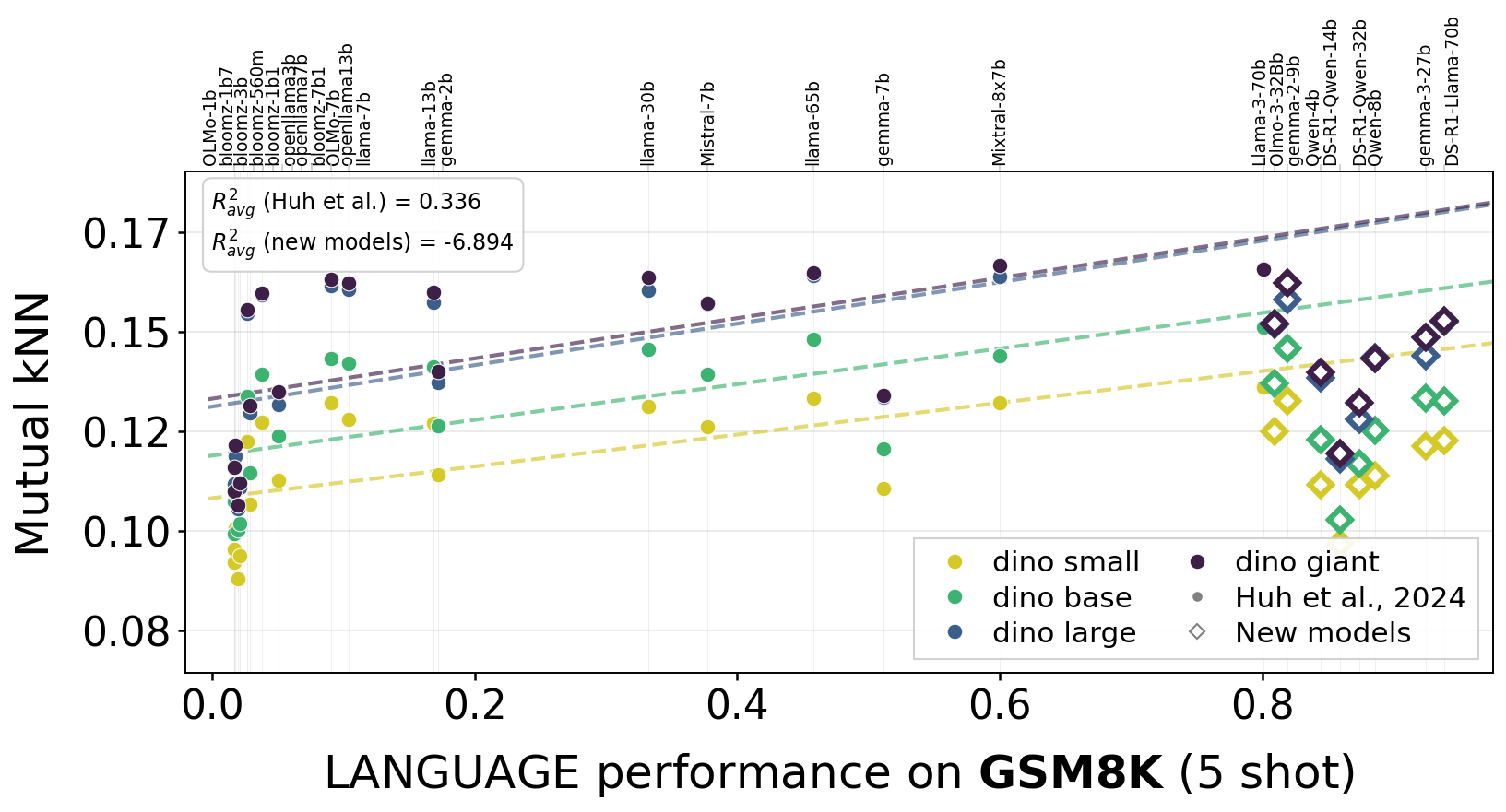}
    \end{minipage}\hspace{0.02\textwidth}%
    \begin{minipage}[t]{0.48\textwidth}
      \centering
      \includegraphics[width=\linewidth]{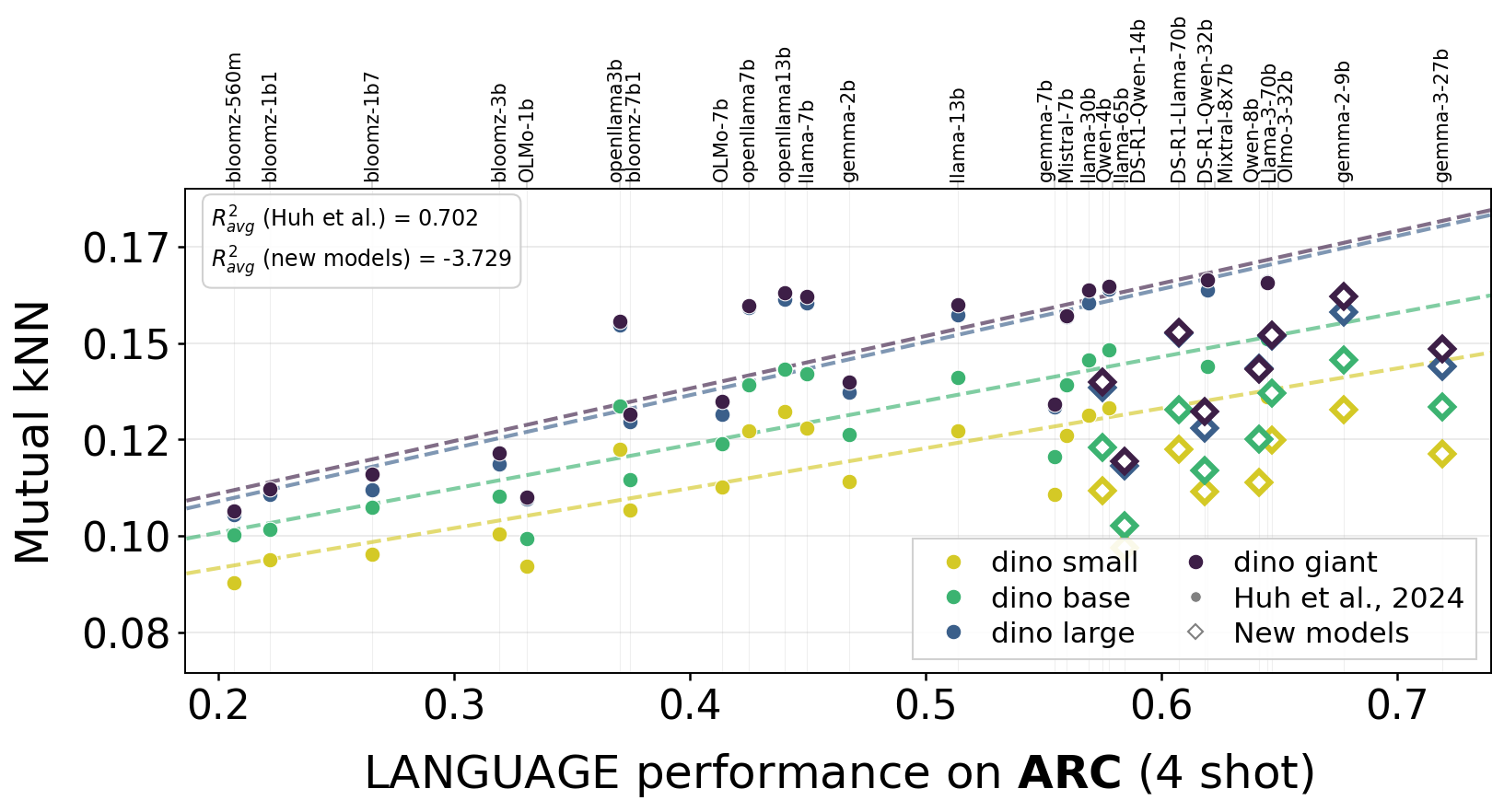}
    \end{minipage}

    \vspace{0.8em}
    \begin{minipage}[t]{0.48\textwidth}
      \centering
      \includegraphics[width=\linewidth]{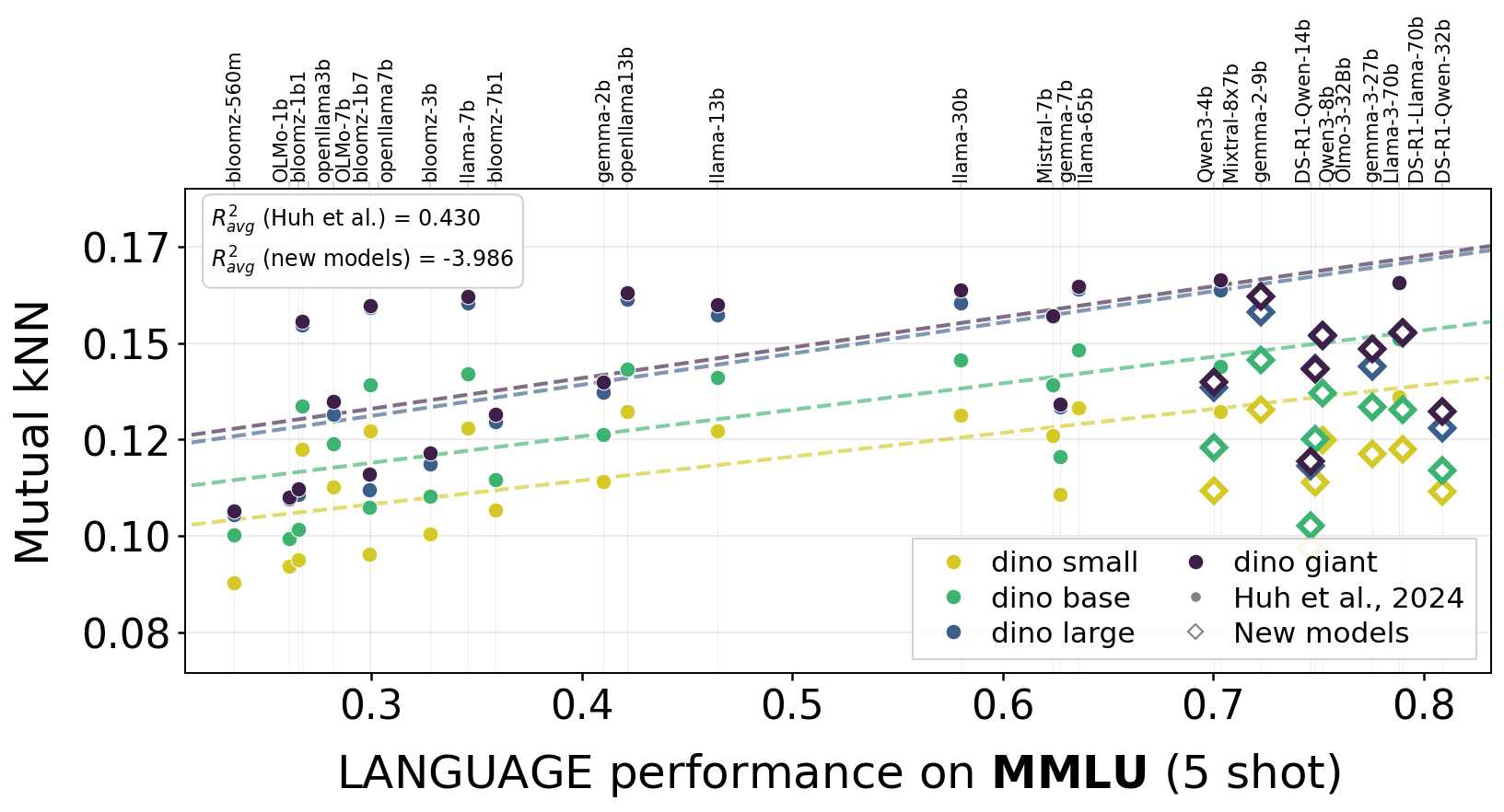}
    \end{minipage}\hspace{0.02\textwidth}%
    \begin{minipage}[t]{0.48\textwidth}
      \centering
      \includegraphics[width=\linewidth]{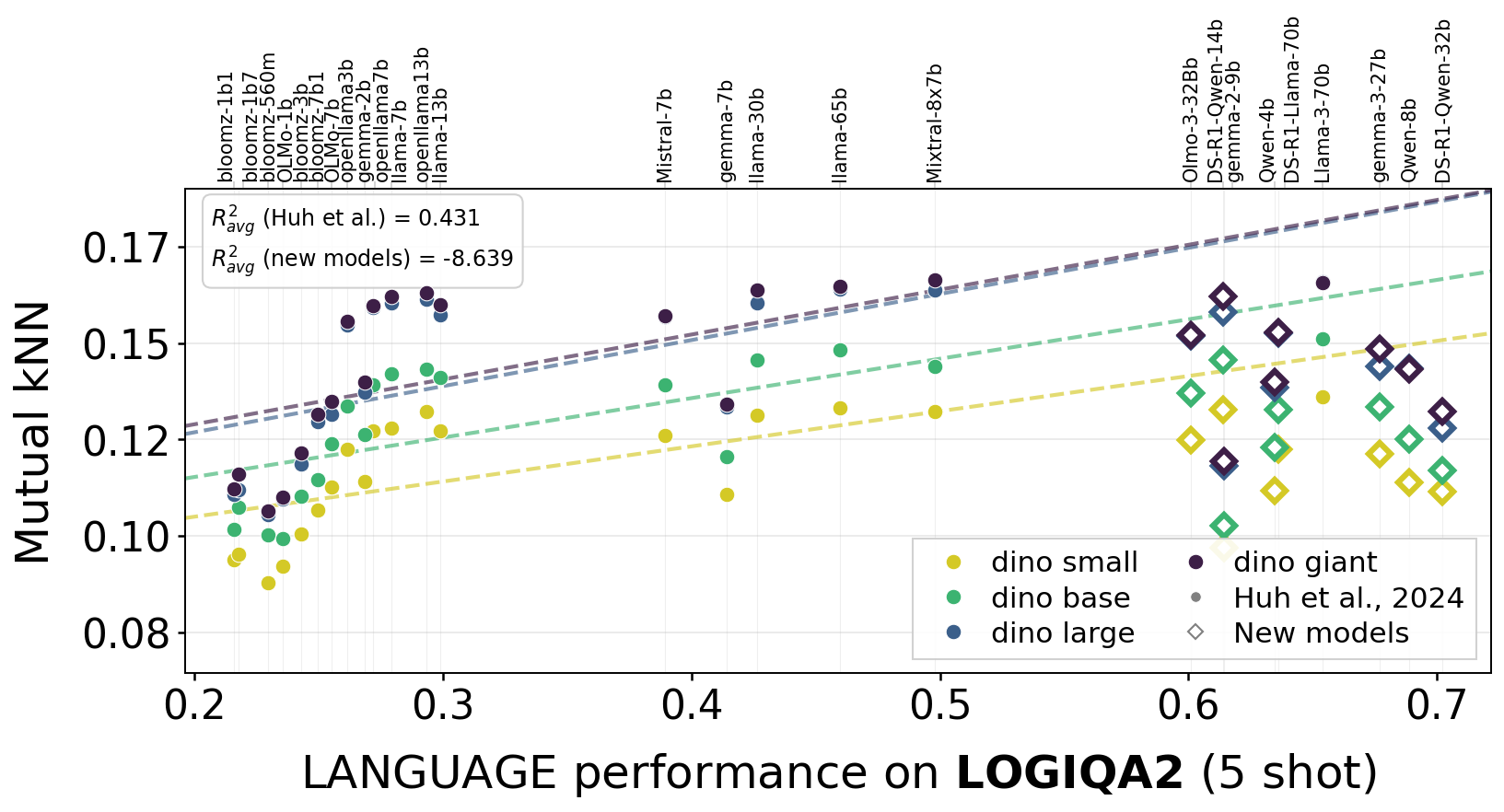}
    \end{minipage}
\vspace{-.05in}
    \caption{
    Testing whether the alignment-LLM performance trend from~\cite{huh2024prh} (dashed lines), tested on WIT-1024, holds for recent LLMs on ARC, GSM8K, MMLU, and LogiQA2.
    Recent LLMs (diamonds) do not follow the trend extrapolated from the models in \citet{huh2024prh}, alignment with DINOv2 saturates.
    }
    \label{fig:trends_combined}
    \vspace{-1.3em}
\end{figure}

\subsubsection{Trend check: Are the predictions from \cite{huh2024prh} holding up so far?}\label{sec:trend-check}
\cite{huh2024prh} predict that as LLMs become stronger, their representations align more with vision representations. This claim is evaluated using three proxies for language performance: HellaSwag~\citep{zellers2019hellaswag}, GSM8K~\citep{cobbe2021gsm8k}, and $(1-\text{bitsperbyte})$ (similar to \cite{huh2024prh}).

In this section, we revisit this trend analysis for an extended set of models and benchmarks. We follow the original experimental setup from \cite{huh2024prh} to evaluate 55 LLMs, spanning from BLOOMZ~\citep{muennighoff2023crosslingualgeneralizationmultitaskfinetuning} to recently released models. We use the ARC Challenge~\citep{clark2018think}, MMLU~\citep{hendrycks2021mmlu}, and LogiQA2~\citep{liu2023logiqa2} benchmarks. These probe scientific question answering (ARC), general knowledge (MMLU), and logical reasoning (LogiQA2), complementing GSM8K's arithmetic reasoning. We present results for models that surpass Llama-3-70B~\citep{grattafiori2024llama} (the strongest model in~\cite{huh2024prh}) on at least one benchmark, testing whether the alignment-capability trend continues. The complete results for all models and benchmarks are included in \cref{sec:llm_eval}.

Our observations depend on the benchmark used to measure language-model performance.
For the benchmarks used in~\cite{huh2024prh}, such as HellaSwag and Wikitext, the trend partially extends to newer models, with positive $R^2_{\text{new}}$ of $0.30$ and $0.49$.
This partial continuation is specific to the local metric. Under
calibrated RSA and CKA, HellaSwag, and Wikitext also fail to extrapolate
(\cref{sec:e4-global-metrics}).
Those benchmarks measure next-token prediction and commonsense language understanding, closely tied to the autoregressive pretraining objective. However, these benchmarks are increasingly saturated for recent LLMs. We therefore focus on broader reasoning and knowledge benchmarks.

On reasoning benchmarks (\cref{fig:trends_combined}), we observe that recent models do not continue the scaling lines extrapolated from the model set from \cite{huh2024prh}. Instead, their points do not follow the predicted alignment trend and hint at saturation with respect to DINOv2 features.
The $R^2$ ranges from $-1.75$ to $-0.58$. The linear fit from the base models is a
worse predictor of alignment for recent models than simply predicting
their mean, i.e., the trend does not extrapolate.  Restricting the new-model set to base pretrained checkpoints (excluding instruction-tuned and reasoning-distilled models) leaves the saturation unchanged ($R^2_{\text{new{base}}}$ ranges from $-5.11$ to $-1.34$, see \cref{tab:r2_benchmarks}), so the failure to extrapolate is not an artifact of the post-training regime.

We note that the trend check holds the vision side fixed. A fixed
encoder can itself cap alignment (\cref{sec:diagnosis}), but the
saturation appears against all four DINOv2 variants up to ViT-G/14
(\cref{fig:trends_bubble_page1,fig:trends_bubble_page2}). The joint scaling
of both modalities is left to future work.

\finding{The alignment--capability trend saturates on reasoning benchmarks. Recent LLMs do not continue the scaling lines extrapolated from \cite{huh2024prh} for base and post-trained models alike.}

\subsubsection{What happens when the data is not bijective?}
\label{sec:no-bijecton-experriments}

In practice, the relationship between modalities, such as image and text, is inherently many-to-many: a single image can be described by countless text descriptions, and a single text caption can correspond to a large set of visually distinct images. More fundamentally, modalities often differ in information content.
Specifically, images encode spatial and textural structure that text captures only to a limited extent. On the other hand, text encodes abstraction, negation, and compositional semantics that images do not.
One could, in principle, bridge this gap trivially. For instance, one could encode pixel values as text or render captions as images and establish a bijection between those. These preserve the information, but the inductive structure (the specific properties that make each modality useful) of each modality is lost.

\begin{wrapfigure}{r}{0.36\textwidth}
    \centering
    \vspace{-1em}
\includegraphics[width=0.98\linewidth, trim=170 170 0 10, clip]{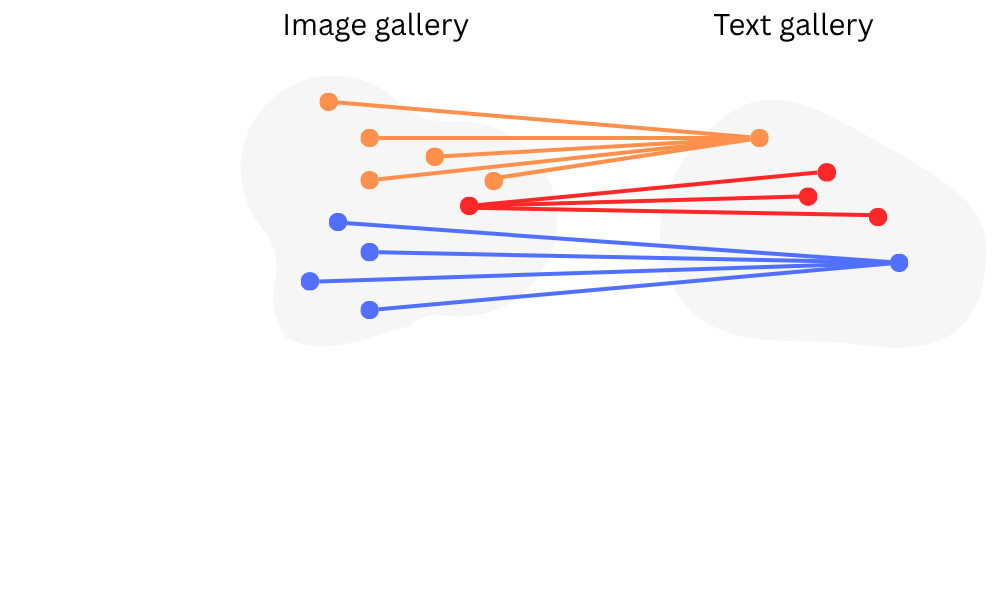}
\vspace{-.14in}
    \caption{Non-bijective correspondence. The nearest neighbor of a text caption for one image (blue) is a caption for a different image (red). The nearest image neighbor for a given image may be another image with the same caption.}
    \label{fig:surjection_example}
    \vspace{-1.3em}
\end{wrapfigure}
To test what happens when bijectivity is relaxed, we use the CycleReward dataset~\citep{bahng2025cycle} which pairs each real sample with multiple synthetic candidates. The I2T subset contains 11 generated captions per real image, and T2I consists of 12 synthetic images for each text prompt. This directly breaks the bijection, i.e.\ one-to-one matching, that our earlier analysis assumes.
We evaluate mutual $k$NN by densifying one modality at a time: for T2I we keep the text fixed and increase the number of generated images per prompt, and for I2T we keep the image fixed and add generated captions.
For illustration, let us consider the T2I experiments where the closest neighbor in the densified modality is more likely to be a similar image that is associated with the same caption,  while in the sparse text space the NN will be a caption for a different image (see \cref{fig:surjection_example}).

\begin{figure}[t]
    \centering
    \begin{subfigure}[t]{0.47\linewidth}
        \centering \includegraphics[width=\linewidth]{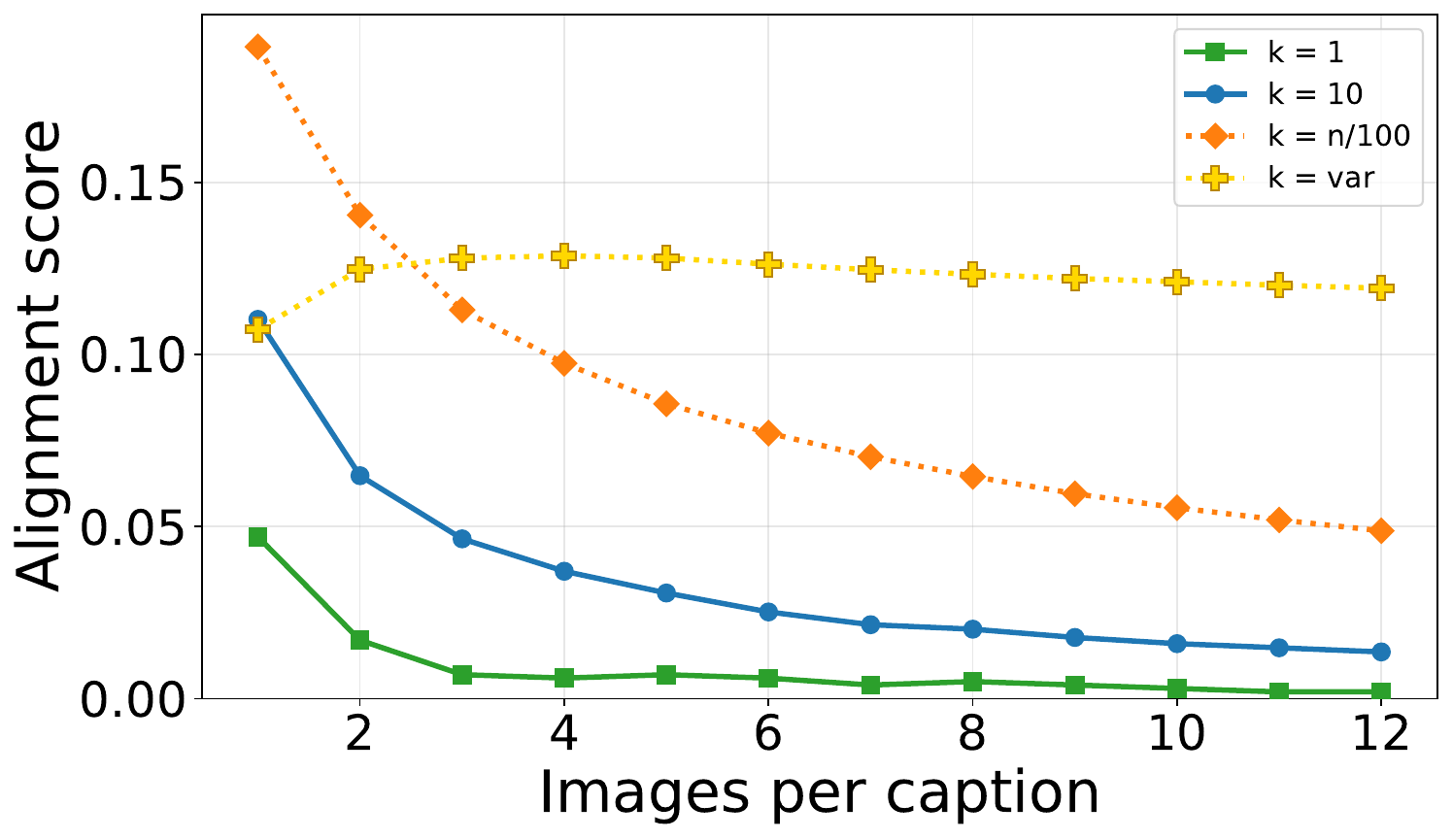}
        \vfill
    \end{subfigure}
    \hfill
    \begin{subfigure}[t]{0.47\linewidth}
        \centering
        \includegraphics[width=\linewidth]{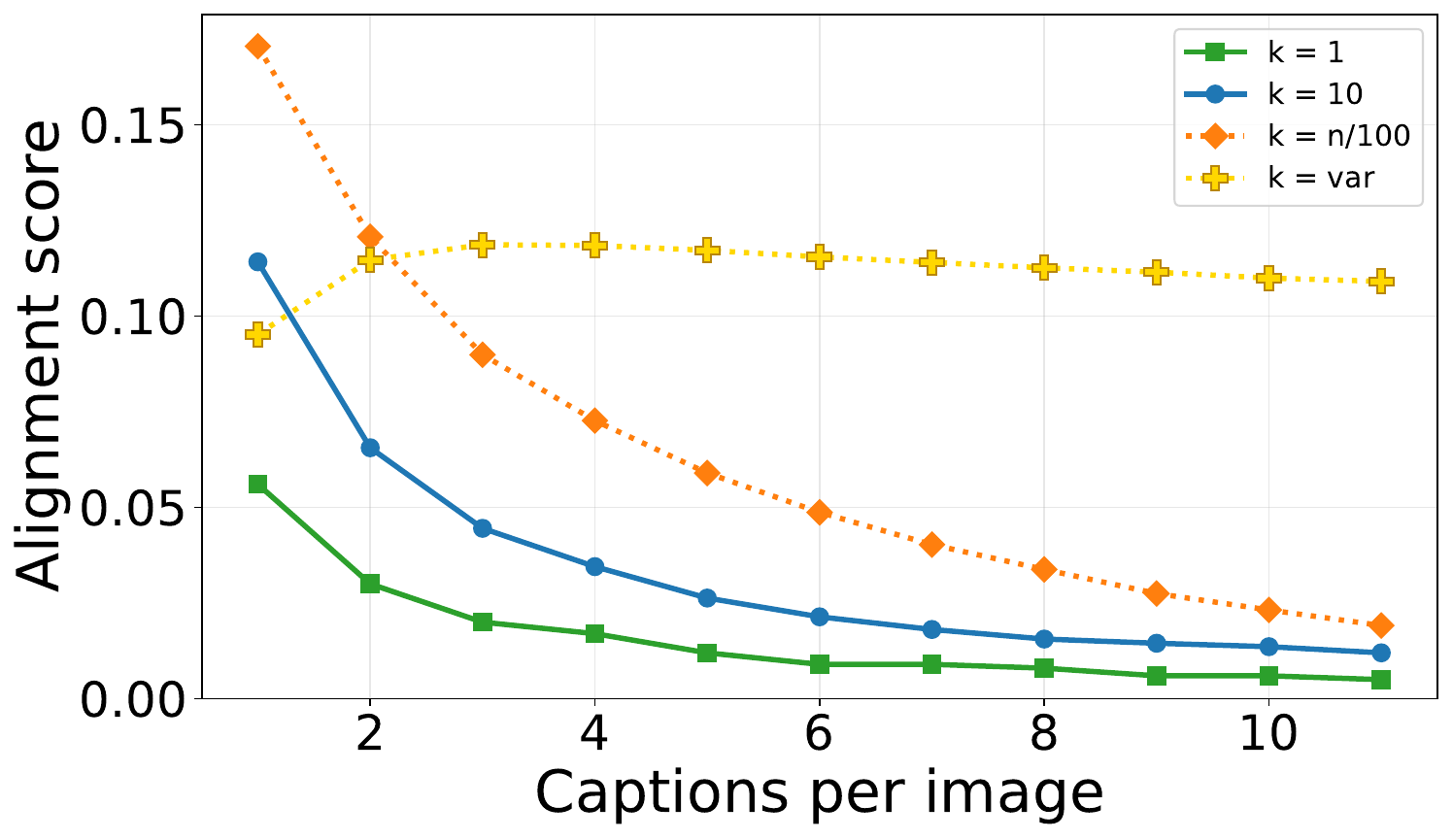}
        \vfill
    \end{subfigure}
\vspace{-.1in}
    \caption{Effect of relaxing the bijective assumption on text-image alignment, using the CycleReward dataset~\citep{bahng2025cycle}. We densify one modality by adding more images per caption (left) or captions per image (right) while keeping the other fixed. Mutual $k$NN alignment decreases for both $k{=}1$, $k{=}10$, and $k=\frac{n}{100}$. Our variable $k$ adaptation of mutual $k$NN remains stable.}
    \label{fig:densifying_modalities}
    \vspace{-1.3em}
\end{figure}

We adapt the mutual $k$NN metric so that a match is counted when the retrieved item corresponds to the source sample, even if it is not the exact same caption or image.
In \cref{fig:densifying_modalities}, we see that the mutual $k$NN scores still decrease as the one-to-one assumption is relaxed.
The convergence evidence in \cite{huh2024prh} rests on a bijective pairing that real-world multi-modal data rarely satisfies, and the mutual $k$NN metric does not distinguish between actual misalignment and differences in density across modalities. Hence, we adapt the mutual $k$NN metric to a variable $k$ that measures a constant neighborhood radius around a query as the gallery grows, whilst collapsing non-bijective data into bijections when computing the metric (similar to the fixed $k$ extension, see \cref{sec:q2g_duplicates}).

\finding{As the one-to-one assumption is relaxed, the standard mutual $k$NN variants decay. On non-bijective data, it reflects correspondence density as much as alignment, revealing a limitation of the evaluation protocol.}

\section{What kind of structure do vision and language models share?}
\label{sec:diagnosis}

\Cref{sec:probing_alignment} established that the agreement
surviving densification is coarse. Here, we characterize both what is shared
and what is not, on the ImageNet validation set~\citep{deng2009imagenet}, which operates in a retrieval
regime comparably dense to WIT-1M (see \cref{tab:nn_quality}). The ImageNet class labels additionally let us isolate fine
within-class ordering.

\begin{figure}[t]
  \centering
  \includegraphics[width=\linewidth]{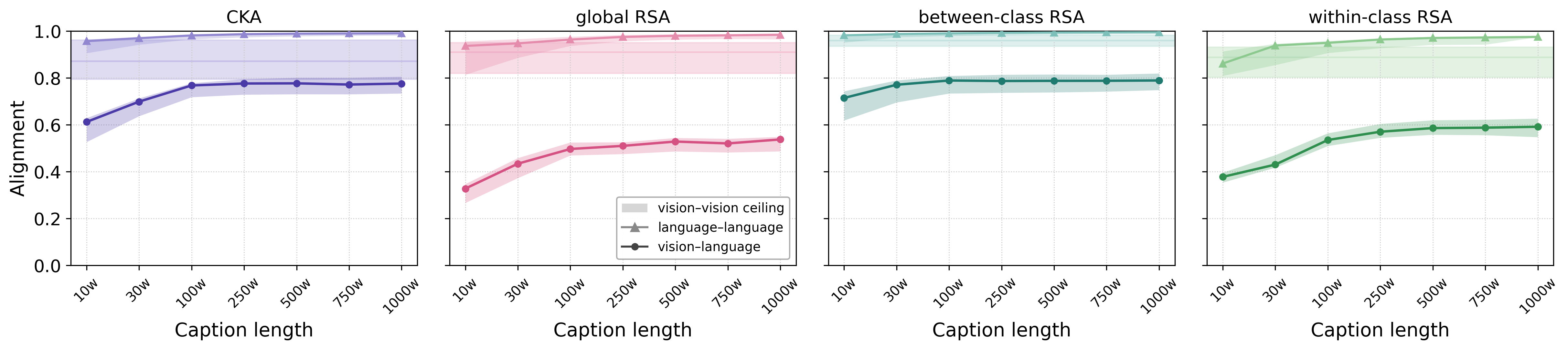}
  \vspace{-.25in}
  \caption{Unimodal alignment ceilings for different caption lengths on the ImageNet validation set, for CKA, global RSA, and between- and within-class RSA. Vision--vision (horizontal band) and language--language
  (triangles) ceilings stay high across caption length, while the
  vision--language (circles; DINOv2-base against 12 LLMs) band saturates by $100$--$250$
  words well below (particularly for RSA).}
  \label{fig:ceiling_bands}
  \vspace{-1.3em}
\end{figure}

\begin{wrapfigure}{r}{0.45\textwidth}
  \centering
  \vspace{-1em}
  \includegraphics[width=0.42\textwidth]{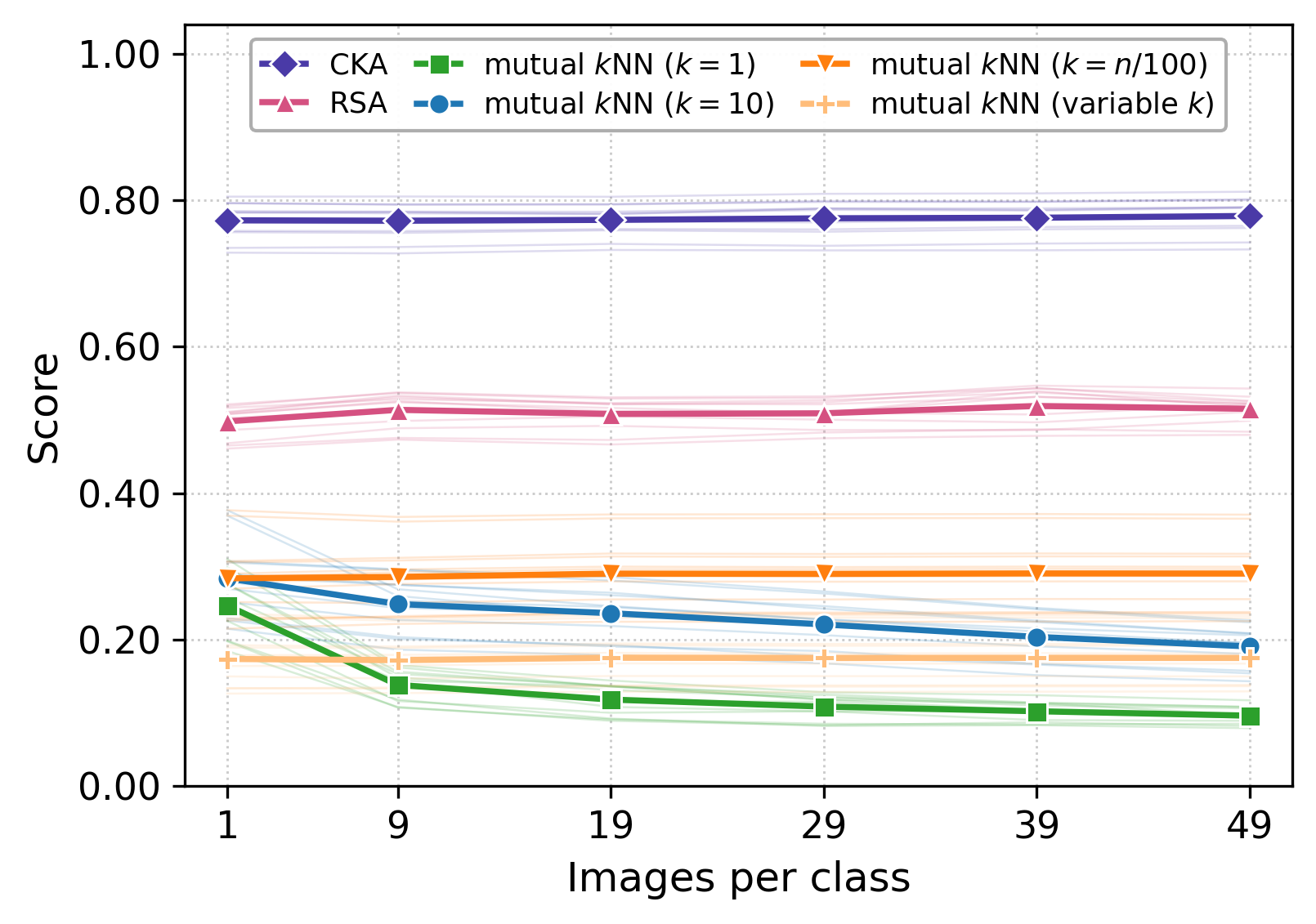}
  \vspace{-.1in}
  \caption{Cross-modal alignment on the ImageNet validation set
  (500-word captions). As the gallery grows, fixed-$k$ mutual $k$NN
  decays, while $k{=}n/100$, variable-$k$, RSA, and CKA are stable.}
  \label{fig:ipc_invariance}
  \vspace{-1.4em}
\end{wrapfigure}
\myparagraph{Using global instead of local metrics.}
We complement fixed-$k$ mutual $k$NN with four additional metrics.
RSA is the Spearman correlation between the two models' pairwise
cosine-distance vectors over the same samples. Specifically, within-class RSA
computes it separately inside each ImageNet class and measures the fine same-category ordering. Linear CKA compares the
dominant directions of variation~\citep{kornblith2019similarity}. The
variable-$k$ mutual $k$NN sets each query's neighborhood size from its
own local density (all gallery items within the query's self-NN
distance) and scores neighborhood IoU. The differences between these metrics provide information about agreement at different granularities. Details are in
\cref{sec:supp_metrics}.

\cref{fig:ipc_invariance} considers the galleries from $1$ to $49$
images per class from the ImageNet validation
set with $1000$ queries (one per class), and $500$-word per-image Gemini captions
(\cref{sec:captioning}).
Mutual $k$NN at
$k{=}n/100$ and variable-$k$, RSA variants, and linear CKA stay stable. Consequently, they read the geometry the models impose
rather than being gallery-dependent measures. In the following, we focus on RSA and CKA.

\myparagraph{The gap between cross-modal and unimodal alignment is organizational.}
We compare the calibrated cross-modal scores to how strongly two good models agree when they observe the \emph{same} modality.
The unimodal ceiling is a band (\cref{fig:ceiling_bands}): vision pairs reach RSA $\approx\!0.95$ same-family and $\approx\!0.82$ cross-family (DINOv2 $\times$ DINOv3~\citep{simeoni2025dinov3}), and 6 LLMs share the within-class fine ordering at $0.94$--$0.98$ across families.
The fine structure is successfully encoded in each modality.
However, cross-modal agreement behaves differently. If
missing descriptions were the bottleneck, richer captions would close the gap. Instead, the cross-modal band stalls by
$100$--$250$ words, while  unimodal agreement is
high at every caption length (\cref{fig:ceiling_bands}).
The size of the cross-modal gap depends on the
granularity a metric reads. Cross-modal alignment reaches
$83$--$92\%$ of the vision ceiling in CKA, which is dominated by the
coarse, shared directions of variation, but only $51$--$58\%$ in RSA,
which grades the full ordering.

\begin{wrapfigure}{r}{0.45\textwidth}
  \centering
  \vspace{-2.3em}
  \includegraphics[width=0.42\textwidth]{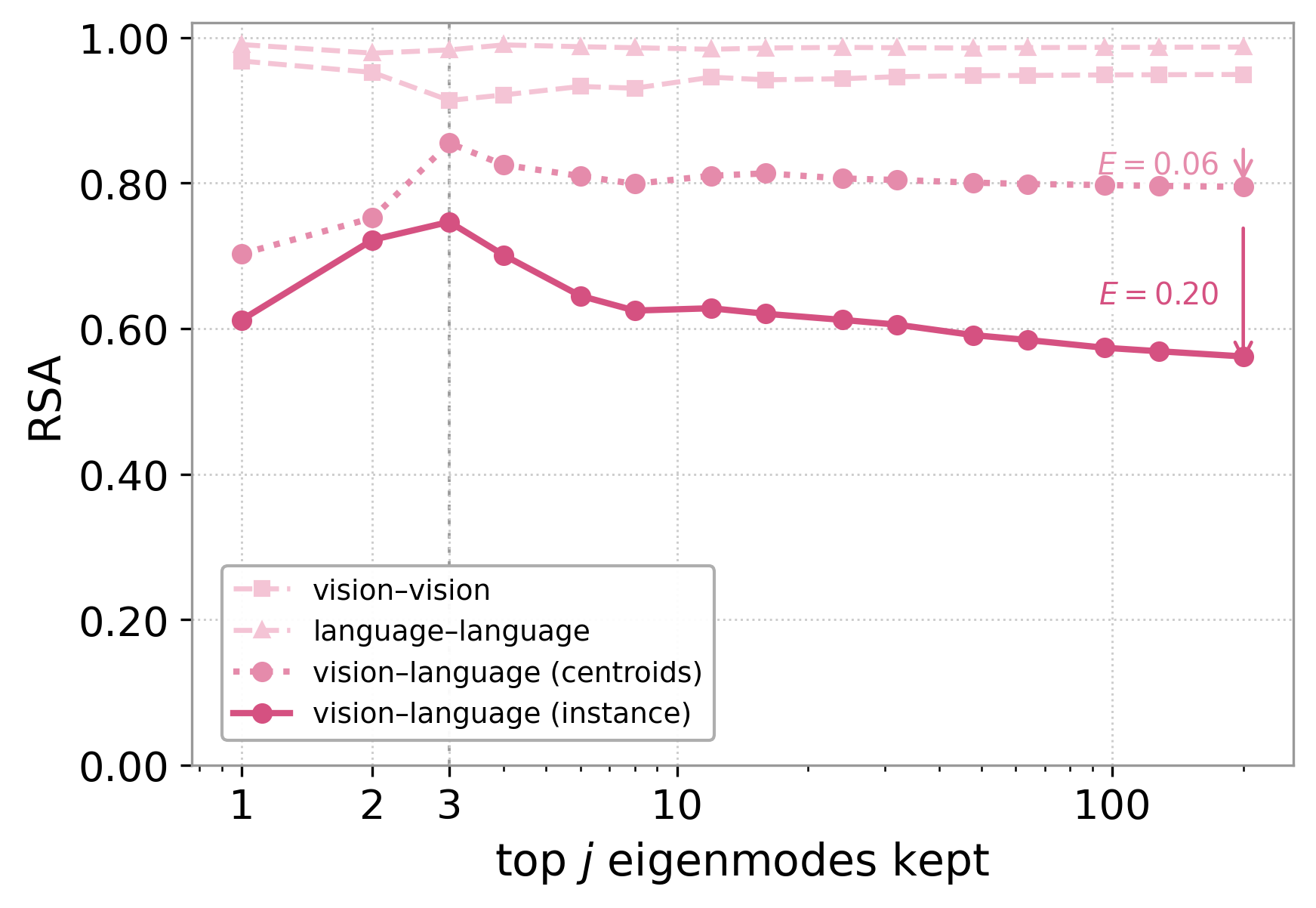}
\vspace{-.1in}
  \caption{Truncated RSA (top-$j$ eigenmodes) on ImageNet (validation). Cross-modal RSA erodes
  with more modes, unimodal is flat. Collapsing to class centroids shrinks erosion ($E$). Agreement is in the top 3 eigenmodes, disagreement is within classes.}
  \label{fig:eigenmodes_main}
  \vspace{-1.8em}
\end{wrapfigure}

The modalities agree on the coarse layout and then organize the fine structure differently.
Truncating both representations to their top
$j$ eigenmodes locates this directly (Fig.~\ref{fig:eigenmodes_main}).
Cross-modal RSA peaks at $j{=}3$, where three modes carrying 80\% of the CKA weight already give the peak agreement ($0.75$), and erodes to $0.55$ as the tail is added.
The tail is not noise. Each model's agreement with its own full representation keeps rising over the same range (\cref{sec:eigenspace_threshold}), so the tail carries structure each modality orders consistently and the other never matches.

The gap is largest where the structure is most fine-grained, and is smaller when measuring coarse structure.
Within-class cross-modal RSA reaches $77$--$85\%$ of the vision ceiling, against $51$--$58\%$ for the global ordering, saturating by ${\sim}100$ words (\cref{fig:ceiling_bands}). Collapsing each class to its centroid leaves only between-class geometry. Between-class coarse cross-modal RSA
is $0.79$ against the weakest vision pair's $0.94$. Coarse
alignment has the smallest gap, but not a closed one.

Using von Uexküll’s \textit{Umwelt}, we distinguish informational ceilings from organizational alignment.
Short captions resemble an informational ceiling by omitting visual details. Enriching them narrows the cross-modal gap while the unimodal ceilings stay flat (\cref{fig:ceiling_bands}). However, the narrowing saturates by $100$--$250$ words, and even with multi-paragraph descriptive captions, cross-modal RSA stalls well below unimodal levels.
Vision and language models occupy distinct \textit{Umwelten}.

\finding{The cross-modal gap is organizational. Both modalities represent fine structure consistently but arrange it differently, so the gap is smallest for CKA, which reads coarse shared structure, and largest for RSA, which grades the full ordering.}

\section{Discussion and future work}
\label{sec:discussion}
The Platonic Representation Hypothesis suggests that models trained on different modalities align and eventually converge toward the same representation of reality as they scale. Our results separate claims about the representations themselves from limitations of the original experimental evidence.

The drop of fixed-$k$ mutual
$k$NN alignment as data scales traces how far cross-modally shared structure extends. Same-modality pairs remain far more stable under identical densification. Even in the original small-scale regime, the reported trend that stronger language models align better with vision does not extrapolate: for recent models, alignment saturates on reasoning benchmarks. The reduction of mutual $k$NN under non-bijective correspondences, by contrast, is a limitation of the protocol: it conflates genuine misalignment with the many-to-many structure of real image-text data. Measuring what the modalities share therefore requires separating coarse from fine structure.

On ImageNet, class labels provide this separation. Cross-modal agreement approaches the same-modality ceiling at coarse level, but falls short on the fine ordering within a class. \cite{huh2024prh} anticipate that non-bijective observation functions cap alignment, but leave open whether the cap matters.
Our measurements show that it does: richer captions increase alignment, but the gain saturates well before the longest captions we generate, leaving cross-modal agreement below the same-modality ceiling at both coarse and fine granularity.

\textbf{The text and image \emph{Umwelten} overlap in coarse structure but diverge in fine-grained organization.} A cross-modal metric therefore measures translation between two differently organized maps, not progress toward a common form.

\myparagraph{Future work: in search of bijection.}
Prior work on the Platonic Representation Hypothesis evaluates alignment under a one-to-one correspondence assumption between modalities.
However, real-world image–text data is inherently many-to-many, and the extent to which any approximate bijection exists at the level of representations remains unclear. A key direction for future work is to directly test whether language can serve as a lossless bottleneck for image reconstruction (i.e.\ an image-text-image autoencoder). Recent round-trip image--text diffusion~\citep{guo2026bit} and panoptic captioning~\citep{lin2025panoptic} provide frameworks for such a test.
If a lossless bottleneck proves elusive in realistic settings (e.g. text bottlenecks under a thousand words), the next question is which part of the text-image space (intersection of a Venn diagram) admits a bijection, and how to disentangle it from the rest.

\paragraph{Acknowledgments.}
This work was in part supported by the BMFTR (FKZ: 16IS24060), the DFG (SFB 1233, project number: 276693517), NSF IIS-2403305, and ONR MURI. This research utilized compute resources at the Tübingen Machine Learning Cloud. The authors thank all Efros group members for valuable discussions that shaped this work, and particularly Tyler Bonnen and Amil Dravid for proofreading the draft. Lastly, we thank Phillip Isola for the thought-provoking hypothesis, and for discussing and engaging openly with our disagreements -- a rare kind of intellectual generosity.

\putbib[main]
\end{bibunit}

\clearpage
\begin{bibunit}[tmlr]

\setcounter{figure}{0}
\setcounter{table}{0}
\setcounter{footnote}{0}
\setcounter{equation}{0}
\setcounter{finding}{0}
\renewcommand{\thefigure}{S\arabic{figure}}
\renewcommand{\theHfigure}{S\arabic{figure}}
\renewcommand{\theHtable}{S\arabic{table}}
\renewcommand{\theHfootnote}{S\arabic{footnote}}

\appendix
\setcounter{section}{0}
\renewcommand{\thesection}{\Alph{section}}

\thispagestyle{empty}
\begin{center}
    {\Large \textbf{Supplementary Material}}\\[0.5em]
    {\Large \textbf{Back into Plato's Cave: Examining Cross-modal\\ Representational Convergence at Scale}}
\end{center}

\section{Is the drop in mutual $k$NN alignment at scale caused by the metric, layer selection, caption quality, or by specific model choices?}
In this section, we provide additional experiments to verify that the main findings in the paper are not due to a confounding variable.

\subsection{Sanity check: does mutual $k$NN inherently drop at scale even within modalities?}
We test whether the alignment drop with increasing gallery size (\cref{fig:both} in the main paper) is merely an artifact of the mutual $k$NN metric being harder at scale.

Specifically, we measure within-modality alignment for two pairs of models: two language models of different scales (OpenLlama-3b and OpenLlama-13b), and, separately, two vision model pairs (DINOv2-base / DINOv2-giant and DINOv2-base / CLIP-B16). If mutual $k$NN alignment collapses for dense galleries regardless of the models being compared, the cross-modal drop observed in the paper would be uninformative. If within-modality alignment remains more stable, the cross-modal drop is meaningful.

\begin{figure}[h]
    \centering
    \begin{subfigure}[t]{0.32\linewidth}
        \centering
        \includegraphics[width=\linewidth]{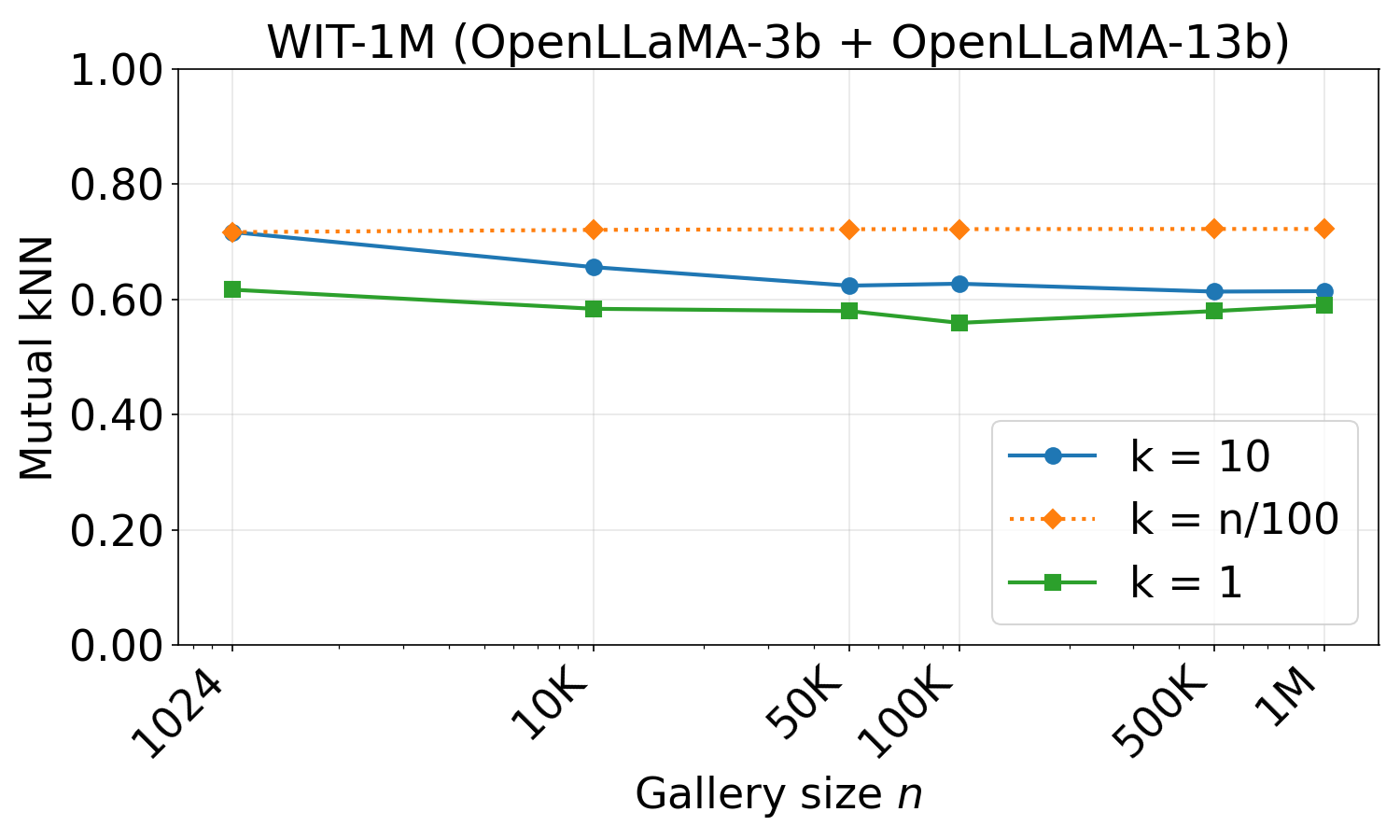}
        \caption{OpenLlama-3b and OpenLlama-13b}\label{supp_fig:fig_llama}
    \end{subfigure}
    \hfill
    \begin{subfigure}[t]{0.32\linewidth}
        \centering
        \includegraphics[width=\linewidth]{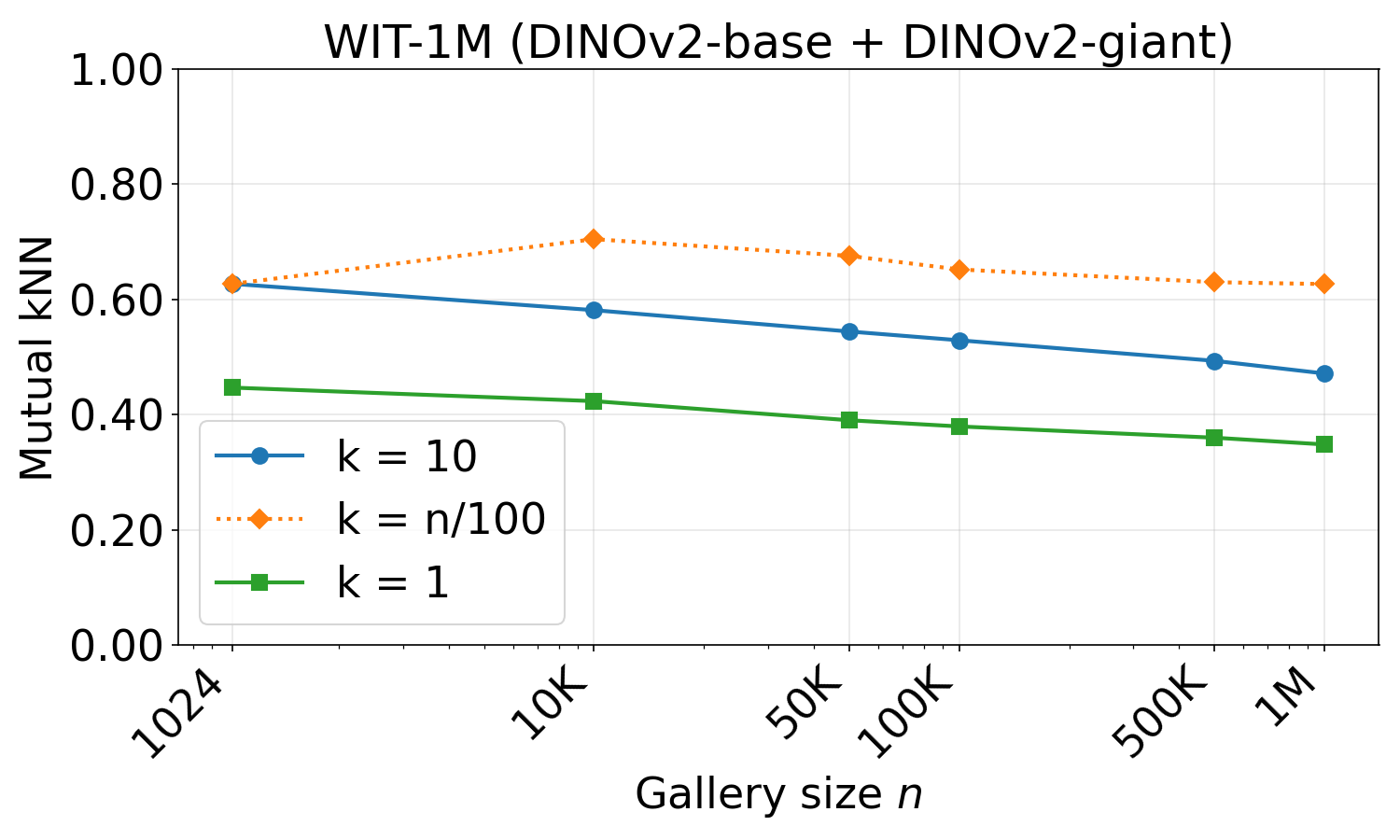}
        \caption{DINOv2-base and DINOv2-giant}\label{supp_fig:fig_dino}
    \end{subfigure}
    \hfill
    \begin{subfigure}[t]{0.32\linewidth}
        \centering
        \includegraphics[width=\linewidth]{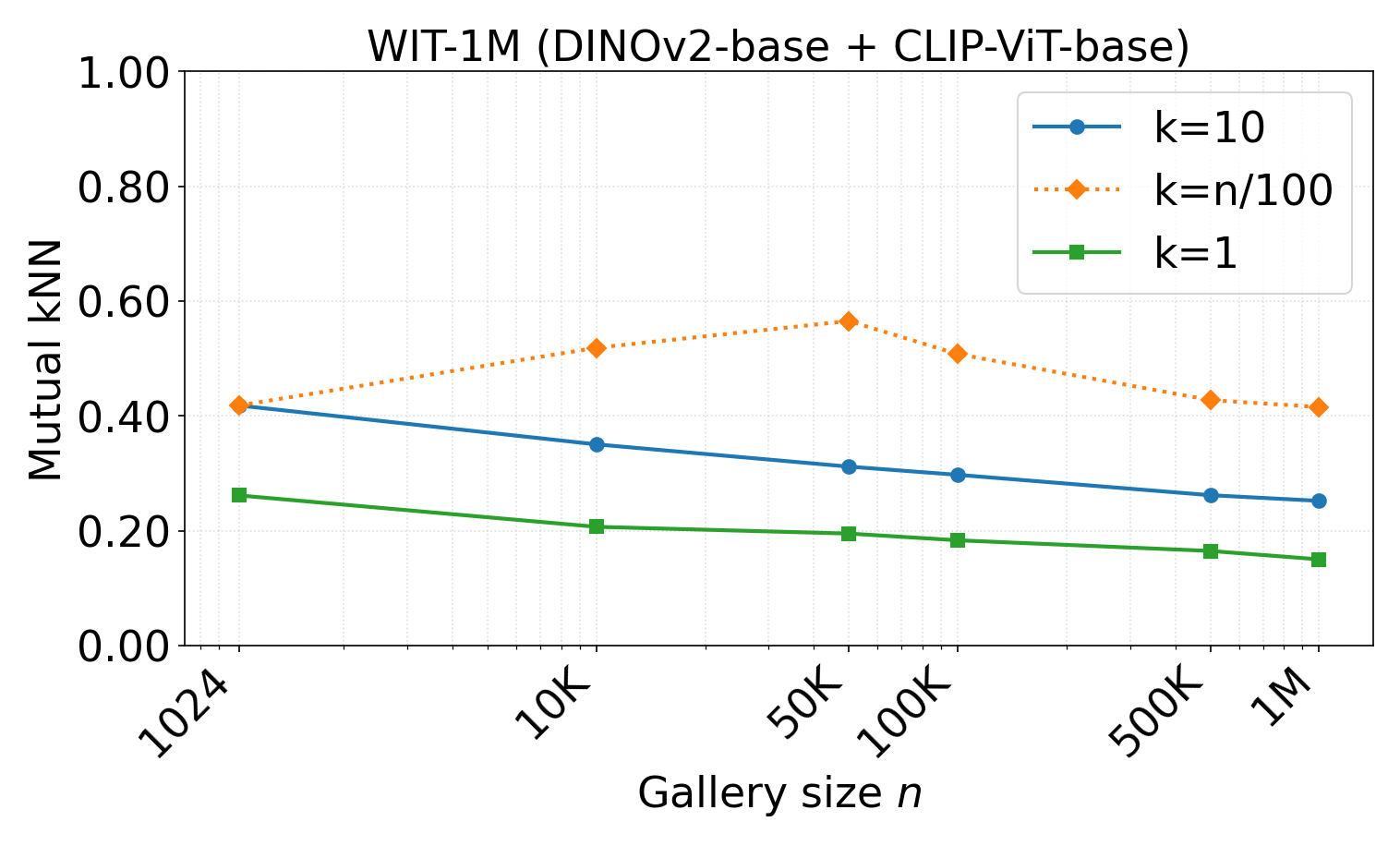}
        \caption{DINOv2-base and CLIP-B/16}\label{supp_fig:fig_dino_clip}
    \end{subfigure}
    \vspace{-.05in}
    \caption{Unimodal mutual $k$NN alignment as a function of gallery
    size on WIT-1M, for same-family pairs (a, b) and a cross-family
    vision pair with no shared training recipe (c). \textit{In contrast
    to cross-modal alignment (\cref{fig:both}), unimodal alignment,
    including for the cross-family pair, remains significantly more
    stable across scales.}}
    \label{fig:uni_dino_alignment}
    \vspace{-0.5em}
\end{figure}

As shown in \cref{fig:uni_dino_alignment}, unimodal alignment remains much more stable across gallery sizes than the cross-modal alignment reported in the main paper. For the OpenLlama pair, mutual $k$NN at $k{=}1$ stays between $[0.59, 0.62]$, and for the DINOv2 pair between $[0.35, 0.45]$, across all gallery scales. Cross-family alignment is lower than same-family, as
expected, but remains more stable than cross-modal alignment.
\textbf{Unimodal mutual $k$NN decays by at most ~2× over the same range where cross-modal decays by ~50× ($k{=}1$). The cross-modal decay reflects that the structure shared across modalities does not extend to the fine granularity probed by small $k$ at large $n$.}

\begin{figure}[h]
    \centering
    \includegraphics[width=0.6\linewidth]
{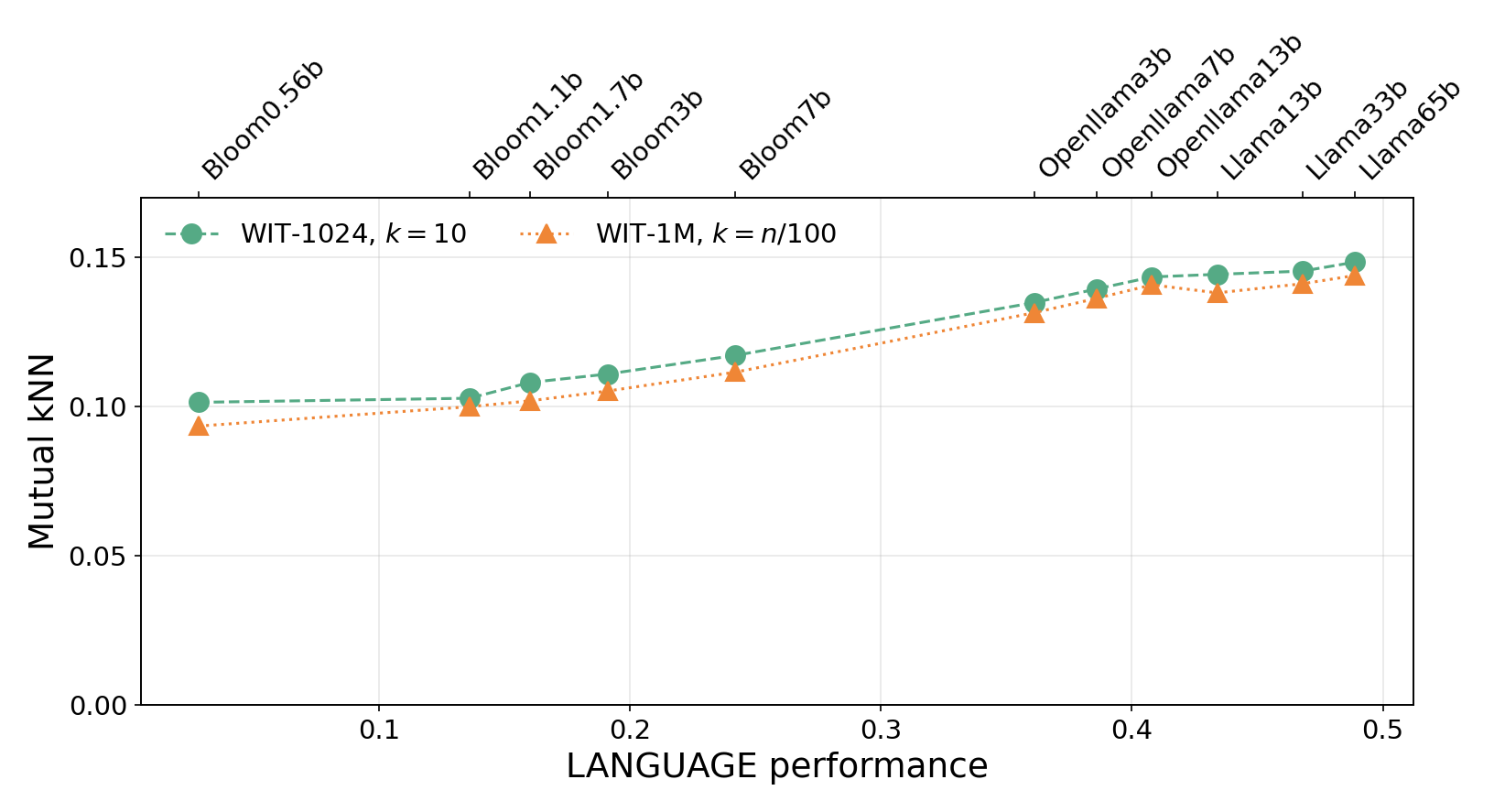}
    \caption{Alignment vs.\ LLM performance: the WIT-1024 trend at $k{=}10$
    compared to WIT-1M at $k{=}\frac{n}{100}$. \textit{Unlike the fixed-$k$ case
    (\cref{fig:second_plot} in the main paper), the capability trend persists for
    $k{=}n/100$, consistent with stable coarse overlap rather
    than fine-grained convergence.}}
    \label{fig:llm_trend_kpct}
        \vspace{-0.5em}
\end{figure}

We additionally verify that the alignment trend occurs at the
coarse $k{=}\frac{n}{100}$ level: \cref{fig:llm_trend_kpct} compares the
WIT-1024 trend at $k{=}10$ with the WIT-1M trend at $k{=}\frac{n}{100}$, and
the two largely match. This is consistent with the original evidence using mutual $k$NN reflecting coarse semantic-category overlap rather than fine-grained representational convergence.

\subsection{Sanity check: is the alignment drop an artifact of the layer pair selected at WIT-1024?}\label{sec:supp_layer_selection}
For our experiments, we select the best layer pair on the 1024-sample subset of WIT and reuse this pair across larger-scale evaluations. Since our central claim concerns how alignment behaves as the gallery densifies, we test whether the observed patterns persist when the layer pair is re-selected at each gallery scale.

\begin{figure}[h]
    \centering
     \begin{subfigure}[t]{0.48\linewidth}
        \centering
        \includegraphics[width=\linewidth]{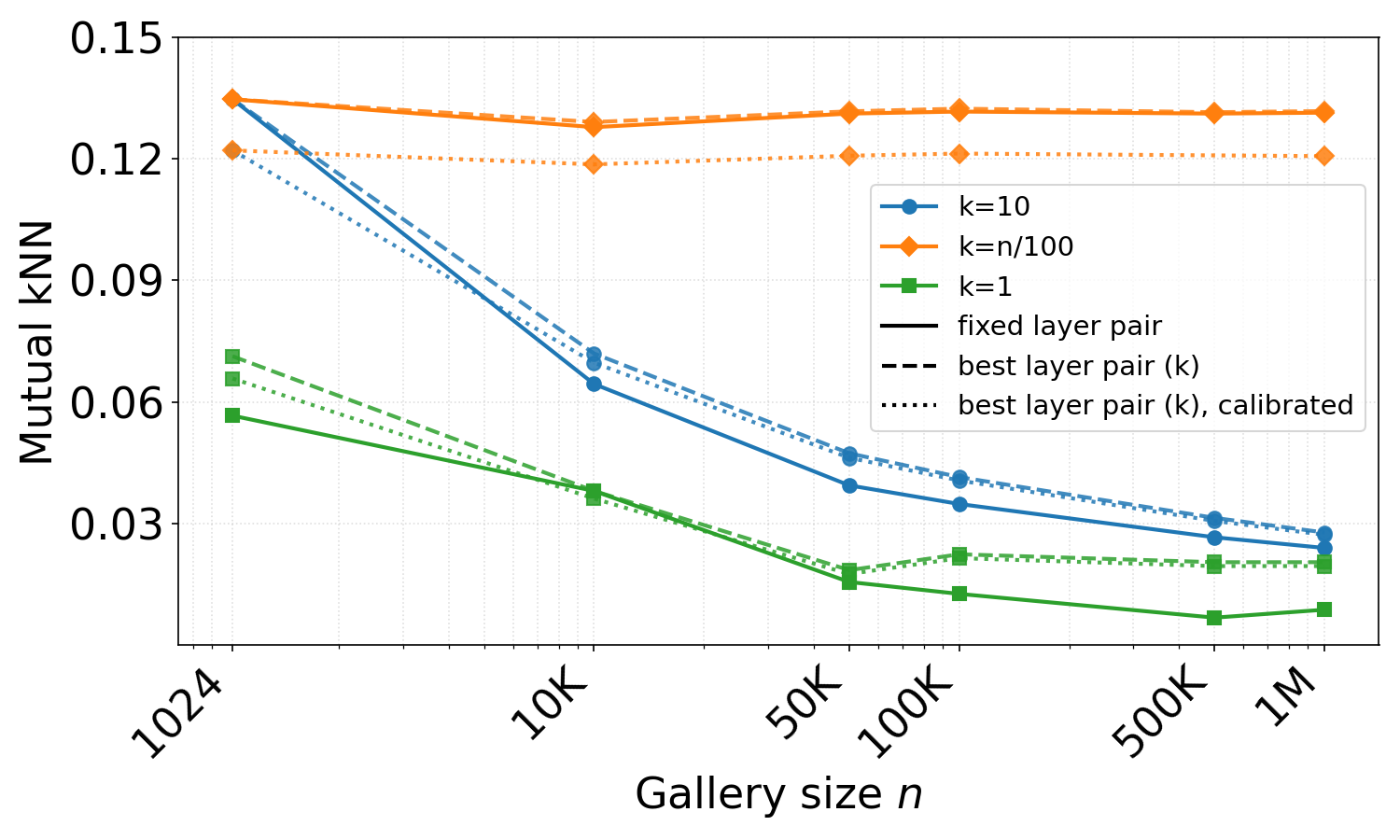}
        \caption{Per-scale layer re-selection performed independently for each $k$. The alignment trends are preserved.}\label{supp_fig:layer_per_k}
    \end{subfigure}
        \hfill
    \begin{subfigure}[t]{0.48\linewidth}
        \centering
         \includegraphics[width=\linewidth]{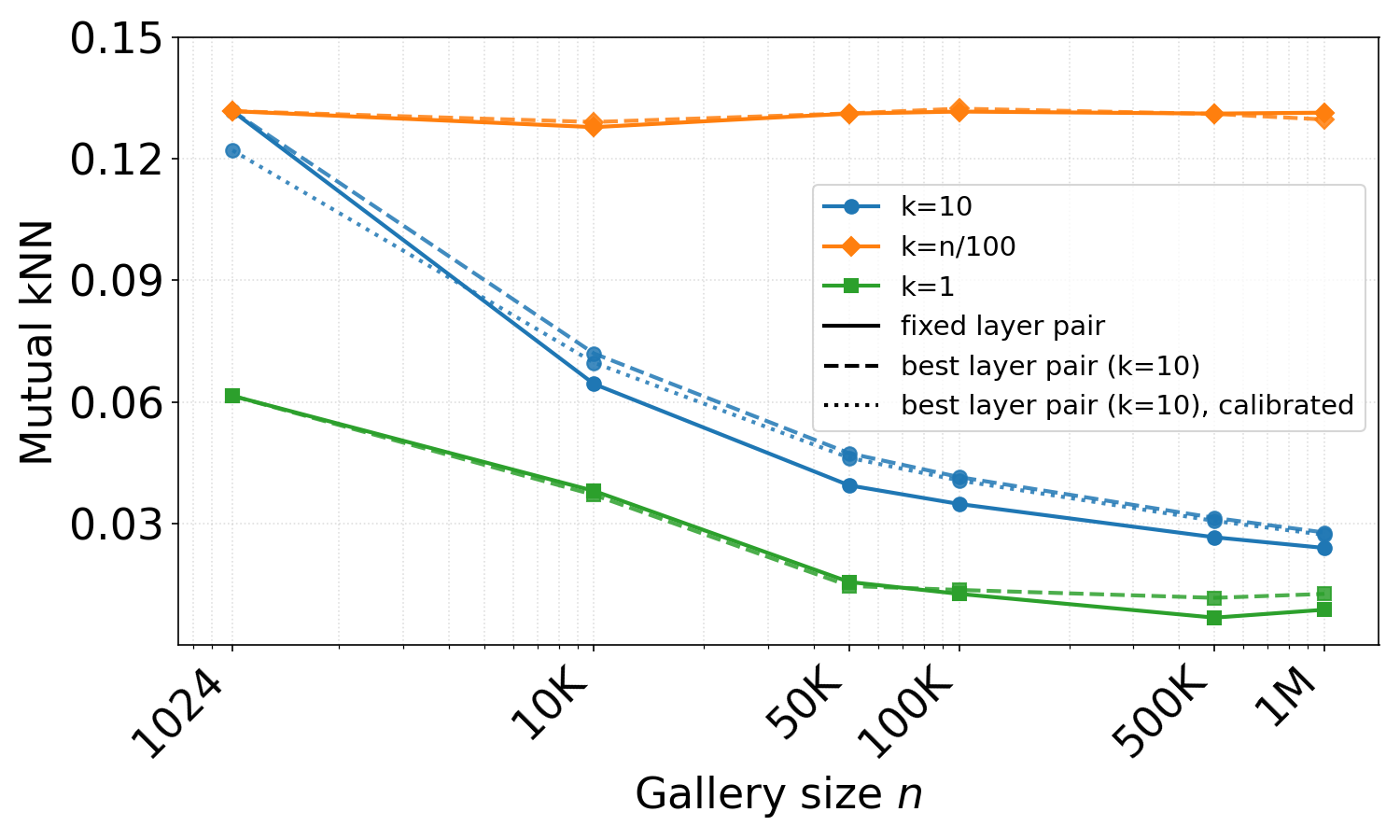}
        \caption{Per-scale layer re-selection performed at $k{=}10$ and applied to all $k$ values. The same trend is observed.}\label{supp_fig:layer_k10}
    \end{subfigure}
    \caption{Cross-modal mutual $k$NN alignment on WIT-1M with the layer pair re-selected at each gallery scale, rather than fixed at WIT-1024. Dotted lines show calibrated scores following \cite{groger2026revisitingplatonicrepresentationhypothesis}. \textit{The qualitative scaling trend is unchanged, indicating that the alignment drop is not driven by suboptimal layer choices.}}
    \label{fig:layer_selection_ablation}
    \vspace{-0.5em}
\end{figure}

Specifically, we re-run the layer search at each gallery size in $\{1024, 10\text{K}, 50\text{K}, 100\text{K}, 500\text{K}, 1\text{M}\}$ and report the resulting mutual $k$NN alignment. \cref{fig:layer_selection_ablation} shows the comparison: (a) re-selecting layers per scale and per $k$ (i.e., the optimal layer pair for $k{=}1$, $k{=}10$, and $k{=}\frac{n}{100}$ separately at each gallery size), and (b) re-selecting layers per scale at fixed $k{=}10$ and applying the same layer pair to other $k$ values. Exhaustive layer-pair search inflates alignment scores by selection. We follow the null calibration proposed by \cite{groger2026revisitingplatonicrepresentationhypothesis}, subtracting the expected alignment under permuted image-text pairings to correct for this bias (shown as dotted lines).

In both cases, the alignment trend with increasing gallery size is preserved. \textbf{This indicates that the alignment drop for fixed small $k$ reported in the main paper does not suffer from an artifact of the layer pair being fixed at WIT-1024 scale.}

\subsection{WIT-1M-recap: Is the alignment drop caused by poor captions?}

\begin{figure}[h]
    \centering
    \includegraphics[width=0.5\linewidth]{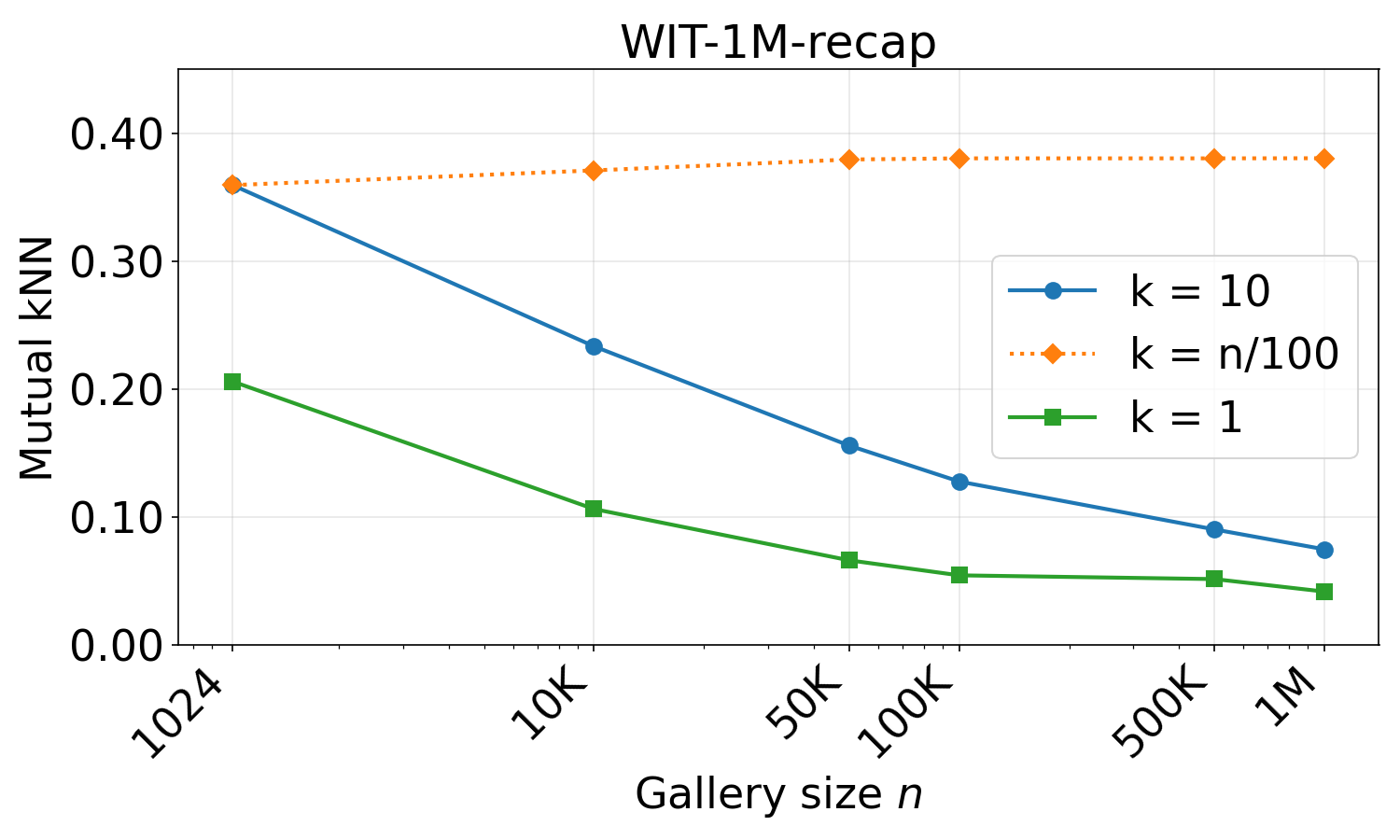}
    \caption{Cross-modal mutual $k$NN alignment on images recaptioned using gemini-3-flash-preview (WIT-1M-recap) as the gallery grows to 1M samples. \textit{Detailed captions result in overall higher mutual $k$NN scores, but do not prevent the drop in scores.}}
    \label{fig:wit_recap_scaling}
\end{figure}

One might hypothesize that the alignment drop at scale is driven by the low quality of the WIT captions rather than by a fundamental cross-modal difference. To test this, we recaption WIT-1M using gemini-3-flash-preview as described in \cref{sec:captioning}. The resulting WIT-1M-recap dataset contains visually detailed descriptions of around 500 words per image. As shown in \cref{fig:wit_recap_scaling}, mutual $k$NN alignment still drops with gallery size. More detailed captions give overall higher mutual $k$NN scores, but do not prevent the decline in scores. \textbf{This suggests that caption quality is not the primary driver of the mutual $k$NN alignment drop}.

\begin{figure}[t]
    \centering
     \begin{subfigure}[t]{0.48\linewidth}
        \centering
        \includegraphics[width=\linewidth]{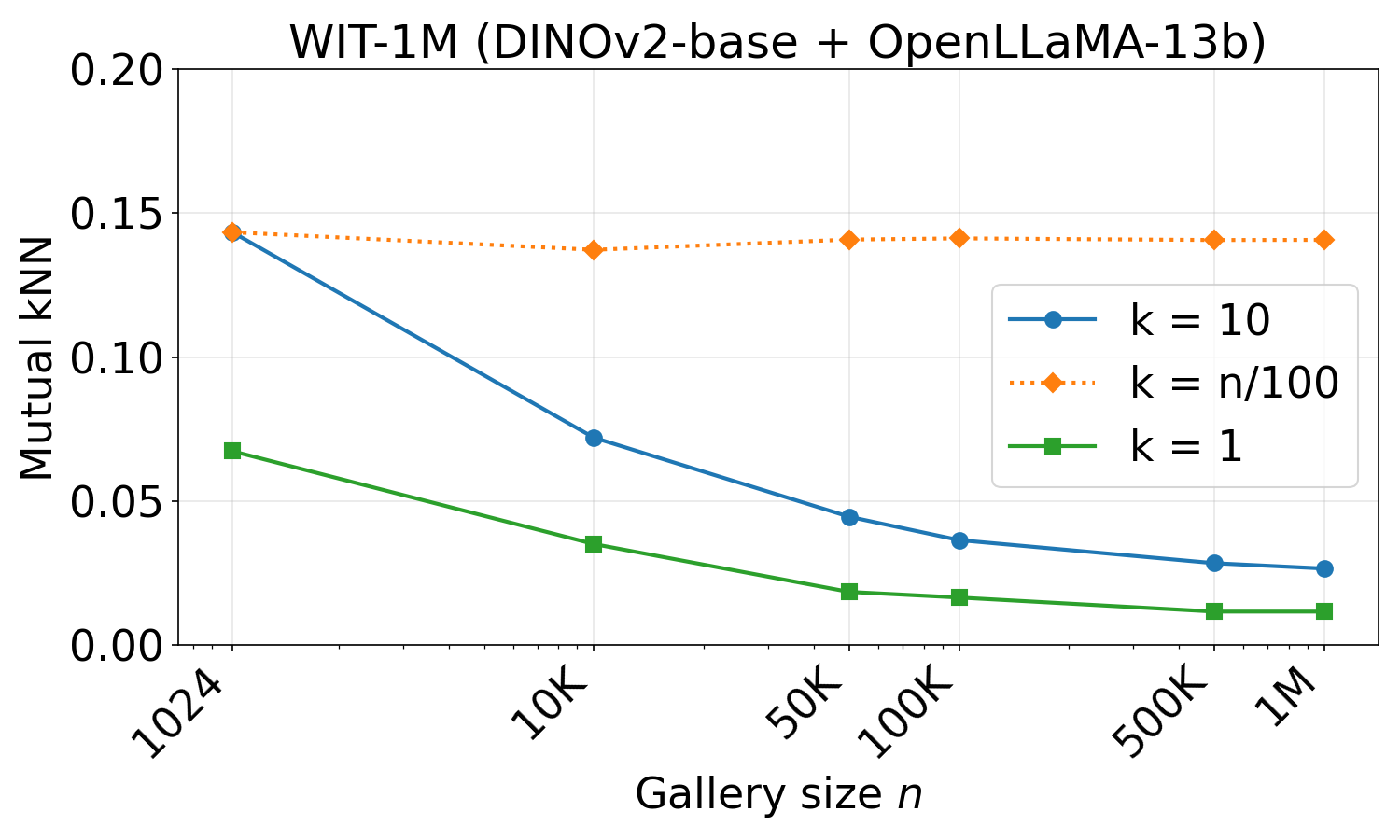}
        \caption{DINOv2-base and OpenLlama-13b}\label{supp_fig:fig_5}
    \end{subfigure}
        \hfill
    \begin{subfigure}[t]{0.48\linewidth}
        \centering
        \includegraphics[width=\linewidth]{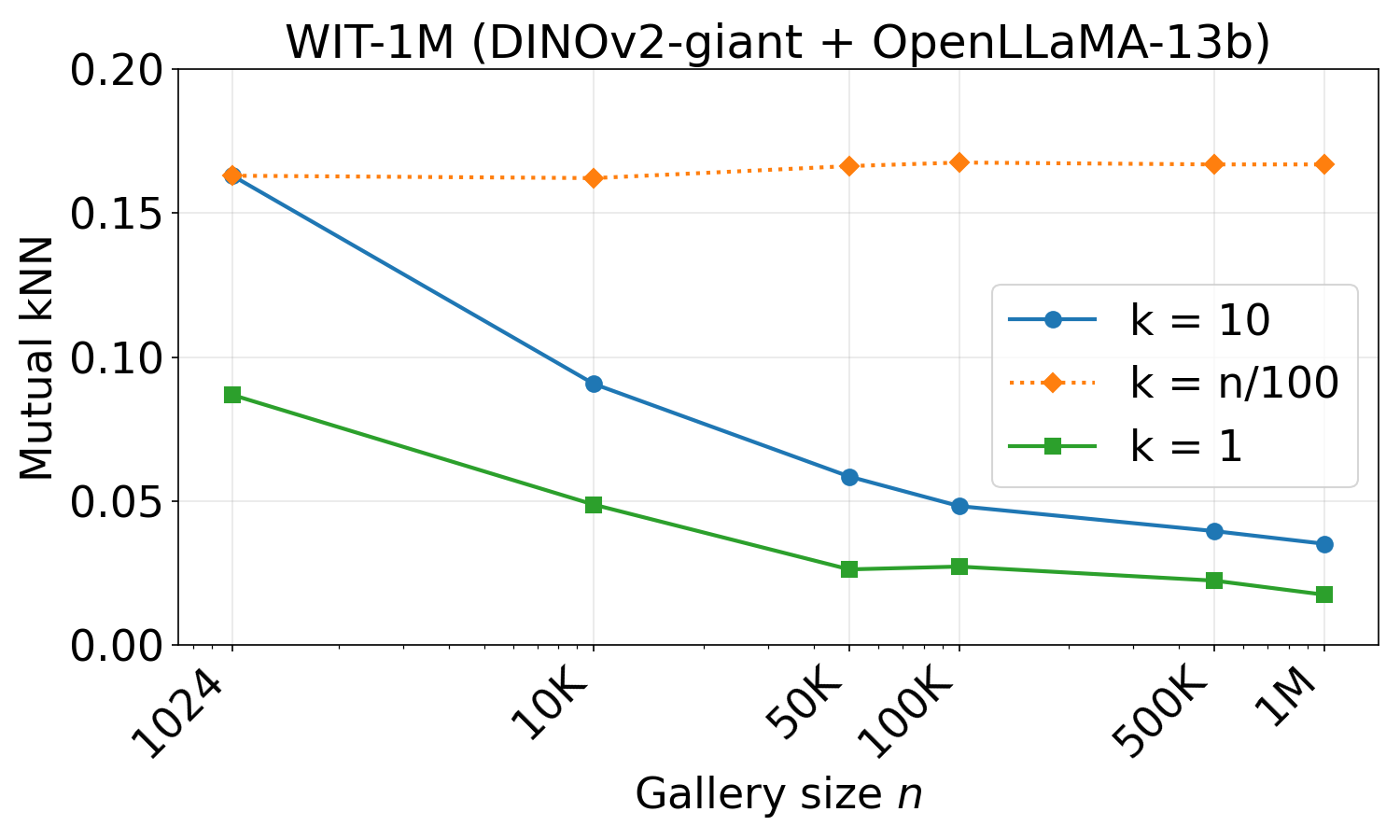}
        \caption{DINOv2-giant and OpenLlama-13b}\label{supp_fig:fig_6}
    \end{subfigure}
    \caption{Cross-modal mutual $k$NN alignment as gallery grows from WIT-1024 to WIT-1M for additional, stronger model pairs. Replacing DINOv2-base with the stronger DINOv2-giant and OpenLlama-3b (Fig.\ 4 in the paper) with OpenLlama-13b does not prevent the drop. \textit{This suggests that the degradation in mutual $k$NN alignment was not a result of the limitation of any individual model}.}
    \label{fig:scaling_wit}
        \vspace{-0.8em}
\end{figure}

\begin{figure}[t]
    \centering
    \begin{subfigure}[t]{0.48\linewidth}
        \centering
        \includegraphics[width=\linewidth]{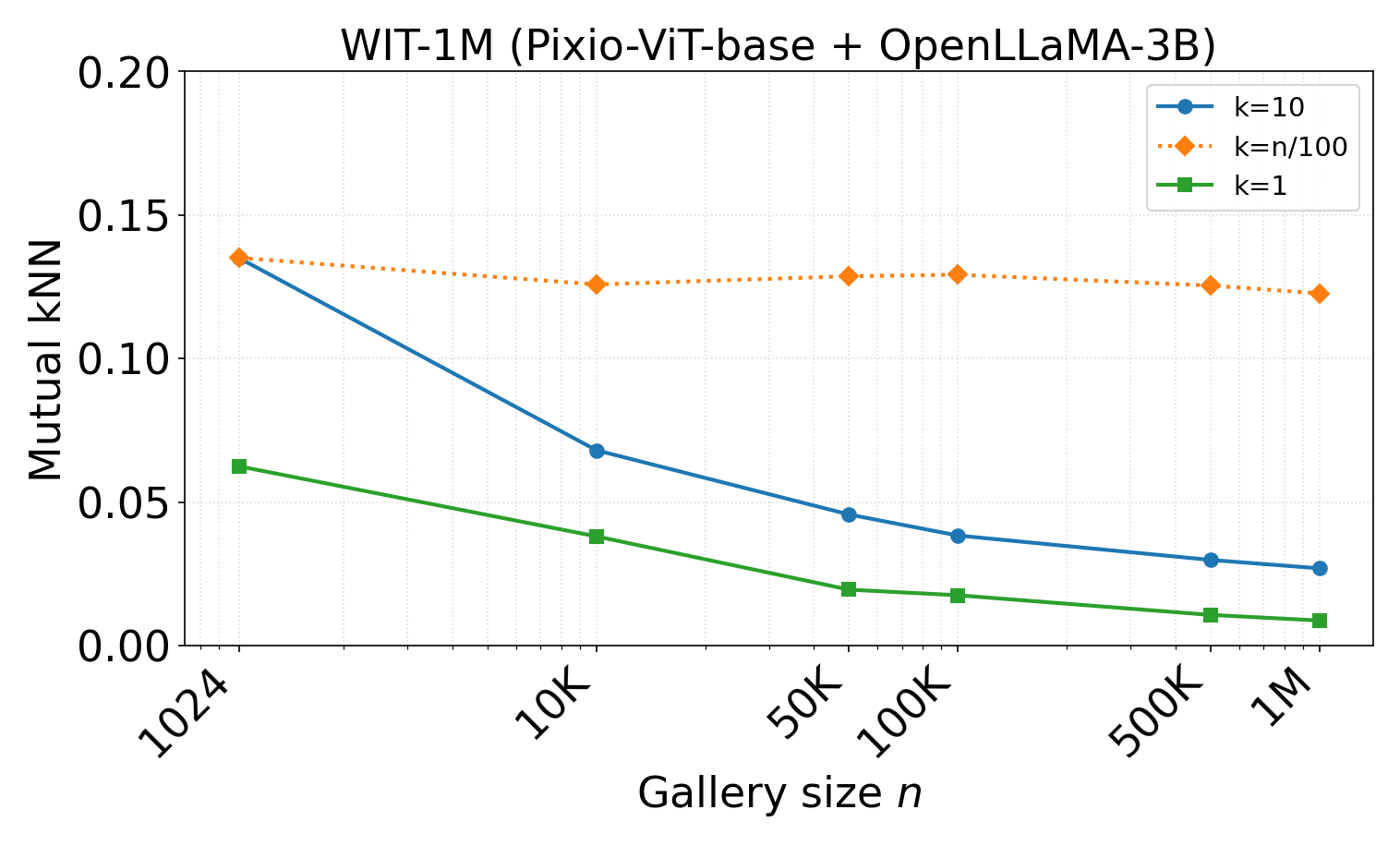}
        \caption{Pixio-ViT-B/16 and OpenLlama-3B}
        \label{fig:scaling_pixio}
    \end{subfigure}
    \hfill
    \begin{subfigure}[t]{0.48\linewidth}
        \centering
        \includegraphics[width=\linewidth]{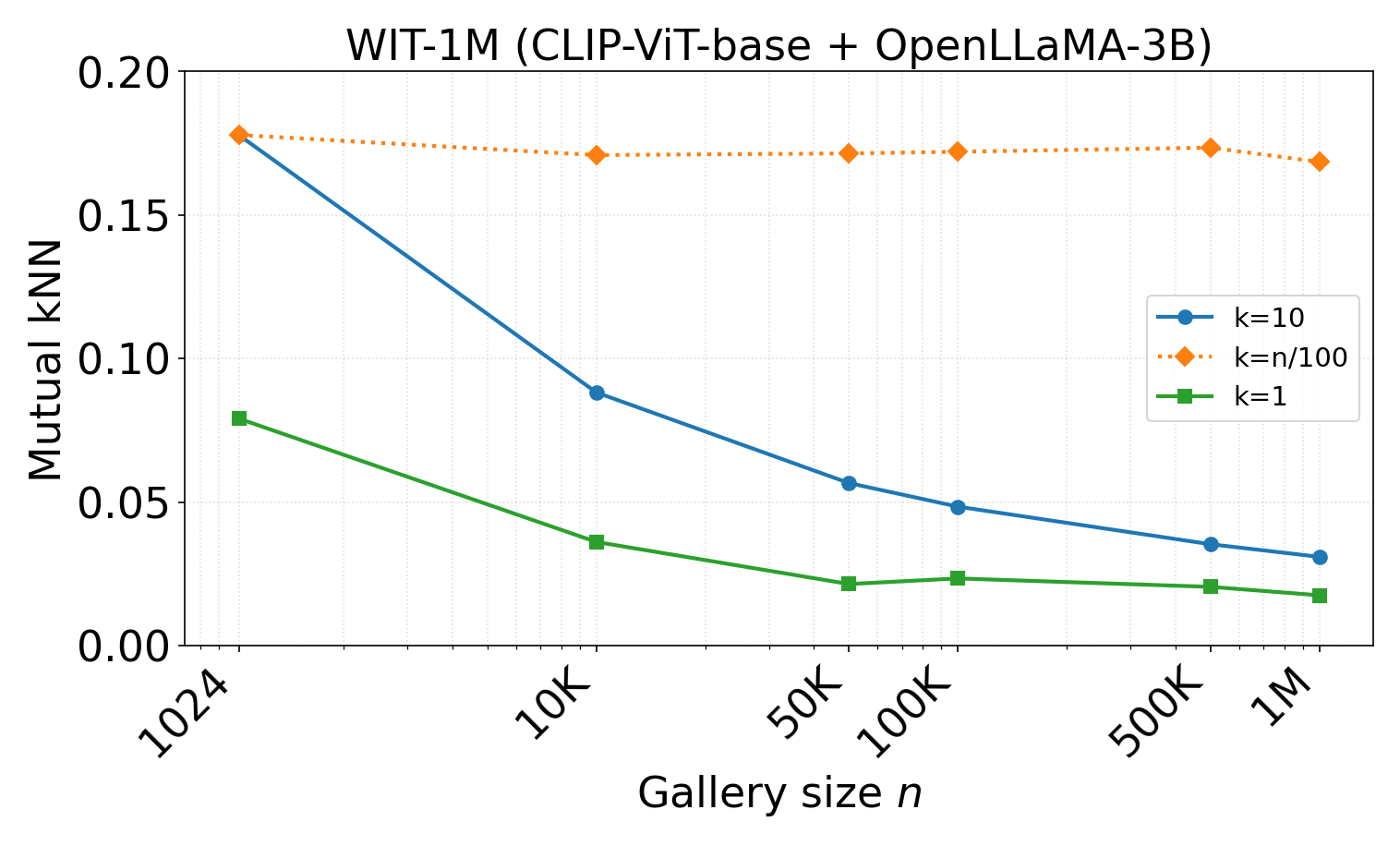}
        \caption{CLIP-B/16 and OpenLlama-3B}
        \label{fig:scaling_clip}
    \end{subfigure}
    \caption{Cross-modal mutual $k$NN alignment as the WIT gallery grows from 1024 to 1M, for vision encoders trained with different objectives (Pixio-style SSL~\citep{yang2026pursuit} and CLIP contrastive image-text pretraining), paired with OpenLlama-3B. \textit{The same scaling pattern holds: alignment at fixed small $k$ drops with gallery size while $k{=}n/100$ stays flat.}}
    \label{fig:scaling_vision_families}
    \vspace{-0.8em}
\end{figure}

\begin{figure}[t]
    \centering
    \begin{subfigure}[t]{0.48\linewidth}
        \centering
        \includegraphics[width=\linewidth]{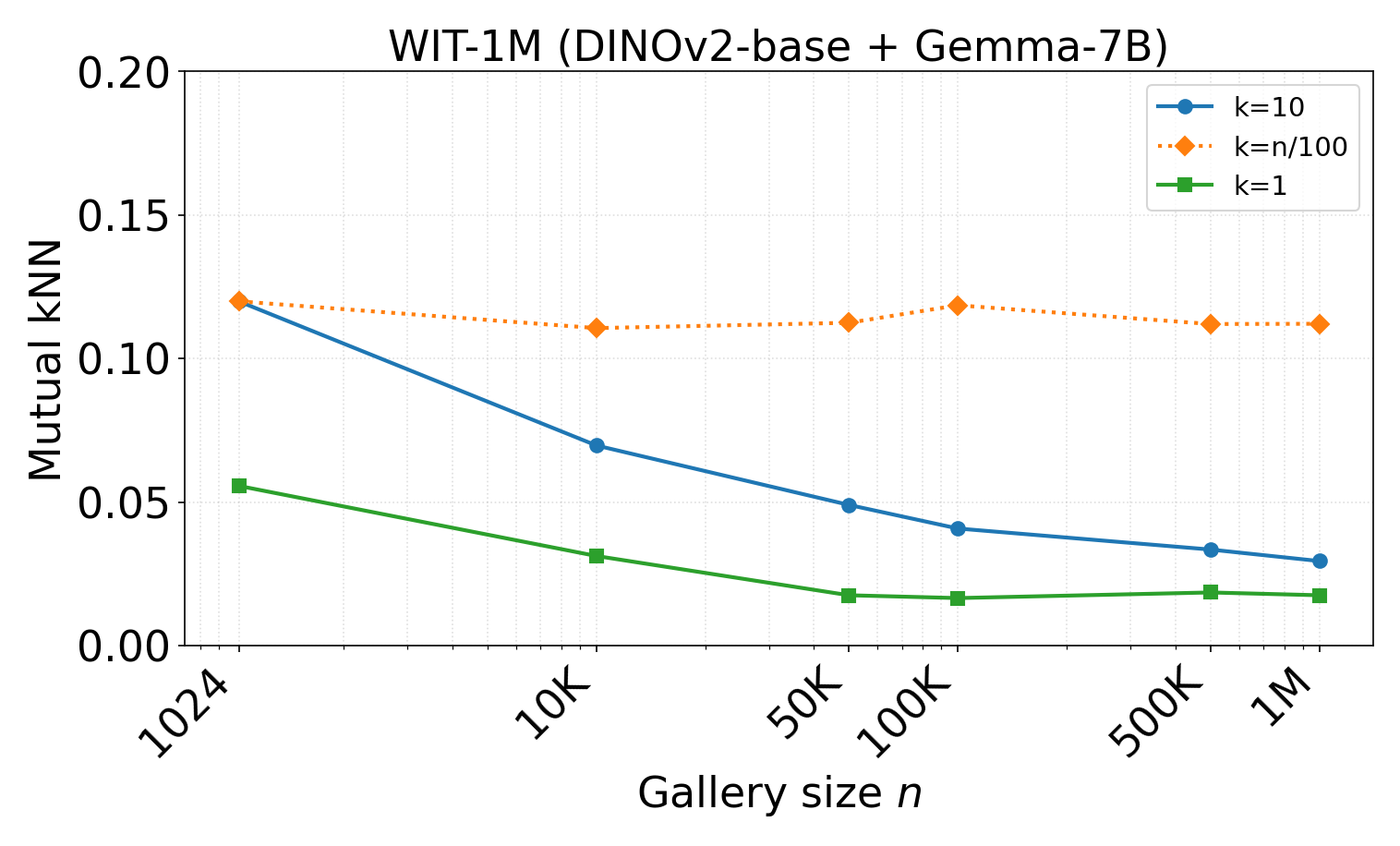}
        \caption{DINOv2-base and Gemma-7B}
        \label{fig:scaling_gemma}
    \end{subfigure}
    \hfill
    \begin{subfigure}[t]{0.48\linewidth}
        \centering
        \includegraphics[width=\linewidth]{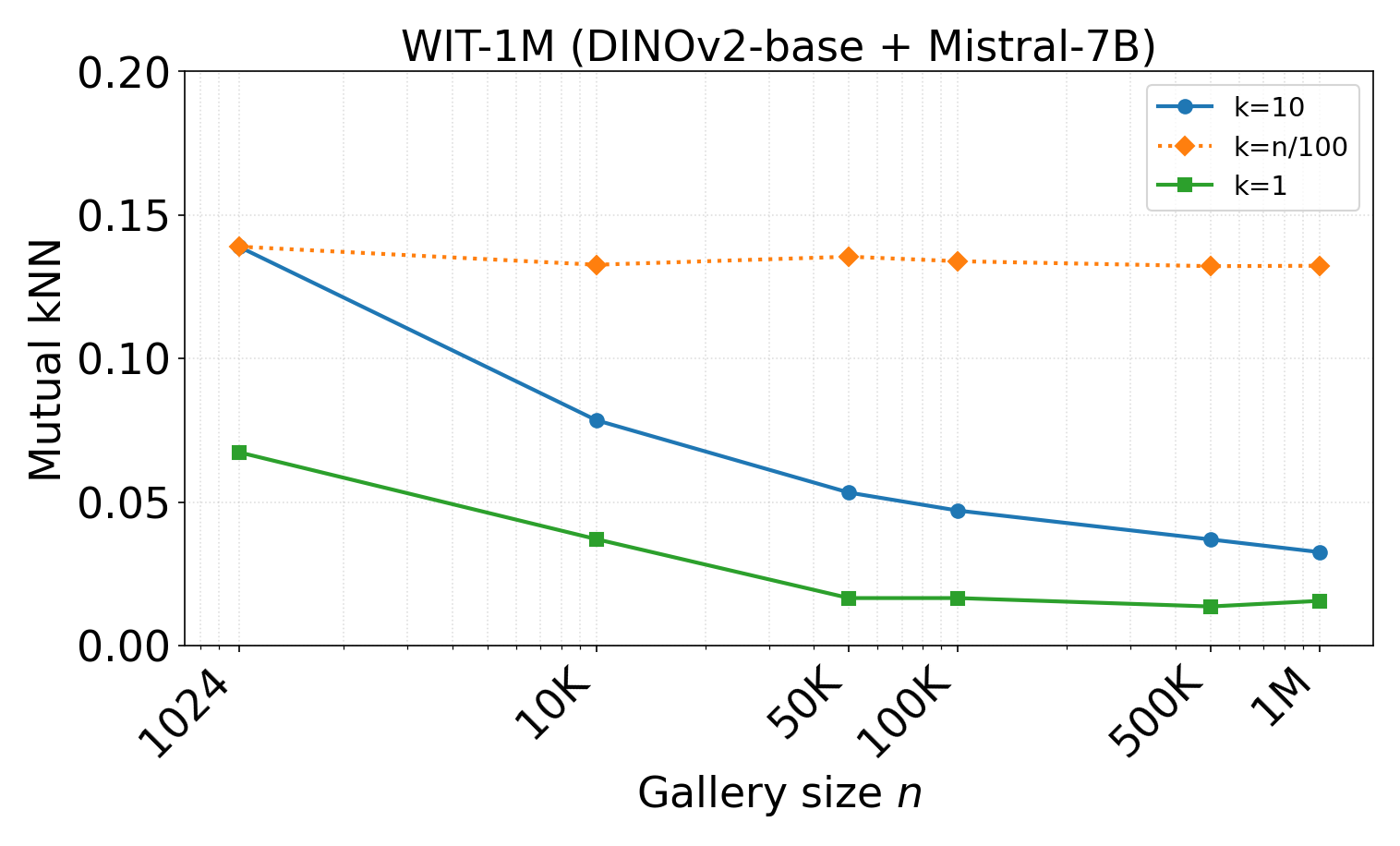}
        \caption{DINOv2-base and Mistral-7B}
        \label{fig:scaling_mistral}
    \end{subfigure}
    \caption{Cross-modal mutual $k$NN alignment as the WIT gallery grows from 1024 to 1M, with DINOv2-base paired with LLMs from different families (Gemma and Mistral). \textit{The scaling pattern is consistent across LLM families: drop at fixed small $k$, stability at $k{=}n/100$.}}
    \label{fig:scaling_llm_families}
\end{figure}
\subsection{Mutual cross-modal $k$NN alignment drops at scale across model pairs}\label{sec:additional_models}
The cross-modal alignment drop reported in Fig.\ 4 in the paper uses DINOv2-base and OpenLlama-3b. Here, we examine whether similar patterns hold for stronger models. In \cref{fig:scaling_wit}, we repeat the scaling experiment for two additional model pairs: DINOv2-base with OpenLlama-13b, and DINOv2-giant with OpenLlama-13b.

Replacing DINOv2-base with the stronger DINOv2-giant and OpenLlama-3b with OpenLlama-13b does not change the pattern. We observe that mutual $k$NN still drops at scale.
This is consistent with Fig. 2b in the paper, which shows low alignment scores across different LLMs at WIT-1M scale.

To test whether the pattern is specific to the DINOv2/OpenLlama choice, we evaluate two further axes of variation. First, we vary the vision encoder while keeping OpenLlama-3B fixed (\cref{fig:scaling_vision_families}): we replace DINOv2 with Pixio-ViT-B/16~\citep{yang2026pursuit} and with CLIP-B/16 (image-text contrastive pretraining). Second, we vary the LLM while keeping DINOv2-base fixed (\cref{fig:scaling_llm_families}): we pair it with Gemma-7B~\citep{gemma_2024} and Mistral-7B~\citep{jiang2023mistral} from different model families.
Across all four combinations, the alignment drop at fixed small $k$ persists, while alignment at $k=\frac{n}{100}$ remains stable.

\textbf{This indicates that the degradation reported in the paper is not specific to that particular choice of models}.

\section{Additional ImageNet experiments}
Here, we verify that the ImageNet validation set used in \cref{sec:diagnosis} in the main paper serves as a suitable test bed. Furthermore, we confirm that our observations are not limited to our choice of models or metric settings.

\subsection{The ImageNet validation set is denser than WIT-1024}
A natural question is how the gallery density in our ImageNet experiments compares to the WIT data. As shown in \cref{tab:nn_quality}, nearest-neighbor cosine similarities on the ImageNet validation set are substantially higher than on WIT-1024 and comparable to WIT-1M. Even for  only one image per class in the gallery (ipc=1), the ImageNet validation set provides a denser retrieval setting than WIT-1024.

\textbf{This confirms that the ImageNet experiments operate in a denser retrieval regime comparable to WIT-1M, making this a meaningful test bed.}

\begin{table*}[t]
  \centering
  \caption{Nearest-neighbor proximity across gallery sizes. ImageNet, even with only one image per class in the gallery (ipc=1), has neighbor proximities comparable to WIT-1M, confirming that it operates in a similarly dense retrieval regime.}\label{tab:nn_quality}
  \setlength{\tabcolsep}{10pt}
   \resizebox{0.5\textwidth}{!}{%
  \begin{tabular}{l@{\hskip 16pt}l@{\hskip 16pt}r@{\hskip 16pt}c@{\hskip 16pt}c}
    \toprule
    Gallery & Model & Dim & $k{=}1$ & $k{=}10$ \\
    \midrule
    WIT-1024       & DINOv2-base     & 768  & 0.799 & 0.717 \\
    WIT-1024       & OpenLlama3b & 3200 & 0.502 & 0.400 \\
    \midrule
    WIT-1M         & DINOv2-base     & 768  & 0.906 & 0.888 \\
    WIT-1M         & OpenLlama3b & 3200 & 0.757 & 0.701 \\
    \midrule
    ImageNet ipc=1  & DINOv2-base     & 768  & 0.823 & 0.763 \\
    ImageNet ipc=1  & DINOv2-giant     & 1536 & 0.609 & 0.496 \\
    ImageNet ipc=1  & OpenLlama3b & 3200 & 0.928 & 0.904 \\
    ImageNet ipc=1  & LLaMA-65B    & 8192 & 0.890 & 0.858 \\
    \midrule
    ImageNet ipc=49 & DINOv2-base     & 768  & 0.887 & 0.861 \\
    ImageNet ipc=49 & DINOv2-giant     & 1536 & 0.749 & 0.690 \\
    ImageNet ipc=49 & OpenLlama3b & 3200 & 0.954 & 0.944 \\
    ImageNet ipc=49 & LLaMA-65B    & 8192 & 0.926 & 0.912 \\
    \bottomrule
    \bottomrule
  \end{tabular}
  }
\end{table*}

\subsection{Ceiling bands: label-free replication (WIT-1024) and the mutual $k$NN metrics}
\label{sec:supp_ceiling_bands_additional}

\begin{figure}[t]
  \centering
  \includegraphics[width=\linewidth]{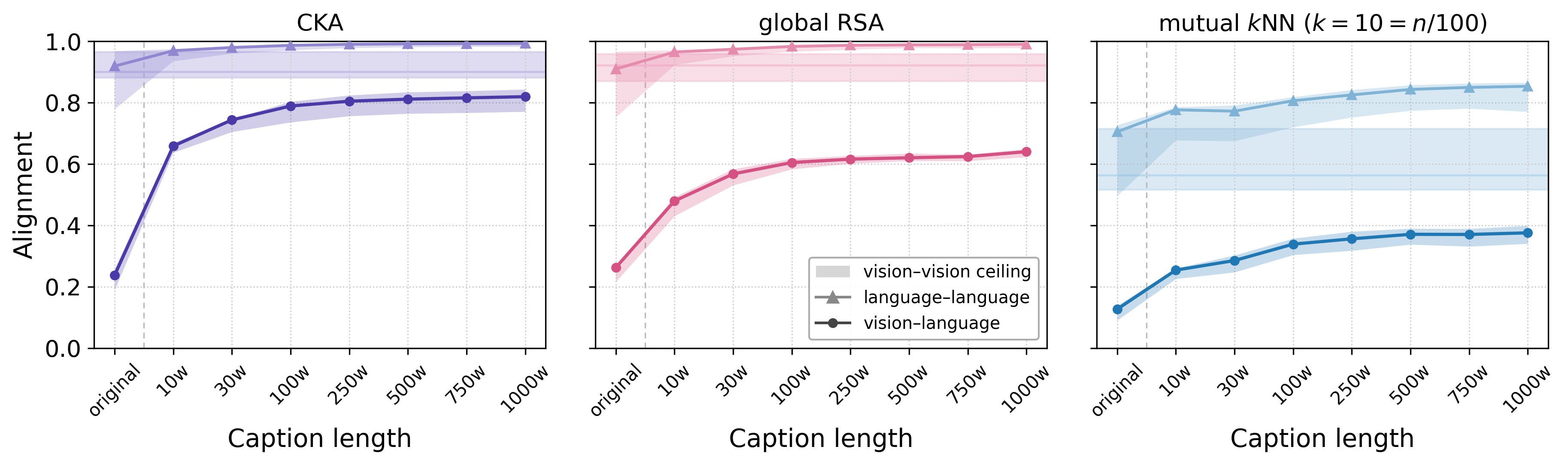}
  \caption{Alignment bands on WIT-1024 across caption types: original
  captions, then Gemini captions of increasing length ($n=1024$; all cells
  calibrated, own 2D layer search; curves are medians over pairs, shading
  the min--max range). Horizontal band: six vision--vision pairs
  (caption-independent). Triangles: four language--language pairs per
  caption type. Circles: twelve validation LLMs against DINOv2-B. The
  mutual $k$NN panel is reference-only at this sample size.}
  \label{fig:wit1024_ceiling_bands}
\end{figure}

\begin{figure}[t]
\vspace{-1em}
  \centering
  \includegraphics[width=0.64\linewidth]{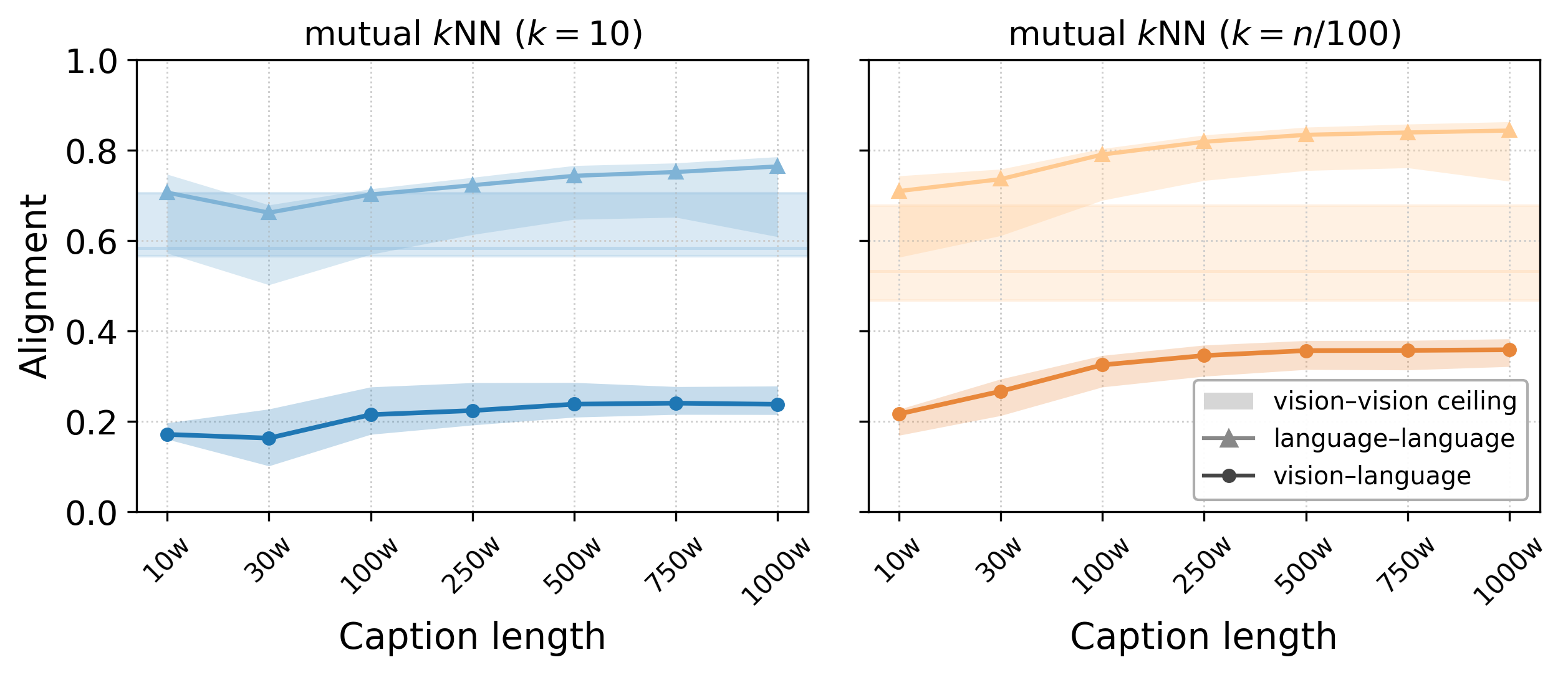}
  \caption{Mutual-$k$NN ceiling bands across caption lengths on the
  ImageNet validation set, the local-metric companion of
  \cref{fig:ceiling_bands} (same cells and calibration; curves are medians
  over pairs, shading the min--max range).}
  \label{fig:ceiling_bands_knn}
\end{figure}

\paragraph{WIT-1024.}
WIT-1024 has no labels, and the ImageNet picture replicates without them
(\cref{fig:wit1024_ceiling_bands}): the language ceiling is flat and high
at every Gemini length (RSA $0.92$--$0.99$) while cross-modal alignment
saturates by ${\sim}250$ words far below it (RSA $0.48 \to 0.64$, CKA
$0.66 \to 0.82$). The original WIT-1024 captions sit below every Gemini length and are the one
caption type that drags down the language ceiling itself (cross-family
pair $0.75$): noise in the observation, not in the models. Since the calibrated re-examination of \citet{groger2026revisitingplatonicrepresentationhypothesis} evaluates on these original terse captions, the alignment it can measure is capped by the observation before calibration enters. On the same data, cross-modal alignment sits substantially higher once captions are rich.

\paragraph{The local metric.}
The mutual-$k$NN bands (\cref{fig:ceiling_bands_knn} for ImageNet, right
panel of \cref{fig:wit1024_ceiling_bands} for WIT-1024) repeat the
cross-modal pattern of the global metrics, with two properties specific to
reading top-$k$ neighborhoods. The vision--vision band is wide: local
agreement is graded by model scale and family rather than forming a single
ceiling, with same-family pairs at the top and small-model and
cross-family pairs at the bottom. And the language--language band lies
above the vision band: LLMs reading the same captions share their local
neighborhoods more than two differently trained vision backbones share
theirs.

\paragraph{Neighborhood composition.}
The band width is set by what neighborhoods are made of. Gallery size is irrelevant (vision--vision overlap is flat from
$1$k to $49$k gallery images), but replaying the WIT protocol on ImageNet
with one image per class reproduces the wide WIT band almost cell for cell
($0.48$--$0.68$), while twenty images per class compress it to
$0.73$--$0.82$. With class siblings available every model fills the
top-$10$ with siblings and the pairs agree; without them the metric reads
fine cross-item distinctions, where vision pairs separate by scale and
family. None of this affects the cross-modal comparison, which stays far
below both bands at every caption length.

\textbf{The ceiling-band structure also emerges without using class labels. For WIT-1024 the
language ceiling stays at $0.92$--$0.99$ at every caption length while
cross-modal alignment saturates at $0.64$. The wide vision m-$k$NN band is
neighborhood composition: one-image-per-class ImageNet reproduces it, and
class-sibling density restores a tight band.}

\subsection{Cross-modal agreement lives in the top three eigenmodes}
\label{sec:eigenspace_threshold}

\begin{figure}[t]
  \centering
  \includegraphics[width=0.44\linewidth]{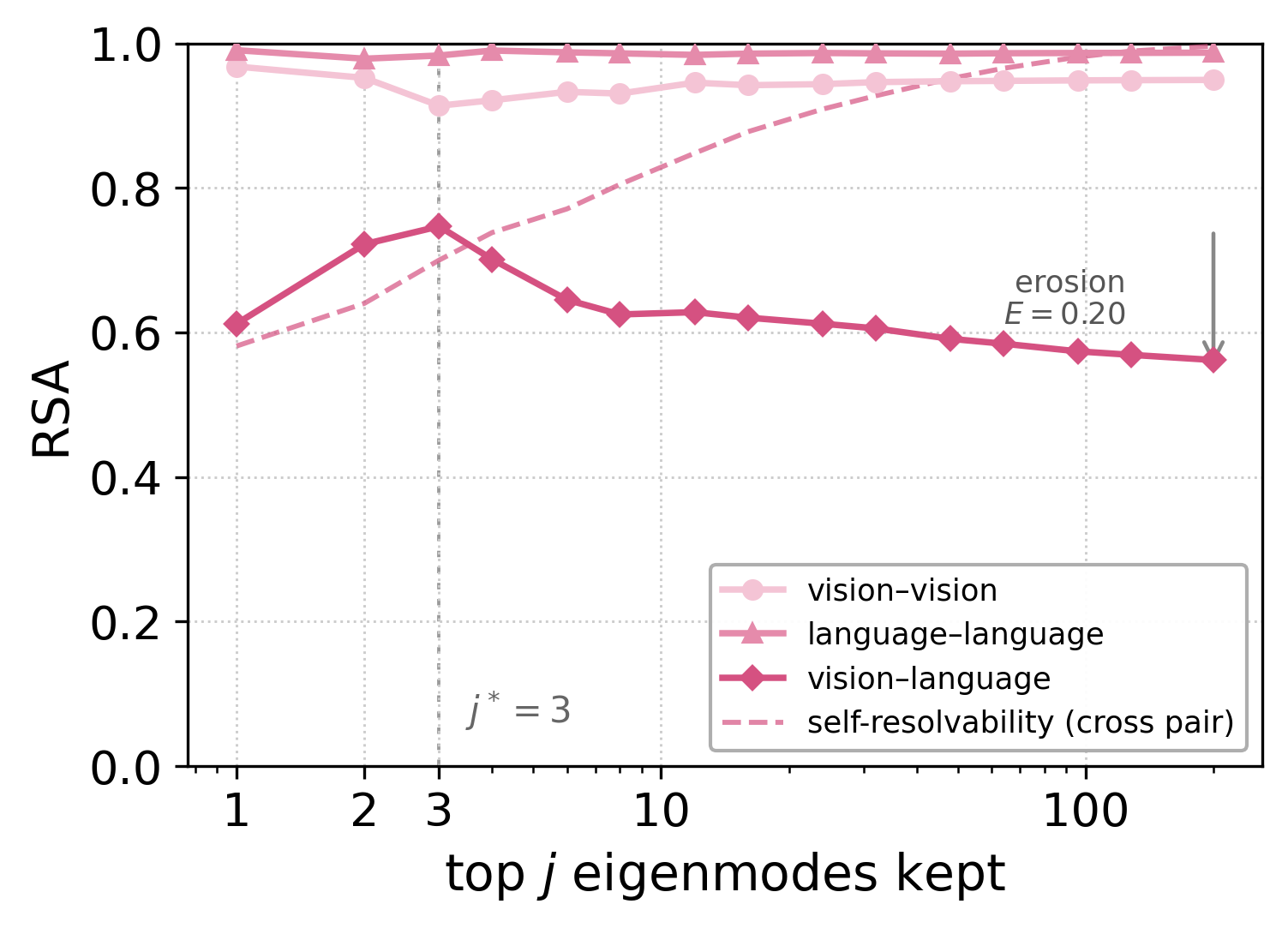}
  \hfill
  \includegraphics[width=0.44\linewidth]{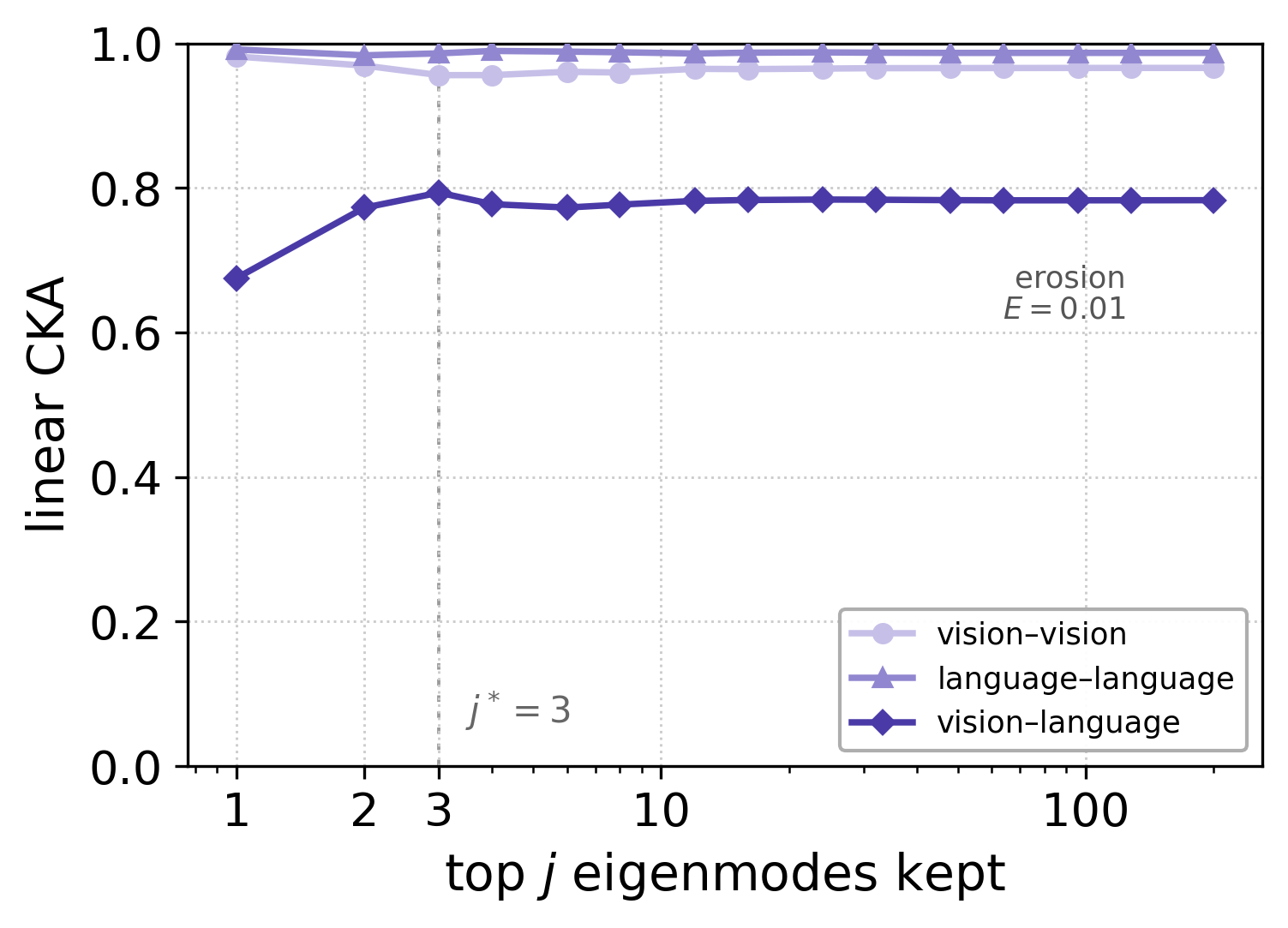}
  \caption{Truncated-rank RSA (left) and truncated-rank CKA (right) between
  the top-$j$ eigenmode reconstructions of both encoders (instance level,
  500-word captions, balanced $n=5000$; RSA-optimal layers). The
  cross-modal curve peaks at $j^\ast = 3$ and erodes under RSA
  ($E = 0.20$) but not under CKA ($E = 0.01$); unimodal pairs erode under
  neither.
  Dashed: self-resolvability of the cross pair, the Spearman
agreement between the pairwise distances of each side's truncation and
those of its own full representation (averaged over the two sides), which
keeps rising where the cross-modal curve falls.}
  \label{fig:eigenspace_threshold}
\end{figure}

Section 4 reports the truncated-rank analysis locating cross-modal agreement in the top three eigenmodes (Fig.~\ref{fig:eigenmodes_main}). Here we give the formal setup, the self-resolvability control, the CKA companion, and the population-level statistics.

CKA yields much higher cross-modal scores than RSA. We can measure the source of this difference explicitly.

Let $X \in \mathbb{R}^{n \times
D_X}$ and $Y \in \mathbb{R}^{n \times D_Y}$ be the column-centered feature
matrices of two encoders on the same $n$ samples, with PC variances
$\lambda_i$ and score directions $u_i \in \mathbb{R}^n$. The score
directions live in the shared sample space, so mode cosines
$u_i^{X\top} u_j^{Y}$ are defined across modalities even though
$D_X \neq D_Y$. Linear CKA weights the agreement of a mode pair by its
variance product, $\mathrm{CKA} = \sum_{ij} \hat\lambda_i^X
\hat\lambda_j^Y (u_i^{X\top} u_j^{Y})^2$ with $\hat\lambda =
\lambda/\|\lambda\|_2$, so its mass concentrates
on the top few modes. Rank-based RSA instead counts every pair-order
decision equally, and near-tied orderings are decided by the low-variance
tail. To locate where cross-modal agreement lives, we truncate both sides
to their top $j$ modes and track $\rho_{\times}(j)$, the Spearman
correlation between the two models' pairwise cosine-distance vectors
computed from the truncations $X_j$ and $Y_j$, with
$\rho_{\mathrm{self}}(j)$ the same statistic between $X_j$ and the full
$X$. How much of a model's own final ordering do its top $j$ modes already
determine? We report the peak $j^\ast =
\arg\max_j \rho_{\times}(j)$ and the erosion $E = \rho_{\times}(j^\ast) -
\rho_{\times}(\infty)$ (\cref{fig:eigenspace_threshold}).

At the RSA-optimal instance layers (500-word captions, balanced $n = 5000$),
the cross-modal pair (DINOv2-B $\times$ OpenLLaMA-3B) peaks at $j^\ast = 3$. The top
three eigenmodes carry $27\%$/$31\%$ of the two sides' variance and
$80\%$ of the CKA weight. Adding the
remaining modes erodes the agreement, down to $0.55$ at full rank
($E = 0.20$, \cref{fig:eigenspace_threshold} left). The tail modes are not
noise: self-resolvability is only $0.71$/$0.69$ at $j^\ast$ and keeps
rising, so the modes added beyond the threshold carry real structure that
each model orders consistently and the other modality never matches.

On the identical truncations CKA hardly changes
(\cref{fig:eigenspace_threshold} right). Truncated CKA peaks at the same
$j^\ast = 3$ and erodes by $0.01$ ($0.00$ on centroids), the measured
form of the weighting asymmetry.

The unimodal controls show negligible erosion at the same layers and sample with $E = 0.02$ for vision $\times$ vision and $E = 0.003$ for language $\times$
language, their curves within $0.05$ of their full-rank value at every $j$
and above the cross-modal curve even at its peak. The peak-then-erode
shape is a property of the cross-modal pair, not of the truncation.
The pattern observed in \cref{fig:eigenmodes_main}
is population-level, not a property of one pair: across all twelve
validation LLMs, $j^\ast = 3$ in $23$ of $24$ cells at both granularities,
with instance-level erosion $0.10$--$0.25$ and centroid-level erosion
$0.05$--$0.08$ throughout.

\textbf{Cross-modal rank agreement peaks at $j^\ast = 3$ eigenmodes and
erodes by $0.20$ as the remaining modes are added. On the same truncations
CKA erodes by $0.01$, and unimodal pairs by at most $0.02$ in either
metric. CKA reads the peak, RSA captures the erosion that lives in the fine-grained structure.}

\section{Alignment metrics and calibration}
\label{sec:supp_metrics}

The mutual $k$NN metric of \cref{sec:experiments} treats every sample as both a query and a gallery item: the neighborhoods $\mathcal{N}^\mathbf{a}_k(i)$ and $\mathcal{N}^\mathbf{b}_k(i)$ are retrieved from the same $n$ points that are scored. When we densify a gallery, either by adding samples (\cref{sec:alignment_across_scales,sec:text_audio_video}) or by adding valid correspondences per sample (\cref{sec:no-bijecton-experriments}), this couples two effects: the neighborhoods of a fixed query change, and the set of queries being averaged over changes as well. To separate the two, we use a \emph{query-to-gallery} (Q2G) variant of the metric that keeps the query set fixed while the gallery grows. All results in \cref{fig:densifying_modalities,fig:audio_video_scaling,fig:ipc_invariance} are computed with this protocol.

\myparagraph{Setup.}
Unless a metric is rank-based, features are prepared following
\cite{huh2024prh}: per-dimension outlier clamping at the $q{=}0.95$
quantile, then L2 normalization. CKA and mutual $k$NN require this
step. RSA operates on ranks and is unaffected by it.

\myparagraph{Query and Gallery.} Let $\mathcal{Q}$ be a fixed query set of $q$ samples and $\mathcal{G}\supseteq\mathcal{Q}$ a gallery of $n$ samples, with features $\mathbf{a}_i\in\mathbb{R}^{d_1}$ and $\mathbf{b}_i\in\mathbb{R}^{d_2}$ from the two models for every $i\in\mathcal{G}$. Galleries are nested, $\mathcal{G}_1\subset\mathcal{G}_2\subset\cdots$, and each contains the queries. Feature preparation (0.95-quantile clamping followed by L2 normalization, \cref{sec:supp_metrics}) is performed once over the full pool rather than per gallery, so the geometry of the two spaces is fixed and only the number of gallery candidates changes between levels. In the scaling experiments of \cref{sec:text_audio_video}, $\mathcal{Q}$ is a fixed random subset of $1$k samples and $\mathcal{G}$ grows from $1$k to $100$k. In the bijection experiments of \cref{sec:no-bijecton-experriments}, $\mathcal{Q}$ holds one correspondence per source sample (one caption per image for I2T, one image per caption for T2I) and $\mathcal{G}$ adds the remaining correspondences one at a time.

\myparagraph{Mutual $k$NN (Fixed $k$).}
For every query $i\in\mathcal{Q}$, each model retrieves its $k$ nearest neighbors from the gallery, excluding the query itself:
$$\mathcal{N}^\mathbf{a}_k(i) = \operatorname{argtopk}_{j \in \mathcal{G}\setminus\{i\}} \mathbf{a}_i^\top \mathbf{a}_j, \qquad \mathcal{N}^\mathbf{b}_k(i) = \operatorname{argtopk}_{j \in \mathcal{G}\setminus\{i\}} \mathbf{b}_i^\top \mathbf{b}_j,$$
and the score is the mean per-query overlap,
$$s_{k} = \frac{1}{q}\sum_{i\in\mathcal{Q}} \frac{|\mathcal{N}^\mathbf{a}_k(i) \cap \mathcal{N}^\mathbf{b}_k(i)|}{k}.$$
For $\mathcal{Q}=\mathcal{G}$ this is exactly the mutual $k$NN score of \cref{sec:experiments}. As in the main paper, we report $k{=}1$, $k{=}10$, and the gallery-proportional $k{=}\frac{n}{100}$, where $n$ is the size of the current gallery, while query size stays fixed 1000 or 1024 samples (\cref{sec:alignment_across_scales}). Because the queries are fixed, any change of $s_k$ across gallery levels is due to the neighborhoods of the same queries being filled by closer candidates, not to a change in which samples are being averaged.

\myparagraph{Mutual $k$NN (Variable $k$).}
Fixed-$k$ scores probe a progressively finer granularity as $n$ grows, since $\frac{k}{n}$ shrinks. As a density-matched control, we let each query define its own neighborhood radius. For a reference size $k_{\text{ref}}$, the threshold of query $i$ in each space is its similarity to its $k_{\text{ref}}$-th nearest \emph{other query},
$$\beta^\mathbf{a}_i = \operatorname{topk}^{(k_{\text{ref}})}_{j\in\mathcal{Q}\setminus\{i\}} \mathbf{a}_i^\top\mathbf{a}_j, \qquad \beta^\mathbf{b}_i = \operatorname{topk}^{(k_{\text{ref}})}_{j\in\mathcal{Q}\setminus\{i\}} \mathbf{b}_i^\top\mathbf{b}_j,$$
where $\operatorname{topk}^{(k)}$ denotes the $k$-th largest value. The neighborhoods are all gallery items within that radius,
$$\mathcal{A}_i = \{\, j\in\mathcal{G}\setminus\{i\} : \mathbf{a}_i^\top\mathbf{a}_j \geq \beta^\mathbf{a}_i \,\}, \qquad \mathcal{B}_i = \{\, j\in\mathcal{G}\setminus\{i\} : \mathbf{b}_i^\top\mathbf{b}_j \geq \beta^\mathbf{b}_i \,\},$$
and, since $|\mathcal{A}_i|$ and $|\mathcal{B}_i|$ are no longer equal, the score is the mean intersection-over-union,
$$s_{\text{var}} = \frac{1}{q}\sum_{i\in\mathcal{Q}} \frac{|\mathcal{A}_i \cap \mathcal{B}_i|}{|\mathcal{A}_i \cup \mathcal{B}_i|}.$$
The thresholds depend only on $\mathcal{Q}$, so they stay fixed while the gallery grows. Adding gallery samples then enlarges both neighborhoods at a rate set by the local density of each space, rather than shrinking the neighborhood fraction as fixed $k$ does. On the smallest gallery ($\mathcal{G}=\mathcal{Q}$) each neighborhood contains exactly $k_{\text{ref}}$ items and $s_{\text{var}}$ is a monotone rescaling of $s_{k_{\text{ref}}}$; the two statistics only diverge once the gallery is denser than the query set. We use $k_{\text{ref}}\in\{1, 10, \frac{q}{100}\}$, matching the fixed-$k$ settings.

\myparagraph{RSA.}
Global RSA is the Spearman correlation between the two models'
condensed pairwise cosine-distance vectors over the same samples
(up to $2{,}000$ samples in the images-per-class sweep. The
caption-ladder cells use a canonical class-balanced pool, $5$ images
per class for the layer search and $15$ per class for the reported
score). Within-class RSA computes the same correlation separately
inside each ImageNet class and averages over classes. The layer search
uses a canonical $150$-class, $30$-images-per-class sample and the
reported score covers all $1{,}000$ classes at the selected layers.

\myparagraph{CKA.}
The images-per-class sweep uses linear CKA on prepared
features~\citep{kornblith2019similarity}. We use RBF CKA on centered, normalized Gram matrices with the median heuristic bandwidth computed per modality, on up to $4{,}000$ samples.

\myparagraph{Layer search and calibration.}
Every model pair receives its own 2D layer search (all vision layers
$\times$ all LLM layers). Each reported score $s$ is calibrated
against a pairing-shuffle null: captions are randomly re-assigned to
images, the statistic is recomputed, and the null is made
search-aware by re-running the full layer-grid maximum per
permutation. With $\tau$ the null's upper $\alpha{=}0.05$ quantile
over $K=200$ permutations, we report
$s_{\mathrm{cal}}=(s-\tau)/(1-\tau)$
\citep{groger2026revisitingplatonicrepresentationhypothesis}, so
chance-level structure reads as zero.

\subsection{Duplicated feature rows in many-to-many data}
\label{sec:q2g_duplicates}

The many-to-many datasets are stored as one row per correspondence. If a caption is paired with $L$ images, its text feature occurs in $L$ bit-identical rows while the corresponding image features differ. Conversely, when an image is paired with $L$ captions, its image feature is repeated while the text features differ
(\cref{fig:surjection_example}).

A naive row-level implementation of mutual $k$NN is unsuitable in this setting. In the repeated modality, several copies of the same feature can occupy multiple positions in a neighborhood, e.g. feature's closest neighbors are bit-identical copies. The other modality distinguishes these rows and may instead retrieve several distinct correspondences.

For each modality $\mathbf{m}\in\{\mathbf{a},\mathbf{b}\}$, let $\mathcal{R}^{\mathbf{m}}\subseteq\mathcal{G}$ contain one representative, the first occurrence of every bit-identical feature group, and let $r^{\mathbf{m}}(i)$ be the representative of the group containing row $i$. We also let $\pi(j)$ denote the source group of correspondence row $j$: the shared caption in the T2I setting or the shared image in the I2T setting. For bijective data, every row forms its own source group and $\pi(j)=j$.

\myparagraph{Fixed $k$.}
Before fixed-$k$ retrieval, each modality's gallery is reduced to its unique representatives. The query's duplicate group is excluded by removing $r^{\mathbf{m}}(i)$, and the neighborhood is
\[
\widetilde{\mathcal{N}}^{\mathbf{m}}_k(i)
=
\operatorname{argtopk}_{
r\in\mathcal{R}^{\mathbf{m}}\setminus\{r^{\mathbf{m}}(i)\}
}
\mathbf{m}_i^\top\mathbf{m}_r.
\]
This prevents repeated copies of the same feature from occupying several of the $k$ positions. In the modality whose rows are already distinct, the gallery is unchanged. Alignment is then computed using the standard fixed-$k$ neighborhood-overlap formula. This is the procedure used in the bijection experiments of \cref{sec:no-bijecton-experriments}.

\myparagraph{Variable $k$.}
For variable-$k$, collapsing the modalities independently before thresholding would define their neighborhoods over different subsets of correspondence rows. We therefore apply the similarity threshold to the original rows and map the selected rows to their common source groups afterward:
\[
\mathcal{N}^{\mathbf{m}}_i
=
\left\{
\pi(j):
j\in\mathcal{G}\setminus\{i\},
\quad
\mathbf{m}_i^\top\mathbf{m}_j
\geq
\beta_i^{\mathbf{m}}
\right\},
\]
where $\beta_i^{\mathbf{m}}$ is the query-specific threshold defined by the variable-$k$ protocol.

A source group is included if any of its correspondence rows lies within the query's similarity radius. Mapping rows to source groups prevents repeated copies from being counted multiple times. Only the query row itself is excluded; another correspondence from the same source remains a valid neighbor.

Because the two neighborhoods may have different cardinalities, their overlap is measured using intersection-over-union:
\[
s_{\mathrm{var}}
=
\frac{1}{|\mathcal{Q}|}
\sum_{i\in\mathcal{Q}}
\frac{
\left|\mathcal{N}^{\mathbf{a}}_i\cap\mathcal{N}^{\mathbf{b}}_i\right|
}{
\left|\mathcal{N}^{\mathbf{a}}_i\cup\mathcal{N}^{\mathbf{b}}_i\right|
}.
\]

\section{What happens with non-synthetic data that is not bijective?}\label{sec:wit_bijection}

In the main paper (\cref{sec:no-bijecton-experriments}), we use the CycleReward dataset~\citep{bahng2025cycle} to test alignment when the bijective (one-to-one) assumption is relaxed with \emph{synthetic} multi-modal correspondences. Here, we complement this analysis using non-synthetic many-to-many correspondences from the WIT dataset~\citep{srinivasan2021wit}.

\subsection{Non-synthetic dataset with many-to-many correspondences}

\noindent \textbf{Natural duplicates in WIT.}
The WIT dataset naturally contains many-to-many correspondences
between images and captions.

The same caption can describe many visually distinct images, and the same image is reused across Wikipedia articles with different corresponding text. Specifically, 7.1\% of the captions are associated with more than one image, and 24.6\% of the images have more than one caption before deduplication (see \cref{sec:wit_laion}).
These naturally occurring one-to-many and many-to-one
correspondences provide a complementary test bed for relaxing the
bijective (one-to-one) setting without relying on generated images or captions.

\noindent \textbf{Within-group deduplication.}
Grouping by caption text (for T2I) or by image (for I2T) can include \emph{within-group} duplicates, i.e.\ a caption group may contain duplicate images, and an image group may contain repeated captions.
After within-group deduplication, the number of qualifying one-to-many samples decreases from 7{,}844 to 4{,}975 for T2I and from 38{,}254 to 24{,}853 for I2T (\cref{tab:wit_dup_stats}).

\begin{wraptable}{r}{0.37\textwidth}
  \centering  
  \caption{Natural duplicates in the WIT dataset after within-group deduplication.}\label{tab:wit_dup_stats}
  \vspace{-0.5em}
  \setlength{\tabcolsep}{4pt}
  \small
    \resizebox{0.25\textwidth}{!}{
  \begin{tabular}{lcc}
    \toprule
     & \textbf{T2I} & \textbf{I2T} \\
    \midrule
    Unique elements     & 3.2\,M & 2.5\,M \\
    Appearing $>$1 time & 7.1\%  & 24.6\% \\
    \midrule
    \multicolumn{3}{l}{\textit{Groups with $\geq$5 corresp.}} \\
    \quad Before dedup  & 7{,}844 & 38{,}254 \\
    \quad After dedup   & 4{,}975 & 24{,}853 \\
    \bottomrule
    \bottomrule
  \end{tabular}
  }
\end{wraptable}
  \vspace{-1em}
We construct two complementary one-to-many datasets to mirror the experimental setup in \cref{sec:no-bijecton-experriments} of the paper:

\noindent  \textit{T2I (text-to-images):} We select all 4{,}975 captions that are associated with at least 5 unique images. For each caption, we take 5 images, yielding a flat dataset of 24{,}875 image-text pairs.

\noindent  \textit{I2T (image-to-texts):} We identify 24{,}853 unique images that are paired with at least 5 distinct captions (by exact string matching). To match the T2I dataset size, we randomly subsample 4{,}975 images. For each image, we take 5 captions, again yielding 24{,}875 samples.

\subsection{Mutual $k$NN also decreases on non-synthetic data when the bijective assumption is relaxed}
\label{sec:app-many2many-wit}

\begin{figure}[h]
    \centering
    \begin{subfigure}[t]{0.48\linewidth}
        \centering
        \includegraphics[width=\linewidth]{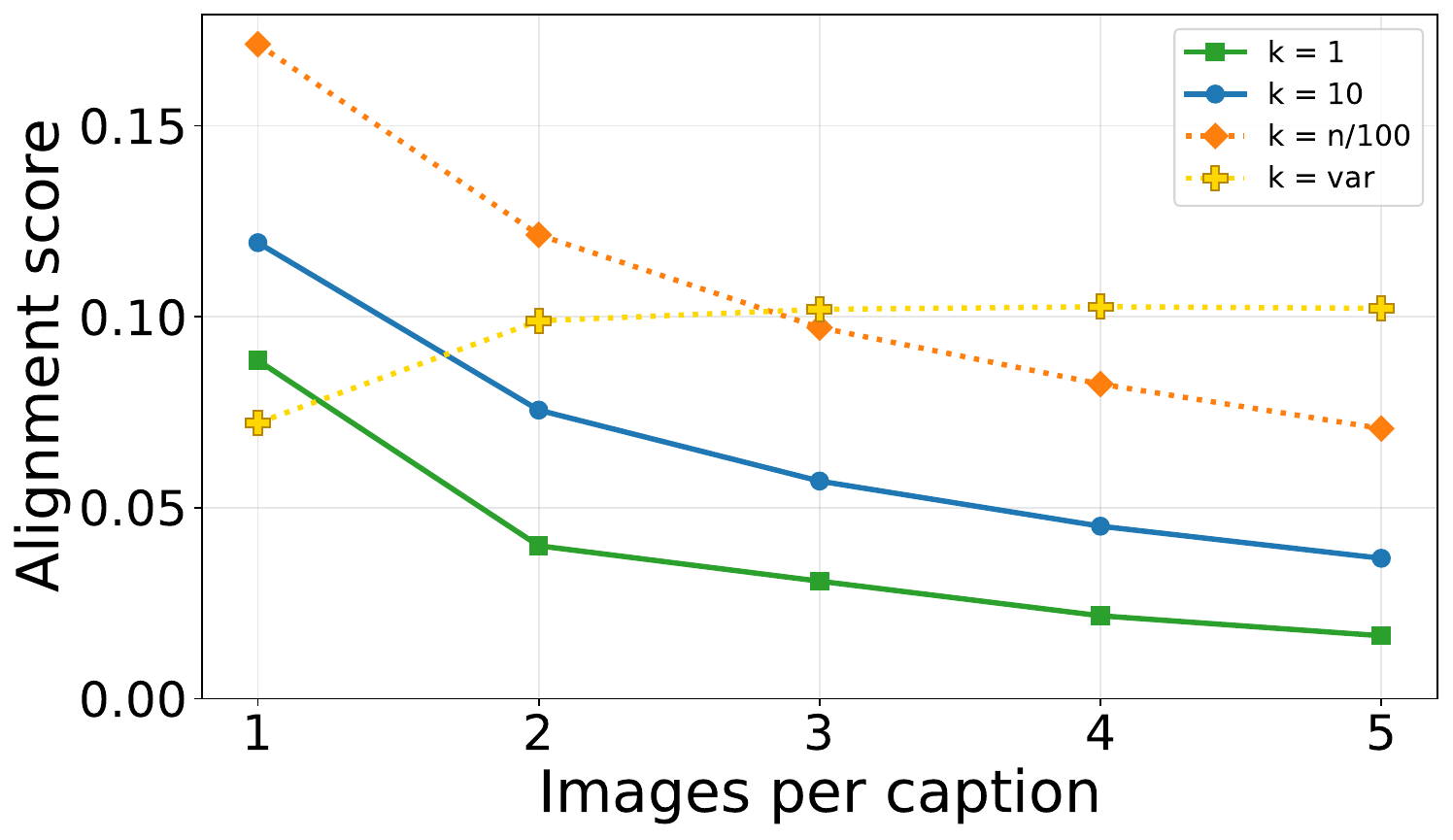}
        \caption{T2I: increasing the number of unique images per caption from 1 (bijective) to 5.}\label{fig:wit_bij_t2i}
    \end{subfigure}
    \hfill
    \begin{subfigure}[t]{0.48\linewidth}
        \centering
        \includegraphics[width=\linewidth]{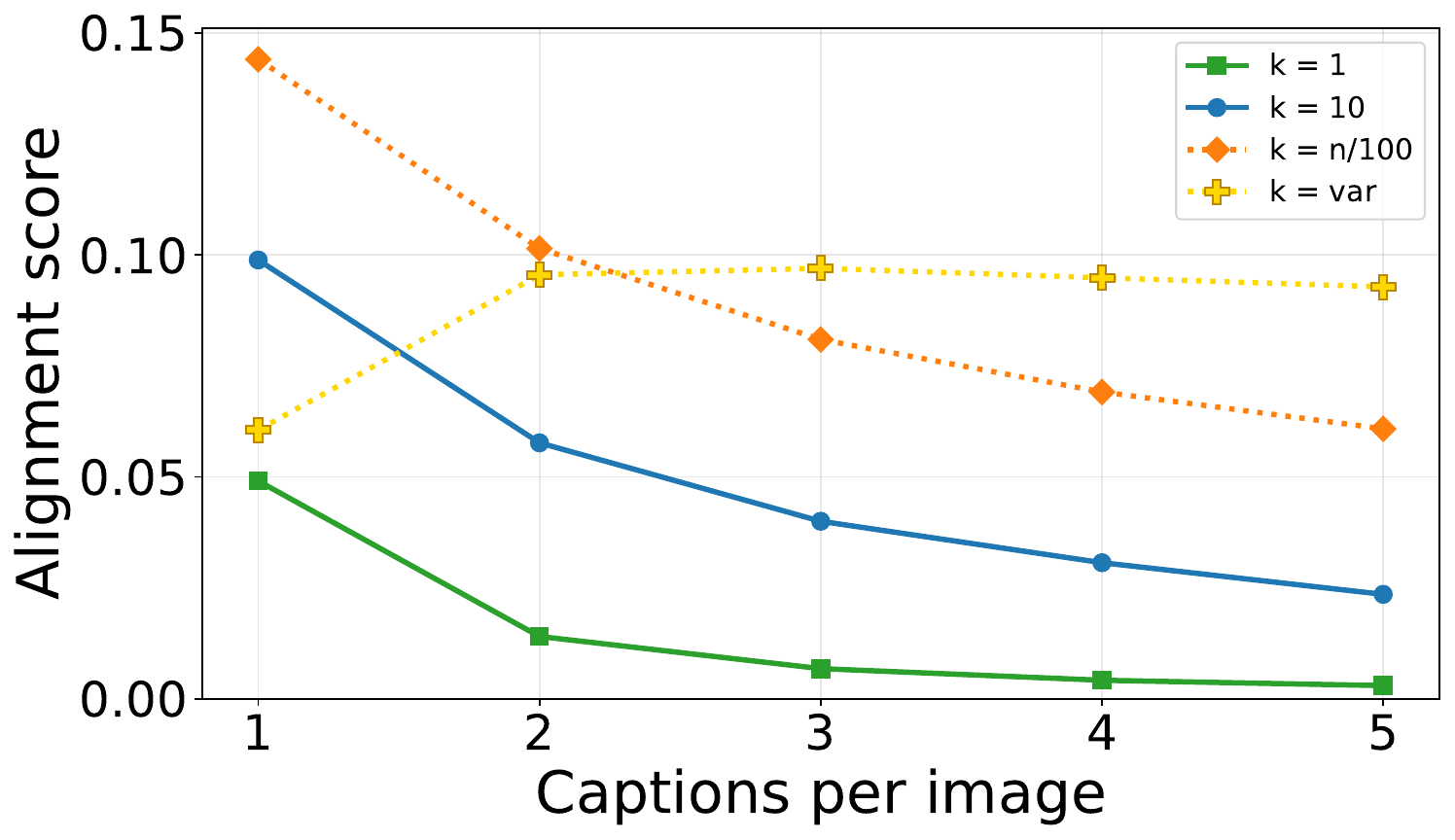}
        \caption{I2T: increasing the number of unique captions per image from 1 (bijective) to 5.}\label{fig:wit_bij_i2t}
    \end{subfigure}
    \caption{Effect of relaxing the bijective assumption on mutual $k$NN alignment using many-to-many correspondences in the WIT data (I2T and T2I subset). \textit{Mutual $k$NN alignment on non-synthetic data drops consistently as we increase the number of images per caption and vice versa. This confirms that the pattern observed on CycleReward is not an artifact of synthetic data.}}
    \label{fig:wit_bijection}
\end{figure}

We evaluate alignment between DINOv2-base~\citep{oquab2024dinov2learningrobustvisual} and OpenLlama-3b~\citep{openlm2023openllama} on the WIT-based T2I and I2T datasets. \cref{fig:wit_bijection} shows mutual $k$NN alignment for $k{=}1$ and $k{=}10$ as the number of images per caption and vice versa increases from 1 to 5.

In both directions, alignment decreases as bijectivity is relaxed. This is consistent with the results on the CycleReward dataset in the main paper (\cref{fig:densifying_modalities}) and reinforces the conclusion that mutual $k$NN alignment is sensitive to the bijective assumption. When multiple valid correspondences exist for a query, the two modalities are less likely to agree on the same nearest neighbor, even if each individually retrieves a good match. \textbf{This confirms that the observed drop in alignment for the non-bijective setting is not an artifact of \textit{synthetic data}.}

\section{Is the many-to-many drop a property of the mutual $k$NN metric?}
\label{sec:bijection_global_metrics}

In \cref{sec:no-bijecton-experriments}, we observed that mutual $k$NN alignment decreases consistently as the one-to-one assumption is relaxed, and noted that the metric cannot distinguish between genuine misalignment and many-to-many correspondence. Here, we separate the two. We repeat the CycleReward densification experiments with two global metrics, RSA and linear CKA (\cref{sec:supp_metrics}), on the identical feature pairs (DINOv2-base and OpenLlama-3b). If the drop reflected genuinely weaker alignment, i.e.\ a change in how the two models organize the densified data relative to one another, global measures of the pairwise structure should register it. If it is instead a property of the neighborhood-based reading, they should stay flat.

\begin{figure}[t]
    \centering
    \begin{subfigure}[t]{0.48\linewidth}
        \centering
        \includegraphics[width=\linewidth]{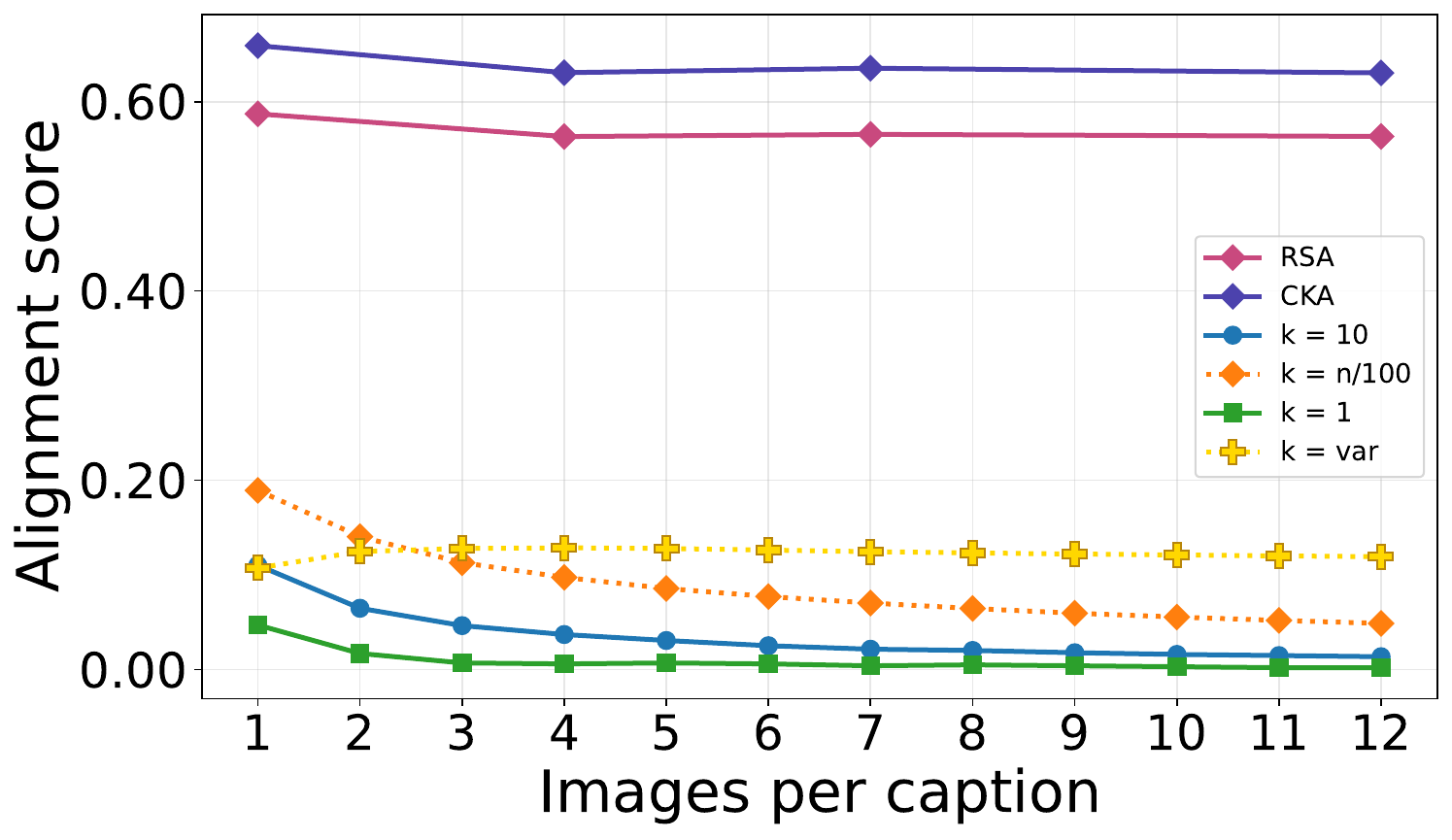}
        \caption{T2I: increasing the number of images per caption.}\label{fig:bij_global_rsa_t2i}
    \end{subfigure}
    \hfill
    \begin{subfigure}[t]{0.48\linewidth}
        \centering
        \includegraphics[width=\linewidth]{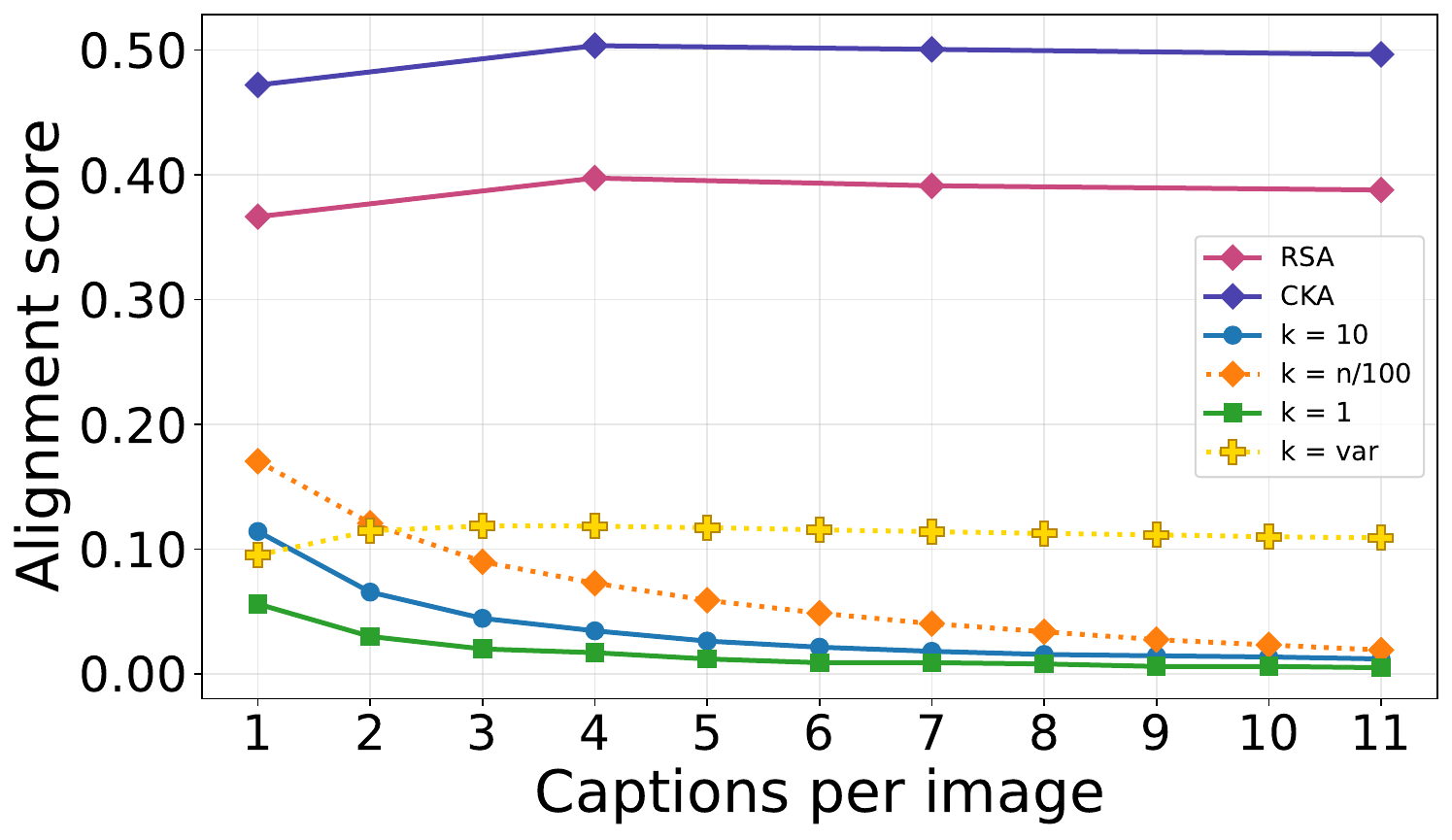}
        \caption{I2T: increasing the number of captions per image.}\label{fig:bij_global_rsa_i2t}
    \end{subfigure}
    \caption{Effect of relaxing the bijective assumption on local and global metrics, for DINOv2-base and OpenLlama-3b on the CycleReward dataset~\citep{bahng2025cycle}, using the same densified sets as \cref{fig:densifying_modalities}. RSA and linear CKA remain stable as correspondences per sample are added, while mutual $k$NN on the identical feature pairs drops by an order of magnitude (\cref{fig:densifying_modalities}).}
    \label{fig:bijection_global}
\end{figure}

The global metrics stay flat (\cref{fig:bijection_global}). While mutual $k$NN drops by an order of magnitude as correspondences are added ($k{=}10$: $0.119 \to 0.012$ for T2I and $0.110 \to 0.008$ for I2T), CKA remains stable across the full densification range ($[0.63, 0.66]$ for T2I, $[0.47, 0.50]$ for I2T), and RSA likewise stays within a narrow band ($[0.56, 0.59]$ and $[0.37, 0.40]$, respectively) as the number of correspondences per sample grows. Adding valid correspondences thus does not change the cross-modal agreement on pairwise structure. What changes is only the composition of the top-$k$ neighborhoods: as \cref{fig:surjection_example} illustrates, the densified modality fills its neighborhoods with siblings of the same source sample while the sparse modality retrieves items from other sources, and mutual $k$NN counts this disagreement as misalignment.

This mirrors the density control on the ImageNet validation set (\cref{fig:ipc_invariance}). There, growing the gallery from 1 to 49 images per class left RSA, CKA, and the density-matched mutual $k$NN variants ($k{=}n/100$, variable-$k$) stable while fixed-$k$ mutual $k$NN decayed. Consequently, the global metrics read the geometry the models impose rather than being a gallery-dependent measure, and fixed-$k$ mutual $k$NN is the density-sensitive statistic. The bijection experiments show the same dissociation along a different axis of densification. Whether neighborhoods are densified by adding more samples (\cref{sec:alignment_across_scales}) or by adding more valid correspondences per sample, the representational geometry that the two models share appears to be unchanged, and only the neighborhood-based structures changes. We argue that both metric families are useful, as they measure different perspectives of alignment.

\textbf{Relaxing bijectivity dissociates the two metric families. Mutual-kNN and the global metrics read different properties of the same representations: mutual-kNN reads top-k neighborhood overlap, which presupposes a unique correct match per query, while RSA and CKA read agreement on pairwise structure, which is defined regardless of how many valid correspondences a sample has.}

\section{Does the alignment vs performance trend predicted by \cite{huh2024prh} continue with recent LLMs?}\label{sec:llm_eval}

To assess whether the alignment vs performance trend predicted by \cite{huh2024prh} continues with recent language models, we evaluate 55 LLMs (see \cref{sec:all-llm-models} for full list) on six standard benchmarks using the LM Evaluation Harness framework~\citep{eval-harness}. \cite{huh2024prh} originally used three benchmarks to measure language capability: HellaSwag~\citep{zellers2019hellaswag}, GSM8K~\citep{cobbe2021gsm8k}, and $(1-\texttt{bitsperbyte})$ on OpenWebText~\citep{Gokaslan2019OpenWeb}. We replace OpenWebText with Wikitext~\citep{merity2016pointer} and extend this analysis to three additional benchmarks that probe different aspects of language understanding: ARC Challenge~\citep{clark2018think}, MMLU~\citep{hendrycks2021mmlu}, and LogiQA2~\citep{liu2023logiqa2}.

\subsection{Benchmarks and metrics.}
\label{app:benchmarks-and-metrics}

\cref{tab:benchmarks} summarizes the evaluation configuration for each benchmark used in \cref{sec:trend-check} of the paper (we used the default configurations from \cite{eval-harness}).

\begin{table}[h]
  \centering
    \caption{Overview of language model benchmarks used in \cref{sec:trend-check} of the paper.}\label{tab:benchmarks}
  \setlength{\tabcolsep}{8pt}
  \resizebox{0.8\textwidth}{!}{
  \begin{tabular}{llcl}
    \toprule
    \textbf{Benchmark} & \textbf{Capability} & \textbf{Few-shot} & \textbf{Metric} \\
    \midrule
    HellaSwag~\citep{zellers2019hellaswag}   & Commonsense inference & 0 & Accuracy \\
    Wikitext~\citep{merity2016pointer}       & Language modeling      & 0 & $1 - \text{bits per byte}$ \\
    ARC Challenge~\cite{clark2018think}   & Science QA             & 4 & Accuracy \\
    GSM8K~\citep{cobbe2021gsm8k}              & Math reasoning         & 5 & Exact match \\
    MMLU~\citep{hendrycks2021mmlu}            & General knowledge      & 5 & Accuracy \\
    LogiQA2~\citep{liu2023logiqa2}            & Logical reasoning      & 5 & Accuracy \\
    \bottomrule
        \bottomrule
  \end{tabular}
  }
  \vspace{0.5em}
\end{table}

\subsection{Does the alignment vs performance trend hold?}
\label{sec:sup-trends}
\begin{figure*}[p]
    \centering

    \includegraphics[width=\textwidth,height=0.24\textheight,keepaspectratio]{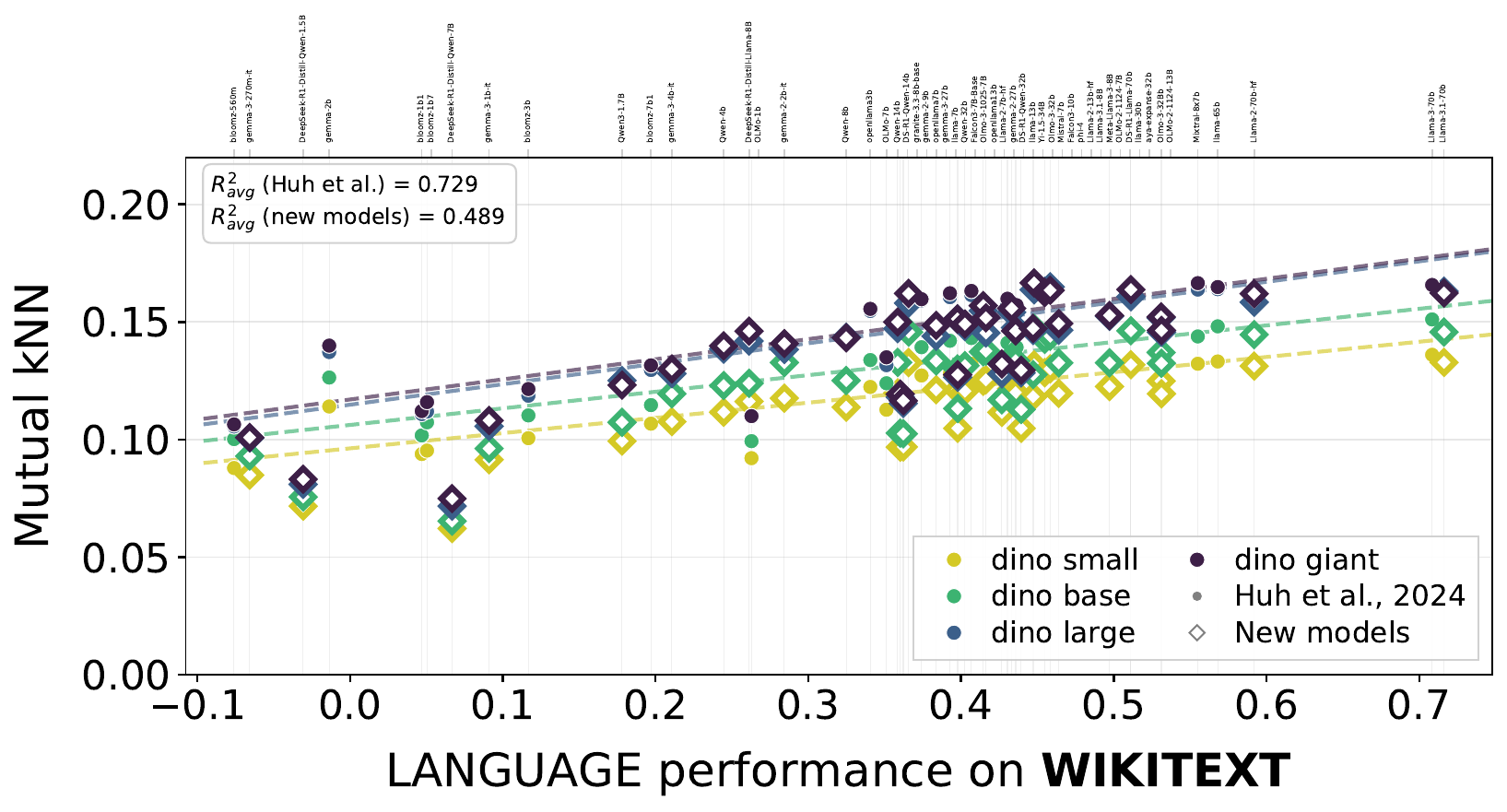}

    \vspace{0.5em}
    \includegraphics[width=\textwidth,height=0.24\textheight,keepaspectratio]{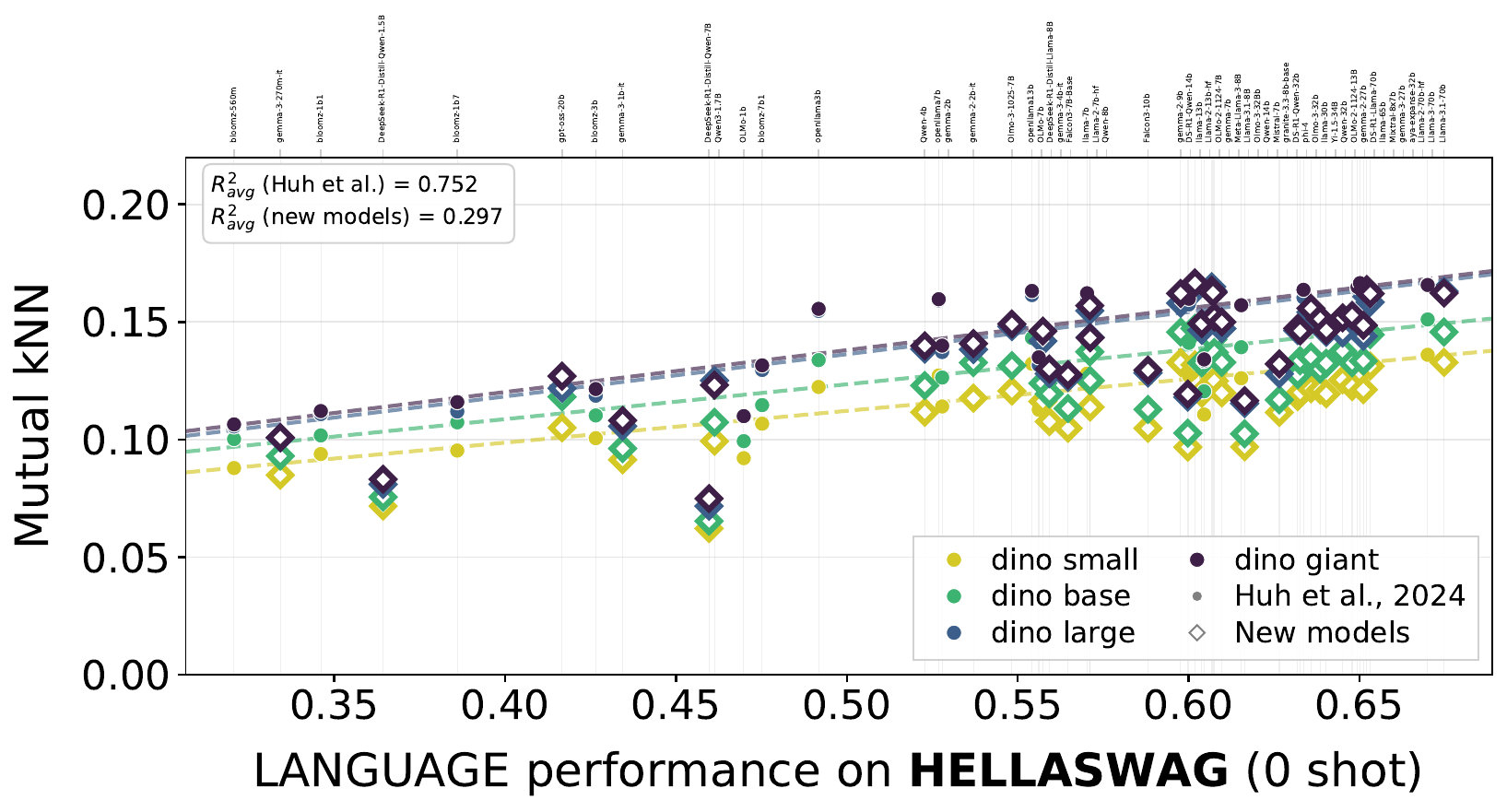}

    \vspace{0.5em}
    \includegraphics[width=\textwidth,height=0.24\textheight,keepaspectratio]{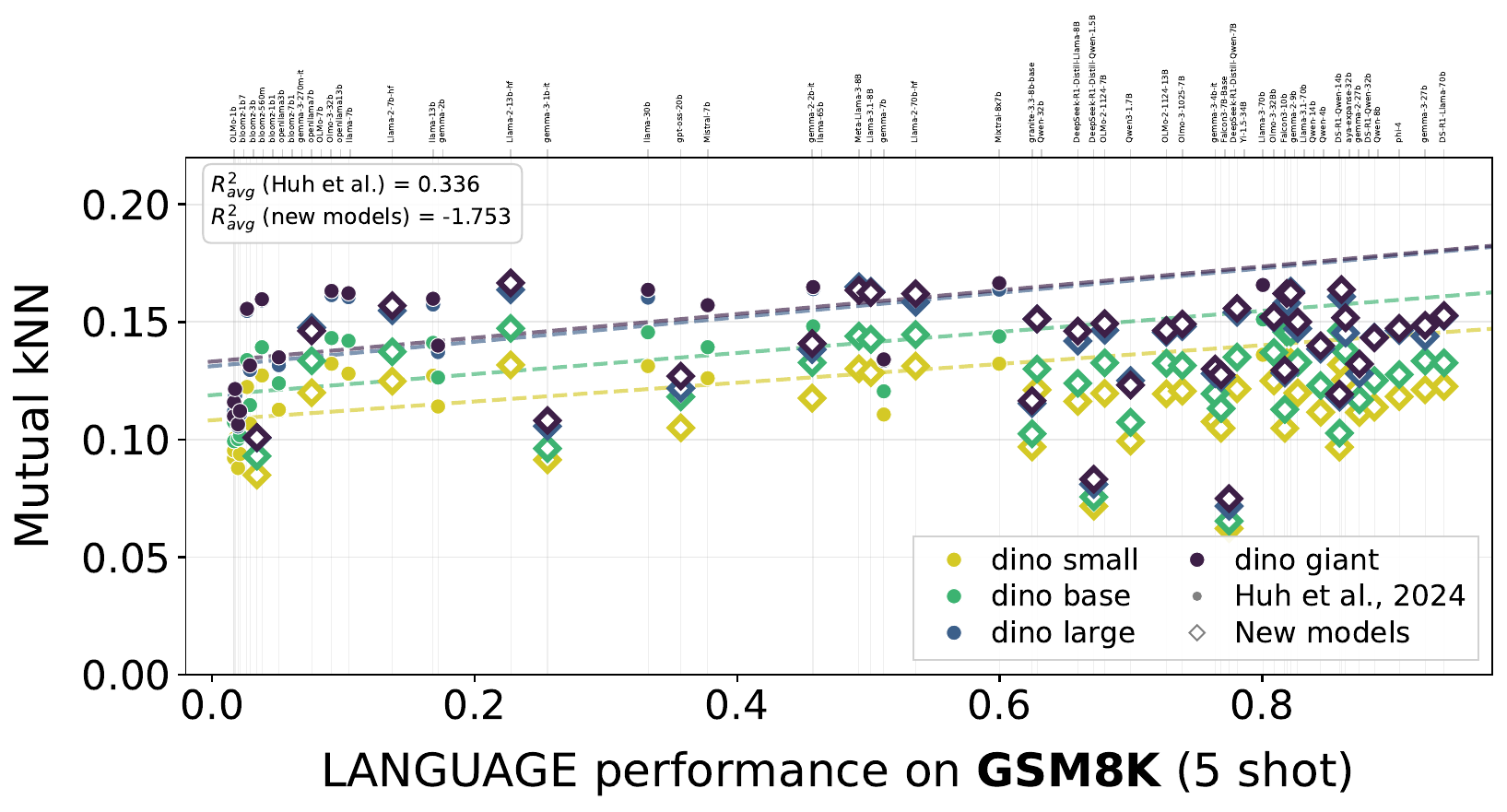}

    \caption{Mutual $k$NN alignment vs.\ language benchmark performance for 55 LLMs across four DINOv2 variants, on WikiText, HellaSwag, and GSM8K. Dashed lines show the linear trend fit to the 19 \textit{base models} from~\cite{huh2024prh}. For WikiText and HellaSwag (top two plots), recent models roughly follow the trend. For GSM8K (bottom plot), the trend is not followed.}
    \label{fig:trends_bubble_page1}
\end{figure*}

\begin{figure*}[p]
    \centering

    \includegraphics[width=\textwidth,height=0.24\textheight,keepaspectratio]{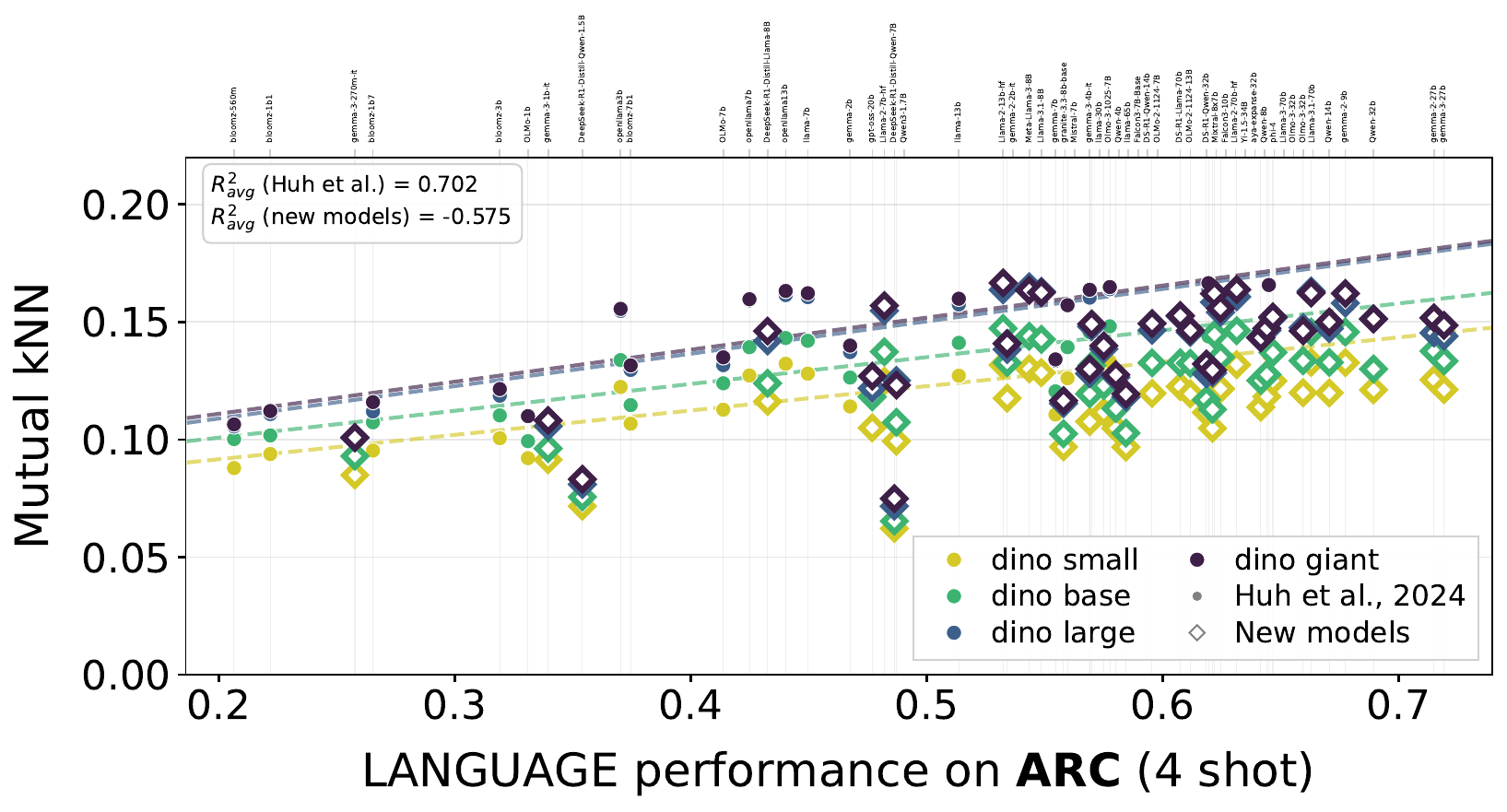}

    \vspace{0.5em}
    \includegraphics[width=\textwidth,height=0.24\textheight,keepaspectratio]{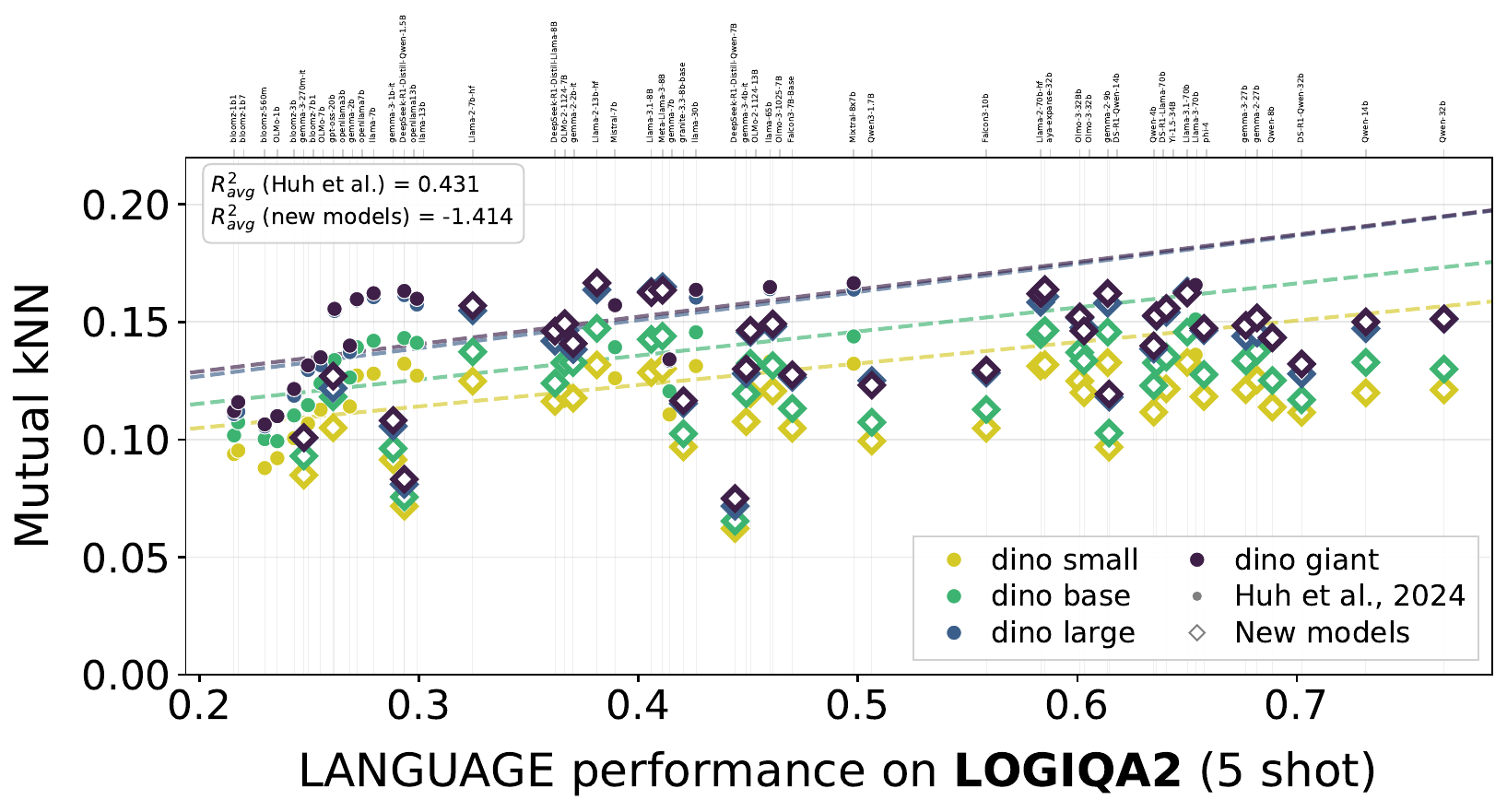}

    \vspace{0.5em}
    \includegraphics[width=\textwidth,height=0.24\textheight,keepaspectratio]{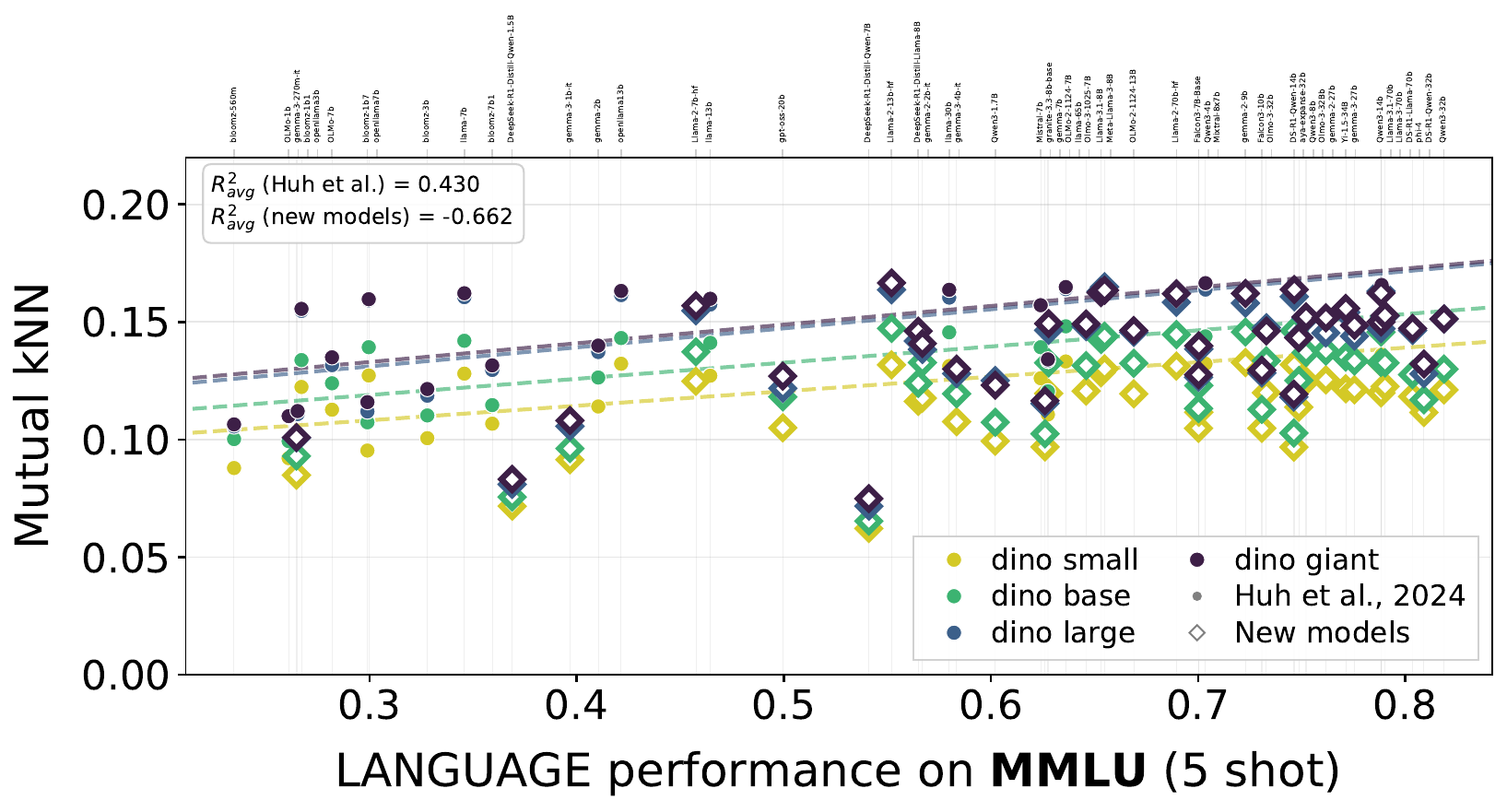}

    \caption{Mutual $k$NN alignment vs.\ language benchmark performance for 55 LLMs across four DINOv2 variants, on ARC, LogiQA2, and MMLU. As with GSM8K, the alignment-performance trend from~\cite{huh2024prh} does not extrapolate to recent models on any of these reasoning benchmarks. Stronger models do not appear to show higher mutual $k$NN alignment with DINOv2 features.}
    \label{fig:trends_bubble_page2}
    \vspace{4em}
\end{figure*}

For each benchmark and each DINOv2 variant, we fit a linear regression on the \textit{base models} used in \cite{huh2024prh}, predicting mutual $k$NN alignment $a$ from benchmark performance $p$. We then evaluate how well this trend describes two populations:
\\\\
\noindent\textbf{$R^2$ (Huh~\etal):} The standard coefficient of determination on the data is used to fit the regression, i.e.\ $R^2(\text{Huh~\etal}) = r^2$, where $r$ is the Pearson correlation between mutual $k$NN alignment and language modelling benchmark score across the 19 \textit{base models} from \cite{huh2024prh}.
\\\\
\noindent\textbf{$R^2$ (new):} We apply the line fitted on the \textit{base models} to the 36 recent models and compute the generalized $R^2$:
$$R^2(\text{new}) =
1 - \frac{\sum_{i \in \mathcal{M}_\text{new}} (a_i - \hat{a}_i)^2}{\sum_{i \in \mathcal{M}_\text{new}} (a_i - \bar{a}_\text{new})^2},$$
where $\hat{a}_i$ are the linear regression alignment predictions based on language performance $p_i$, $a_i$ is the alignment score for the $i$-th model and  $\bar{a}_\text{new}$ is the mean alignment of the new models. When $R^2(\text{new}) > 0$, the relation between alignment and language performance predicted in Huh~\etal\ extrapolates. When $R^2(\text{new}) < 0$, the regression line is a worse predictor than simply predicting the average $\bar{a}_\text{new}$. $R^2_\text{avg}$ values are reported in \cref{tab:r2_benchmarks} and \cref{fig:trends_bubble_page1,fig:trends_bubble_page2}.

\begin{table}[t]
  \centering
    \caption{Average $R^2$ of the linear regression (fitted on the 19 models from \cite{huh2024prh}) evaluated on the models themselves and on the 36 recent models, across all four DINOv2 variants. Positive $R^2_{\text{avg}}(\text{new})$ indicates that the trend from the \textit{base models} is a good predictor for the new models. Negative values indicate that the regression line is a worse predictor than the mean. The last two columns show $R^2$ computed on 25 recent \textit{base models} and 20 recent \textit{post-trained models}.}\label{tab:r2_benchmarks}
  \setlength{\tabcolsep}{10pt}
  \begin{tabular}{lcccc}
    \toprule
    \textbf{Benchmark} & $R^2_\text{avg}$ \textbf{(\text{Huh~\etal})} & $R^2_\text{avg}$ \textbf{(new)} & $R^2_\text{avg}$ \textbf{(base)} & $R^2_\text{avg}$ \textbf{(post-trained)} \\
    \midrule
    HellaSwag  &     0.752 &     0.297 &  $-$0.231 &  $-$1.113 \\
    Wikitext   &     0.729 &     0.489 &  $-$0.266 &  $-$0.479 \\
    \midrule
    ARC        &     0.702 &  $-$0.575 &  $-$1.339 &  $-$3.611 \\
    GSM8K      &     0.336 &  $-$1.753 &  $-$5.110 & $-$11.011 \\
    MMLU       &     0.430 &  $-$0.662 &  $-$1.801 &  $-$4.551 \\
    LogiQA2    &     0.431 &  $-$1.414 &  $-$3.160 & $-$12.539 \\
    \bottomrule
    \bottomrule
  \end{tabular}
  \vspace{0.5em}
  \label{tab:r2-new}
\end{table}

The results reveal a split across language modelling benchmarks. For HellaSwag and Wikitext, the relation between alignment and language performance observed by Huh~\etal partially extends to recent models: the $R^2_\text{avg}$ on new models remains positive (0.297 and 0.489, respectively), indicating that stronger language models according to these benchmarks have higher mutual $k$NN alignment with DINOv2. Both benchmarks primarily measure next-token prediction quality and commonsense language understanding, which are closely related to the pretraining objective of autoregressive LLMs.

In contrast, for the four benchmarks that probe more specialized reasoning abilities: ARC (science QA), GSM8K (arithmetic), MMLU (general knowledge), and LogiQA2 (logical reasoning), the relation between alignment and language performance predicted from the \textit{base models}~\citep{huh2024prh} does not appear to hold for this set of recent models.

Specifically, the $R^2_\text{avg}$ on new models is consistently negative, ranging from $-0.575$ (ARC) to $-1.753$ (GSM8K). This means that the linear fit from the \textit{base models} from \cite{huh2024prh} is a worse predictor of alignment for recent models than simply predicting the mean. In \cref{fig:trends_bubble_page1,fig:trends_bubble_page2}, we observe that recent models that are stronger than the best \textit{base model} (Meta-Llama-3-70B) do not show higher mutual $k$NN alignment with DINOv2 features. Instead, their alignment scores seem to saturate or decrease.

We note that the 36 added (new) models are heterogeneous. They include new models trained on next-token prediction (pre-training), instruction-tuned models, and reasoning-distilled models (e.g.\ DeepSeek-R1-Distill). We treat them as a single population to test whether the trend extrapolates to recent LLMs.

The above results support and extend the finding from \cref{sec:trend-check} of the main paper.
\textbf{The relationship between alignment and language performance from~\cite{huh2024prh} holds for core language modelling benchmarks, but does not seem to generalize to reasoning benchmarks.}

\subsection{Is the failure to extrapolate an artifact of post-training?}

\label{sec:e3-posttraining}

We note that the negative $R^2_{\mathrm{avg}}$ of \cref{sec:sup-trends} might be attributed to post-training rather than to capability. Instruction-tuned and reasoning-distilled models spread mostly at the end of the reasoning benchmarks \cref{fig:trends_bubble_page2}, and their representations are shaped by objectives such as preference optimization and distillation from a teacher's reasoning traces. That differs from models used to predict the trend~\citep{huh2024prh}, although one must note that \cite{huh2024prh} uses the bloomz family, which is also post-trained. Nevertheless, if post-training moved representations away from vision, the saturation would be a result of the post-training objectives rather than a breakdown of the capability–alignment relation.  To separate the two, we re-evaluate the generalized $R^2$ on the recent models split by training regime, \emph{base-only} against \emph{post-trained}, holding the regression line fixed to the \cite{huh2024prh} base-model fit, \cref{tab:r2-new}. For these experiments, we extended the model set from \cite{huh2024prh} with 25 base-only models and 20 post-trained models.

Both subsets are negative on every benchmark: the \emph{base-only} recent models already fail to extrapolate. The failure is mildest on the core language-modelling benchmarks (WikiText, HellaSwag), where $R^2_{\mathrm{avg}}$ stays close to zero, and largest on the reasoning and math benchmarks (ARC, MMLU, GSM8K, LogiQA2), where the \emph{base-only} $R^2_{\mathrm{avg}}$ ranges from $-1.34$ to $-5.11$ and the \emph{post-trained} from $-3.61$ to $-12.54$. \footnote{$R^2_{\mathrm{avg}}(\text{base})$ and $R^2_{\mathrm{avg}}(\text{post-trained})$ values differ from $R^2_{\mathrm{avg}}(\text{new})$ in Table~\ref{tab:r2-new} because the ``new'' subset contains a mix of base and post-trained, splitting the population changes $R^2_{\mathrm{avg}}$ values.}

Recent \emph{base} models that surpass the strongest base model in \cite{huh2024prh} do not exhibit higher mutual $k$-NN alignment with DINOv2 features either. Moreover, the gap between the two subsets has no consistent sign. Post-trained and base models are both more negative on ARC-C, MMLU, GSM8K, and LogiQA2, so the divergence does not track post-training in a systematic way. \textbf{The failure to extrapolate is therefore not an artifact of post-training: recent \emph{base} models saturate against DINOv2 features on their own.}

\subsection{Is the failure to extrapolate an artifact of the mutual $k$NN metric?}
\label{sec:e4-global-metrics}

\begin{table}[t]
  \centering
    \caption{Average $R^2$ of the linear regression (fitted on the 19 \textit{base models} from \cite{huh2024prh}) evaluated on the models themselves and on the recent models, for calibrated global RSA and calibrated linear CKA, averaged across the four DINOv2 variants. As in \cref{tab:r2_benchmarks}, positive $R^2_{\text{avg}}(\text{new})$ indicates that the trend fitted on the \textit{base models} extrapolates to the recent models. Negative values indicate that the regression line is a worse predictor than the mean. In contrast to mutual $k$NN (\cref{tab:r2_benchmarks}), $R^2_{\text{avg}}(\text{new})$ is negative on all six benchmarks for both global metrics, including the core language-modelling benchmarks.}\label{tab:r2_global_metrics}
  \setlength{\tabcolsep}{10pt}
  \begin{tabular}{lcccc}
    \toprule
    & \multicolumn{2}{c}{\textbf{RSA}} & \multicolumn{2}{c}{\textbf{CKA}} \\
    \cmidrule(lr){2-3} \cmidrule(lr){4-5}
    \textbf{Benchmark} & $R^2_\text{avg}$ \textbf{(Huh~\etal)} & $R^2_\text{avg}$ \textbf{(new)} & $R^2_\text{avg}$ \textbf{(Huh~\etal)} & $R^2_\text{avg}$ \textbf{(new)} \\
    \midrule
    HellaSwag  & 0.602 & $-$0.264 & 0.208 & $-$0.416 \\
    Wikitext   & 0.685 & $-$0.073 & 0.315 & $-$0.313 \\
    \midrule
    ARC        & 0.568 & $-$0.628 & 0.179 & $-$0.754 \\
    GSM8K      & 0.236 & $-$1.426 & 0.052 & $-$1.218 \\
    MMLU       & 0.262 & $-$0.621 & 0.047 & $-$0.487 \\
    LogiQA2    & 0.329 & $-$1.112 & 0.092 & $-$1.280 \\
    \bottomrule
    \bottomrule
  \end{tabular}
  \vspace{0.5em}
\end{table}

The trend analyses in \cref{sec:sup-trends,sec:e3-posttraining} measure alignment
with mutual $k$NN, a local metric that reads top-$k$ neighborhoods. One might
therefore attribute the failure to extrapolate to the metric rather than to the
representations. \cite{groger2026revisitingplatonicrepresentationhypothesis}
report that local and global similarity measures can behave differently under
calibration, and our granularity analysis (\cref{sec:alignment_across_scales})
shows that mutual $k$NN with $k{=}10$ at $n{=}1024$ probes the coarse regime. If
the saturation were a property of how mutual $k$NN reads neighborhoods, global
measures of representational similarity should still continue the trend. To
separate the metric from the finding, we repeat the trend check with two global
measures: RSA and linear CKA (\cref{sec:supp_metrics}), both calibrated against
the pairing-shuffle null of
\cite{groger2026revisitingplatonicrepresentationhypothesis}. The protocol is
otherwise unchanged from \cref{sec:sup-trends}: for each benchmark and each
DINOv2 variant, we fit a linear regression on the 19 \textit{base models} from
\cite{huh2024prh} and evaluate the generalized $R^2$ on the recent models.

\begin{figure*}[p]
    \centering
    \setlength{\tabcolsep}{1.5pt}
    \begin{tabular}{@{}>{\centering\arraybackslash}m{1.2em}@{\hskip 2pt}m{0.315\textwidth}m{0.315\textwidth}m{0.315\textwidth}@{}}
    \rotatebox[origin=c]{90}{\scriptsize RSA, linear fit} &
    \includegraphics[width=\linewidth]{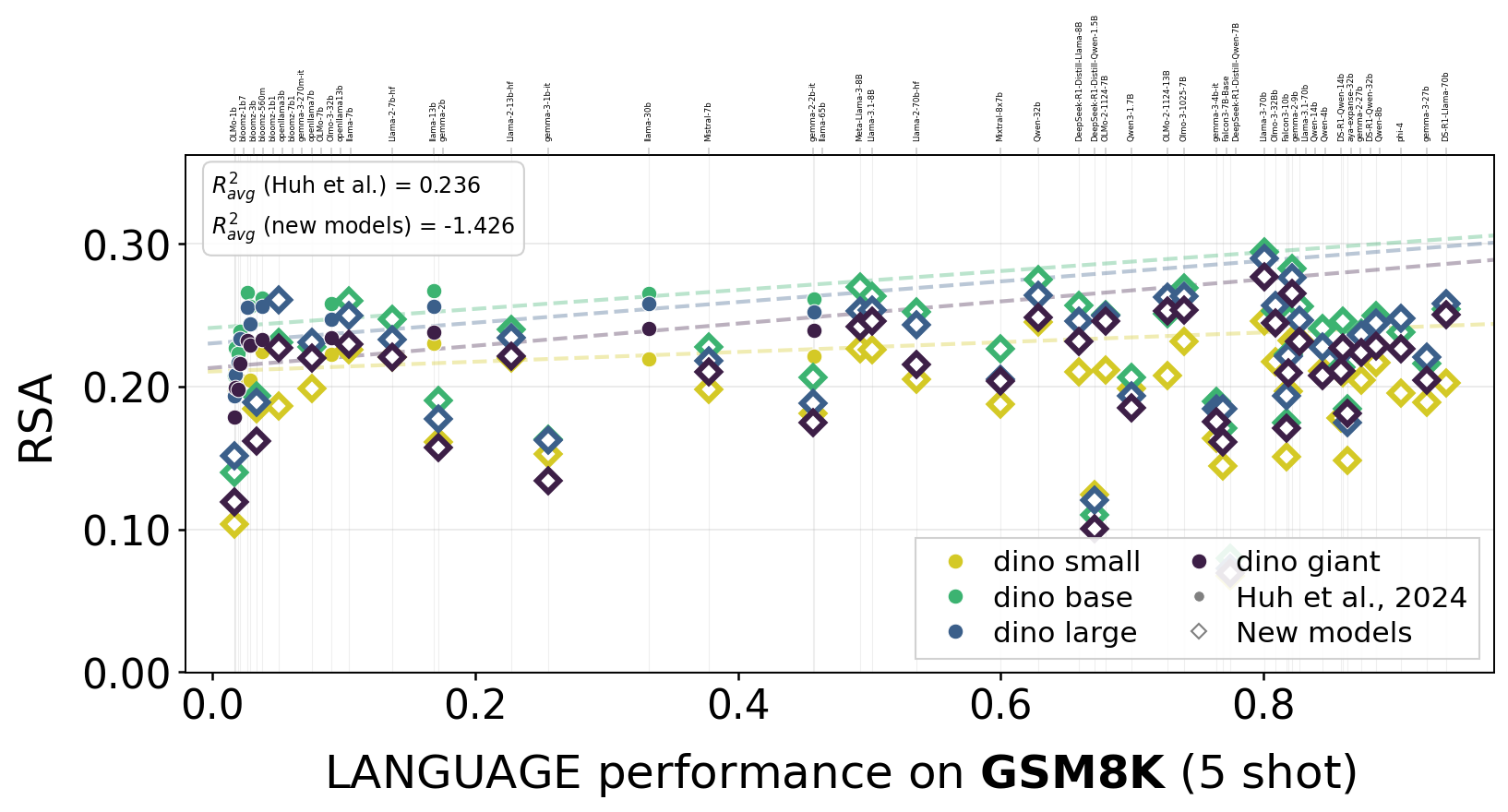} &
    \includegraphics[width=\linewidth]{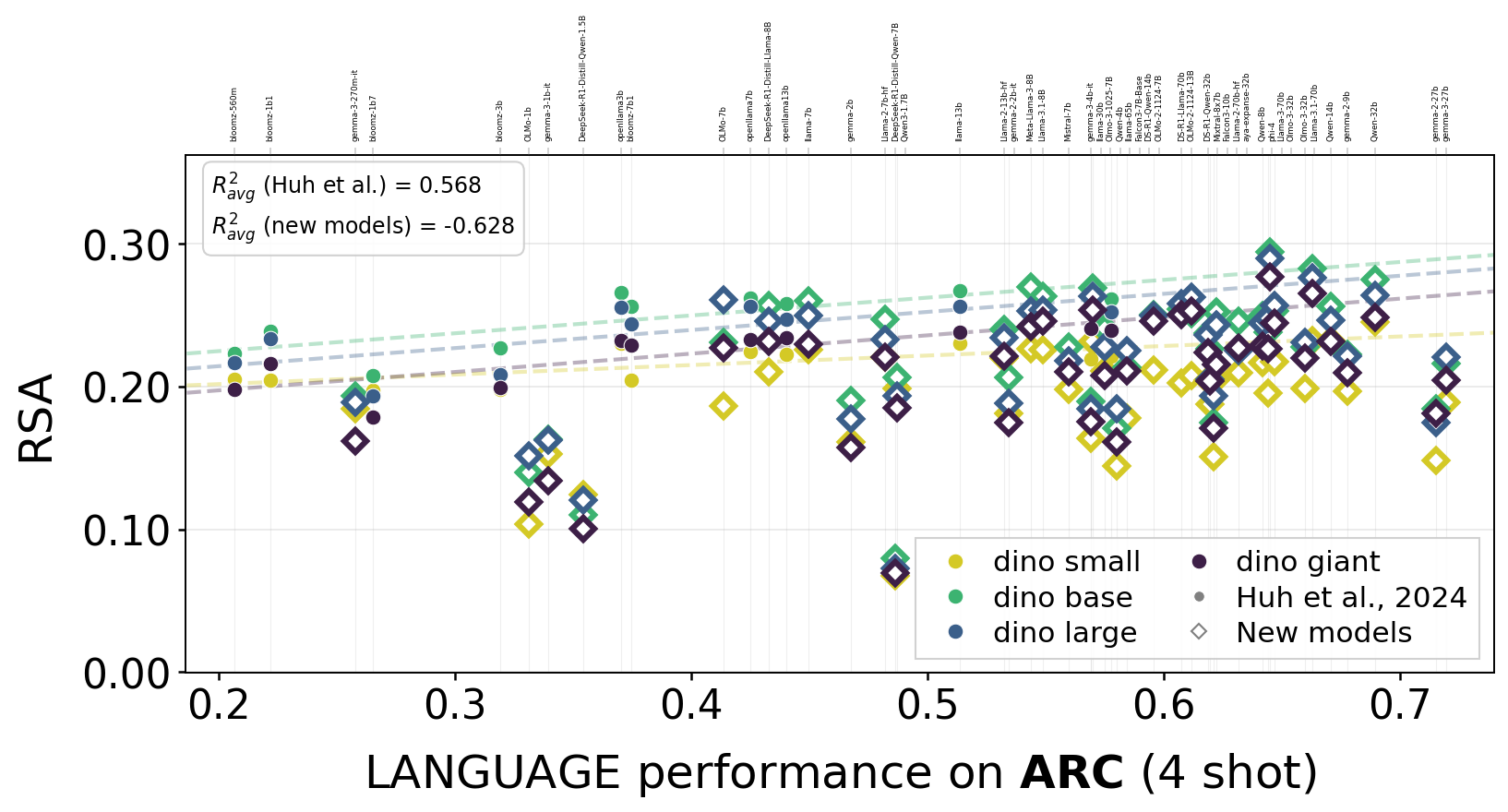} &
    \includegraphics[width=\linewidth]{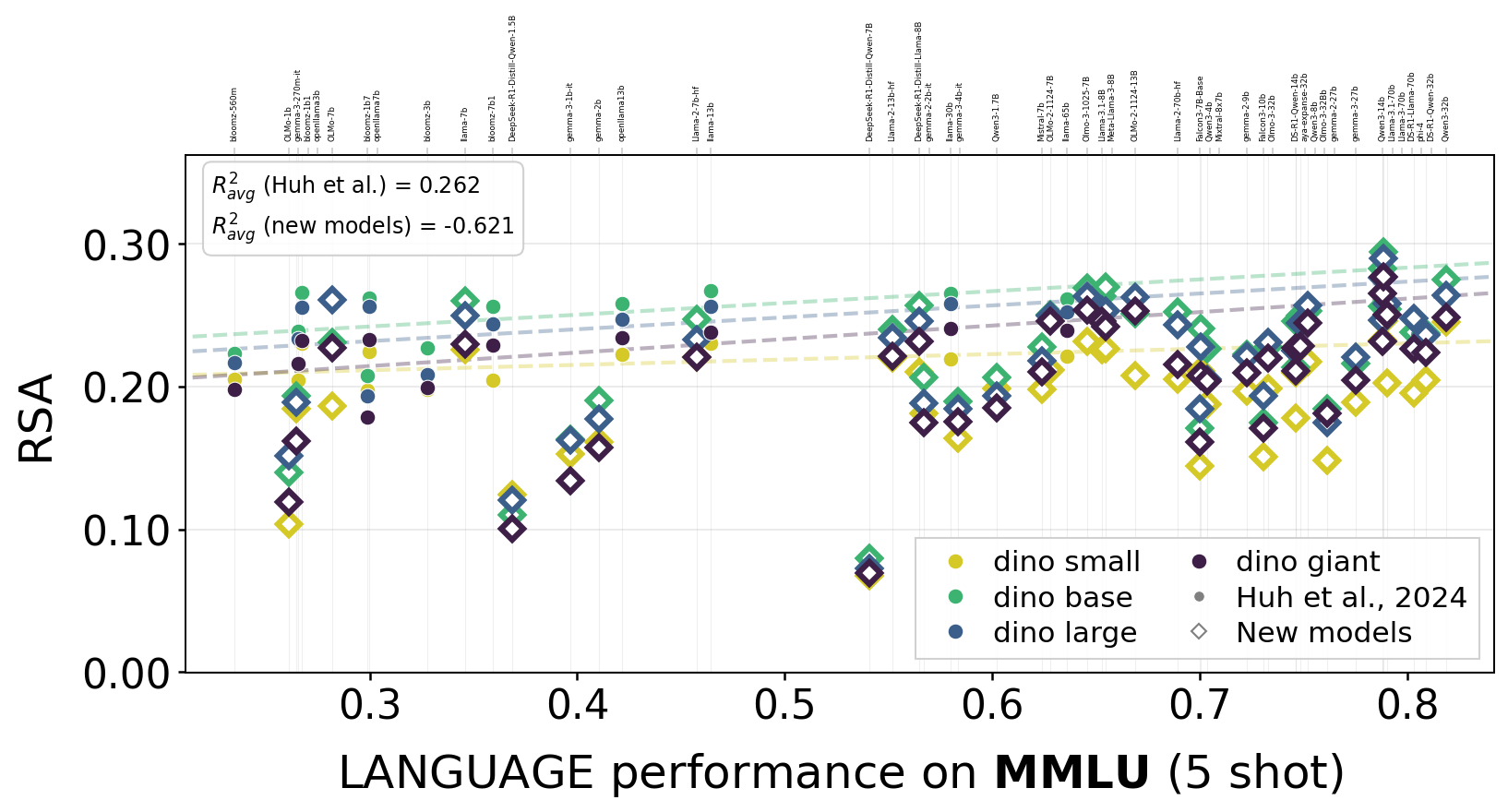} \\[2pt]
    \rotatebox[origin=c]{90}{\scriptsize CKA, linear fit} &
    \includegraphics[width=\linewidth]{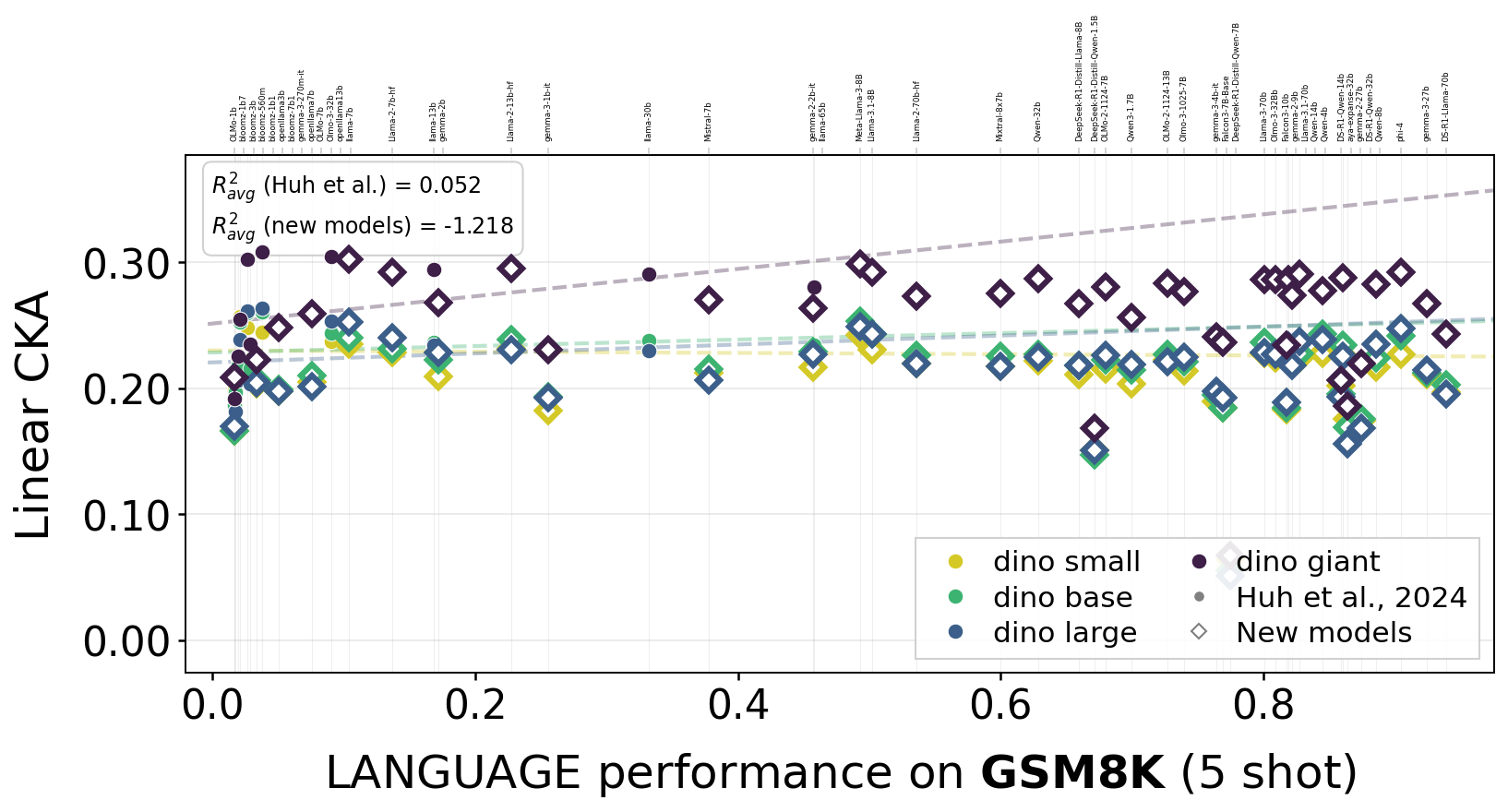} &
    \includegraphics[width=\linewidth]{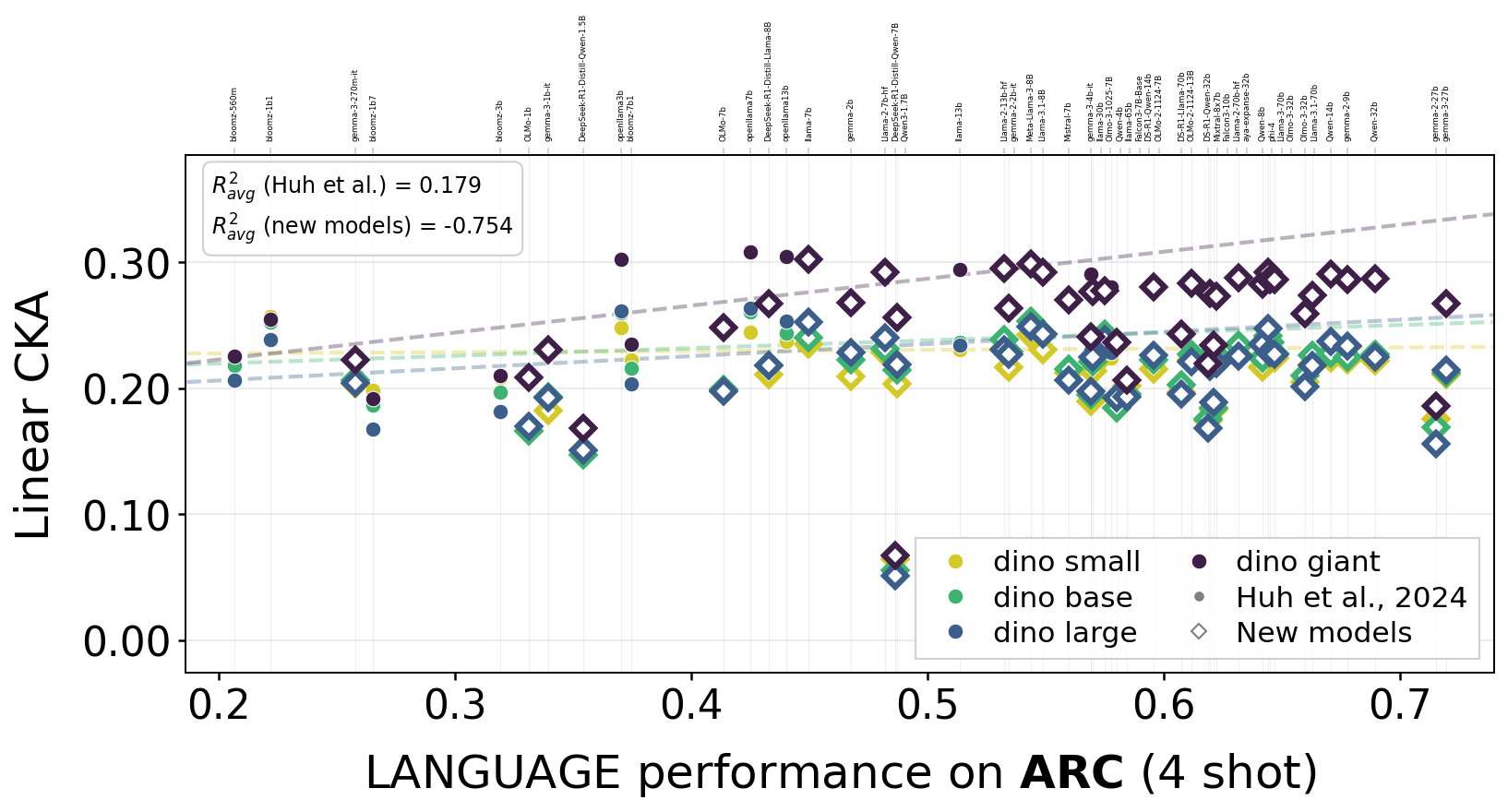} &
    \includegraphics[width=\linewidth]{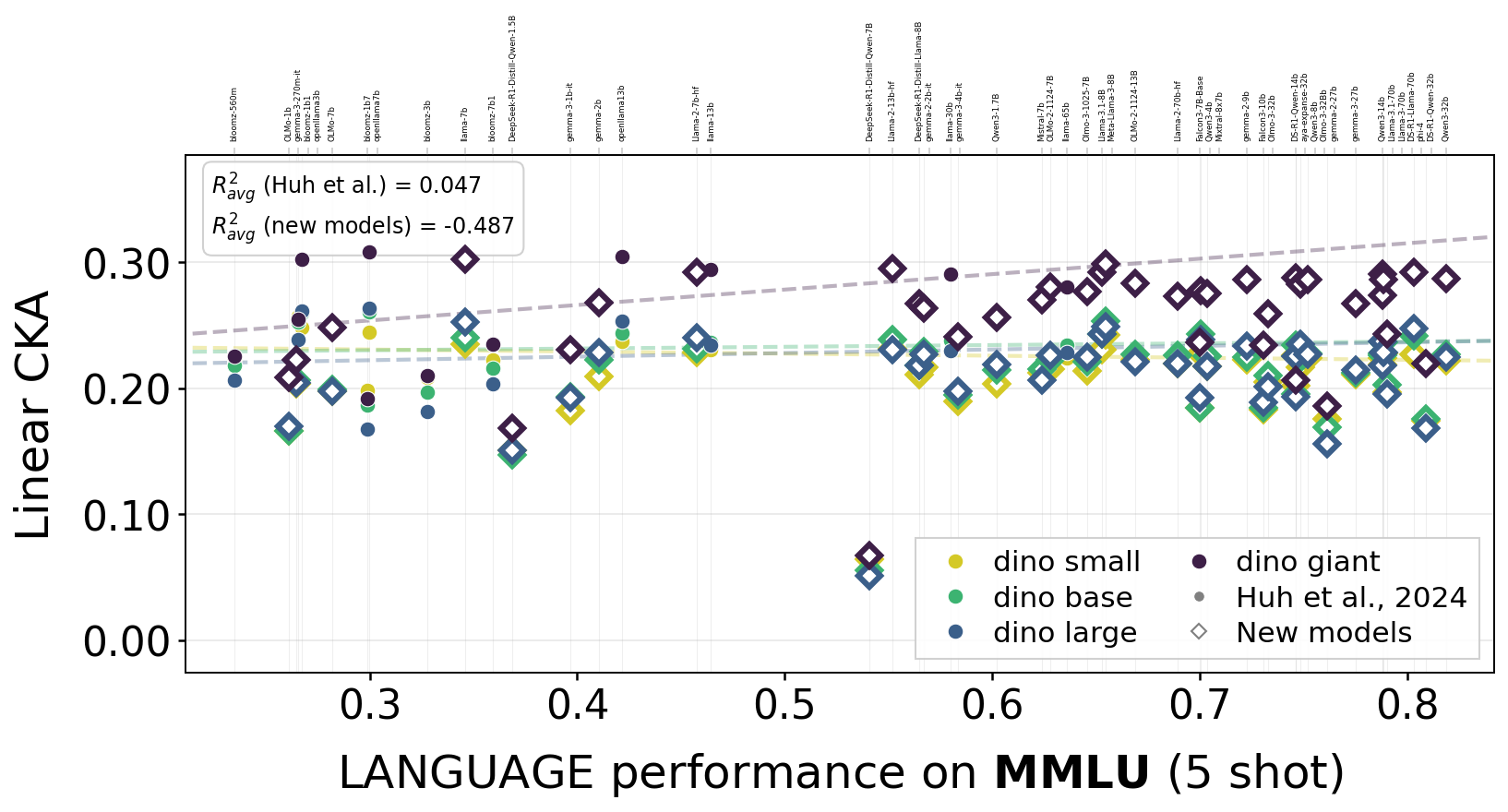} \\[2pt]
    \rotatebox[origin=c]{90}{\scriptsize RSA, LOWESS} &
    \includegraphics[width=\linewidth]{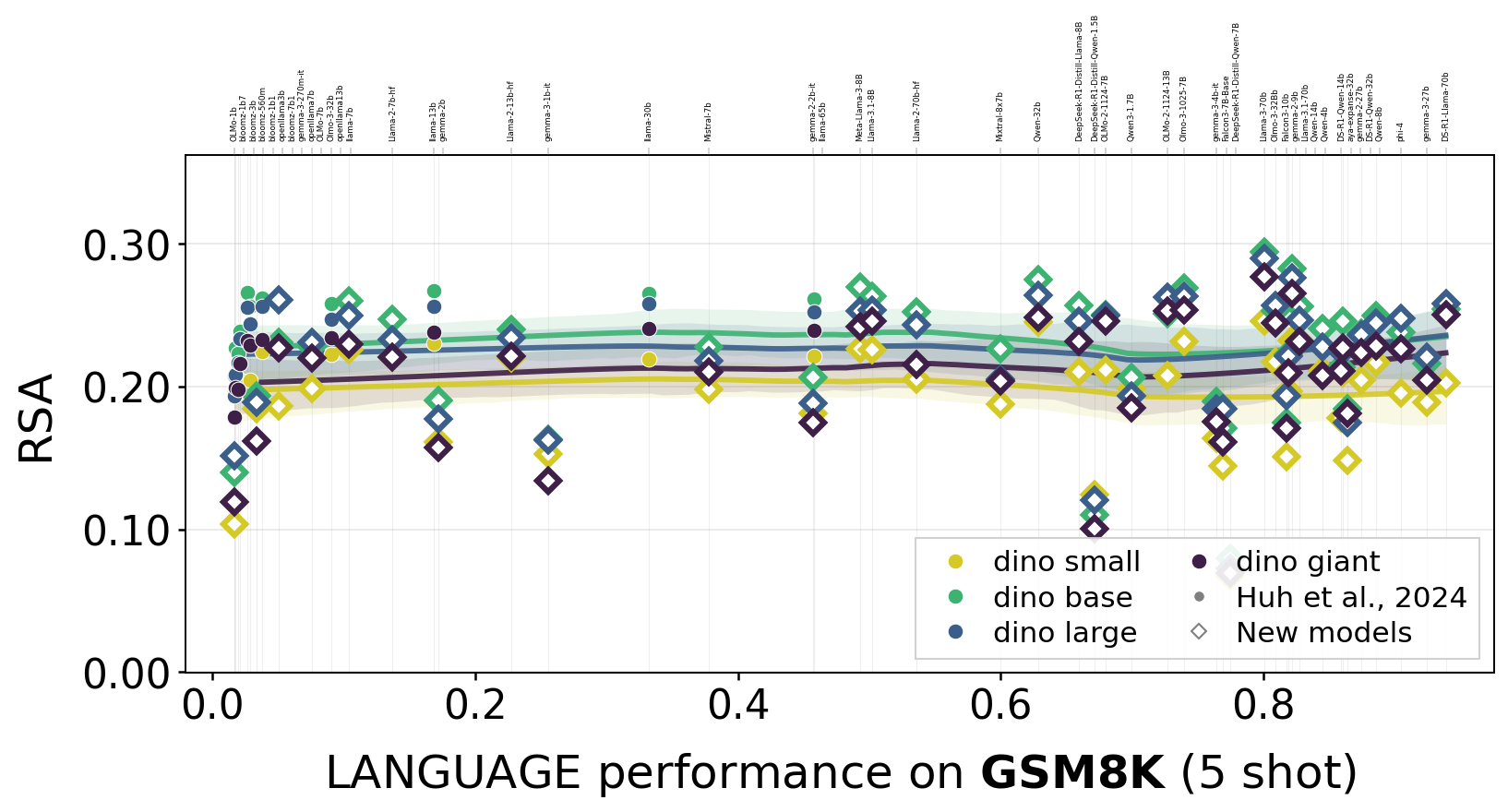} &
    \includegraphics[width=\linewidth]{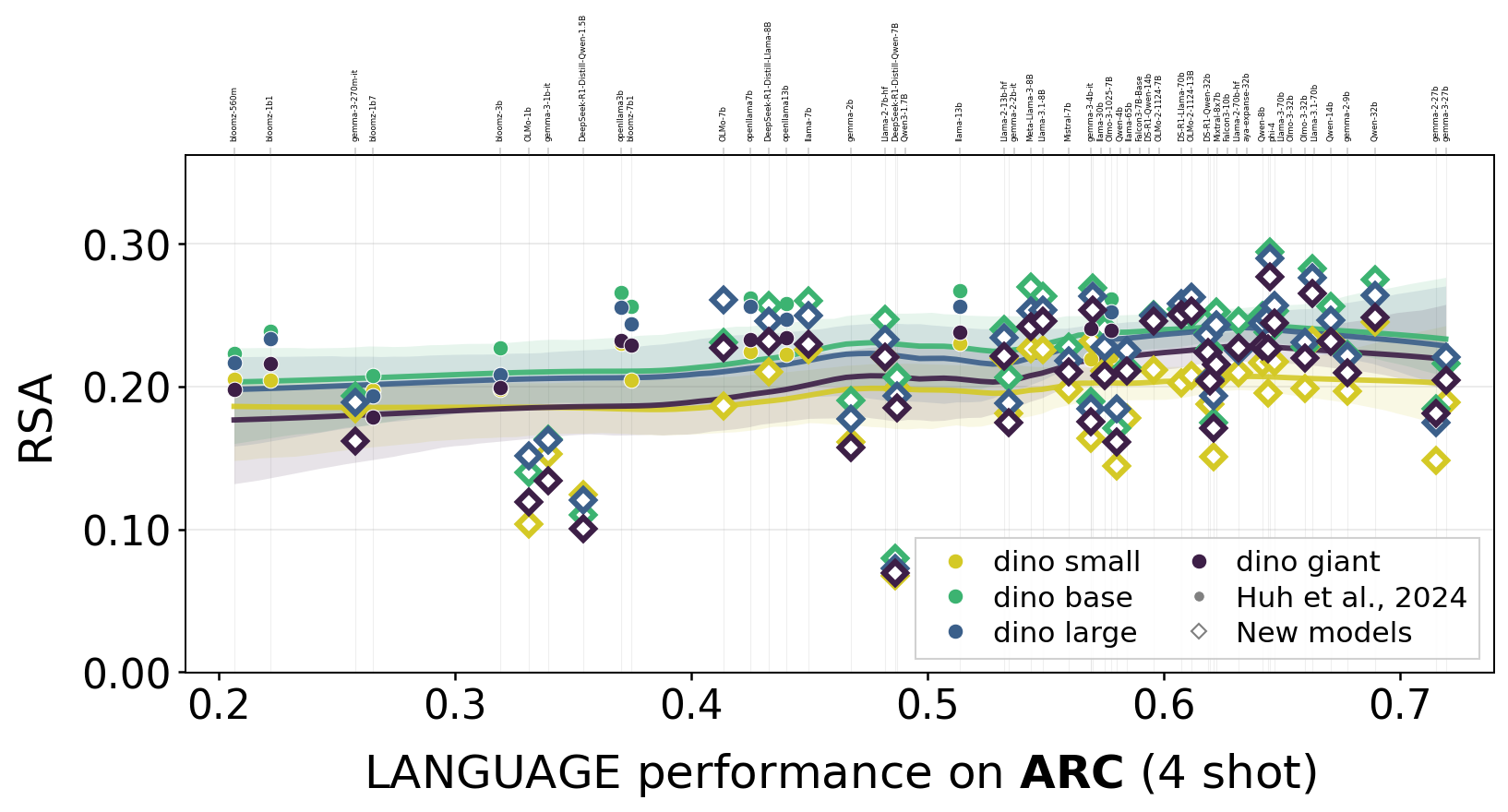} &
    \includegraphics[width=\linewidth]{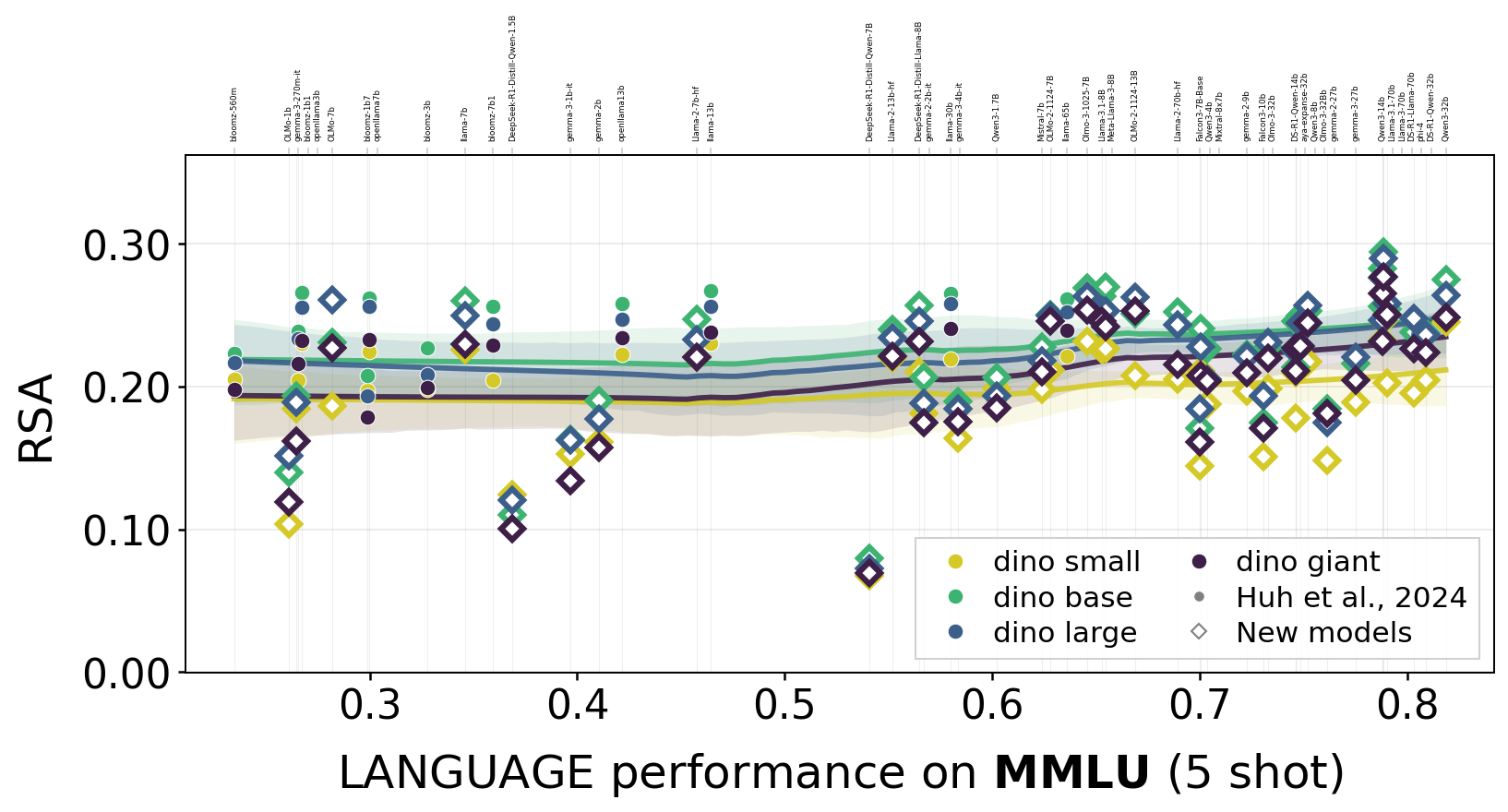} \\[2pt]
    \rotatebox[origin=c]{90}{\scriptsize CKA, LOWESS} &
    \includegraphics[width=\linewidth]{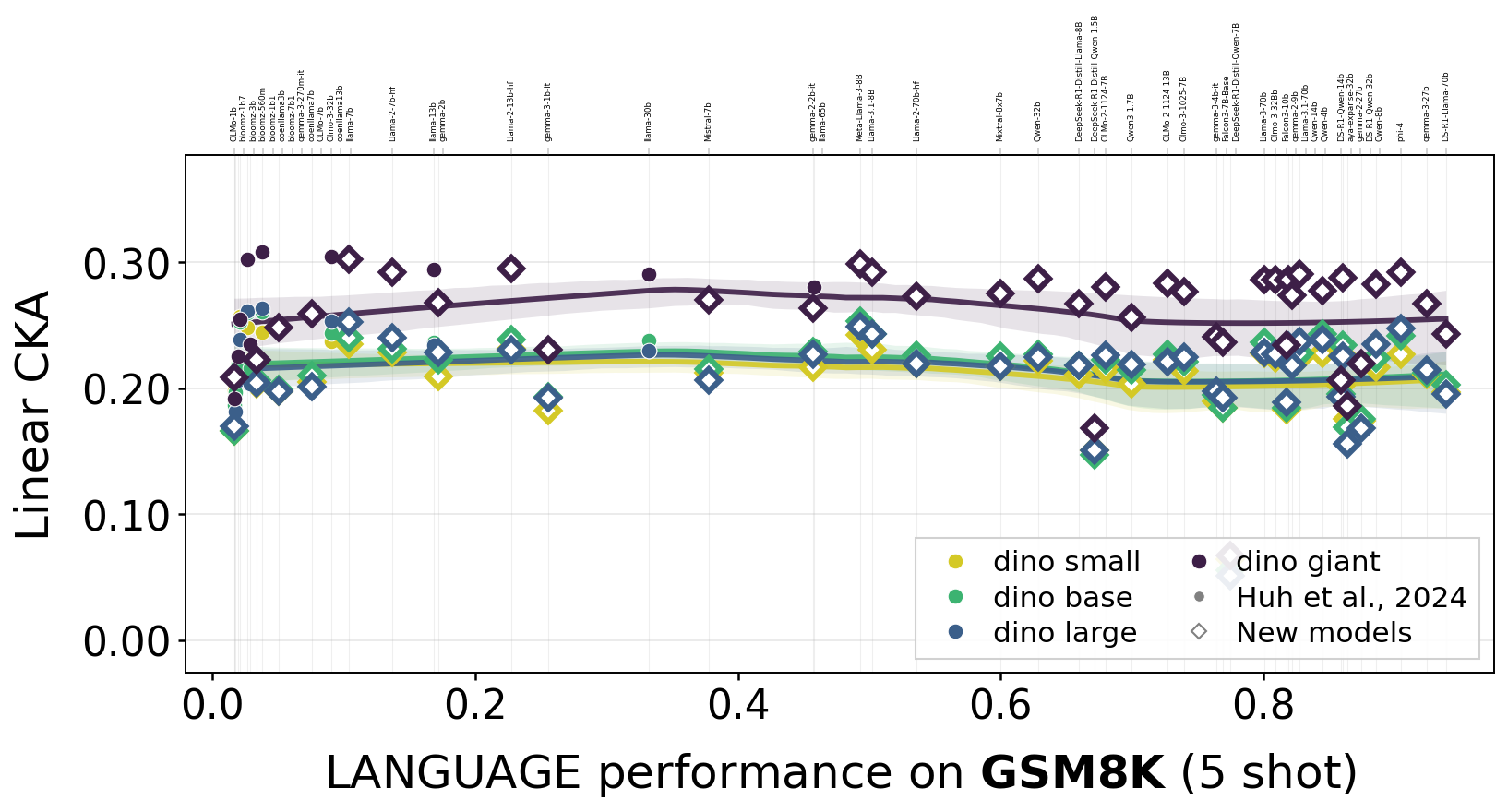} &
    \includegraphics[width=\linewidth]{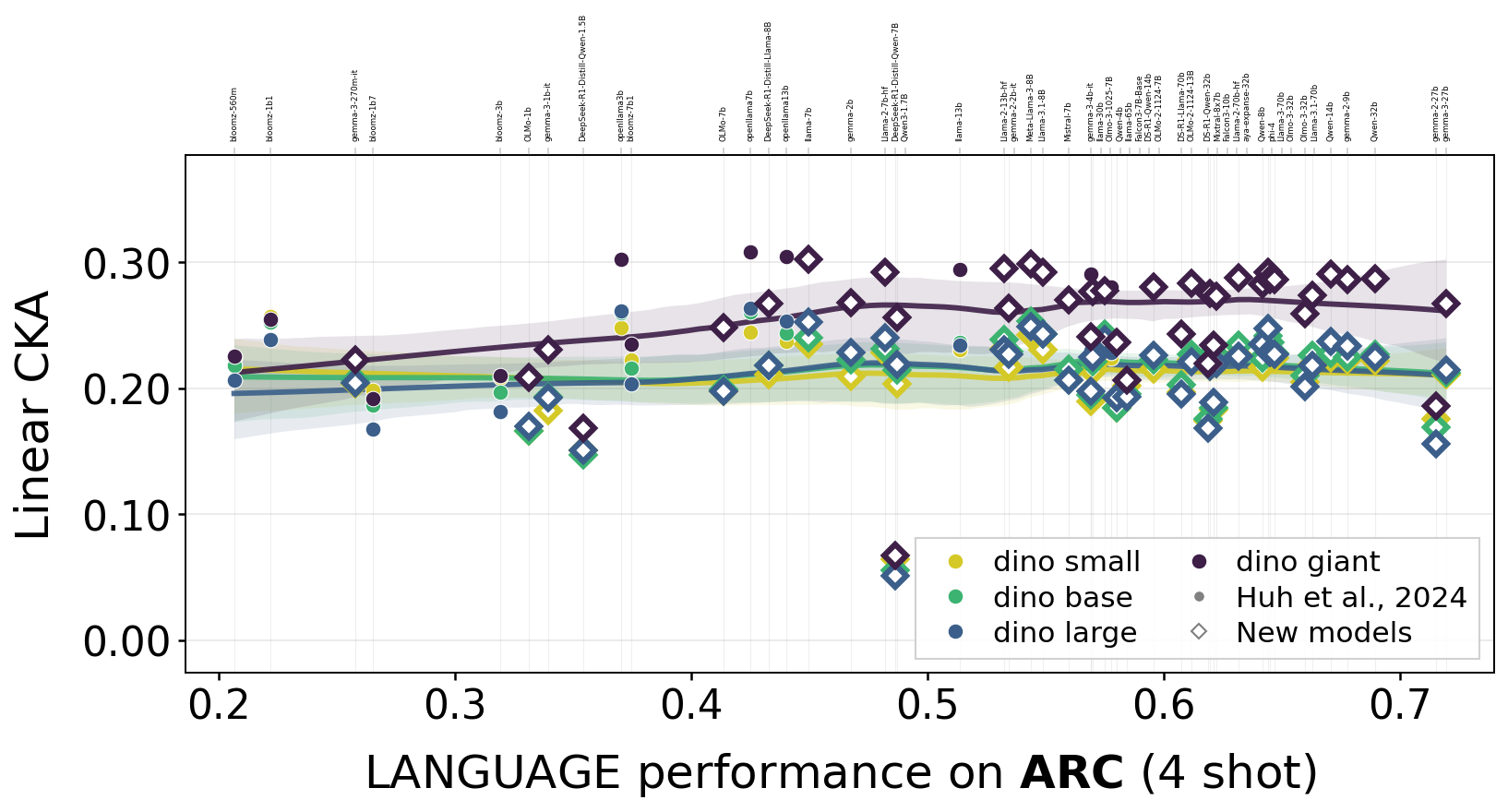} &
    \includegraphics[width=\linewidth]{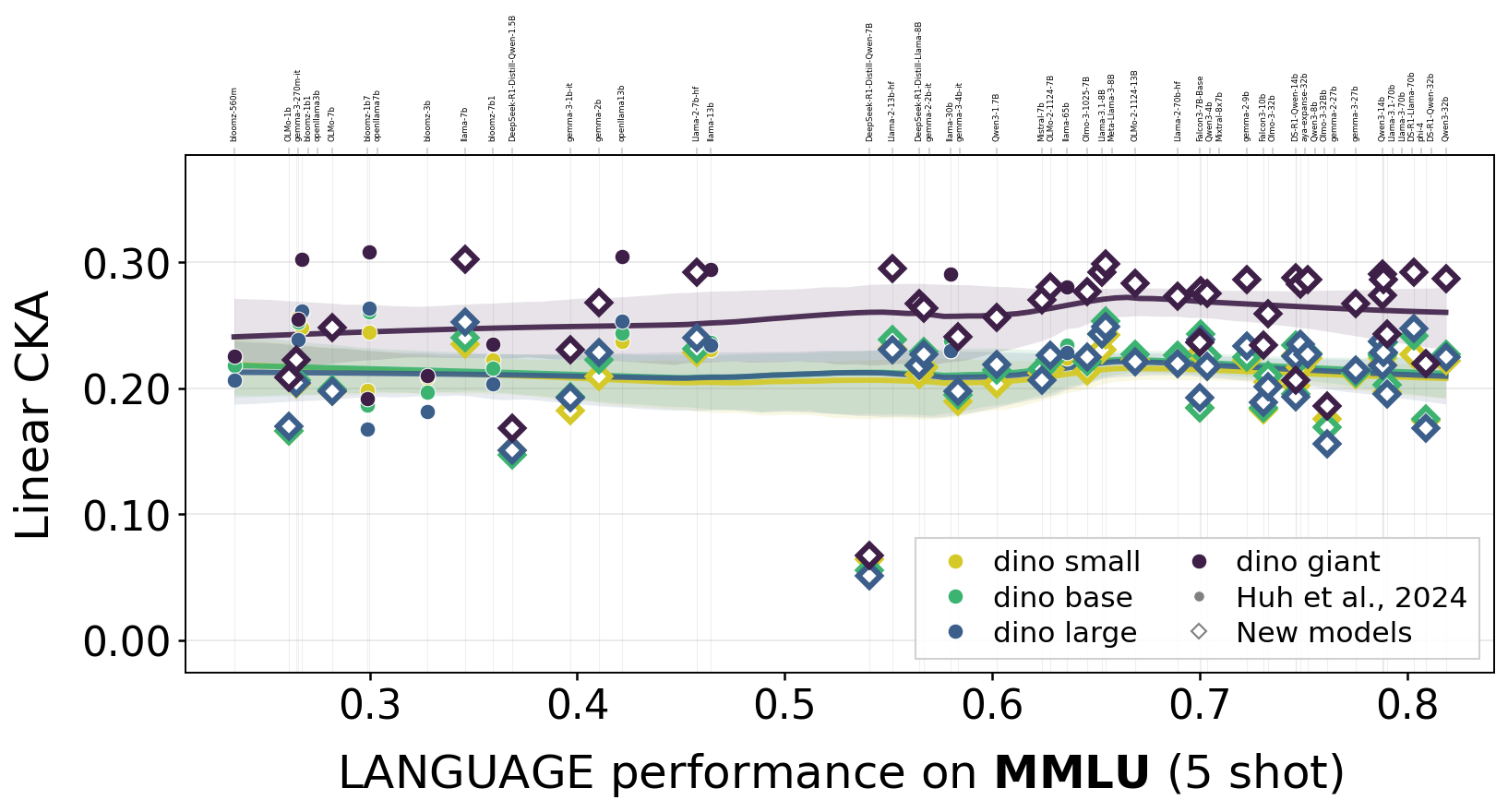} \\
    \end{tabular}
    \caption{Calibrated global RSA (rows 1, 3) and calibrated linear CKA (rows 2, 4) between LLMs and four DINOv2 variants on WIT-1024, plotted against language performance on the reasoning benchmarks GSM8K, ARC, and MMLU (columns). The top two rows show linear fits to the 19 \textit{base models} from~\cite{huh2024prh} (circles; dashed lines), evaluated on the recent models (diamonds). $R^2_{\text{avg}}$ values are averaged over the four DINOv2 variants. The bottom two rows show LOWESS fits over all 55 models with bootstrap confidence bands. \textit{As with mutual $k$NN (\cref{fig:trends_bubble_page1,fig:trends_bubble_page2}), recent models do not continue the extrapolated trend under either global metric: $R^2_{\text{avg}}(\text{new})$ is negative throughout, and the LOWESS fits flatten over the capability range covered by the recent models.}}
    \label{fig:trends_global_page1}
\end{figure*}

\begin{figure*}[p]
    \centering
    \setlength{\tabcolsep}{1.5pt}
    \begin{tabular}{@{}>{\centering\arraybackslash}m{1.2em}@{\hskip 2pt}m{0.315\textwidth}m{0.315\textwidth}m{0.315\textwidth}@{}}
    \rotatebox[origin=c]{90}{\scriptsize RSA, linear fit} &
    \includegraphics[width=\linewidth]{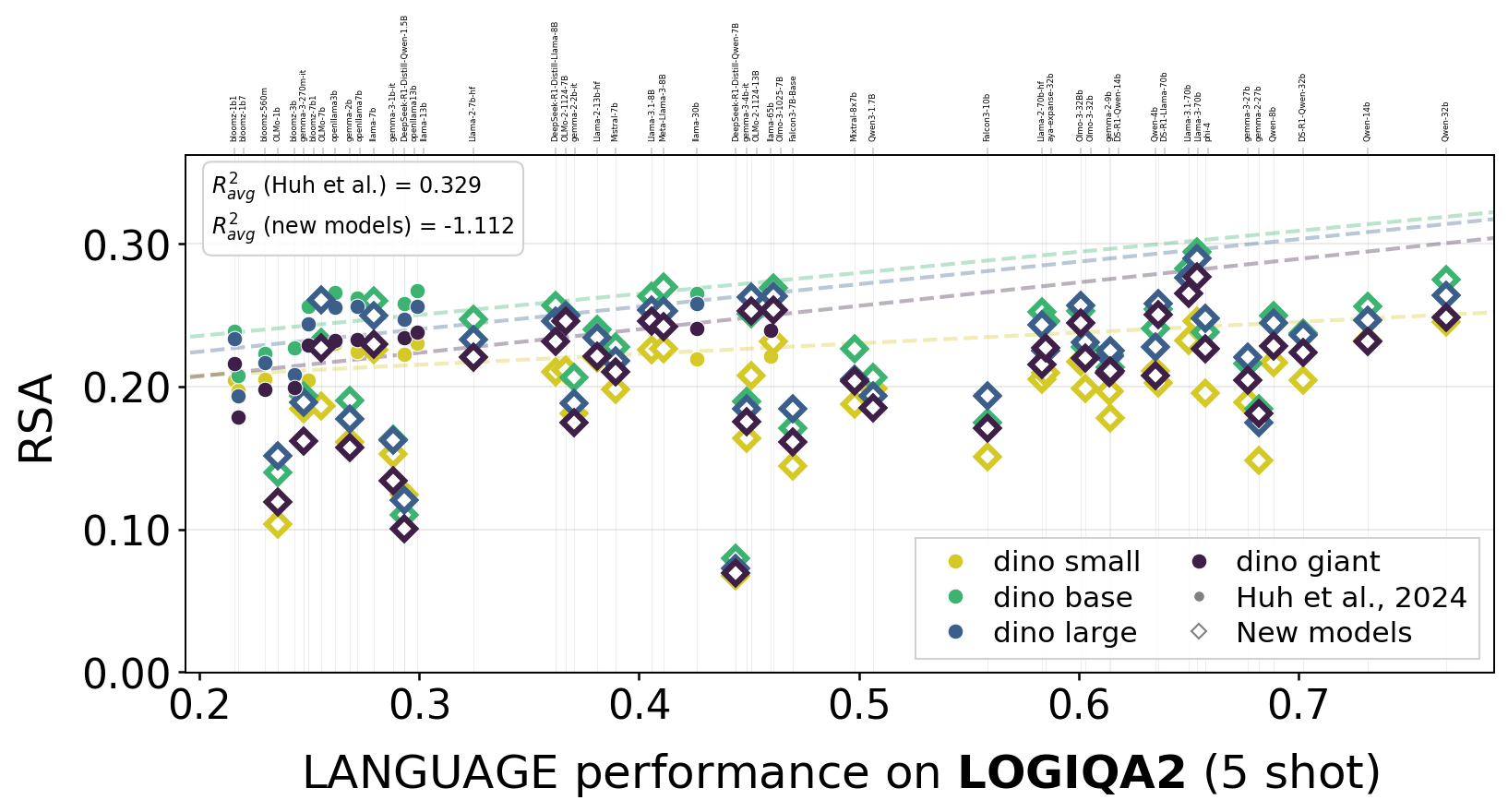} &
    \includegraphics[width=\linewidth]{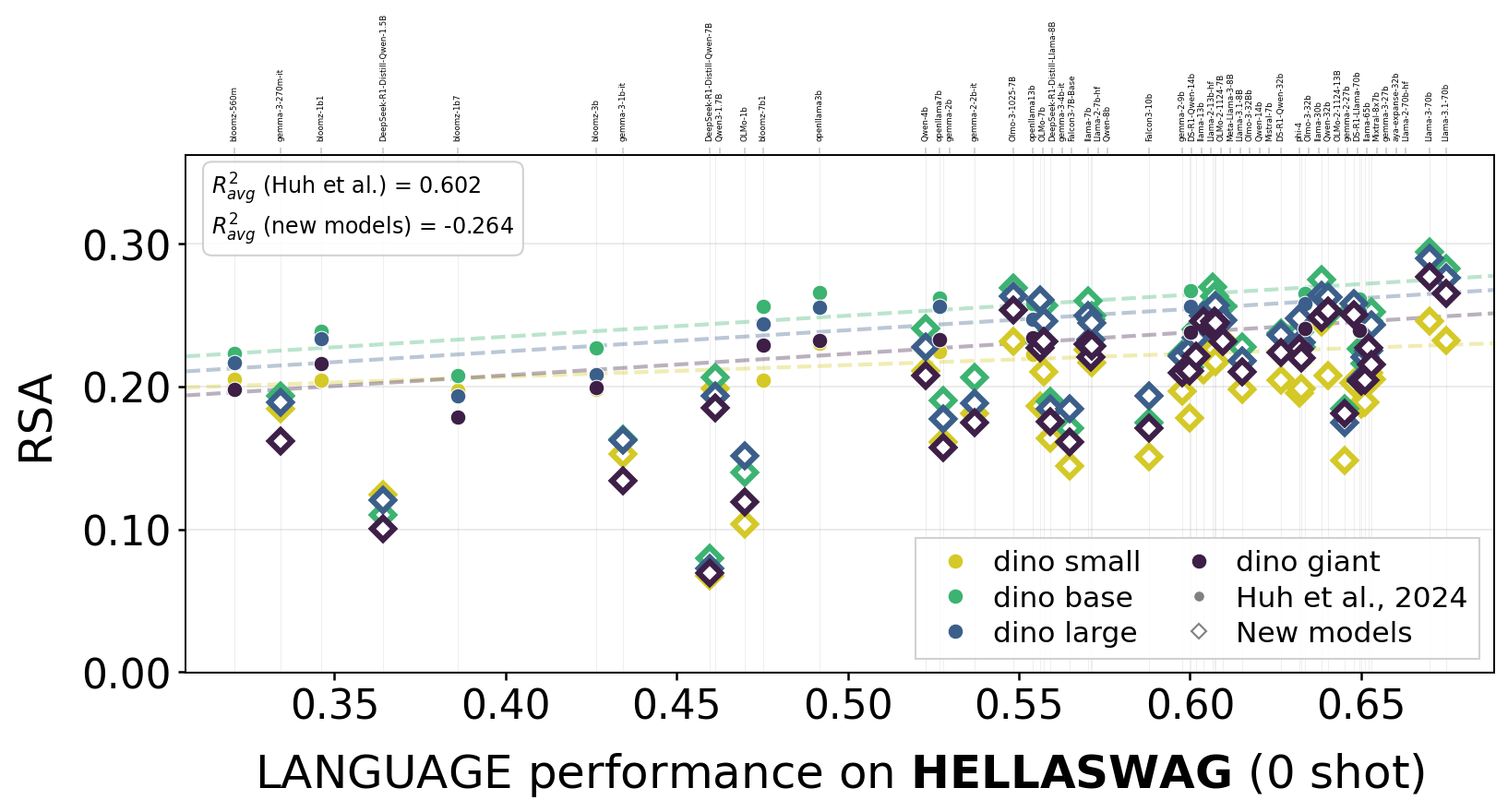} &
    \includegraphics[width=\linewidth]{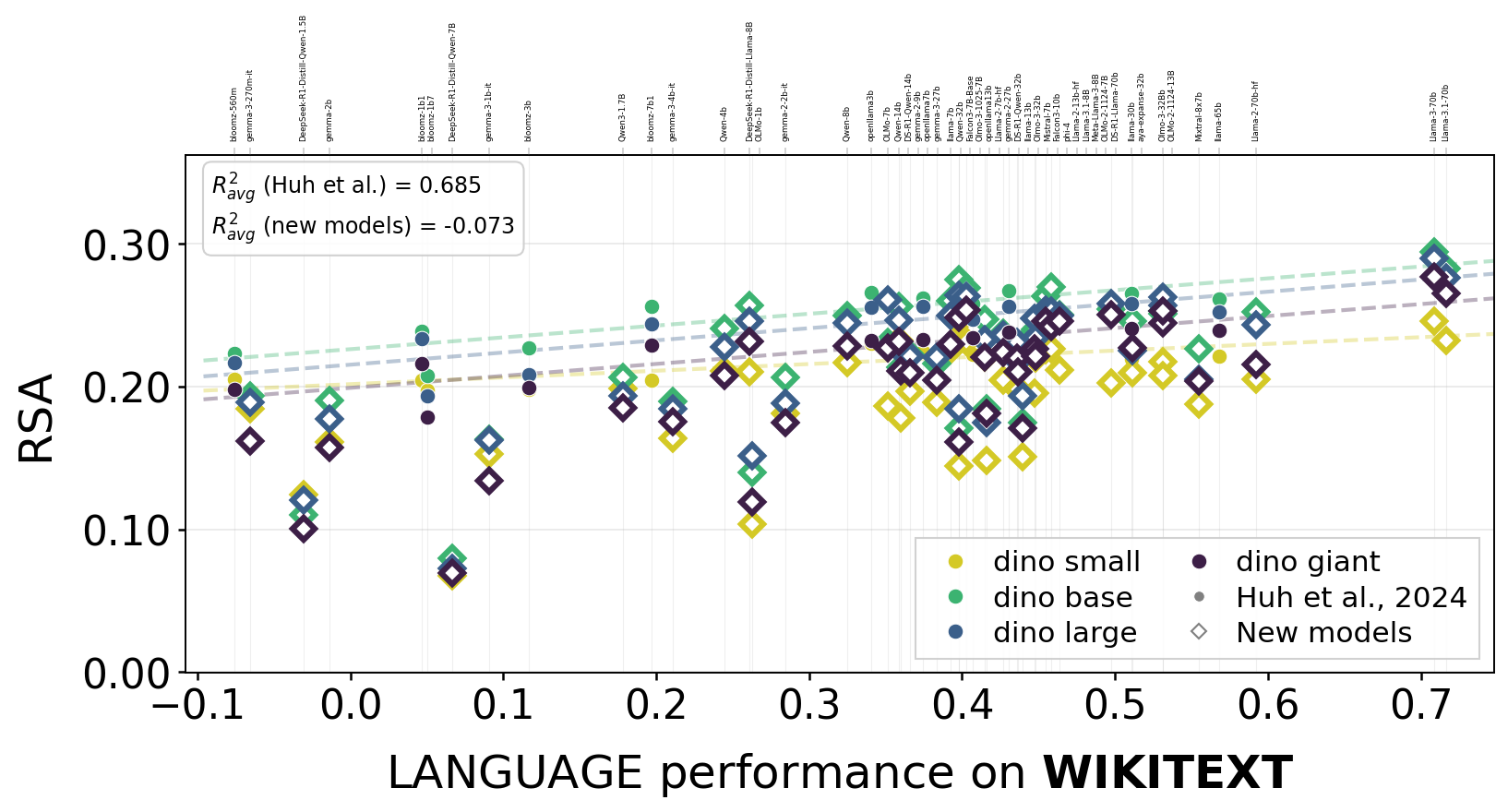} \\[2pt]
    \rotatebox[origin=c]{90}{\scriptsize CKA, linear fit} &
    \includegraphics[width=\linewidth]{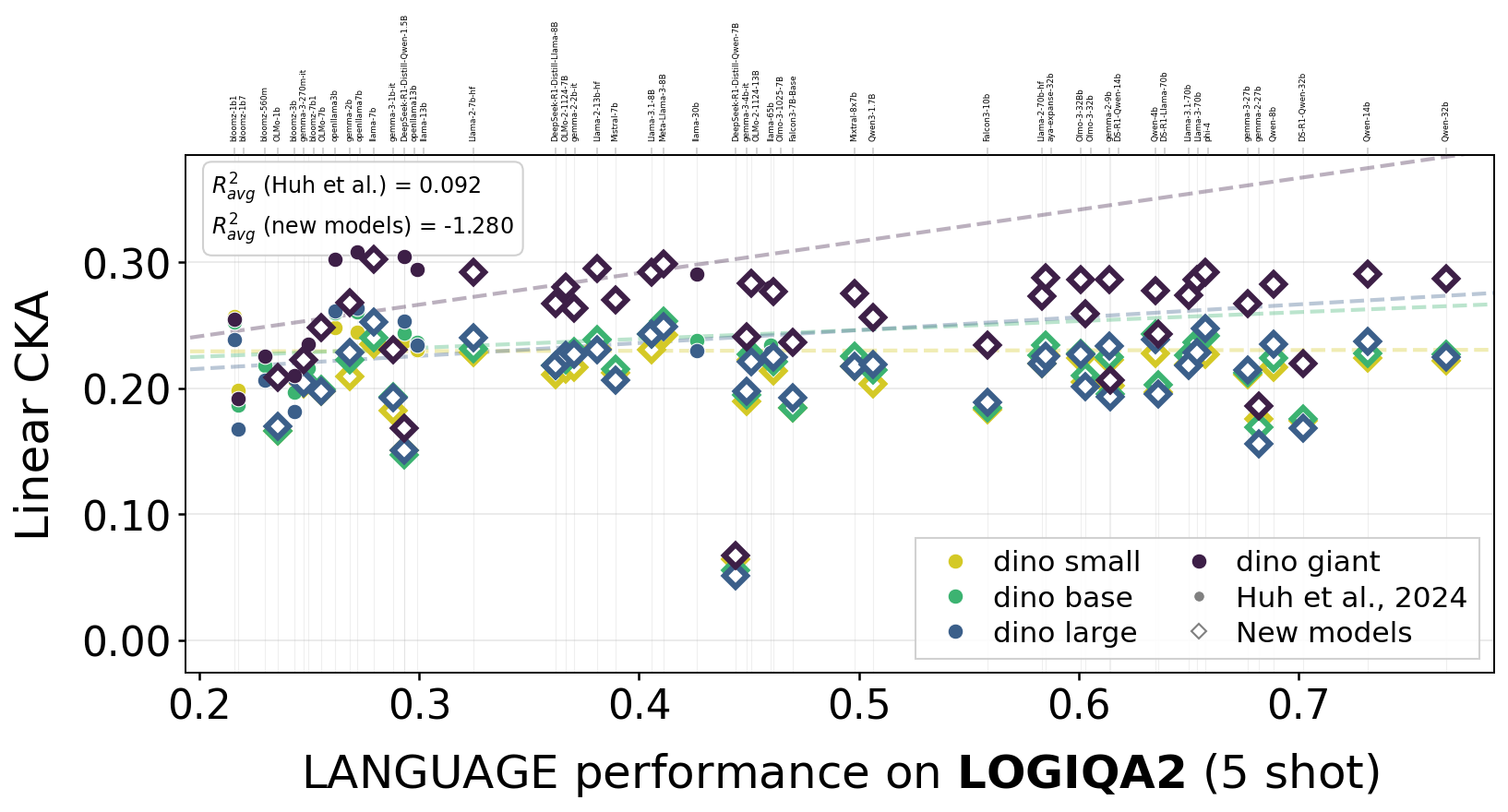} &
    \includegraphics[width=\linewidth]{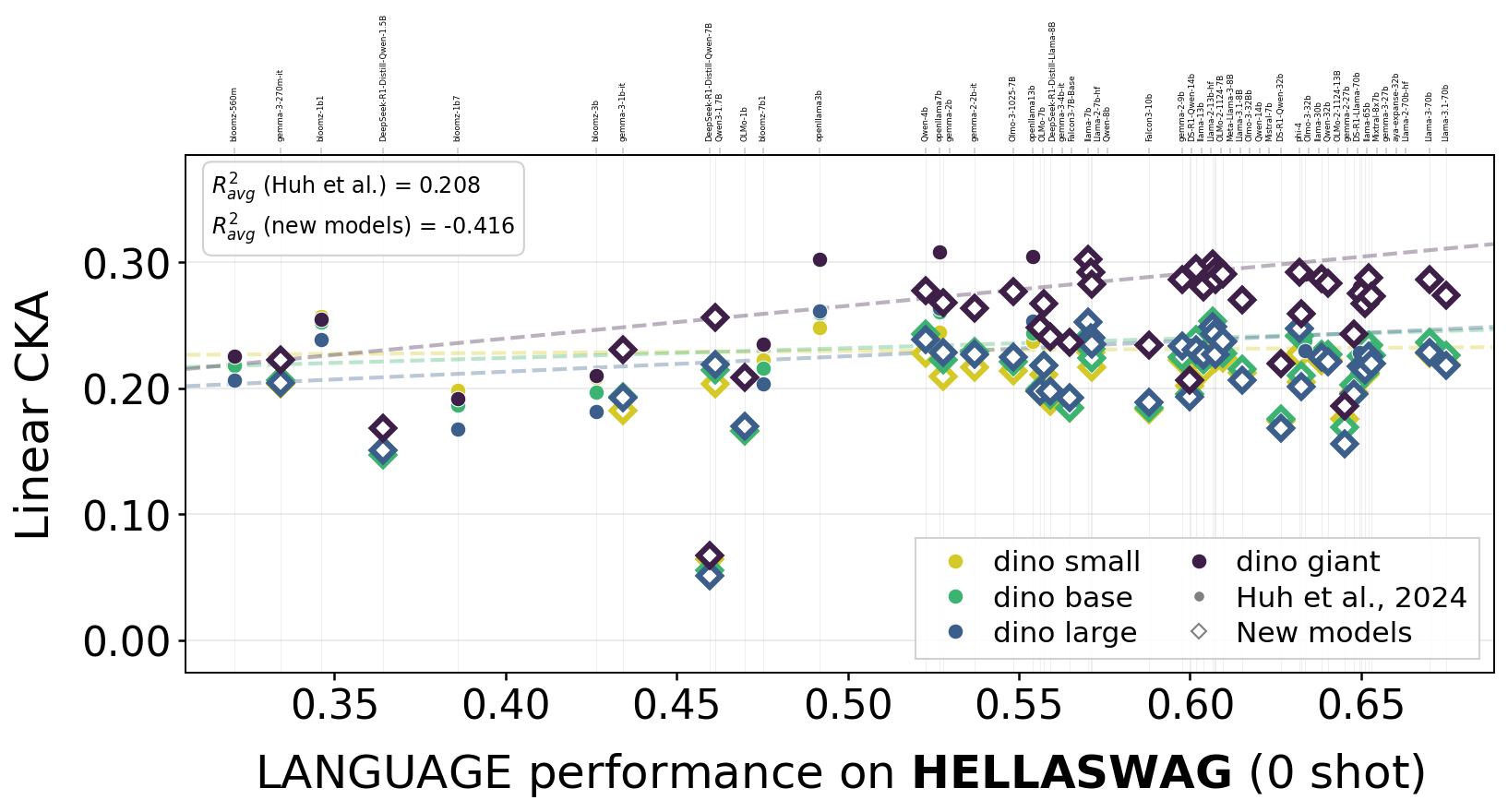} &
    \includegraphics[width=\linewidth]{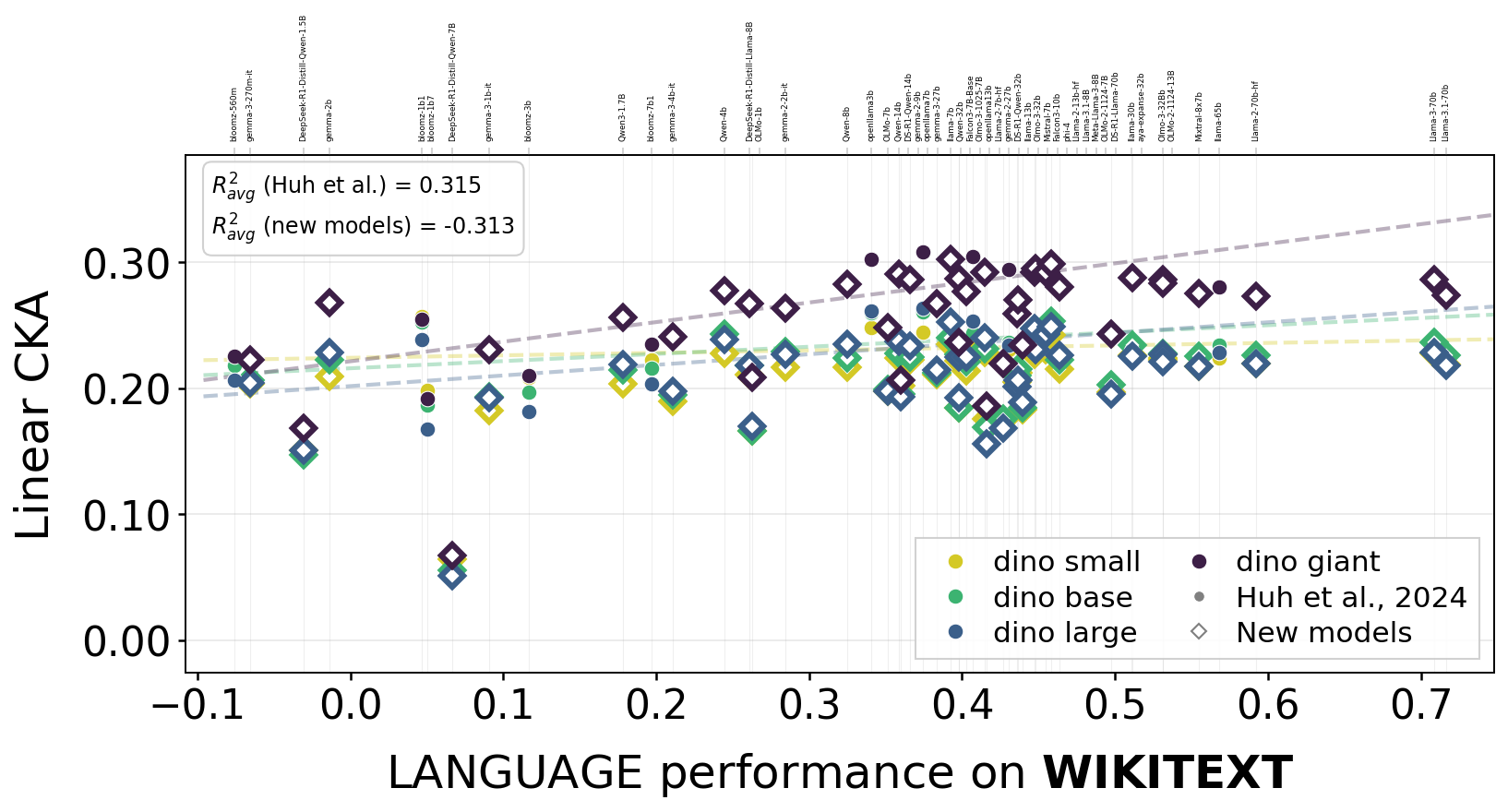} \\[2pt]
    \rotatebox[origin=c]{90}{\scriptsize RSA, LOWESS} &
    \includegraphics[width=\linewidth]{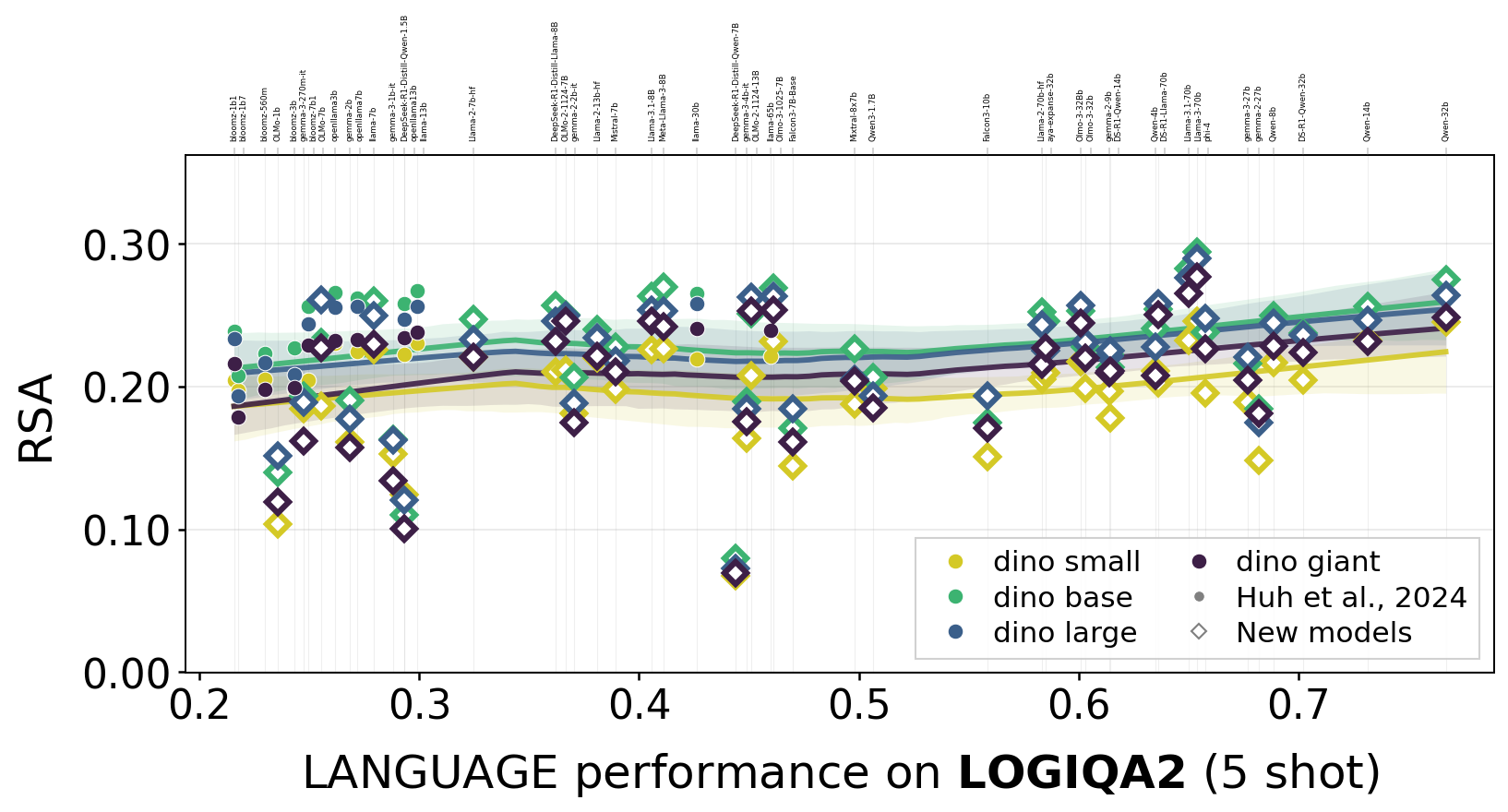} &
    \includegraphics[width=\linewidth]{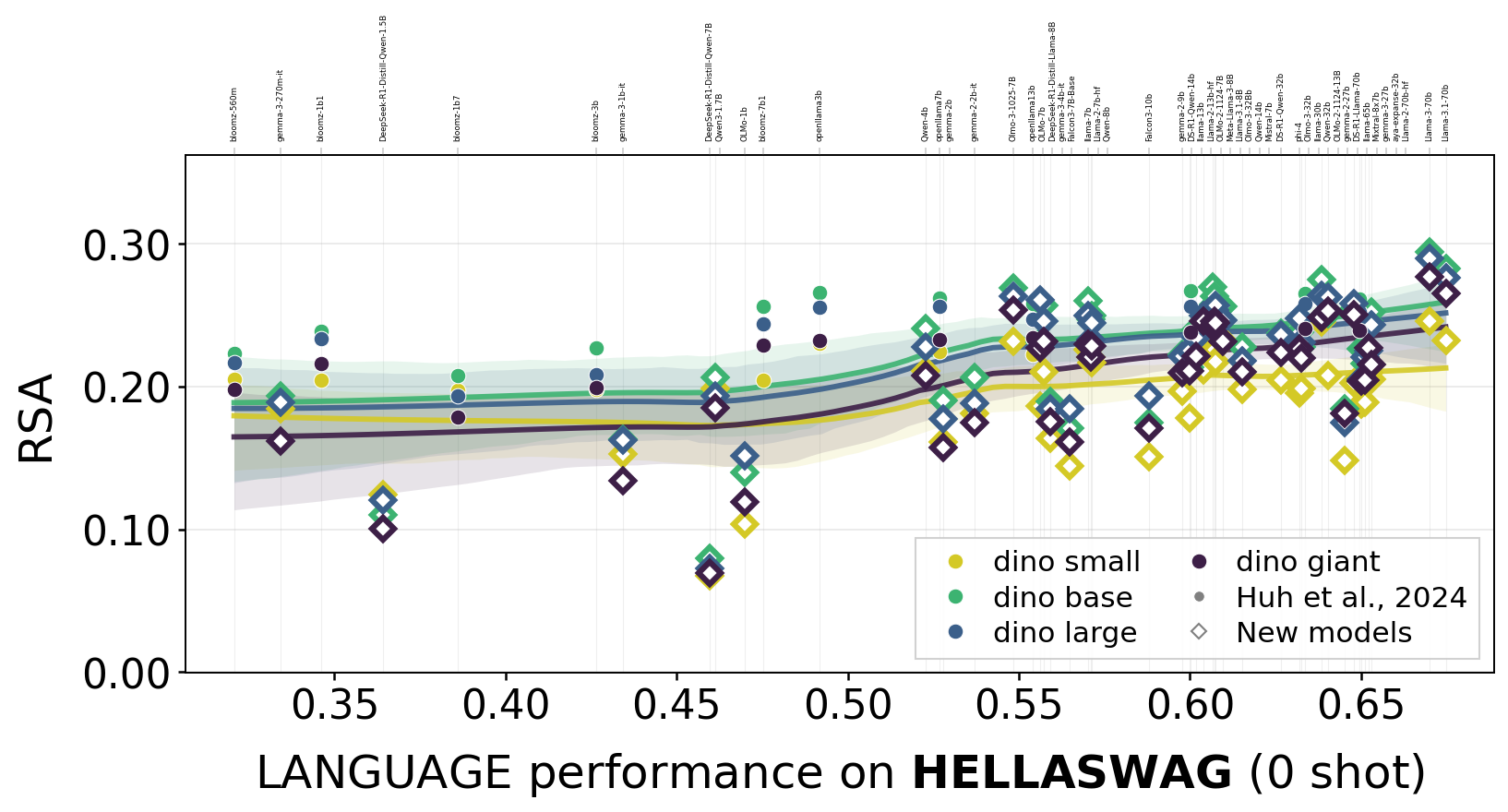} &
    \includegraphics[width=\linewidth]{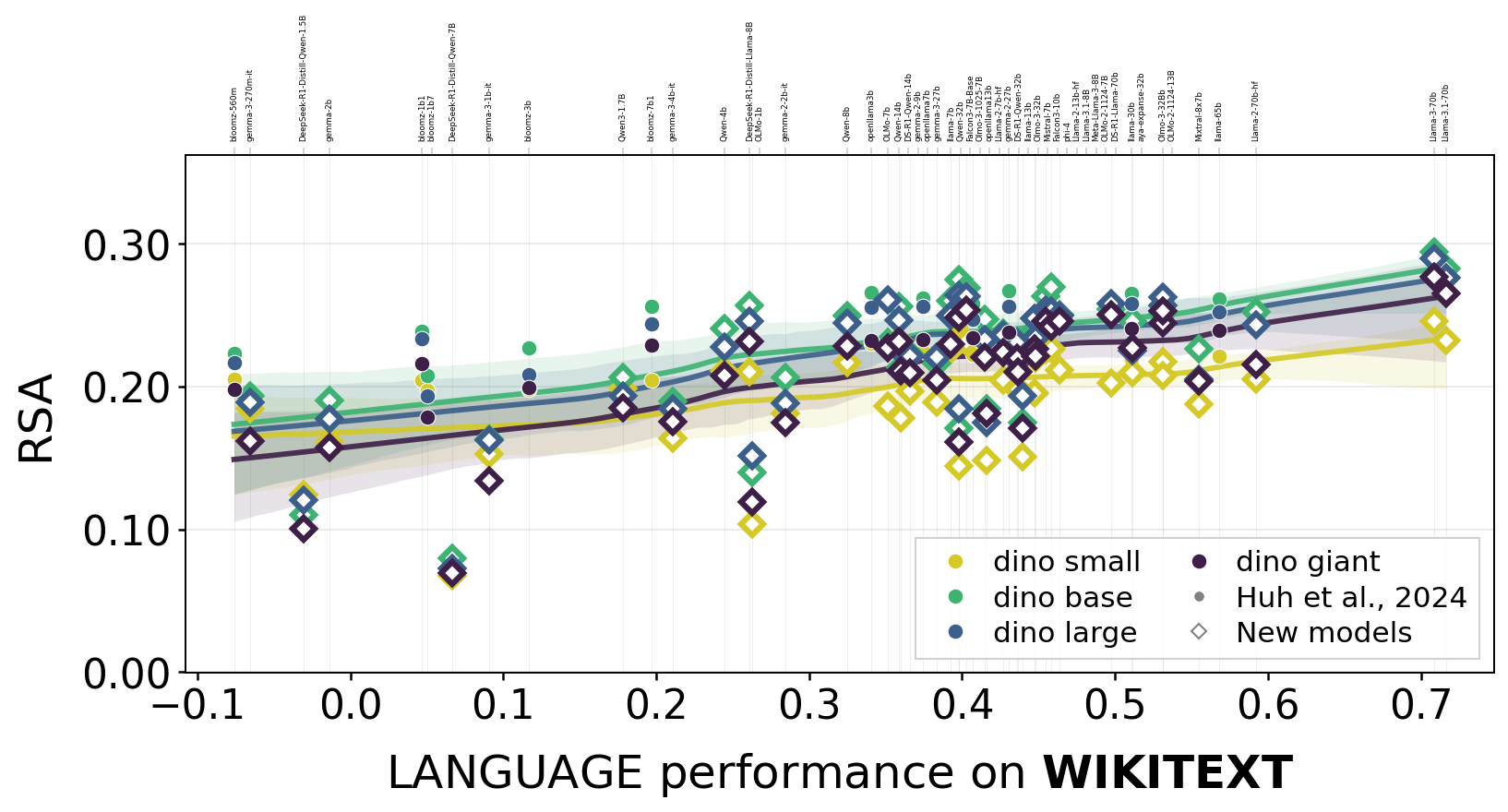} \\[2pt]
    \rotatebox[origin=c]{90}{\scriptsize CKA, LOWESS} &
    \includegraphics[width=\linewidth]{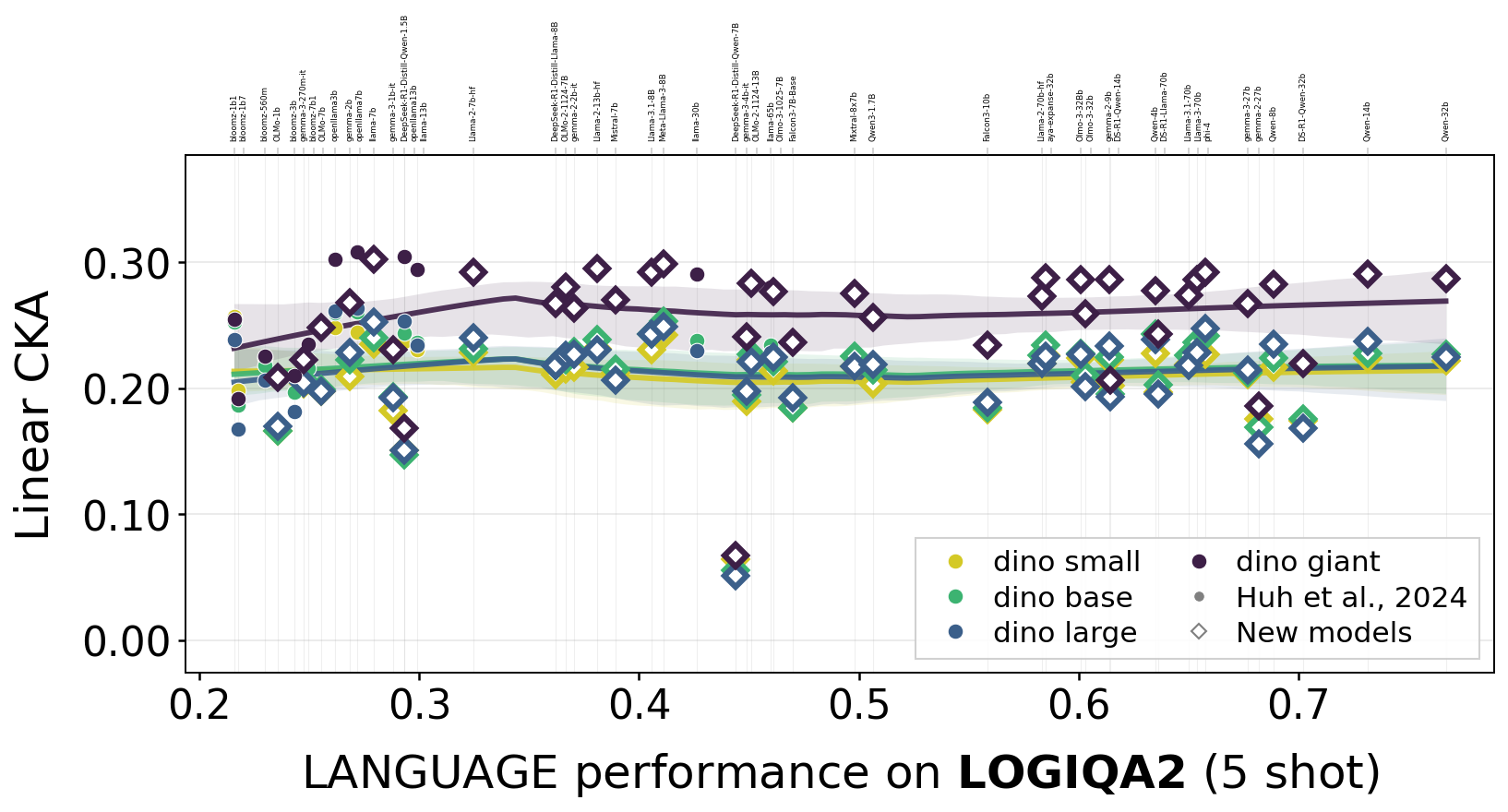} &
    \includegraphics[width=\linewidth]{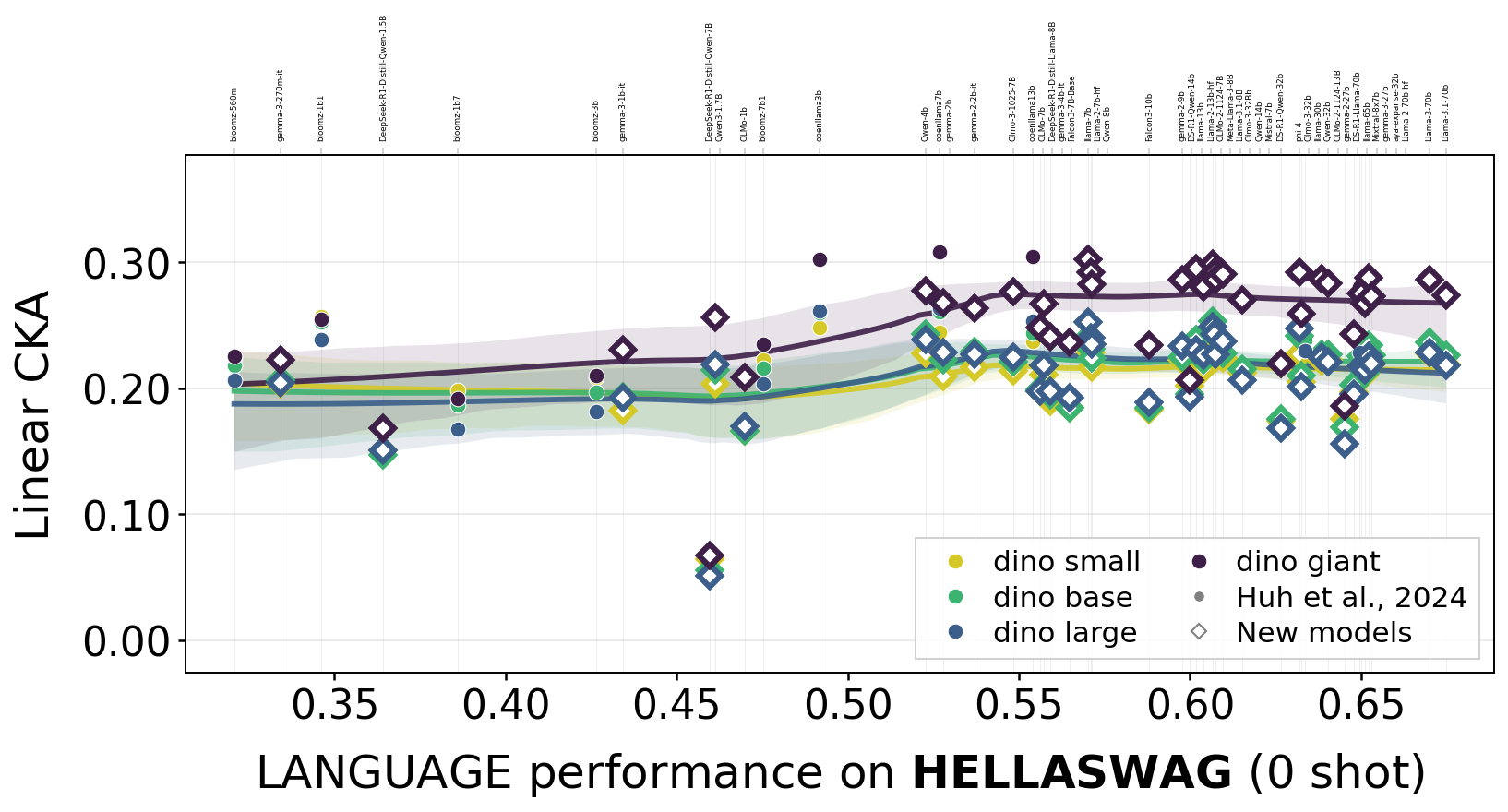} &
    \includegraphics[width=\linewidth]{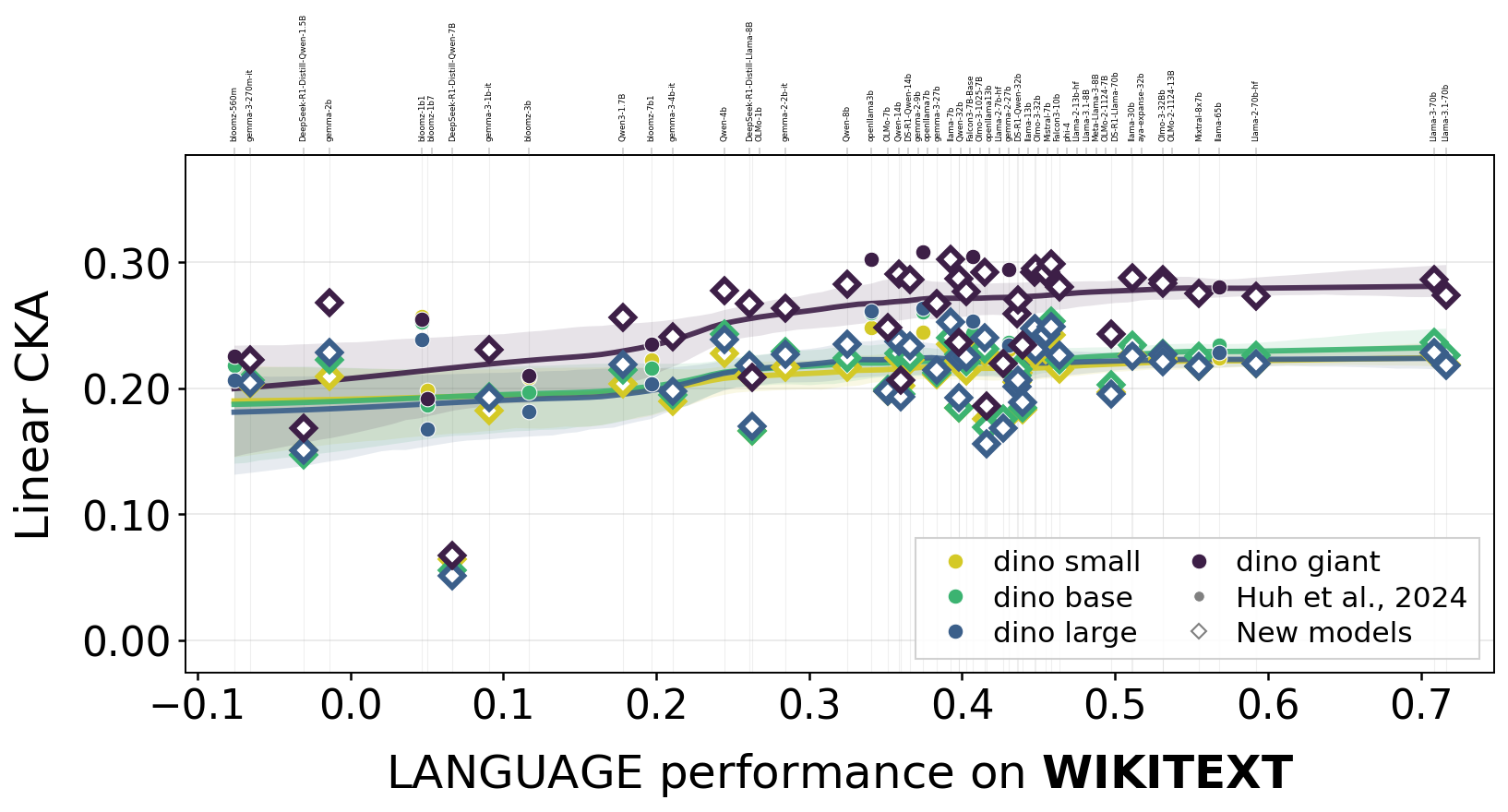} \\
    \end{tabular}
    \caption{Calibrated global RSA (rows 1, 3) and calibrated linear CKA (rows 2, 4) between LLMs and four DINOv2 variants on WIT-1024, against language performance on LogiQA2 and the core language-modelling benchmarks HellaSwag and Wikitext (columns); layout as in \cref{fig:trends_global_page1}. \textit{Unlike mutual $k$NN, for which HellaSwag and Wikitext partially extended the trend to recent models ($R^2_{\text{avg}}(\text{new})$ of $0.30$ and $0.49$, \cref{tab:r2_benchmarks}), the global metrics fail to extrapolate on the core language-modelling benchmarks as well.}}
    \label{fig:trends_global_page2}
\end{figure*}

The saturation persists under both global metrics
(\cref{fig:trends_global_page1,fig:trends_global_page2},
\cref{tab:r2_global_metrics}). On the reasoning benchmarks,
$R^2_{\text{avg}}(\text{new})$ is consistently negative, ranging from $-0.621$
(MMLU) to $-1.426$ (GSM8K) for RSA and from $-0.487$ (MMLU) to $-1.280$
(LogiQA2) for CKA. The linear fit from the \textit{base models} is again a worse
predictor of alignment for recent models than simply predicting their mean.
Notably, the global metrics fail to extrapolate even on the core
language-modelling benchmarks, where mutual $k$NN partially extended the trend
($R^2_{\text{avg}}(\text{new})$ of $0.30$ for HellaSwag and $0.49$ for Wikitext,
\cref{tab:r2_benchmarks}): under RSA and CKA, HellaSwag and Wikitext drop to
$-0.264$/$-0.073$ and $-0.416$/$-0.313$, respectively. As a nonparametric check
that does not presuppose a linear relationship, we additionally fit LOWESS
curves over all models (bottom rows). The smoothers flatten over the
capability range covered by the recent models on every benchmark, for both
metrics and all four DINOv2 variants, confirming that the plateau is not an
artifact of the linear functional form.
The LOWESS estimation also makes it easier to see that there is no growing trend across the majority of benchmarks.

We note that for CKA the trend on the \textit{base models} was weak to begin
with ($R^2_{\text{avg}}(\text{Huh~\etal})$ between $0.047$ and $0.315$),
consistent with \cite{huh2024prh}, who found CKA to reveal only a ``weak trend
of alignment between models'' and introduced mutual $k$NN for this reason. The
CKA result should therefore be read not as a strong trend breaking down, but as
no capability trend emerging under CKA for recent models either. For RSA, the
\textit{base-model} trend is substantial on the benchmarks used by
\cite{huh2024prh} ($R^2_{\text{avg}}(\text{Huh~\etal})$ of $0.602$ on HellaSwag
and $0.685$ on Wikitext), and it is this trend that the recent models do not
continue.

\textbf{The failure to extrapolate is not an artifact of the mutual $k$NN
metric. Calibrated global RSA and CKA show the same saturation for recent models
on all six benchmarks, and, unlike mutual $k$NN, they fail to extrapolate even
on the core language-modelling benchmarks.}

\section{Experimental setup}
In \cref{sec:wit_laion}, we provide additional details about the deduplication pipeline for the WIT-1M and LAION-15M datasets. We then describe the captioning pipeline used for WIT-1M-recap and for the ImageNet validation set in \cref{sec:captioning}.

\subsection{WIT-1M and LAION-15M datasets}\label{sec:wit_laion}
\subsubsection{Image deduplication.} We deduplicate the gallery pools from WIT~\citep{srinivasan2021wit} and LAION400M~\citep{schuhmann2021laion} at the image level using perceptual hashing~\citep{zauner2010implementation}.  For each image, a 64-bit hash $\mathbf{h}_i \in \{0,1\}^{64}$ is computed. For this, we first convert the image to grayscale, resize it to $32\times32$, apply a 2D Discrete Cosine Transform, and threshold the top-left $8\times8$ low-frequency coefficients against their median. This produces a binary fingerprint that is robust to minor changes, e.g.\ due to recompression. We consider images $i$ and $j$ duplicates if their Hamming distance satisfies
$$d_{\text{H}}(\mathbf{h}_i, \mathbf{h}_j) \leq 2,$$ measuring the number of bit positions at which two hashes differ: $$d_{\text{H}}(\mathbf{h}_i, \mathbf{h}_j) = \sum_{b=1}^{64} \mathbf{1}[h_{i,b} \neq h_{j,b}].$$

We perform deduplication of the gallery against the WIT-1024 images, and within the gallery by keeping the first occurrence in the case of image duplicates.

\subsubsection{Caption deduplication.}
In addition to image deduplication, we do a text deduplication pass to remove gallery samples with captions identical to another gallery sample or to a WIT-1024 query caption. Duplicate captions are undesirable because they allow trivial text-based query-gallery matching, inflating retrieval scores regardless of visual representations.\\

We use exact string matching and remove any gallery samples that match WIT-1024 captions. Among the gallery samples, we discard duplicates and keep only the first occurrence of a duplicate caption-sample.

\begin{table}[t]
\centering
\caption{Deduplication statistics for the WIT-1M and LAION-15M gallery pools.
  Image duplicates are detected via perceptual hashing (pHash) with Hamming distance $\leq 2$.
  Caption duplicates are detected by exact string matching. WIT-1M and LAION-15M are 1M and 15M image-caption pairs randomly sampled from the remaining final pool.}
\label{tab:dedup_stats}
  \resizebox{0.5\textwidth}{!}{
\begin{tabular}{l@{\hspace{1.5em}}c@{\hspace{1.5em}}c}
\toprule
 & \textbf{WIT-1M} & \textbf{LAION-15M} \\
\midrule
Raw pool & 3{,}582{,}610 & 20{,}000{,}000 \\
\midrule
\multicolumn{3}{l}{\textit{Image deduplication}} \\
\quad Duplicates (with WIT-1024)   &      2{,}847 &          53 \\
\quad Duplicates (within gallery) & 2{,}164{,}343 & 3{,}371{,}128 \\
Pool after image deduplication     & 2{,}486{,}852 & 17{,}941{,}016 \\
\midrule
\multicolumn{3}{l}{\textit{Caption deduplication}} \\
\quad Duplicates (with WIT-1024)   &          52 &          14 \\
\quad Duplicates (within gallery) &     97{,}654 &    642{,}895 \\
\midrule
\textbf{Final pool size}                 & \textbf{2{,}389{,}146} & \textbf{17{,}298{,}107} \\
\bottomrule
\bottomrule
\end{tabular}
}
\end{table}

\begin{table}[t]
\centering
\caption{Distribution of caption duplicates in the WIT and LAION galleries after image deduplication. Note that the ``unique captions'' include some captions that are removed as query matches for the final pool.
}
\label{tab:textdup_combined}
\setlength{\tabcolsep}{12pt}
\resizebox{0.8\linewidth}{!}{%
\begin{tabular}{l@{\hspace{20pt}}cc@{\hspace{20pt}}cc}
\toprule
 & \multicolumn{2}{c}{\textbf{WIT}} & \multicolumn{2}{c}{\textbf{LAION}} \\
\cmidrule(lr){2-3} \cmidrule(lr){4-5}
Copies per caption & Unique captions & Total samples & Unique captions & Total samples \\
\midrule
1            & 2{,}357{,}353 & 2{,}357{,}353 & 16{,}897{,}221 & 16{,}897{,}221 \\
2            &    22{,}799   &    45{,}598   &    316{,}109   &    632{,}218   \\
3            &     3{,}807   &    11{,}421   &     48{,}864   &    146{,}592   \\
4            &     1{,}529   &     6{,}116   &     15{,}422   &     61{,}688   \\
5            &       787     &     3{,}935   &      6{,}932   &     34{,}660   \\
6--10        &     1{,}520   &    11{,}431   &      9{,}596   &     69{,}614   \\
11--100      &     1{,}283   &    29{,}765   &      3{,}838   &     76{,}200   \\
101--1{,}000 &       118     &    27{,}164   &        103     &     12{,}376   \\
$>$1{,}000   &         2     &     5{,}077   &          4     &     14{,}588   \\
\midrule
Total        & 2{,}389{,}198 & 2{,}486{,}852 & 17{,}298{,}111 & 17{,}941{,}016 \\
\bottomrule
\bottomrule
\end{tabular}}
\end{table}

\subsubsection{WIT-1M.} We obtain a raw pool of 3,582,610 samples from the English-text WIT dataset~\citep{srinivasan2021wit}. To construct the English-only subset of the dataset, we used a subset of \cite{singh2022flava}.
Since \cite{pmd_datset} only contain the image URLs, we retrieved the corresponding images from \cite{srinivasan2021wit}.
The raw pool undergoes our deduplication pipeline, resulting in 2,389,146 samples. We randomly sampled 1 million samples for the WIT-1M dataset. Deduplication statistics are provided in \cref{tab:dedup_stats} and \cref{tab:textdup_combined}. In WIT, 31{,}845 captions appear repeatedly,
accounting for 129{,}499 samples (5.21\%).
The most frequent captions are \texttt{coat of arms} (2716$\times$)
and \texttt{Town hall} (2361$\times$).

\subsubsection{LAION-15M.} We randomly sample 20M samples from the LAION-400M dataset~\citep{schuhmann2021laion} which consists of English image-text pairs. We randomly sample 20M samples as our raw pool which undergoes our deduplication pipeline. Finally, we randomly sample 15M from the final pool after deduplication, resulting in our LAION-15M data pool. Deduplication statistics are provided in \cref{tab:dedup_stats} and \cref{tab:textdup_combined}. In LAION, 400{,}890 unique captions appear repeatedly, accounting for 1{,}043{,}795 samples (5.82\%). The most frequent captions are \texttt{Patent Drawing} (10027$\times$)
and \texttt{Throw Pillow} (3246$\times$).

\subsection{Captioning pipeline for the ImageNet validation set and WIT-1M-recap}\label{sec:captioning}

We used gemini-3-flash-preview~\citep{team2023gemini,pichai2025gemini3} for captioning the images in the ImageNet validation~\citep{deng2009imagenet} set and in WIT-1M. Specifically, we used the following text prompt for each image.
\medskip

\begin{lstlisting}[
  basicstyle=\ttfamily\scriptsize,
  breaklines=true,
  frame=single,
  backgroundcolor=\color{gray!10},
  rulecolor=\color{gray!40}
]
You are a precise image description system. Describe the image in the following JSON format.
Return ONLY a valid JSON object with exactly these 7 keys. No text before or after the JSON.
{
  "one_sentence": "<exactly one sentence, strictly fewer than 15 words>",
  "short":        "<2-3 sentences, 20-40 words total>",
  "100w":         "<a paragraph, approximately 100 words>",
  "250w":         "<several paragraphs, approximately 250 words>",
  "500w":         "<detailed description, approximately 500 words>",
  "750w":         "<thorough description covering all visual details, approximately 750 words>",
  "extreme_long": "<maximally detailed description covering every visible element, texture, color, spatial relationship, lighting, and context. YOU MUST WRITE AT LEAST 1000 WORDS. If your draft is under 1000 words, keep adding more detail about textures, materials, lighting, spatial layout, colors, and any other visible elements until you reach at least 1000 words.
  Target 1000-1500 words.>"
}
Be factual and visual. Describe what you actually see: objects, people, animals, colors, textures, spatial relationships, background, lighting, and mood. Do not invent information not visible in the image.
\end{lstlisting}
\medskip

For WIT-1M-recap, we caption the 1 million images in the WIT-1M dataset using the \texttt{500w} variant for computational efficiency. This resulted in captions for 999,971 images (29 images did not get processed by gemini-3-flash-preview) of on average 478 words.

\subsection{Models and feature extraction pipeline}\label{sec:models_extraction}

Our feature extraction pipeline is based on the experimental protocol from \cite{huh2024prh}, which we extend to include additional LLMs.

\subsubsection{Vision models.}
On the vision side, we use four DINOv2~\citep{oquab2024dinov2learningrobustvisual} variants: ViT-S/14 (384-d), ViT-B/14 (768-d), ViT-L/14 (1024-d), and ViT-G/14 (1536-d) \citep{dosovitskiy2021an}, loaded via the \texttt{timm} \citep{rwightman2019timm} library. For each image, we extract the CLS token representation from every transformer layer, yielding a per-sample feature tensor of shape $L \times d$, where $L$ is the number of layers and $d$ the feature dimensionality.

\subsubsection{Language models.}\label{sec:all-llm-models}
We evaluate 55 large language models spanning 13 model families. The first group comprises the 19 \textit{base models} used by \cite{huh2024prh}: BLOOMZ (560M--7.1B)~\citep{muennighoff2023crosslingualgeneralizationmultitaskfinetuning}, OpenLlama (3B--13B)~\citep{openlm2023openllama}, LLaMA (7B--65B)~\citep{touvron2023llama}, OLMo (1B, 7B)~\citep{groeneveld2024olmoacceleratingsciencelanguage}, Gemma (2B, 7B)~\citep{gemma_2024}, Mistral-7B~\citep{jiang2023mistral}, Mixtral-8$\times$7B~\citep{jiang2024mixtral_experts}, and Meta-Llama-3-70B~\citep{grattafiori2024llama}.\\
For the trend analysis in \cref{sec:llm_eval}, we extend this set with 36 recent models: LLaMA-2 (7B--70B)~\citep{touvron2023llama2}, Llama-3/3.1~\citep{grattafiori2024llama}, OLMo-2/3 ~\citep{olmo20252olmo2furious}, Gemma-2 (2B--27B)\citep{gemma_2_2024}, Gemma-3 (270M--27B) \citep{gemma_3_2025}, DeepSeek-R1-Distill (1.5B--70B)~\citep{deepseek_r1_2025}, Qwen3 (1.7B--32B)~\citep{qwen3_2025}, Falcon3 (7B, 10B)~\citep{Falcon3}, 01.AI Yi-1.5 (34B)~\citep{yi2024}, IBM Granite (8B)~\citep{granite33_2025}, Phi-4 (15B) \citep{abdin2024phi4technicalreport},  Cohere Aya Expanse (32B) \citep{dang2024ayaexpansecombiningresearch} and OpenAI GPT-OSS (20B)~\citep{gpt_oss_2025}.\\
For each model, we extract hidden-state representations from all layers. Following \cite{huh2024prh}, we apply average pooling over non-padding tokens to obtain a single vector per layer.

\section{Additional qualitative results}
\begin{figure}[h]
    \centering
\includegraphics[width=1\textwidth]{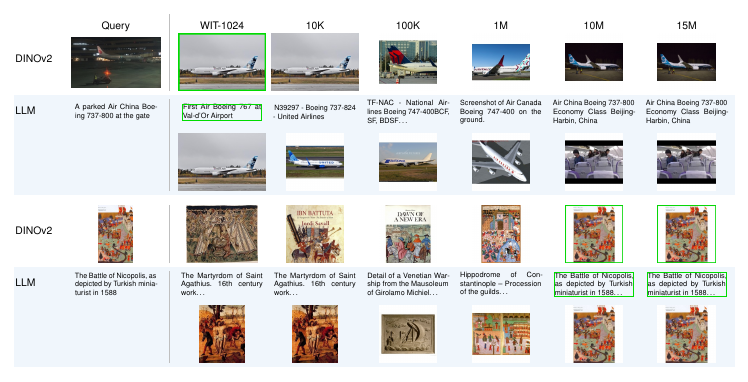}
    \vspace{-2em}
    \caption{Nearest-neighbor ($k{=}1$) examples with DINOv2 and OpenLlama across gallery scales on LAION-15M. \textit{As the gallery densifies, each model finds closer but different matches (top example). The match at 15M (bottom right) is a near-duplicate that survived our deduplication pipeline.}}
    \label{fig:laion_transfer_examples}
    \vspace{-1.3em}
\end{figure}

\begin{figure}[t]
    \centering
   \includegraphics[width=1\textwidth]{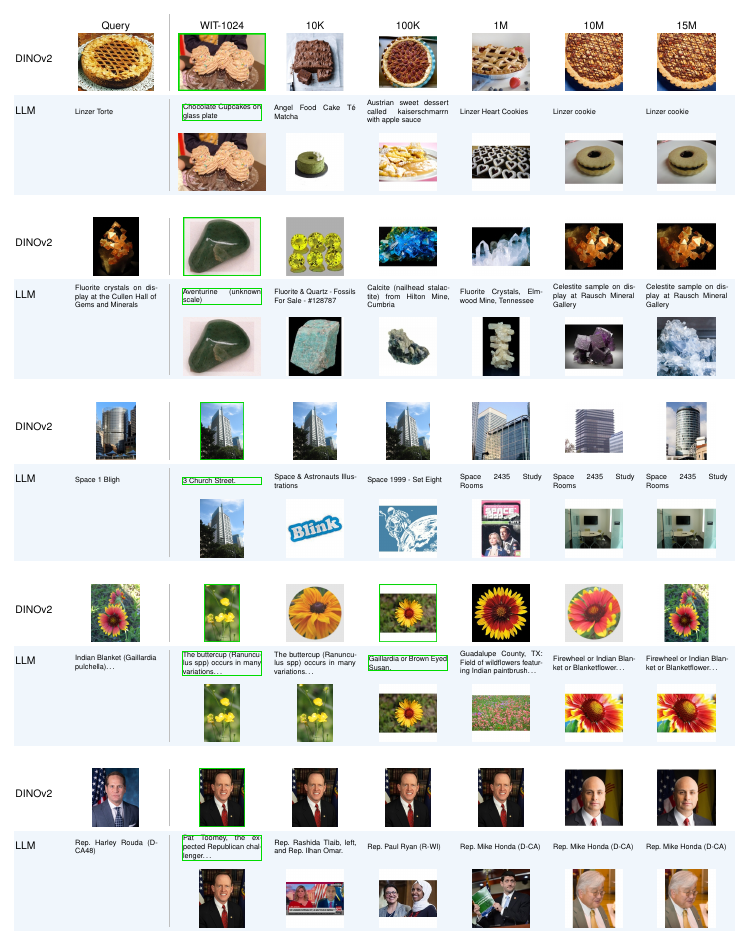}
   \vspace{-2em}
    \caption{Additional nearest-neighbor examples with DINOv2 and OpenLlama-3b for $k{=}1$ across gallery scales on the LAION-15M dataset. Green-bordered captions and images indicate a mutual $k$NN match across modalities.}
    \label{fig:laion_transfer_examples_2}
\end{figure}

\begin{figure}[t]
    \centering
   \includegraphics[width=1\textwidth]{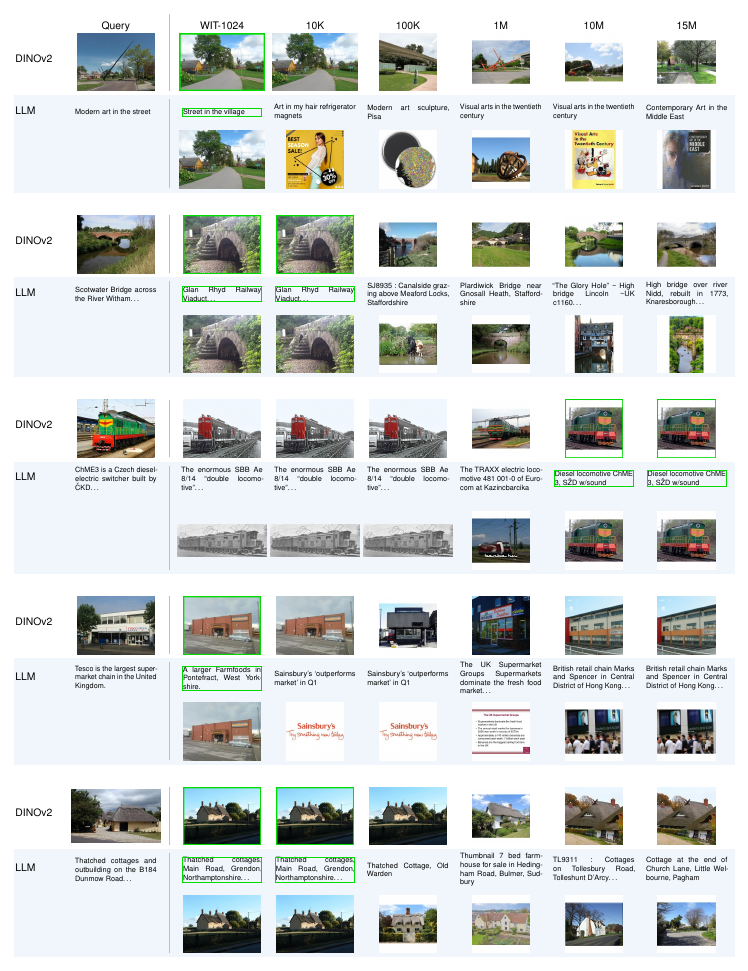}
   \vspace{-2em}
    \caption{Additional nearest-neighbor examples with DINOv2 and OpenLlama-3b for $k{=}1$ across gallery scales on the LAION-15M dataset. Green-bordered captions and images indicate a mutual $k$NN match across modalities.}
    \label{fig:laion_transfer_examples_3}
\end{figure}

\begin{figure}[t]
    \centering
   \includegraphics[width=1\textwidth]{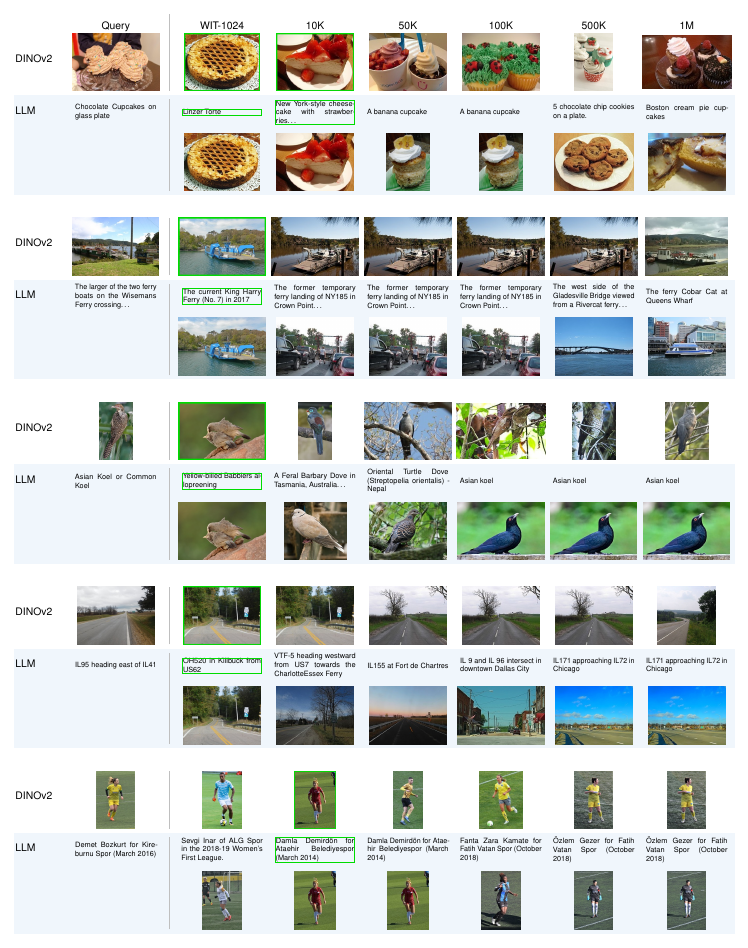}
   \vspace{-2em}
    \caption{Additional nearest-neighbor examples with DINOv2 and OpenLlama-3b for $k{=}1$ across gallery scales on the WIT-1M dataset. For OpenLlama-3b, we show the (partial) retrieved captions along with the corresponding reference image (LLM-ref) for visualization. Green-bordered captions and images indicate a mutual $k$NN match across modalities.}
    \label{fig:wit_transfer_examples}
\end{figure}

\begin{figure}[t]
    \centering
   \includegraphics[width=1\textwidth]{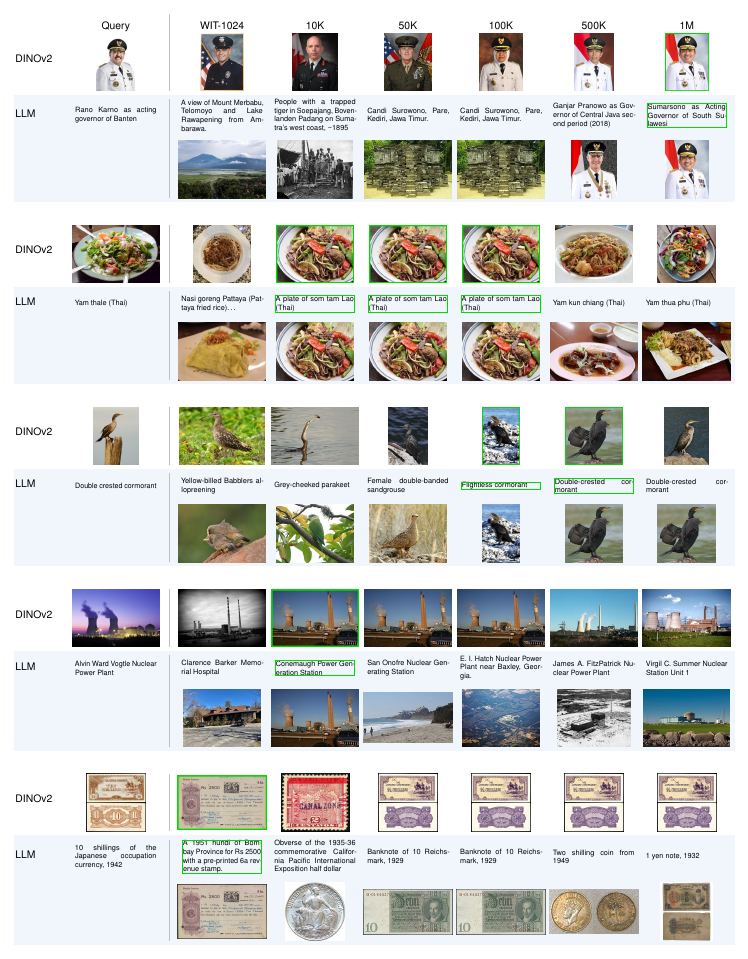}
   \vspace{-2em}
    \caption{Additional nearest-neighbor examples with DINOv2 and OpenLlama-3b for $k{=}1$ across gallery scales on the WIT-1M dataset. Green-bordered captions and images indicate a mutual $k$NN match across modalities.}
    \label{fig:wit_transfer_examples_2}
\end{figure}

We present additional qualitative results for nearest-neighbor retrieval at different gallery scales on LAION-15M and WIT-1M in \cref{fig:laion_transfer_examples,fig:laion_transfer_examples_2,fig:laion_transfer_examples_3} and \cref{fig:wit_transfer_examples,fig:wit_transfer_examples_2} respectively. Here, we show further examples for cross-modal agreement (green-bordered matches) at scale when the modalities happen to select the same neighbor. Others show agreement at WIT-1024 that breaks down as the gallery densifies. In those cases, each modality individually finds a better match at scale, but they no longer agree on the same one.

\clearpage
\newpage
\putbib[main]
\end{bibunit}


\begin{thebibliography}{87}
\providecommand{\natexlab}[1]{#1}
\providecommand{\url}[1]{\texttt{#1}}
\expandafter\ifx\csname urlstyle\endcsname\relax
  \providecommand{\doi}[1]{doi: #1}\else
  \providecommand{\doi}{doi: \begingroup \urlstyle{rm}\Url}\fi

\bibitem[Ahn et~al.(2022)Ahn, Brohan, Brown, Chebotar, Cortes, David, Finn, Fu, Gopalakrishnan, Hausman, et~al.]{ahn2022can}
Michael Ahn, Anthony Brohan, Noah Brown, Yevgen Chebotar, Omar Cortes, Byron David, Chelsea Finn, Chuyuan Fu, Keerthana Gopalakrishnan, Karol Hausman, et~al.
\newblock Do as i can, not as i say: Grounding language in robotic affordances.
\newblock \emph{arXiv preprint arXiv:2204.01691}, 2022.

\bibitem[Aky{\"u}rek et~al.(2024)Aky{\"u}rek, Damani, Zweiger, Qiu, Guo, Pari, Kim, and Andreas]{akyurek2024surprising}
Ekin Aky{\"u}rek, Mehul Damani, Adam Zweiger, Linlu Qiu, Han Guo, Jyothish Pari, Yoon Kim, and Jacob Andreas.
\newblock The surprising effectiveness of test-time training for few-shot learning.
\newblock \emph{arXiv preprint arXiv:2411.07279}, 2024.

\bibitem[Bahng et~al.(2025)Bahng, Chan, Durand, and Isola]{bahng2025cycle}
Hyojin Bahng, Caroline Chan, Fredo Durand, and Phillip Isola.
\newblock Cycle consistency as reward: Learning image-text alignment without human preferences.
\newblock \emph{arXiv preprint arXiv:2506.02095}, 2025.

\bibitem[Bai et~al.(2025)Bai, Cai, Chen, Chen, Chen, Cheng, Deng, Ding, Gao, Ge, et~al.]{baiQwen3VLTechnicalReport2025}
Shuai Bai, Yuxuan Cai, Ruizhe Chen, Keqin Chen, Xionghui Chen, Zesen Cheng, Lianghao Deng, Wei Ding, Chang Gao, Chunjiang Ge, et~al.
\newblock Qwen3-vl technical report.
\newblock \emph{arXiv preprint arXiv:2511.21631}, 2025.

\bibitem[Balestriero et~al.(2018)]{balestriero2018spline}
Randall Balestriero et~al.
\newblock A spline theory of deep learning.
\newblock In \emph{ICML}, 2018.

\bibitem[Bansal et~al.(2021)Bansal, Nakkiran, and Barak]{bansal2021revisitingmodelstitchingcompare}
Yamini Bansal, Preetum Nakkiran, and Boaz Barak.
\newblock Revisiting model stitching to compare neural representations.
\newblock In \emph{NeurIPS}, 2021.

\bibitem[Bender \& Koller(2020)Bender and Koller]{bender-koller-2020-climbing}
Emily~M. Bender and Alexander Koller.
\newblock Climbing towards {NLU}: {On} meaning, form, and understanding in the age of data.
\newblock In Dan Jurafsky, Joyce Chai, Natalie Schluter, and Joel Tetreault (eds.), \emph{Proceedings of the 58th Annual Meeting of the Association for Computational Linguistics}, 2020.

\bibitem[Bolya et~al.(2025)Bolya, Huang, Sun, Cho, Madotto, Wei, Ma, Zhi, Rajasegaran, Rasheed, et~al.]{bolya2025perception}
Daniel Bolya, Po-Yao Huang, Peize Sun, Jang~Hyun Cho, Andrea Madotto, Chen Wei, Tengyu Ma, Jiale Zhi, Jathushan Rajasegaran, Hanoona Rasheed, et~al.
\newblock Perception encoder: The best visual embeddings are not at the output of the network.
\newblock \emph{arXiv preprint arXiv:2504.13181}, 2025.

\bibitem[Browning \& LeCun(2022)Browning and LeCun]{Yann22}
Jacob Browning and Yann LeCun.
\newblock Ai and the limits of language.
\newblock \emph{Noema Magazine}, 2022.

\bibitem[Chai et~al.(2025)Chai, Song, Du, Meng, Madhavan, Bar-Tal, Hwang, Xie, and Manning]{chai2025auroracapefficientperformantvideo}
Wenhao Chai, Enxin Song, Yilun Du, Chenlin Meng, Vashisht Madhavan, Omer Bar-Tal, Jenq-Neng Hwang, Saining Xie, and Christopher~D. Manning.
\newblock Auroracap: Efficient, performant video detailed captioning and a new benchmark.
\newblock In \emph{ICLR}, 2025.

\bibitem[Chollet(2019)]{chollet2019measureintelligence}
François Chollet.
\newblock On the measure of intelligence.
\newblock \emph{arXiv preprint arXiv:1911.01547}, 2019.

\bibitem[Clark et~al.(2018)Clark, Cowhey, Etzioni, Khot, Sabharwal, Schoenick, and Tafjord]{clark2018think}
Peter Clark, Isaac Cowhey, Oren Etzioni, Tushar Khot, Ashish Sabharwal, Carissa Schoenick, and Oyvind Tafjord.
\newblock Think you have solved question answering? try arc, the ai2 reasoning challenge.
\newblock \emph{arXiv preprint arXiv:1803.05457}, 2018.

\bibitem[Cobbe et~al.(2021)Cobbe, Kosaraju, Bavarian, Chen, Jun, Kaiser, Plappert, Tworek, Hilton, Nakano, Hesse, and Schulman]{cobbe2021gsm8k}
Karl Cobbe, Vineet Kosaraju, Mohammad Bavarian, Mark Chen, Heewoo Jun, Lukasz Kaiser, Matthias Plappert, Jerry Tworek, Jacob Hilton, Reiichiro Nakano, Christopher Hesse, and John Schulman.
\newblock Training verifiers to solve math word problems.
\newblock \emph{arXiv preprint arXiv:2110.14168}, 2021.

\bibitem[Dar(2025)]{dar2026minivec2vecscalinguniversalgeometry}
Guy Dar.
\newblock mini-vec2vec: Scaling universal geometry alignment with linear transformations.
\newblock \emph{arXiv preprint arXiv:2510.02348}, 2025.

\bibitem[Deng et~al.(2009)Deng, Dong, Socher, Li, Li, and Fei-Fei]{deng2009imagenet}
Jia Deng, Wei Dong, Richard Socher, Li-Jia Li, Kai Li, and Li~Fei-Fei.
\newblock Imagenet: A large-scale hierarchical image database.
\newblock In \emph{CVPR}, 2009.

\bibitem[Dinkel et~al.(2024)Dinkel, Yan, Wang, Zhang, Wang, and Wang]{dinkel2024dasheng}
Heinrich Dinkel, Zhiyong Yan, Yongqing Wang, Junbo Zhang, Yujun Wang, and Bin Wang.
\newblock Scaling up masked audio encoder learning for general audio classification.
\newblock In \emph{Interspeech}, 2024.

\bibitem[Douze et~al.(2024)Douze, Guzhva, Deng, Johnson, Szilvasy, Mazaré, Lomeli, Hosseini, and Jégou]{douze2024faiss}
Matthijs Douze, Alexandr Guzhva, Chengqi Deng, Jeff Johnson, Gergely Szilvasy, Pierre-Emmanuel Mazaré, Maria Lomeli, Lucas Hosseini, and Hervé Jégou.
\newblock The faiss library.
\newblock \emph{arXiv preprint arXiv:2401.08281}, 2024.

\bibitem[Dravid et~al.(2023)Dravid, Gandelsman, Efros, and Shocher]{dravid2023rosetta}
Amil Dravid, Yossi Gandelsman, Alexei~A Efros, and Assaf Shocher.
\newblock Rosetta neurons: Mining the common units in a model zoo.
\newblock In \emph{ICCV}, 2023.

\bibitem[Edelman(1998)]{edelman1998representation}
Shimon Edelman.
\newblock Representation is representation of similarities.
\newblock \emph{Behavioral and brain sciences}, 1998.

\bibitem[Gemma~Team(2024)]{gemma_2_2024}
Google~DeepMind Gemma~Team.
\newblock Gemma 2: Improving open language models at a practical size.
\newblock \emph{arXiv preprint arXiv:2408.00118}, 2024.

\bibitem[Geng \& Liu(2023)Geng and Liu]{openlm2023openllama}
Xinyang Geng and Hao Liu.
\newblock Openllama: An open reproduction of llama, 2023.
\newblock URL \url{https://github.com/openlm-research/open_llama}.

\bibitem[Gibson(1979)]{gibson1979ecological}
James~J Gibson.
\newblock \emph{The Ecological Approach to Visual Perception}.
\newblock Houghton Mifflin, Boston, 1979.
\newblock ISBN 978-0898593019.

\bibitem[Grattafiori et~al.(2024)Grattafiori, Dubey, Jauhri, Pandey, Kadian, Al-Dahle, Letman, Mathur, Schelten, Vaughan, et~al.]{grattafiori2024llama}
Aaron Grattafiori, Abhimanyu Dubey, Abhinav Jauhri, Abhinav Pandey, Abhishek Kadian, Ahmad Al-Dahle, Aiesha Letman, Akhil Mathur, Alan Schelten, Alex Vaughan, et~al.
\newblock The llama 3 herd of models.
\newblock \emph{arXiv preprint arXiv:2407.21783}, 2024.

\bibitem[Gröger et~al.(2026)Gröger, Wen, and Brbić]{groger2026revisitingplatonicrepresentationhypothesis}
Fabian Gröger, Shuo Wen, and Maria Brbić.
\newblock Revisiting the platonic representation hypothesis: An aristotelian view.
\newblock \emph{arXiv preprint arXiv:2602.14486}, 2026.

\bibitem[Gu et~al.(2023)Gu, Clark, and Kembhavi]{gu2023can}
Sophia Gu, Christopher Clark, and Aniruddha Kembhavi.
\newblock I can't believe there's no images! learning visual tasks using only language supervision.
\newblock In \emph{ICCV}, 2023.

\bibitem[Guo et~al.(2026)Guo, Litman, Sornwanee, Blanchet, and Ermon]{guo2026bit}
Gabe Guo, Elon Litman, Thanawat Sornwanee, Jose Blanchet, and Stefano Ermon.
\newblock There and back again: Bidirectional diffusion bridges for multimodality translation.
\newblock \emph{arXiv preprint arXiv:2608.27885}, 2026.

\bibitem[Gupta et~al.(2026)Gupta, Kansal, Jegelka, Isola, and Garg]{gupta2026canonicalizingmultimodalcontrastiverepresentation}
Sharut Gupta, Sanyam Kansal, Stefanie Jegelka, Phillip Isola, and Vikas Garg.
\newblock Canonicalizing multimodal contrastive representation learning.
\newblock In \emph{ICLR}, 2026.

\bibitem[Hadgi et~al.(2025)Hadgi, Moschella, Santilli, Gomez, Huang, Rodolà, Melzi, and Ovsjanikov]{hadgi2025escapingplatoscavealignment}
Souhail Hadgi, Luca Moschella, Andrea Santilli, Diego Gomez, Qixing Huang, Emanuele Rodolà, Simone Melzi, and Maks Ovsjanikov.
\newblock Escaping plato's cave: Towards the alignment of 3d and text latent spaces.
\newblock In \emph{CVPR}, 2025.

\bibitem[Hastings~Jr et~al.(1947)Hastings~Jr, Mosteller, Tukey, and Winsor]{hastings1947low}
Cecil Hastings~Jr, Frederick Mosteller, John~W Tukey, and Charles~P Winsor.
\newblock Low moments for small samples: a comparative study of order statistics.
\newblock \emph{The Annals of Mathematical Statistics}, 1947.

\bibitem[Haxby et~al.(2001)Haxby, Gobbini, Furey, Ishai, Schouten, and Pietrini]{haxby2001distributed}
James~V Haxby, M~Ida Gobbini, Maura~L Furey, Alumit Ishai, Jennifer~L Schouten, and Pietro Pietrini.
\newblock Distributed and overlapping representations of faces and objects in ventral temporal cortex.
\newblock \emph{Science}, 2001.

\bibitem[Hendrycks et~al.(2021)Hendrycks, Burns, Basart, Zou, Mazeika, Song, and Steinhardt]{hendrycks2021mmlu}
Dan Hendrycks, Collin Burns, Steven Basart, Andy Zou, Mantas Mazeika, Dawn Song, and Jacob Steinhardt.
\newblock Measuring massive multitask language understanding.
\newblock In \emph{ICLR}, 2021.

\bibitem[Hong et~al.(2025)Hong, Yu, Gu, Wang, Gan, Tang, Cheng, Qi, Ji, Pan, et~al.]{vteam2026glm45vglm41vthinkingversatilemultimodal}
Wenyi Hong, Wenmeng Yu, Xiaotao Gu, Guo Wang, Guobing Gan, Haomiao Tang, Jiale Cheng, Ji~Qi, Junhui Ji, Lihang Pan, et~al.
\newblock Glm-4.5 v and glm-4.1 v-thinking: Towards versatile multimodal reasoning with scalable reinforcement learning.
\newblock \emph{arXiv preprint arXiv:2507.01006}, 2025.

\bibitem[Hotelling(1936)]{hotelling1992relations}
Harold Hotelling.
\newblock Relations between two sets of variates.
\newblock volume~28. [Oxford University Press, Biometrika Trust], 1936.

\bibitem[Hu et~al.(2023{\natexlab{a}})Hu, Storks, Lewis, and Chai]{hu2023context}
Xiaoyang Hu, Shane Storks, Richard~L Lewis, and Joyce Chai.
\newblock In-context analogical reasoning with pre-trained language models.
\newblock In \emph{ACL}, 2023{\natexlab{a}}.

\bibitem[Hu et~al.(2023{\natexlab{b}})Hu, Hua, Yang, Shi, Smith, and Luo]{hu2023promptcap}
Yushi Hu, Hang Hua, Zhengyuan Yang, Weijia Shi, Noah~A Smith, and Jiebo Luo.
\newblock Promptcap: Prompt-guided image captioning for vqa with gpt-3.
\newblock In \emph{ICCV}, 2023{\natexlab{b}}.

\bibitem[Huang et~al.(2023)Huang, Dong, Wang, Hao, Singhal, Ma, Lv, Cui, Mohammed, Patra, Liu, Aggarwal, Chi, Bjorck, Chaudhary, Som, Song, and Wei]{huang2023languageneedaligningperception}
Shaohan Huang, Li~Dong, Wenhui Wang, Yaru Hao, Saksham Singhal, Shuming Ma, Tengchao Lv, Lei Cui, Owais~Khan Mohammed, Barun Patra, Qiang Liu, Kriti Aggarwal, Zewen Chi, Johan Bjorck, Vishrav Chaudhary, Subhojit Som, Xia Song, and Furu Wei.
\newblock Language is not all you need: Aligning perception with language models.
\newblock In \emph{NeurIPS}, 2023.

\bibitem[Huh et~al.(2024)Huh, Cheung, Wang, and Isola]{huh2024prh}
Minyoung Huh, Brian Cheung, Tongzhou Wang, and Phillip Isola.
\newblock The platonic representation hypothesis.
\newblock In \emph{ICML}, 2024.

\bibitem[Isola(2025)]{phil_communication}
Phillip Isola.
\newblock Personal communication, 2025.

\bibitem[Jha et~al.(2025)Jha, Zhang, Shmatikov, and Morris]{jha2025harnessinguniversalgeometryembeddings}
Rishi Jha, Collin Zhang, Vitaly Shmatikov, and John~X. Morris.
\newblock Harnessing the universal geometry of embeddings.
\newblock In \emph{NeurIPS}, 2025.

\bibitem[Jiang et~al.(2024)Jiang, Zhou, and Zhu]{jiang2024tracing}
Jiachen Jiang, Jinxin Zhou, and Zhihui Zhu.
\newblock Tracing representation progression: Analyzing and enhancing layer-wise similarity.
\newblock \emph{arXiv preprint arXiv:2406.14479}, 2024.

\bibitem[Koenderink(2019)]{koenderink2019sentience}
Jan~J. Koenderink.
\newblock \emph{Sentience}.
\newblock De Clootcrans Press, Trajectum, Netherlands, 2019.

\bibitem[Kornblith et~al.(2019)Kornblith, Norouzi, Lee, and Hinton]{kornblith2019similarity}
Simon Kornblith, Mohammad Norouzi, Honglak Lee, and Geoffrey Hinton.
\newblock Similarity of neural network representations revisited.
\newblock In \emph{ICML}, 2019.

\bibitem[Kriegeskorte et~al.(2008)Kriegeskorte, Mur, and Bandettini]{kriegeskorte2008representational}
Nikolaus Kriegeskorte, Marieke Mur, and Peter~A Bandettini.
\newblock Representational similarity analysis-connecting the branches of systems neuroscience.
\newblock \emph{Frontiers in systems neuroscience}, 2008.

\bibitem[Krishna et~al.(2017)Krishna, Zhu, Groth, Johnson, Hata, Kravitz, Chen, Kalantidis, Li, Shamma, et~al.]{krishna2017visual}
Ranjay Krishna, Yuke Zhu, Oliver Groth, Justin Johnson, Kenji Hata, Joshua Kravitz, Stephanie Chen, Yannis Kalantidis, Li-Jia Li, David~A Shamma, et~al.
\newblock Visual genome: Connecting language and vision using crowdsourced dense image annotations.
\newblock \emph{IJCV}, 2017.

\bibitem[Kumar et~al.(2025)Kumar, Clune, Lehman, and Stanley]{kumar2025questioning}
Akarsh Kumar, Jeff Clune, Joel Lehman, and Kenneth~O Stanley.
\newblock Questioning representational optimism in deep learning: The fractured entangled representation hypothesis.
\newblock \emph{arXiv preprint arXiv:2505.11581}, 2025.

\bibitem[LeCun et~al.(2022)]{lecun2022path}
Yann LeCun et~al.
\newblock A path towards autonomous machine intelligence version 0.9. 2, 2022-06-27.
\newblock \emph{Openreview}, 2022.

\bibitem[Lenc \& Vedaldi(2015)Lenc and Vedaldi]{lenc2015understandingimagerepresentationsmeasuring}
Karel Lenc and Andrea Vedaldi.
\newblock Understanding image representations by measuring their equivariance and equivalence.
\newblock In \emph{CVPR}, 2015.

\bibitem[Li et~al.(2024)Li, Kementchedjhieva, Fierro, and S{\o}gaard]{li2024vision}
Jiaang Li, Yova Kementchedjhieva, Constanza Fierro, and Anders S{\o}gaard.
\newblock Do vision and language models share concepts? a vector space alignment study.
\newblock \emph{Transactions of the Association for Computational Linguistics}, 2024.

\bibitem[Li et~al.(2016)Li, Yosinski, Clune, Lipson, and Hopcroft]{li2016convergentlearningdifferentneural}
Yixuan Li, Jason Yosinski, Jeff Clune, Hod Lipson, and John Hopcroft.
\newblock Convergent learning: Do different neural networks learn the same representations?
\newblock In \emph{ICLR}, 2016.

\bibitem[Liang et~al.(2023)Liang, Huang, Xia, Xu, Hausman, Ichter, Florence, and Zeng]{liang2023code}
Jacky Liang, Wenlong Huang, Fei Xia, Peng Xu, Karol Hausman, Brian Ichter, Pete Florence, and Andy Zeng.
\newblock Code as policies: Language model programs for embodied control.
\newblock In \emph{ICRA}, 2023.

\bibitem[Lin et~al.(2025)Lin, Wang, Ren, and Han]{lin2025panoptic}
Kun-Yu Lin, Hongjun Wang, Weining Ren, and Kai Han.
\newblock Panoptic captioning: An equivalence bridge for image and text.
\newblock \emph{NeurIPS}, 2025.

\bibitem[Lin et~al.(2014)Lin, Maire, Belongie, Hays, Perona, Ramanan, Doll{\'a}r, and Zitnick]{lin2014microsoft}
Tsung-Yi Lin, Michael Maire, Serge Belongie, James Hays, Pietro Perona, Deva Ramanan, Piotr Doll{\'a}r, and C~Lawrence Zitnick.
\newblock Microsoft coco: Common objects in context.
\newblock In \emph{ECCV}, 2014.

\bibitem[Liu et~al.(2025)Liu, Zhao, Zhuo, Lin, Xin, Li, Qin, Qiao, Li, and Gao]{liu2025luminamgptilluminateflexiblephotorealistic}
Dongyang Liu, Shitian Zhao, Le~Zhuo, Weifeng Lin, Yi~Xin, Xinyue Li, Qi~Qin, Yu~Qiao, Hongsheng Li, and Peng Gao.
\newblock Lumina-mgpt: Illuminate flexible photorealistic text-to-image generation with multimodal generative pretraining.
\newblock \emph{arXiv preprint arXiv:2408.02657}, 2025.

\bibitem[Liu et~al.(2023)Liu, Liu, Cui, Teng, Duan, Zhou, and Zhang]{liu2023logiqa2}
Hanmeng Liu, Jian Liu, Leyang Cui, Zhiyang Teng, Nan Duan, Ming Zhou, and Yue Zhang.
\newblock Logiqa2.0: The logicqa dataset for logical reasoning.
\newblock \emph{IEEE Transactions on Audio, Speech, and Language Processing}, 2023.

\bibitem[Lu et~al.(2025{\natexlab{a}})Lu, Wang, Xu, Wang, Yang, and Fu]{lu2025representationpotentialsfoundationmodels}
Jianglin Lu, Hailing Wang, Yi~Xu, Yizhou Wang, Kuo Yang, and Yun Fu.
\newblock Representation potentials of foundation models for multimodal alignment: A survey.
\newblock \emph{EMNLP}, 2025{\natexlab{a}}.

\bibitem[Lu et~al.(2025{\natexlab{b}})Lu, Wang, Yang, Zhang, Jenni, and Fu]{lu2026indrarepresentationhypothesismultimodal}
Jianglin Lu, Hailing Wang, Kuo Yang, Yitian Zhang, Simon Jenni, and Yun Fu.
\newblock The indra representation hypothesis for multimodal alignment.
\newblock In \emph{NeurIPS}, 2025{\natexlab{b}}.

\bibitem[Maniparambil et~al.(2024)Maniparambil, Akshulakov, Djilali, Narayan, Seddik, Mangalam, and O'Connor]{maniparambil2024visionlanguageencodersrepresent}
Mayug Maniparambil, Raiymbek Akshulakov, Yasser Abdelaziz~Dahou Djilali, Sanath Narayan, Mohamed El~Amine Seddik, Karttikeya Mangalam, and Noel~E. O'Connor.
\newblock Do vision and language encoders represent the world similarly?
\newblock In \emph{CVPR}, 2024.

\bibitem[Marcos-Manchón \& Fuentemilla(2026)Marcos-Manchón and Fuentemilla]{marcosmanchon2026sharedrepresentationsbrainsmodels}
Pablo Marcos-Manchón and Lluís Fuentemilla.
\newblock Shared representations in brains and models reveal a two-route cortical organization during scene perception.
\newblock \emph{arXiv preprint arXiv:2507.13941}, 2026.

\bibitem[Merullo et~al.(2023)Merullo, Castricato, Eickhoff, and Pavlick]{merullo2023linearlymappingimagetext}
Jack Merullo, Louis Castricato, Carsten Eickhoff, and Ellie Pavlick.
\newblock Linearly mapping from image to text space.
\newblock In \emph{ICLR}, 2023.

\bibitem[{Meta AI}(2025)]{meta_llama4_2025}
{Meta AI}.
\newblock The llama 4 herd: The beginning of a new era of natively multimodal ai innovation, 2025.
\newblock URL \url{https://ai.meta.com/blog/llama-4-multimodal-intelligence/}.

\bibitem[Morcos et~al.(2018)Morcos, Raghu, and Bengio]{morcos2018insightsrepresentationalsimilarityneural}
Ari~S. Morcos, Maithra Raghu, and Samy Bengio.
\newblock Insights on representational similarity in neural networks with canonical correlation.
\newblock In \emph{NeurIPS}, 2018.

\bibitem[Moschella et~al.(2023)Moschella, Maiorca, Fumero, Norelli, Locatello, and Rodolà]{moschella2023relativerepresentationsenablezeroshot}
Luca Moschella, Valentino Maiorca, Marco Fumero, Antonio Norelli, Francesco Locatello, and Emanuele Rodolà.
\newblock Relative representations enable zero-shot latent space communication.
\newblock In \emph{ICLR}, 2023.

\bibitem[Muennighoff et~al.(2023)Muennighoff, Wang, Sutawika, Roberts, Biderman, Scao, Bari, Shen, Yong, Schoelkopf, Tang, Radev, Aji, Almubarak, Albanie, Alyafeai, Webson, Raff, and Raffel]{muennighoff2023crosslingualgeneralizationmultitaskfinetuning}
Niklas Muennighoff, Thomas Wang, Lintang Sutawika, Adam Roberts, Stella Biderman, Teven~Le Scao, M~Saiful Bari, Sheng Shen, Zheng-Xin Yong, Hailey Schoelkopf, Xiangru Tang, Dragomir Radev, Alham~Fikri Aji, Khalid Almubarak, Samuel Albanie, Zaid Alyafeai, Albert Webson, Edward Raff, and Colin Raffel.
\newblock Crosslingual generalization through multitask finetuning.
\newblock In \emph{ACL}, 2023.

\bibitem[Oquab et~al.(2024)Oquab, Darcet, Moutakanni, Vo, Szafraniec, Khalidov, Fernandez, Haziza, Massa, El-Nouby, Assran, Ballas, Galuba, Howes, Huang, Li, Misra, Rabbat, Sharma, Synnaeve, Xu, Jegou, Mairal, Labatut, Joulin, and Bojanowski]{oquab2024dinov2learningrobustvisual}
Maxime Oquab, Timothée Darcet, Théo Moutakanni, Huy Vo, Marc Szafraniec, Vasil Khalidov, Pierre Fernandez, Daniel Haziza, Francisco Massa, Alaaeldin El-Nouby, Mahmoud Assran, Nicolas Ballas, Wojciech Galuba, Russell Howes, Po-Yao Huang, Shang-Wen Li, Ishan Misra, Michael Rabbat, Vasu Sharma, Gabriel Synnaeve, Hu~Xu, Hervé Jegou, Julien Mairal, Patrick Labatut, Armand Joulin, and Piotr Bojanowski.
\newblock Dinov2: Learning robust visual features without supervision.
\newblock \emph{TMLR}, 2024.

\bibitem[Plato(c. 375 BC)]{plato_forms}
Plato.
\newblock Republic.
\newblock c. 375 BC.

\bibitem[Radford et~al.(2021)Radford, Kim, Hallacy, Ramesh, Goh, Agarwal, Sastry, Askell, Mishkin, Clark, Krueger, and Sutskever]{radford2021learningtransferablevisualmodels}
Alec Radford, Jong~Wook Kim, Chris Hallacy, Aditya Ramesh, Gabriel Goh, Sandhini Agarwal, Girish Sastry, Amanda Askell, Pamela Mishkin, Jack Clark, Gretchen Krueger, and Ilya Sutskever.
\newblock Learning transferable visual models from natural language supervision.
\newblock In \emph{ICML}, 2021.

\bibitem[Rosch(1978)]{Rosch1978}
Eleanor Rosch.
\newblock Principles of categorization.
\newblock In Eleanor Rosch and Barbara~Bloom Lloyd (eds.), \emph{Cognition and Categorization}. 1978.

\bibitem[Ruan et~al.(2024)Ruan, Abudula, Liu, Li, Li, Wang, Fan, Ge, Xiao, and Zhu]{ruan2024ndpdistributionpredictionbroad}
Junhao Ruan, Abudukeyumu Abudula, Xinyu Liu, Bei Li, Yinqiao Li, Chenglong Wang, Yuchun Fan, Yuan Ge, Tong Xiao, and Jingbo Zhu.
\newblock Ndp: Next distribution prediction as a more broad target.
\newblock \emph{arXiv preprint arXiv:2408.17377}, 2024.

\bibitem[Schnaus et~al.(2025)Schnaus, Araslanov, and Cremers]{schnaus2025itsblindmatchvisionlanguage}
Dominik Schnaus, Nikita Araslanov, and Daniel Cremers.
\newblock It's a (blind) match! towards vision-language correspondence without parallel data.
\newblock In \emph{CVPR}, 2025.

\bibitem[Schuhmann et~al.(2021)Schuhmann, Vencu, Beaumont, Kaczmarczyk, Mullis, Katta, Coombes, Jitsev, and Komatsuzaki]{schuhmann2021laion}
Christoph Schuhmann, Richard Vencu, Romain Beaumont, Robert Kaczmarczyk, Clayton Mullis, Aarush Katta, Theo Coombes, Jenia Jitsev, and Aran Komatsuzaki.
\newblock Laion-400m: Open dataset of clip-filtered 400 million image-text pairs.
\newblock \emph{arXiv preprint arXiv:2111.02114}, 2021.

\bibitem[Sim{\'e}oni et~al.(2025)Sim{\'e}oni, Vo, Seitzer, Baldassarre, Oquab, Jose, Khalidov, Szafraniec, Yi, Ramamonjisoa, et~al.]{simeoni2025dinov3}
Oriane Sim{\'e}oni, Huy~V Vo, Maximilian Seitzer, Federico Baldassarre, Maxime Oquab, Cijo Jose, Vasil Khalidov, Marc Szafraniec, Seungeun Yi, Micha{\"e}l Ramamonjisoa, et~al.
\newblock Dinov3.
\newblock \emph{arXiv preprint arXiv:2508.10104}, 2025.

\bibitem[Smith et~al.(2025)Smith, Mannering, and Marcu]{pmlr-v267-smith25a}
Damian Smith, Harvey Mannering, and Antonia Marcu.
\newblock Functional alignment can mislead: Examining model stitching.
\newblock In \emph{ICML}, 2025.

\bibitem[Srinivasan et~al.(2021)Srinivasan, Raman, Chen, Bendersky, and Najork]{srinivasan2021wit}
Krishna Srinivasan, Karthik Raman, Jiecao Chen, Michael Bendersky, and Marc Najork.
\newblock Wit: Wikipedia-based image text dataset for multimodal multilingual machine learning.
\newblock In \emph{ACM SIGIR conference on research and development in information retrieval}, 2021.

\bibitem[Sutskever(2023)]{Sutskever2023}
Ilya Sutskever.
\newblock “the mastermind behind gpt-4 and the future of ai” — eye on a.i. (podcast, season 2 episode 118).
\newblock \url{https://podcasts.apple.com/us/podcast/ilya-sutskever-the-mastermind-behind-gpt-4-and/id1438378439?i=1000604382855}, mar 2023.
\newblock Accessed: 2026-02-28.

\bibitem[Tasker et~al.(2026)Tasker, Betser, Gofer, Levi, and Gilboa]{tasker2026universalnormalembedding}
Chen Tasker, Roy Betser, Eyal Gofer, Meir~Yossef Levi, and Guy Gilboa.
\newblock The universal normal embedding.
\newblock In \emph{CVPR}, 2026.

\bibitem[Tjandrasuwita et~al.(2025)Tjandrasuwita, Ekbote, Ziyin, and Liang]{tjandrasuwita2025understandingemergencemultimodalrepresentation}
Megan Tjandrasuwita, Chanakya Ekbote, Liu Ziyin, and Paul~Pu Liang.
\newblock Understanding the emergence of multimodal representation alignment.
\newblock In \emph{ICML}, 2025.

\bibitem[Touvron et~al.(2023)Touvron, Lavril, Izacard, Martinet, Lachaux, Lacroix, Rozi{\`e}re, Goyal, Hambro, Azhar, et~al.]{touvron2023llama}
Hugo Touvron, Thibaut Lavril, Gautier Izacard, Xavier Martinet, Marie-Anne Lachaux, Timoth{\'e}e Lacroix, Baptiste Rozi{\`e}re, Naman Goyal, Eric Hambro, Faisal Azhar, et~al.
\newblock Llama: Open and efficient foundation language models.
\newblock \emph{arXiv preprint arXiv:2302.13971}, 2023.

\bibitem[Uexk{\"u}ll \& Kriszat(1934)Uexk{\"u}ll and Kriszat]{uexkull1934streifzuge}
Johann~von Uexk{\"u}ll and Georg Kriszat.
\newblock \emph{Streifzuge durch die {U}mwelten von {T}ieren und {M}enschen Ein {B}ilderbuch unsichtbarer {W}elten}.
\newblock Springer, 1934.

\bibitem[Von~Ahn et~al.(2007)Von~Ahn, Ginosar, Kedia, and Blum]{von2007improving}
Luis Von~Ahn, Shiry Ginosar, Mihir Kedia, and Manuel Blum.
\newblock Improving image search with phetch.
\newblock In \emph{ICASSP}, 2007.

\bibitem[Wang et~al.(2023{\natexlab{a}})Wang, Huang, Zhao, Tong, He, Wang, Wang, and Qiao]{wang2023videomaev2scalingvideo}
Limin Wang, Bingkun Huang, Zhiyu Zhao, Zhan Tong, Yinan He, Yi~Wang, Yali Wang, and Yu~Qiao.
\newblock Videomae v2: Scaling video masked autoencoders with dual masking.
\newblock In \emph{CVPR}, 2023{\natexlab{a}}.

\bibitem[Wang et~al.(2023{\natexlab{b}})Wang, Zelikman, Poesia, Pu, Haber, and Goodman]{wang2023hypothesis}
Ruocheng Wang, Eric Zelikman, Gabriel Poesia, Yewen Pu, Nick Haber, and Noah~D Goodman.
\newblock Hypothesis search: Inductive reasoning with language models.
\newblock \emph{arXiv preprint arXiv:2309.05660}, 2023{\natexlab{b}}.

\bibitem[Wang et~al.(2019)Wang, Wu, Chen, Li, Wang, and Wang]{wang2019vatex}
Xin Wang, Jiawei Wu, Junkun Chen, Lei Li, Yuan-Fang Wang, and William~Yang Wang.
\newblock Vatex: A large-scale, high-quality multilingual dataset for video-and-language research.
\newblock In \emph{ICCV}, 2019.

\bibitem[Wittgenstein(1953/2009)]{Wittgenstein1953-WITPI-4}
Ludwig Wittgenstein.
\newblock \emph{Philosophical Investigations}.
\newblock Wiley-Blackwell, 4th, revised edition, 1953/2009.

\bibitem[Wu et~al.(2023)Wu, Chen, Zhang, Hui, Berg-Kirkpatrick, and Dubnov]{wu2023large}
Yusong Wu, Ke~Chen, Tianyu Zhang, Yuchen Hui, Taylor Berg-Kirkpatrick, and Shlomo Dubnov.
\newblock Large-scale contrastive language-audio pretraining with feature fusion and keyword-to-caption augmentation.
\newblock In \emph{ICASSP}, 2023.

\bibitem[Zellers et~al.(2019)Zellers, Holtzman, Bisk, Farhadi, and Choi]{zellers2019hellaswag}
Rowan Zellers, Ari Holtzman, Yonatan Bisk, Ali Farhadi, and Yejin Choi.
\newblock Hellaswag: Can a machine really finish your sentence?
\newblock In \emph{ACL}, 2019.

\bibitem[Zhu et~al.(2017)Zhu, Park, Isola, and Efros]{zhu2017unpairedimagetoimagetranslationusing}
Jun-Yan Zhu, Taesung Park, Phillip Isola, and Alexei~A. Efros.
\newblock Unpaired image-to-image translation using cycle-consistent adversarial networks.
\newblock In \emph{ICCV}, 2017.

\bibitem[Zhu et~al.(2026)Zhu, Han, Guibas, Pătrăucean, and Ovsjanikov]{zhu2026dynamicreflectionsprobingvideo}
Tyler Zhu, Tengda Han, Leonidas Guibas, Viorica Pătrăucean, and Maks Ovsjanikov.
\newblock Dynamic reflections: Probing video representations with text alignment.
\newblock In \emph{ICLR}, 2026.

\end{thebibliography}


\begin{thebibliography}{44}
\providecommand{\natexlab}[1]{#1}
\providecommand{\url}[1]{\texttt{#1}}
\expandafter\ifx\csname urlstyle\endcsname\relax
  \providecommand{\doi}[1]{doi: #1}\else
  \providecommand{\doi}{doi: \begingroup \urlstyle{rm}\Url}\fi

\bibitem[Abdin et~al.(2024)Abdin, Aneja, Behl, Bubeck, Eldan, Gunasekar, Harrison, Hewett, Javaheripi, Kauffmann, Lee, Lee, Li, Liu, Mendes, Nguyen, Price, de~Rosa, Saarikivi, Salim, Shah, Wang, Ward, Wu, Yu, Zhang, and Zhang]{abdin2024phi4technicalreport}
Marah Abdin, Jyoti Aneja, Harkirat Behl, Sébastien Bubeck, Ronen Eldan, Suriya Gunasekar, Michael Harrison, Russell~J. Hewett, Mojan Javaheripi, Piero Kauffmann, James~R. Lee, Yin~Tat Lee, Yuanzhi Li, Weishung Liu, Caio C.~T. Mendes, Anh Nguyen, Eric Price, Gustavo de~Rosa, Olli Saarikivi, Adil Salim, Shital Shah, Xin Wang, Rachel Ward, Yue Wu, Dingli Yu, Cyril Zhang, and Yi~Zhang.
\newblock Phi-4 technical report.
\newblock \emph{arXiv preprint arXiv:2412.08905}, 2024.

\bibitem[Bahng et~al.(2025)Bahng, Chan, Durand, and Isola]{bahng2025cycle}
Hyojin Bahng, Caroline Chan, Fredo Durand, and Phillip Isola.
\newblock Cycle consistency as reward: Learning image-text alignment without human preferences.
\newblock \emph{arXiv preprint arXiv:2506.02095}, 2025.

\bibitem[Clark et~al.(2018)Clark, Cowhey, Etzioni, Khot, Sabharwal, Schoenick, and Tafjord]{clark2018think}
Peter Clark, Isaac Cowhey, Oren Etzioni, Tushar Khot, Ashish Sabharwal, Carissa Schoenick, and Oyvind Tafjord.
\newblock Think you have solved question answering? try arc, the ai2 reasoning challenge.
\newblock \emph{arXiv preprint arXiv:1803.05457}, 2018.

\bibitem[Cobbe et~al.(2021)Cobbe, Kosaraju, Bavarian, Chen, Jun, Kaiser, Plappert, Tworek, Hilton, Nakano, Hesse, and Schulman]{cobbe2021gsm8k}
Karl Cobbe, Vineet Kosaraju, Mohammad Bavarian, Mark Chen, Heewoo Jun, Lukasz Kaiser, Matthias Plappert, Jerry Tworek, Jacob Hilton, Reiichiro Nakano, Christopher Hesse, and John Schulman.
\newblock Training verifiers to solve math word problems.
\newblock \emph{arXiv preprint arXiv:2110.14168}, 2021.

\bibitem[Dang et~al.(2024)Dang, Singh, D'souza, Ahmadian, Salamanca, Smith, Peppin, Hong, Govindassamy, Zhao, Kublik, Amer, Aryabumi, Campos, Tan, Kocmi, Strub, Grinsztajn, Flet-Berliac, Locatelli, Lin, Talupuru, Venkitesh, Cairuz, Yang, Chung, Ko, Shi, Shukayev, Bae, Piktus, Castagné, Cruz-Salinas, Kim, Crawhall-Stein, Morisot, Roy, Blunsom, Zhang, Gomez, Frosst, Fadaee, Ermis, Üstün, and Hooker]{dang2024ayaexpansecombiningresearch}
John Dang, Shivalika Singh, Daniel D'souza, Arash Ahmadian, Alejandro Salamanca, Madeline Smith, Aidan Peppin, Sungjin Hong, Manoj Govindassamy, Terrence Zhao, Sandra Kublik, Meor Amer, Viraat Aryabumi, Jon~Ander Campos, Yi-Chern Tan, Tom Kocmi, Florian Strub, Nathan Grinsztajn, Yannis Flet-Berliac, Acyr Locatelli, Hangyu Lin, Dwarak Talupuru, Bharat Venkitesh, David Cairuz, Bowen Yang, Tim Chung, Wei-Yin Ko, Sylvie~Shang Shi, Amir Shukayev, Sammie Bae, Aleksandra Piktus, Roman Castagné, Felipe Cruz-Salinas, Eddie Kim, Lucas Crawhall-Stein, Adrien Morisot, Sudip Roy, Phil Blunsom, Ivan Zhang, Aidan Gomez, Nick Frosst, Marzieh Fadaee, Beyza Ermis, Ahmet Üstün, and Sara Hooker.
\newblock Aya expanse: Combining research breakthroughs for a new multilingual frontier.
\newblock \emph{arXiv preprint arXiv:2412.04261}, 2024.

\bibitem[DeepSeek-AI et~al.(2025)DeepSeek-AI, Guo, et~al.]{deepseek_r1_2025}
DeepSeek-AI, Daya Guo, et~al.
\newblock Deepseek-r1: Incentivizing reasoning capability in llms via reinforcement learning.
\newblock \emph{arXiv preprint arXiv:2501.12948}, 2025.

\bibitem[Deng et~al.(2009)Deng, Dong, Socher, Li, Li, and Fei-Fei]{deng2009imagenet}
Jia Deng, Wei Dong, Richard Socher, Li-Jia Li, Kai Li, and Li~Fei-Fei.
\newblock Imagenet: A large-scale hierarchical image database.
\newblock In \emph{CVPR}, 2009.

\bibitem[Dosovitskiy et~al.(2021)Dosovitskiy, Beyer, Kolesnikov, Weissenborn, Zhai, Unterthiner, Dehghani, Minderer, Heigold, Gelly, Uszkoreit, and Houlsby]{dosovitskiy2021an}
Alexey Dosovitskiy, Lucas Beyer, Alexander Kolesnikov, Dirk Weissenborn, Xiaohua Zhai, Thomas Unterthiner, Mostafa Dehghani, Matthias Minderer, Georg Heigold, Sylvain Gelly, Jakob Uszkoreit, and Neil Houlsby.
\newblock An image is worth 16x16 words: Transformers for image recognition at scale.
\newblock In \emph{ICLR}, 2021.

\bibitem[Gao et~al.(2024)Gao, Tow, Abbasi, Biderman, Black, DiPofi, Foster, Golding, Hsu, Le~Noac'h, Li, McDonell, Muennighoff, Ociepa, Phang, Reynolds, Schoelkopf, Skowron, Sutawika, Tang, Thite, Wang, Wang, and Zou]{eval-harness}
Leo Gao, Jonathan Tow, Baber Abbasi, Stella Biderman, Sid Black, Anthony DiPofi, Charles Foster, Laurence Golding, Jeffrey Hsu, Alain Le~Noac'h, Haonan Li, Kyle McDonell, Niklas Muennighoff, Chris Ociepa, Jason Phang, Laria Reynolds, Hailey Schoelkopf, Aviya Skowron, Lintang Sutawika, Eric Tang, Anish Thite, Ben Wang, Kevin Wang, and Andy Zou.
\newblock The language model evaluation harness.
\newblock \emph{Zenodo}, 07 2024.
\newblock \doi{10.5281/zenodo.12608602}.
\newblock URL \url{https://zenodo.org/records/12608602}.

\bibitem[Gemma~Team(2024{\natexlab{a}})]{gemma_2024}
Google~DeepMind Gemma~Team.
\newblock Gemma: Open models based on gemini research and technology.
\newblock \emph{arXiv preprint arXiv:2403.08295}, 2024{\natexlab{a}}.

\bibitem[Gemma~Team(2024{\natexlab{b}})]{gemma_2_2024}
Google~DeepMind Gemma~Team.
\newblock Gemma 2: Improving open language models at a practical size.
\newblock \emph{arXiv preprint arXiv:2408.00118}, 2024{\natexlab{b}}.

\bibitem[Gemma~Team(2025)]{gemma_3_2025}
Google~DeepMind Gemma~Team.
\newblock Gemma 3 technical report.
\newblock \emph{arXiv preprint arXiv:2503.19786}, 2025.

\bibitem[Geng \& Liu(2023)Geng and Liu]{openlm2023openllama}
Xinyang Geng and Hao Liu.
\newblock Openllama: An open reproduction of llama, 2023.
\newblock URL \url{https://github.com/openlm-research/open_llama}.

\bibitem[Gokaslan \& Cohen(2019)Gokaslan and Cohen]{Gokaslan2019OpenWeb}
Aaron Gokaslan and Vanya Cohen.
\newblock Openwebtext corpus.
\newblock \url{http://Skylion007.github.io/OpenWebTextCorpus}, 2019.

\bibitem[Grattafiori et~al.(2024)Grattafiori, Dubey, Jauhri, Pandey, Kadian, Al-Dahle, Letman, Mathur, Schelten, Vaughan, et~al.]{grattafiori2024llama}
Aaron Grattafiori, Abhimanyu Dubey, Abhinav Jauhri, Abhinav Pandey, Abhishek Kadian, Ahmad Al-Dahle, Aiesha Letman, Akhil Mathur, Alan Schelten, Alex Vaughan, et~al.
\newblock The llama 3 herd of models.
\newblock \emph{arXiv preprint arXiv:2407.21783}, 2024.

\bibitem[Groeneveld et~al.(2024)Groeneveld, Beltagy, Walsh, Bhagia, Kinney, Tafjord, Jha, Ivison, Magnusson, Wang, Arora, Atkinson, Authur, Chandu, Cohan, Dumas, Elazar, Gu, Hessel, Khot, Merrill, Morrison, Muennighoff, Naik, Nam, Peters, Pyatkin, Ravichander, Schwenk, Shah, Smith, Strubell, Subramani, Wortsman, Dasigi, Lambert, Richardson, Zettlemoyer, Dodge, Lo, Soldaini, Smith, and Hajishirzi]{groeneveld2024olmoacceleratingsciencelanguage}
Dirk Groeneveld, Iz~Beltagy, Pete Walsh, Akshita Bhagia, Rodney Kinney, Oyvind Tafjord, Ananya~Harsh Jha, Hamish Ivison, Ian Magnusson, Yizhong Wang, Shane Arora, David Atkinson, Russell Authur, Khyathi~Raghavi Chandu, Arman Cohan, Jennifer Dumas, Yanai Elazar, Yuling Gu, Jack Hessel, Tushar Khot, William Merrill, Jacob Morrison, Niklas Muennighoff, Aakanksha Naik, Crystal Nam, Matthew~E. Peters, Valentina Pyatkin, Abhilasha Ravichander, Dustin Schwenk, Saurabh Shah, Will Smith, Emma Strubell, Nishant Subramani, Mitchell Wortsman, Pradeep Dasigi, Nathan Lambert, Kyle Richardson, Luke Zettlemoyer, Jesse Dodge, Kyle Lo, Luca Soldaini, Noah~A. Smith, and Hannaneh Hajishirzi.
\newblock Olmo: Accelerating the science of language models.
\newblock In \emph{ACL}, 2024.

\bibitem[Gröger et~al.(2026)Gröger, Wen, and Brbić]{groger2026revisitingplatonicrepresentationhypothesis}
Fabian Gröger, Shuo Wen, and Maria Brbić.
\newblock Revisiting the platonic representation hypothesis: An aristotelian view.
\newblock \emph{arXiv preprint arXiv:2602.14486}, 2026.

\bibitem[Hendrycks et~al.(2021)Hendrycks, Burns, Basart, Zou, Mazeika, Song, and Steinhardt]{hendrycks2021mmlu}
Dan Hendrycks, Collin Burns, Steven Basart, Andy Zou, Mantas Mazeika, Dawn Song, and Jacob Steinhardt.
\newblock Measuring massive multitask language understanding.
\newblock In \emph{ICLR}, 2021.

\bibitem[Huh et~al.(2024)Huh, Cheung, Wang, and Isola]{huh2024prh}
Minyoung Huh, Brian Cheung, Tongzhou Wang, and Phillip Isola.
\newblock The platonic representation hypothesis.
\newblock In \emph{ICML}, 2024.

\bibitem[Jiang et~al.(2023)Jiang, Sablayrolles, Mensch, Bamford, Chaplot, de~las Casas, Bressand, Lengyel, Lample, Saulnier, Renard~Lavaud, Lachaux, Stock, Scao, Lavril, Wang, Lacroix, and El~Sayed]{jiang2023mistral}
Albert~Q. Jiang, Alexandre Sablayrolles, Arthur Mensch, Chris Bamford, Devendra~Singh Chaplot, Diego de~las Casas, Florian Bressand, Gianna Lengyel, Guillaume Lample, Lucile Saulnier, L{\'e}lio Renard~Lavaud, Marie-Anne Lachaux, Pierre Stock, Teven~Le Scao, Thibaut Lavril, Thomas Wang, Timoth{\'e}e Lacroix, and William El~Sayed.
\newblock Mistral 7{B}.
\newblock \emph{arXiv preprint arXiv:2310.06825}, 2023.

\bibitem[Jiang et~al.(2024)Jiang, Sablayrolles, Roux, Mensch, Savary, Bamford, Chaplot, Casas, Hanna, Bressand, et~al.]{jiang2024mixtral_experts}
Albert~Q Jiang, Alexandre Sablayrolles, Antoine Roux, Arthur Mensch, Blanche Savary, Chris Bamford, Devendra~Singh Chaplot, Diego de~las Casas, Emma~Bou Hanna, Florian Bressand, et~al.
\newblock Mixtral of experts.
\newblock \emph{arXiv preprint arXiv:2401.04088}, 2024.

\bibitem[Kornblith et~al.(2019)Kornblith, Norouzi, Lee, and Hinton]{kornblith2019similarity}
Simon Kornblith, Mohammad Norouzi, Honglak Lee, and Geoffrey Hinton.
\newblock Similarity of neural network representations revisited.
\newblock In \emph{ICML}, 2019.

\bibitem[Liu et~al.(2023)Liu, Liu, Cui, Teng, Duan, Zhou, and Zhang]{liu2023logiqa2}
Hanmeng Liu, Jian Liu, Leyang Cui, Zhiyang Teng, Nan Duan, Ming Zhou, and Yue Zhang.
\newblock Logiqa2.0: The logicqa dataset for logical reasoning.
\newblock \emph{IEEE Transactions on Audio, Speech, and Language Processing}, 2023.

\bibitem[Merity et~al.(2016)Merity, Xiong, Bradbury, and Socher]{merity2016pointer}
Stephen Merity, Caiming Xiong, James Bradbury, and Richard Socher.
\newblock Pointer sentinel mixture models.
\newblock \emph{arXiv preprint arXiv:1609.07843}, 2016.

\bibitem[Muennighoff et~al.(2023)Muennighoff, Wang, Sutawika, Roberts, Biderman, Scao, Bari, Shen, Yong, Schoelkopf, Tang, Radev, Aji, Almubarak, Albanie, Alyafeai, Webson, Raff, and Raffel]{muennighoff2023crosslingualgeneralizationmultitaskfinetuning}
Niklas Muennighoff, Thomas Wang, Lintang Sutawika, Adam Roberts, Stella Biderman, Teven~Le Scao, M~Saiful Bari, Sheng Shen, Zheng-Xin Yong, Hailey Schoelkopf, Xiangru Tang, Dragomir Radev, Alham~Fikri Aji, Khalid Almubarak, Samuel Albanie, Zaid Alyafeai, Albert Webson, Edward Raff, and Colin Raffel.
\newblock Crosslingual generalization through multitask finetuning.
\newblock In \emph{ACL}, 2023.

\bibitem[OLMo et~al.(2025)OLMo, Walsh, Soldaini, Groeneveld, Lo, Arora, Bhagia, Gu, Huang, Jordan, Lambert, Schwenk, Tafjord, Anderson, Atkinson, Brahman, Clark, Dasigi, Dziri, Ettinger, Guerquin, Heineman, Ivison, Koh, Liu, Malik, Merrill, Miranda, Morrison, Murray, Nam, Poznanski, Pyatkin, Rangapur, Schmitz, Skjonsberg, Wadden, Wilhelm, Wilson, Zettlemoyer, Farhadi, Smith, and Hajishirzi]{olmo20252olmo2furious}
Team OLMo, Pete Walsh, Luca Soldaini, Dirk Groeneveld, Kyle Lo, Shane Arora, Akshita Bhagia, Yuling Gu, Shengyi Huang, Matt Jordan, Nathan Lambert, Dustin Schwenk, Oyvind Tafjord, Taira Anderson, David Atkinson, Faeze Brahman, Christopher Clark, Pradeep Dasigi, Nouha Dziri, Allyson Ettinger, Michal Guerquin, David Heineman, Hamish Ivison, Pang~Wei Koh, Jiacheng Liu, Saumya Malik, William Merrill, Lester James~V. Miranda, Jacob Morrison, Tyler Murray, Crystal Nam, Jake Poznanski, Valentina Pyatkin, Aman Rangapur, Michael Schmitz, Sam Skjonsberg, David Wadden, Christopher Wilhelm, Michael Wilson, Luke Zettlemoyer, Ali Farhadi, Noah~A. Smith, and Hannaneh Hajishirzi.
\newblock 2 olmo 2 furious.
\newblock \emph{arXiv preprint arXiv:2501.00656}, 2025.

\bibitem[OpenAI(2025)]{gpt_oss_2025}
OpenAI.
\newblock Introducing gpt-oss, 2025.
\newblock URL \url{https://openai.com/index/introducing-gpt-oss/}.

\bibitem[Oquab et~al.(2024)Oquab, Darcet, Moutakanni, Vo, Szafraniec, Khalidov, Fernandez, Haziza, Massa, El-Nouby, Assran, Ballas, Galuba, Howes, Huang, Li, Misra, Rabbat, Sharma, Synnaeve, Xu, Jegou, Mairal, Labatut, Joulin, and Bojanowski]{oquab2024dinov2learningrobustvisual}
Maxime Oquab, Timothée Darcet, Théo Moutakanni, Huy Vo, Marc Szafraniec, Vasil Khalidov, Pierre Fernandez, Daniel Haziza, Francisco Massa, Alaaeldin El-Nouby, Mahmoud Assran, Nicolas Ballas, Wojciech Galuba, Russell Howes, Po-Yao Huang, Shang-Wen Li, Ishan Misra, Michael Rabbat, Vasu Sharma, Gabriel Synnaeve, Hu~Xu, Hervé Jegou, Julien Mairal, Patrick Labatut, Armand Joulin, and Piotr Bojanowski.
\newblock Dinov2: Learning robust visual features without supervision.
\newblock \emph{TMLR}, 2024.

\bibitem[Pichai et~al.(2025)Pichai, Hassabis, and Kavukcuoglu]{pichai2025gemini3}
Sundar Pichai, Demis Hassabis, and Koray Kavukcuoglu.
\newblock A new era of intelligence with {Gemini} 3.
\newblock Google Blog (The Keyword), November 2025.
\newblock URL \url{https://blog.google/products-and-platforms/products/gemini/gemini-3/}.
\newblock Accessed: 2026-01-01.

\bibitem[Qwen~Team(2025)]{qwen3_2025}
Alibaba~Cloud Qwen~Team.
\newblock Qwen3 technical report.
\newblock \emph{arXiv preprint arXiv:2505.09388}, 2025.

\bibitem[Research(2025)]{granite33_2025}
IBM Research.
\newblock Granite 3.3 8b base, 2025.
\newblock URL \url{https://huggingface.co/ibm-granite/granite-3.3-8b-base}.

\bibitem[Schuhmann et~al.(2021)Schuhmann, Vencu, Beaumont, Kaczmarczyk, Mullis, Katta, Coombes, Jitsev, and Komatsuzaki]{schuhmann2021laion}
Christoph Schuhmann, Richard Vencu, Romain Beaumont, Robert Kaczmarczyk, Clayton Mullis, Aarush Katta, Theo Coombes, Jenia Jitsev, and Aran Komatsuzaki.
\newblock Laion-400m: Open dataset of clip-filtered 400 million image-text pairs.
\newblock \emph{arXiv preprint arXiv:2111.02114}, 2021.

\bibitem[Singh et~al.()Singh, Hu, Goswami, Couairon, Galuba, Rohrbach, and Kiela]{pmd_datset}
Amanpreet Singh, Ronghang Hu, Vedanuj Goswami, Guillaume Couairon, Wojciech Galuba, Marcus Rohrbach, and Douwe Kiela.
\newblock Public multimodal dataset ({PMD}).
\newblock URL \url{https://huggingface.co/datasets/facebook/pmd}.

\bibitem[Singh et~al.(2022)Singh, Hu, Goswami, Couairon, Galuba, Rohrbach, and Kiela]{singh2022flava}
Amanpreet Singh, Ronghang Hu, Vedanuj Goswami, Guillaume Couairon, Wojciech Galuba, Marcus Rohrbach, and Douwe Kiela.
\newblock Flava: A foundational language and vision alignment model.
\newblock In \emph{CVPR}, 2022.

\bibitem[Srinivasan et~al.(2021)Srinivasan, Raman, Chen, Bendersky, and Najork]{srinivasan2021wit}
Krishna Srinivasan, Karthik Raman, Jiecao Chen, Michael Bendersky, and Marc Najork.
\newblock Wit: Wikipedia-based image text dataset for multimodal multilingual machine learning.
\newblock In \emph{ACM SIGIR conference on research and development in information retrieval}, 2021.

\bibitem[Team(2024)]{Falcon3}
Falcon-LLM Team.
\newblock The falcon 3 family of open models, 2024.
\newblock URL \url{https://huggingface.co/blog/falcon3}.

\bibitem[Team et~al.(2023)Team, Anil, Borgeaud, Alayrac, Yu, Soricut, Schalkwyk, Dai, Hauth, Millican, et~al.]{team2023gemini}
Gemini Team, Rohan Anil, Sebastian Borgeaud, Jean-Baptiste Alayrac, Jiahui Yu, Radu Soricut, Johan Schalkwyk, Andrew~M Dai, Anja Hauth, Katie Millican, et~al.
\newblock Gemini: a family of highly capable multimodal models.
\newblock \emph{arXiv preprint arXiv:2312.11805}, 2023.

\bibitem[Touvron et~al.(2023{\natexlab{a}})Touvron, Lavril, Izacard, Martinet, Lachaux, Lacroix, Rozi{\`e}re, Goyal, Hambro, Azhar, et~al.]{touvron2023llama}
Hugo Touvron, Thibaut Lavril, Gautier Izacard, Xavier Martinet, Marie-Anne Lachaux, Timoth{\'e}e Lacroix, Baptiste Rozi{\`e}re, Naman Goyal, Eric Hambro, Faisal Azhar, et~al.
\newblock Llama: Open and efficient foundation language models.
\newblock \emph{arXiv preprint arXiv:2302.13971}, 2023{\natexlab{a}}.

\bibitem[Touvron et~al.(2023{\natexlab{b}})Touvron, Martin, Stone, Albert, Almahairi, Babaei, Bashlykov, Batra, Bhargava, Bhosale, et~al.]{touvron2023llama2}
Hugo Touvron, Louis Martin, Kevin Stone, Peter Albert, Amjad Almahairi, Yasmine Babaei, Nikolay Bashlykov, Soumya Batra, Prajjwal Bhargava, Shruti Bhosale, et~al.
\newblock Llama 2: Open foundation and fine-tuned chat models.
\newblock \emph{arXiv preprint arXiv:2307.09288}, 2023{\natexlab{b}}.

\bibitem[Wightman(2019)]{rwightman2019timm}
Ross Wightman.
\newblock Pytorch image models, 2019.
\newblock URL \url{https://github.com/huggingface/pytorch-image-models}.

\bibitem[Yang et~al.(2026)Yang, Li, Li, Lei, Wang, Mohamed, Xie, Zhao, He, and Xu]{yang2026pursuit}
Lihe Yang, Shang-Wen Li, Yang Li, Xinjie Lei, Dong Wang, Abdelrahman Mohamed, Saining Xie, Hengshuang Zhao, Kaiming He, and Hu~Xu.
\newblock In pursuit of pixel supervision for visual pre-training.
\newblock In \emph{CVPR}, 2026.

\bibitem[Young et~al.(2024)Young, Chen, Li, Huang, Zhang, Zhang, Wang, Li, Zhu, Chen, et~al.]{yi2024}
Alex Young, Bei Chen, Chao Li, Chengen Huang, Ge~Zhang, Guanwei Zhang, Guoyin Wang, Heng Li, Jiangcheng Zhu, Jianqun Chen, et~al.
\newblock Yi: Open foundation models by 01.ai.
\newblock \emph{arXiv preprint arXiv:2403.04652}, 2024.

\bibitem[Zauner(2010)]{zauner2010implementation}
Christoph Zauner.
\newblock Implementation and benchmarking of perceptual image hash functions.
\newblock 2010.

\bibitem[Zellers et~al.(2019)Zellers, Holtzman, Bisk, Farhadi, and Choi]{zellers2019hellaswag}
Rowan Zellers, Ari Holtzman, Yonatan Bisk, Ali Farhadi, and Yejin Choi.
\newblock Hellaswag: Can a machine really finish your sentence?
\newblock In \emph{ACL}, 2019.

\end{thebibliography}
\end{document}